\documentclass{article}

\usepackage{natbib}

\usepackage[utf8]{inputenc}

\usepackage{amssymb,enumitem}
\usepackage[margin=1in]{geometry}
\usepackage[table]{xcolor}
\usepackage{minitoc}
\usepackage{array}
\usepackage{makecell}
\usepackage{pdfpages}
\usepackage{mathtools}
\usepackage{tikz}
\usepackage{multirow}        
\usepackage{booktabs}

\usepackage{arydshln} 
\usepackage{xcolor}         
\definecolor{DarkGreen}{rgb}{0.1,0.5,0.1}
\definecolor{DarkRed}{rgb}{0.5,0.1,0.1}
\definecolor{DarkBlue}{rgb}{0.1,0.1,0.5}
\definecolor{DarkYellow}{rgb}{.79,.79,0}
\definecolor{unitednationsblue}{rgb}{0.36, 0.57, 0.9}
\definecolor{blue_ppt}{rgb}{0,0.6,0.93}

\definecolor{darkblue_ppt}{rgb}{0.05,0.4,0.8}

\definecolor{orange_ppt}{rgb}{0.82,0.5,0}

\usepackage{color} 
\definecolor{yc}{RGB}{225,0,0}

\usepackage{mathabx}

\usepackage{thm-restate}

\usepackage{hyperref}
\hypersetup{colorlinks,citecolor=blue,filecolor=blue,linkcolor=blue,urlcolor=blue}
\usepackage{cleveref}
\usepackage{tablefootnote} 
\DeclarePairedDelimiter{\ceil}{\lceil}{\rceil}

\newcolumntype{?}{!{\vrule width 1pt}}
\usepackage{amsmath}
\usepackage{amsthm}
\usepackage{algorithm,algorithmic}

\usepackage{graphicx} 
\usepackage{subfigure}
\usepackage{bbm}

\usepackage{extarrows}
\usepackage{bbm}
\theoremstyle{plain}
\newtheorem{lm}{Lemma} 
\newtheorem{definition}{Definition}
\newtheorem{thm}{Theorem}
\newtheorem{prop}{Proposition}
\newtheorem{asmp}{Assumption}

\allowdisplaybreaks
\usepackage{xspace}

\usepackage{amsmath,amsfonts,bm}
\usepackage{algorithm,algorithmic}

\def\ceil#1{\lceil #1 \rceil}
\def\floor#1{\lfloor #1 \rfloor}
\def\1{\bm{1}}

\def\vone{{\bm{1}}}

\DeclareMathAlphabet{\mathsfit}{\encodingdefault}{\sfdefault}{m}{sl}
\SetMathAlphabet{\mathsfit}{bold}{\encodingdefault}{\sfdefault}{bx}{n}

\def\gA{{\mathcal{A}}}
\def\gB{{\mathcal{B}}}
\def\gC{{\mathcal{C}}}
\def\gD{{\mathcal{D}}}
\def\gE{{\mathcal{E}}}
\def\gF{{\mathcal{F}}}

\def\gO{{\mathcal{O}}}
\def\gP{{\mathcal{P}}}
\def\gQ{{\mathcal{Q}}}

\def\gS{{\mathcal{S}}}
\def\gT{{\mathcal{T}}}
\def\gU{{\mathcal{U}}}
\def\gV{{\mathcal{V}}}

\newcommand{\E}{\mathbb{E}}

\newcommand{\R}{\mathbb{R}}

\newcommand{\softmax}{\mathrm{softmax}}

\DeclareMathOperator*{\argmax}{arg\,max}

\DeclareMathOperator{\diag}{diag}

\usepackage{amssymb}

\usepackage{xcolor}
\definecolor{mydarkblue}{rgb}{0,0.08,0.45}
\definecolor{mygreen}{rgb}{0.032, 0.6392, 0.2039}
\definecolor{mypurple}{HTML}{B266FF}

\def\RR{{\mathbb R}}
\def\NN{{\mathbb N}}
\def\EE{{\mathbb E}}
\def\PP{{\mathbb P}}

\def\RR{{\mathbb R}}

\def\NN{{\mathbb N}}
\def\EE{{\mathbb E}}

\newcommand{\norm}[1]{\left\| #1 \right\|}

\newcommand{\sm}{\mathsf{softmax}}

\newcommand{\poly}{\mathsf{poly}}
\newcommand{\up}{\texttt{up}}
\newcommand{\down}{\texttt{down}}

\definecolor{blue_ppt}{rgb}{0,0.47,0.97}
\definecolor{violet}{RGB}{138,43,226}

\newcommand{\tr}{\textnormal{tr}}

\usepackage[utf8]{inputenc} 
\usepackage[T1]{fontenc}    
\usepackage{hyperref}       
\usepackage{xurl}           
\usepackage{url}            
\usepackage{booktabs}       
\usepackage{amsfonts}       
\usepackage{nicefrac}       
\usepackage{microtype}      
\usepackage{xcolor,wrapfig}         

\title{Agentic Transformers Provably Learn to Search \\ via  Reinforcement Learning }

\author{%
 Tong Yang\thanks{Department of Electrical and Computer Engineering, Carnegie Mellon University; email: \texttt{tongyang@andrew.cmu.edu}. } \\
CMU    \\
	\and
	Yu Huang\thanks{Department of Statistics and Data Science, Wharton School, University of Pennsylvania; email: \texttt{yuh42@wharton.upenn.edu}. } \\
 UPenn  \\
 	\and
	Yingbin Liang\thanks{Department of Electrical and Computer Engineering, The Ohio State University; email: \texttt{liang.889@osu.edu}. }\\
	OSU\\
	\and
	Yuejie Chi\thanks{Department of Statistics and Data Science, Yale University; email: \texttt{yuejie.chi@yale.edu}. }\\
	Yale\\ 	
}
\date{\today}

\begin{document}

\maketitle

\begin{abstract}
    Tree search is a central abstraction behind many language-agent reasoning and decision-making tasks: agents must explore actions, remember failures, and backtrack toward promising alternatives. Yet, we lack a theoretical understanding of how transformer-based policies acquire such search capabilities from the training dynamics of reinforcement learning (RL). We study this question in a stochastic $k$-ary tree environment, where an agentic transformer observes only its trajectory history through interaction and receives a terminal reward for reaching a hidden leaf goal node. 
    We first construct a two-head transformer that implements randomized depth-first search (DFS): one head tracks previous actions, while the other detects failure outcomes and triggers backtracking. 
    We then analyze the training dynamics of policy gradient under a depth-wise curriculum, showing that this same DFS mechanism emerges in stages from sparse reinforcement feedback without expert demonstrations. 
    The resulting policy exhibits depth generalization: after training only on depth-$1$ and depth-$2$ trees, it succeeds on deeper full trees. We further show that, under imbalanced goal distributions, discounting the return leads to a ranked DFS policy that prioritizes higher-probability branches. Overall, our results identify a mechanistic normal form for transformer-based search, in which attention heads specialize and cooperate to extract decision-relevant traces from context and convert them into agentic action selection via RL training.
\end{abstract}

\setcounter{tocdepth}{2}
\tableofcontents

\section{Introduction}\label{sec:motivation}

Tree search is a core computational structure in modern learning agents: they must explore different paths, memorize failed branches, and backtrack towards promising alternatives while searching for a target embedded at the leaf of the tree~\citep{browne2012survey,silver2016mastering,silver2018general,lample2022hypertree}. It underlies classical systems such as Monte Carlo tree search in game playing and proof search in automated theorem proving, and has recently resurged as an explicit test-time reasoning mechanism in language agents, such as Tree of Thoughts, planning-based reasoning, Language Agent Tree Search, and search for web agents~\citep{yao2023tree,hao2023reasoning,zhou2024language,koh2024tree,snell2024scaling}. 

In these agentic settings, the search tree is usually not handed to the agent as input. Rather, it is a hidden environment revealed through interaction: each action exposes only local information, and the reward can be highly sparse depending on the hidden goal. As a result, the agent needs to access the entire trajectory to determine the optimal next action. 
Reinforcement learning (RL) has been established as an effective approach to train the agent's policy -- parameterized by autoregressive transformers -- to search via policy gradient methods, even in the absence of explicit supervision of when to branch or backtrack along the tree.

Despite this practical progress, the theory of how transformers~\citep{vaswani2023attentionneed} learn to search remains limited, especially from a training dynamic perspective.
A growing literature shows that transformers can implement or learn graph reasoning tasks, but mostly in supervised or RL settings with static instances, where the graph is specified in the prompt or encoded as a fixed abstraction. The learned reasoning skills therefore confine to the retrieval, reachability, or generation of paths, rather than the exploration of an unknown graph through interactions~\citep{saparov2024transformers,wang2024alpine,ye2026transformers,yang2025multi,ran2026outcome,wang2025benefits}. It remains unclear how a transformer embodies an algorithmic search policy discovered through policy-gradient training despite nonconvexity. We therefore ask:
\begin{center}
\emph{Can transformers learn to search over trees via policy-gradient training \\ when the tree is hidden and must be discovered through stochastic interaction?}
\end{center}

\begin{figure}[t]
  \centering
  \includegraphics[width=0.96\textwidth]{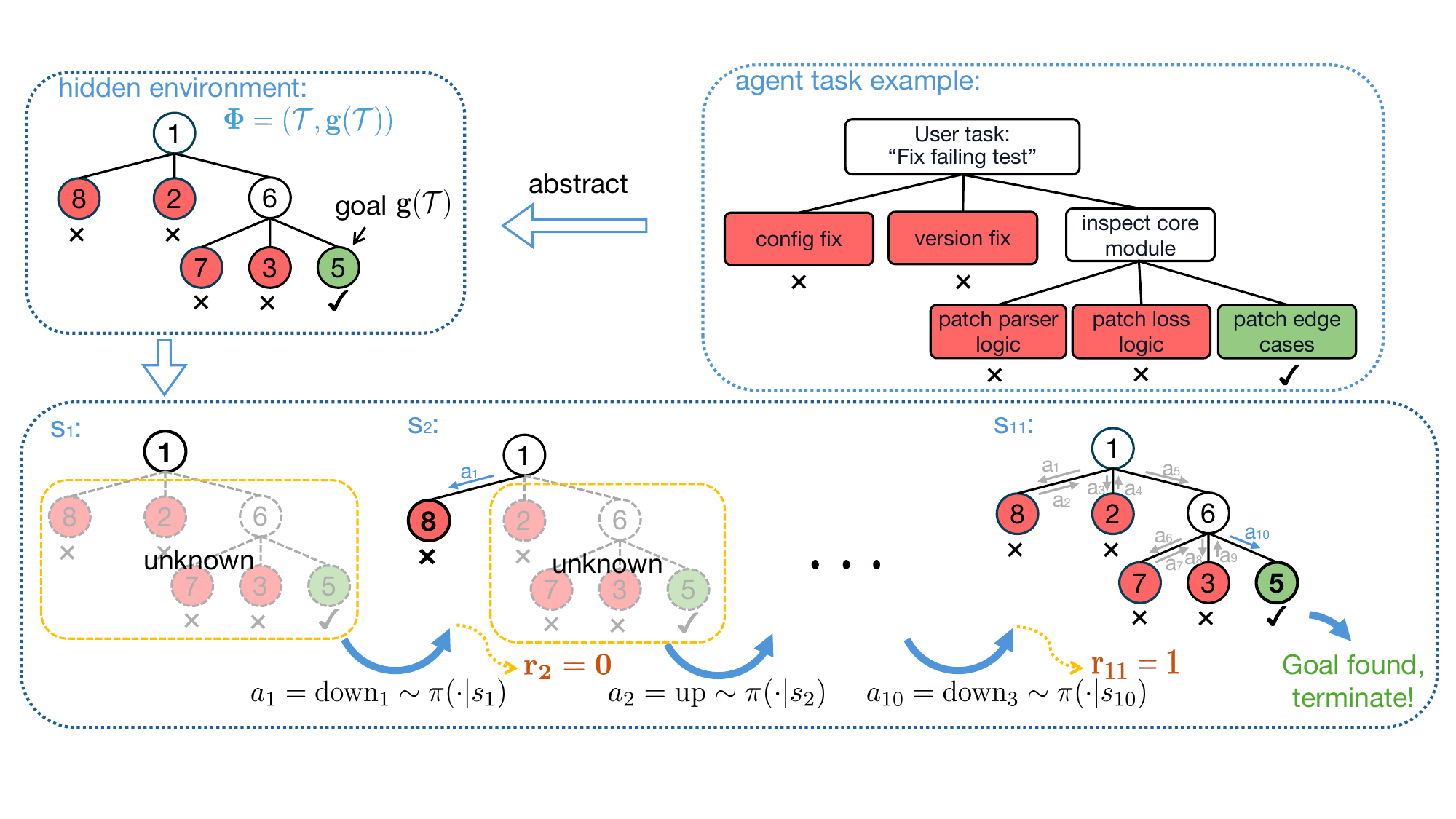}
  \vspace{-0.2in}
  \caption{An illustration of the interactive tree-search task studied in this paper, which can be viewed as an abstraction of agentic tasks. The environment is a hidden stochastic $k$-ary tree (with $k=3$ in the illustration), and at each step the agent chooses a search action based on its interaction history, without observing the full tree as input.}
  \label{fig:agent_task}
\end{figure}

\subsection{Our Contributions}

We study this question in a stylized environment featuring interaction with stochastic hidden $k$-ary trees, illustrated in Figure~\ref{fig:agent_task}. At the start of each episode, the environment samples a tree with distinct node identities and a hidden leaf set as the target goal. Beginning at the root, the agent at each step observes the current node identity together with a local status label indicating whether the node is internal, a non-goal leaf, or the goal; it then chooses a local action, either moving to a child or backtracking to the parent. The agent receives a positive reward only upon reaching the goal. This stylized setting preserves the core of interactive search while abstracting away the burden of opaque representation learning faced by practical LLM agents.

Solving this task requires the agent to search through the tree based on interactive feedbacks from the environment, most naturally, by implementing a depth-first search (DFS) algorithm. In this paper, we investigate how transformers, as the agent policy, acquire precisely the DFS algorithm through RL training. Our main contributions are as follows.
\begin{itemize}
    \item \textbf{A mechanistic transformer construction for interactive search.}
    We give an explicit two-head shallow transformer that implements randomized DFS with head specialization: one head records which child actions have already been tried at the current node, while the other detects search outcomes and triggers backtracking. These two heads cooperate to form the search algorithm.

    \item \textbf{An optimization analysis of policy-gradient learning.}
    We prove that policy-gradient training can induce this search circuit from sparse terminal reward alone, without expert demonstrations or action-level supervision. Under a depth-wise curriculum, the transformer first learns to backtrack from failed leafs and avoid repetitive actions by training on depth-1 trees, and then by further training on depth-2 trees, it learns to compose them and backtrack from internal nodes when their offsprings are all exhausted.

    \item \textbf{Provable depth generalization from shallow training.}
    We show that training on depth-1 and depth-2 trees suffices for the learned policy to succeed on deeper full trees. This demonstrates that the transformer learns compositional search skills rather than memorizing short trajectories.

    \item \textbf{Extension to imbalanced goal distributions.} We extend the analysis beyond the balanced case and show that, when goal probabilities differ across branches, discounting the returns selects a ranked DFS policy that searches more promising branches first and yields the minimum expected path length among all search policies.  
\end{itemize}

These results isolate a core mechanism behind agentic reasoning: using memory of past interaction to control future exploration under partial observability and sparse feedback. By making this mechanism explicit and provable, our work provides a tractable foundation for understanding memory, credit assignment, and algorithmic generalization in agentic ML systems.

\subsection{Related Work}\label{sec:related_work}

\paragraph{Test-time agentic search with language models.} Tree search is a central tool for sequential decision-making, from Monte Carlo tree search and AlphaGo-style systems to neural theorem proving \citep{browne2012survey,silver2016mastering,silver2018general,lample2022hypertree}. In language agents, search has appeared mainly as an inference-time mechanism: ReAct and Reflexion use interaction histories and feedback \citep{yao2022react,shinn2023reflexion}, while Tree-of-Thoughts, RAP, LATS, and web-agent tree search explicitly explore alternative reasoning or environment branches \citep{yao2023tree,hao2023reasoning,zhou2024language,koh2024tree}. Test-time compute scaling further shows the value of verifier- or reward-guided search \citep{snell2024scaling}. Together, these works establish exploration, lookahead, and backtracking as useful primitives for language agents.

\paragraph{Transformer mechanisms for structured search and reasoning.}
A growing literature of theory studies how transformers represent structured search and reasoning, including graph connectivity and path-finding \citep{saparov2024transformers,wang2024alpine}, chain-of-thought \citep{yang2025multi,huang2025transformers}, and learned graph heuristics \citep{ye2026transformers}. These works do model aspects of search, but mostly through static instances encoded in the input. In contrast, our setting is interactive: the tree is hidden, observations are local, and the transformer must infer the search state from history.

\paragraph{RL training dynamics for transformer reasoning.}
Recent theory studies how RL improves transformer reasoning from outcome-level feedback, including planning on graph abstractions \citep{wang2025benefits} and curriculum or reinforcement learning with verifiable rewards (RLVR) dynamics \citep{bu2025provable, ran2026outcome, huang2026learning}. \citet{wang2025benefits} are closest to ours: they analyze policy gradient and Q-learning for language-model planning, modeled as generating paths on a graph. This is a search/planning abstraction, but the environment is a fixed graph and the evolving state is the generated node prefix; process rewards can also check adjacency. We instead study a transformer policy interacting with a hidden stochastic tree, where successful search requires learning when to explore, avoid repeats, and backtrack from local observations.

\paragraph{Notation.} Let $[n]=\{1,\ldots, n\}$. Denote $A_{:,-1}$ or $A(:,-1)$ as the last column of matrix $A$. $I_n$ denotes the identity matrix of size $n\times n$, $1_{m\times n}$, $0_{m\times n}$ denote the all-ones matrix and all-zeros matrix of size $m\times n$, respectively. Let $\norm{\cdot}_p$ represent the $\ell_p$ norm of a vector. In addition, $g(x)\lesssim f(x)$ or $g(x)=O(f(x))$ (resp. $g(x)\gtrsim f(x)$ or $f(x)=O(g(x))$) means that $g(x)\leq c\cdot f(x)$ (resp. $g(x)\geq c\cdot f(x)$) for some constant $c>0$ and all $x$ in some range; $g(x)\asymp f(x)$ means that $g(x)\lesssim f(x)$ and $g(x)\gtrsim f(x)$. We write $g(x)=o(f(x))$ if $g(x)/f(x)\to 0$ in the relevant limit.

\section{Agentic Tree Search via Reinforcement Learning}\label{sec:formulation}

We define the tree search task and the corresponding RL formulation.

\subsection{Tree search via RL}

 Throughout this work, we consider full $k$-ary ($k\geq 2$) trees $\gT$ of depth $l(\gT)\in [L]$ ($L\geq 2$) with \textit{distinct} nodes, i.e., each node in the tree has either $k$ children or no children. The nodes of the trees are generated by sampling randomly from the set $[N]$ without replacement, where we assume $N\geq \frac{k^{L+1}-1}{k-1}$ to ensure any such $k$-ary tree has distinct nodes.
Given a tree $\gT$, we choose one of its leaf nodes as the goal node $g(\gT)$. The task of the agent is to find the goal node by searching over the tree without knowing its topology a priori.

\paragraph{Environment setup.} We first fix
an \emph{environment distribution} $\gP$ over  the environment parameters $\Phi=(\gT,g(\gT))$. At the beginning of each episode, we sample $\Phi\sim\gP$ independently, and the agent starts from the root node $n_1 = r (\gT)$. At the $h$-th timestep, the agent arrives at node $n_h\in[N]$ and takes one of the following actions from an action space $\gA$:
\begin{align}\label{eq:action_set}
  a_h \in \gA\coloneqq \big\{\up \big\}\cup \gA_{\texttt{down}} = \big\{\up \big\}\cup \big\{\down_i:i\in[k] \big\},
\end{align}
where $\up$ denotes moving to the parent node and $\down_i$ denotes moving to the $i$-th child node. Actions are \emph{illegal} if: (i) $\up$ at the root, (ii) $\down_i$ at a leaf, or (iii) the action that has already been queried at $n_h$. 
On the other hand, if $a_h$ is legal, the agent deterministically moves to the next node $n_{h+1}$ according to $a_h$. For $h\geq 1$, the agent also receives observation $o_{h}=(n_{h},c_{h})$, where the label $c_{h}$ of $n_{h}$ is defined as
\begin{align}\label{eq:label}
c_{h} \in \{0, \times, \checkmark\},\quad
c_{h}=
\begin{cases}
0, & \text{if }n_{h}\text{ is not a leaf,}\\
\times, & \text{if }n_{h}\text{ is a leaf and }n_{h}\neq g(\gT),\\
\checkmark, & \text{if }n_{h}=g(\gT).
\end{cases}\quad
\end{align}
Consequently, we note that $o_1=(n_1,c_1)=(r(\gT),0)$. The episode terminates when $c_h=\checkmark$ or an illegal action is taken.

\paragraph{Markov decision process (MDP).}
The episode evolves deterministically given $\Phi$ and the agent's actions, and the maximum length of the episode is set as $H=2N$.\footnote{Since in our setting, each edge can be visited at most twice (down/up).} We cast the interaction as a Markov decision process (MDP) across episodes by taking the \emph{history} as the state: we set the state at timestep $h$ as 
 \begin{align}
 \label{eq:state}
 s_h \coloneqq
 (o_1,a_1,\dots,o_{h-1},a_{h-1},o_h).
 \end{align}
Let $\gS$ be the set of all possible states,  $\gS_\checkmark\in\gS$ denote the set of all success states, i.e., if $c_h=\checkmark$, then $s_h\in\gS_\checkmark$. In addition, we introduce an absorbing terminal state $s_\perp$. The transition kernel is given by the posterior predictive distribution~\citep{duff2002optimal}:
\begin{align*}
  \PP(s_{h+1}\mid s_h,a_h)&= \begin{cases}\sum_{o\in\gO} \EE_{\Phi\sim\gP(\cdot|s_h)}[\mathbbm{1}\{o=O(s_h,a_h\mid\Phi)\}] \mathbbm{1}\{s_{h+1}=(s_h,a_h,o)\}, \\
  & \hspace{-1in} \text{if }s_h\notin \gS_\checkmark,\,  \,a_h\text{ legal at }s_h\\
    \mathbbm{1}\{s_{h+1}=s_\perp\}, & \hspace{-1in} \text{otherwise}
  \end{cases},
\end{align*}
where $\gO$ is the set of all possible observations, $O(s_h,a_h\mid\Phi)$ is the deterministic observation produced by taking action $a_h$ in state $s_h$ given a hidden parameter $\Phi$, and $\gP(\cdot|s_h)$ denotes the posterior distribution over $\Phi$ given $s_h$. 
The reward obtained after taking action $a_h$ and transitioning to $s_{h+1}$ is $r_{h+1}=\mathbbm{1}\{s_{h+1}\in \gS_\checkmark\}$. The objective of the agent is to learn a policy $\pi$ that maximizes the expected return, i.e.,
\begin{align}\label{eq:value_objective}
 J^{\pi}(\gP) &\coloneqq \EE_{s_1\sim\rho,  a_h\sim \pi(\cdot\mid s_h), \atop s_{h+1}\sim\PP(\cdot\mid s_h,a_h), h\in[H]}\left[\sum_{h=1}^H r_{h+1}\right]= \EE_{\Phi\sim\gP,\pi}\left[\sum_{h=1}^H r_{h+1}\right],
\end{align}
where $\rho(s) \coloneqq \PP_{\Phi\sim\gP}[s_1(\Phi)=s]$
is the initial state distribution. In addition, let $R(\pi;\Phi)$ denote the expected success rate of policy $\pi$ in environment $\Phi=(\gT,g(\gT))$, and  denote
\begin{align}\label{eq:success_rate_l}
    R(\pi;\gP)\coloneqq \E_{\Phi\sim \gP}\left[R(\pi;\Phi)\right]=J^\pi(\gP)
\end{align}
as the expected success rate of policy $\pi$ on trees sampled from the environment distribution $\gP$. Therefore, maximizing $J^\pi(\gP)$ is equivalent to maximizing the success rate $R(\pi;\gP)$.

\subsection{Transformer Architecture}\label{sec:construction}

We set our agent as a two-layer transformer-based model, which maps its trajectory observations to a probability distribution over the action space. To formally state our model, we first describe the state and action embeddings, followed by the attention model with simplified parameters for the tractability of analysis. 

\paragraph{State and action embeddings.} For each node $i\in[N]$, let $u_i\in\R^{d_1}$ denote its token embedding. For the action set,  let $v_i\in \R^{d_2}$ denote the embedding of action $\down_i$ for $i\in[k]$, and
 $v_{k+1}\in \R^{d_2}$ denote the embedding of action $\up$. 
For the label set, let $z_0,z_1,z_2\in\R^{d_3}$ denote the embeddings of labels $0, \times, \checkmark$ respectively. For notational convenience, let $\overline{n}_h\in\R^{d_1}$, $\overline{a}_h\in\R^{d_2}$ and $\overline{c}_h\in\R^{d_3}$ denote the embedding of node $n_h$, action $a_h$, and label $c_h$ at step $h$, respectively. Given any tree $\gT$, the state embedding $E^{(h)}=E^{(h)}(\gT)$ of state $s_h$ is given (auto-regressively) as
\begin{align}\label{eq:state_embedding}
  E^{(h)} \coloneqq \begin{pmatrix}
    E_n^{(h)} \\
    E_a^{(h)} \\
    E_{n'}^{(h)} \\
    E_c^{(h)} \\
  \end{pmatrix}\coloneqq 
  \begin{pmatrix}
    u_0 & \overline{n}_{1} & \cdots & \overline{n}_{h-1} \\
    v_0 & \overline{a}_{1} & \cdots & \overline{a}_{h-1} \\
    \overline{n}_{1} & \overline{n}_{2} & \cdots & \overline{n}_{h} \\
    \overline{c}_1 & \overline{c}_2 & \cdots & \overline{c}_{h}
  \end{pmatrix} =  \begin{pmatrix}
    E^{(h-1)}, & \begin{pmatrix}
      \overline{n}_{h-1} \\
      \overline{a}_{h-1} \\
      \overline{n}_{h} \\
      \overline{c}_{h}
    \end{pmatrix}
  \end{pmatrix} \in \R^{(2d_1+d_2+d_3)\times h}
\end{align}
with filler embeddings $u_0\in\R^{d_1}$ and $v_0\in\R^{d_2}$ for the first step. 
Note that $E^{(h)}$ contains all the information in $s_h$ in a matrix form, and each column $i$ of $E^{(h)}$ represents the embedding of the transition tuple $(n_{i-1},a_{i-1},n_i,c_i)$ at step $i$. Let $\overline{\gS}$ denote the set of all possible state embeddings.

\paragraph{Transformer architecture and multi-turn reasoning.} Our transformer-based agent maps the state to a policy over the action space, i.e. $\pi_\theta:\overline{\gS}\mapsto\Delta(\gA)$. Following standard simplification in theoretical analyses \citep{huang2023context,yang2024context,nichani2024transformers}, for an input state embedding $E = [ E_n^{\top}, E_a^{\top}, E_{n'}^{\top}, E_c^{\top}]^{\top}$, we define a two-head transformer model $f:\overline{\gS}\to\R^{k+1}$ as
\begin{align}\label{eq:transformer_architecture}
  f(E) =  P \cdot    \texttt{head}_1(E) + Q \cdot   \texttt{head}_2(E) 
\end{align}
where $P\in\R^{(k+1)\times d_2}$, $Q\in\R^{(k+1)\times d_3}$, and
\begin{align}
\texttt{head}_1(E) = E_a \cdot\sm\big(E^\top {W_1^{KQ}}  E\big), \quad \texttt{head}_2(E) = E_c \cdot\sm\big(E^\top {W_2^{KQ}}  E\big), 
\end{align}
with the key and query matrices merged into a single matrix $W_i^{KQ}$. We further simplify the parameterization by setting
\begin{align*}
  W_1^{KQ} = \begin{pmatrix}
    0_{d_1\times d_1} & 0_{d_1\times d_2} &  B & 0_{d_1\times d_3} \\
    0_{d_2\times d_1} & 0_{d_2\times d_2} & 0_{d_2\times d_1} & 0_{d_2\times d_3} \\
    0_{d_1\times d_1} & 0_{d_1\times d_2} & 0_{d_1\times d_1} & 0_{d_1\times d_3} \\
    0_{d_3\times d_1} & 0_{d_3\times d_2} & 0_{d_3\times d_1} & 0_{d_3\times d_3}
  \end{pmatrix},\,
  W_2^{KQ} = \begin{pmatrix}
    0_{d_1\times d_1} & 0_{d_1\times d_2} &  0_{d_1\times d_1} & 0_{d_1\times d_3} \\
    0_{d_2\times d_1} & 0_{d_2\times d_2} & 0_{d_2\times d_1} & 0_{d_2\times d_3} \\
    0_{d_1\times d_1} & 0_{d_1\times d_2} & 0_{d_1\times d_1} & 0_{d_1\times d_3} \\
    0_{d_3\times d_1} & 0_{d_3\times d_2} & 0_{d_3\times d_1} & C
 \end{pmatrix},
\end{align*}
where $B\in\R^{d_1\times d_1}$ and $C\in\R^{d_3\times d_3}$, leading to 
\begin{equation}
f_{\theta}(E) : = f(E) = P E_a \cdot\sm\big(E_n^{\top} B  E_{n'}\big) + Q E_c \cdot\sm\big(E_c^\top C E_c \big),
\end{equation}
where $\theta \coloneqq \{ P, Q, B, C \}$ collects the parameters of the transformer. The policy $\pi_\theta$ is then given by applying  softmax to the last column of $f_{\theta}(E)$:
\begin{align}\label{eq:pi_theta}
  \pi_\theta(E)\coloneqq \pi_\theta(\cdot\mid E)\coloneqq  \begin{pmatrix}
    \pi_\theta(\down_1\mid E) \\
    \vdots \\
    \pi_\theta(\down_k\mid E) \\
    \pi_\theta(\up\mid E)
  \end{pmatrix}
  \coloneqq \sm\left( (f_\theta(E))_{:,-1}\right)\in\Delta(\gA).
\end{align}
During inference, at each step $h$, given the state embedding $E^{(h)}$, the agent samples an action $a_h$ according to 
\begin{align}\label{eq:pi_theta_simplified}
  \pi_\theta(E^{(h)}) = \sm\left(P E_a^{(h)}\sm\big( E_n^{(h)\top} B \overline{n}_h\big) + Q E_c^{(h)}\sm\big( E_c^{(h)\top} C \overline{c}_h \big)\right),
\end{align}
and observes $o_{h+1}=(n_{h+1},c_{h+1})$. The next state embedding $E^{(h+1)}$ is then updated auto-regressively following \eqref{eq:state_embedding}
as long as the episode is not terminated, and the agent moves to the next timestep $h+1$. 



 \subsection{Transformer Construction for Random DFS} 
We collect the embeddings in matrix forms as
\begin{align}\label{eq:U_V_Z}
U& \coloneqq (u_0,\cdots,u_N),
\,\quad V \coloneqq (v_0,\cdots,v_{k+1}),
    \,\quad Z \coloneqq (z_0,z_1,z_2),
\end{align}
and define induced matrices $A_M$ $(M\in\{B,C,P,Q\})$ as
\begin{align}\label{eq:induced_matrices_main}
  A_B\coloneqq U^\top B U\in\R^{(N+1)\times (N+1)}, \quad
  A_C\coloneqq Z^\top CZ\in\R^{3\times 3}, \notag\\
  A_P\coloneqq P V\in\R^{(k+1)\times (k+2)}, \quad
  A_Q\coloneqq Q Z\in\R^{(k+1)\times 3}.
\end{align}

We first present a set of matrices $\{B, C, P, Q\}$ in Theorem~\ref{thm:transformer_construction} that enable the transformer to perform the search task, thereby clarifying its underlying mechanism. This construction is not unique; our particular choice is intended to align more closely with the training dynamics while having a simple form; see Appendix~\ref{sec:proof_construction_balanced} for the proof.
\begin{thm}[Transformer construction for random DFS]\label{thm:transformer_construction}
  Assume $U,V,Z$ all have full column rank. Then there exists $\theta=\{B, C, P, Q\}$ such that
  \begin{align}\label{eq:A_B_C_P_Q}
    A_B & = 
  \begin{pmatrix}
    0 & a_{b,0}1_{1\times N} \\
    0_{N\times 1} & a_{b,1}I_{N}
    \end{pmatrix},
  \quad \quad A_C = \begin{pmatrix}
    a_{c,0} & -a_{c,1} & 0 \\
    -a_{c,0} & a_{c,1} & 0 \\
    0 & 0 & 0
\end{pmatrix},\notag\\
 A_{P} & = 
\begin{pmatrix}
    0_{k\times 1} & -a_{p,1}I_{k} & 0_{k\times 1} \\
    0 & 0_{1\times k} & 0
    \end{pmatrix},
\quad A_Q = 
\begin{pmatrix}
  \frac{a_{q,0}}{k}1_{k\times 1} & -\frac{a_{q,\times}}{k}1_{k\times 1} &  0_{k\times 1} \\
  -a_{q,0} & a_{q,\times} & 0
  \end{pmatrix}
  \end{align}
  for any $a_{b,0}, a_{b,1}, a_{c,0}, a_{c,1}, a_{p,1}, a_{q,0}, a_{q,\times}\in\R$. For any $k\geq 2$, $L\in\NN_+$, $N\geq \frac{k^{L+1}-1}{k-1}$ and precision $\epsilon>0$, there exists an absolute constant $C>0$ such that when 
  \begin{align}\label{eq:construction_balanced}
    e^{a_{c,0}}\geq \ln k, \quad e^{a_{c,1}}\geq \sqrt{N}\ln k,\quad e^{a_{b,0}}\geq N\ln k,\quad e^{a_{b,1}}\geq \ln k e^{a_{b,0}},\notag\\
    \frac{a_{p,1}}{k}\asymp a_{q,0}\asymp a_{q,\times},\quad a_{q,0},\, a_{q,\times}\geq C\ln\left(N/\epsilon\right),\quad
    \frac{a_{p,1}}{k}\geq C\left(\ln\left(N/\epsilon\right)+a_{q,0}\right),
  \end{align}
  we have 
  \begin{align}
  R(\pi_\theta;\Phi)\geq 1-\epsilon
  \end{align}
  for any $\Phi=(\gT,g(\gT))$, where $\gT$ is a full $k$-ary tree with depth $l\in[L]$, and $g(\gT)$ is an arbitrary leaf of $\gT$.
\end{thm}

\begin{figure}[ht]
  \centering
  \includegraphics[width=0.85\textwidth]{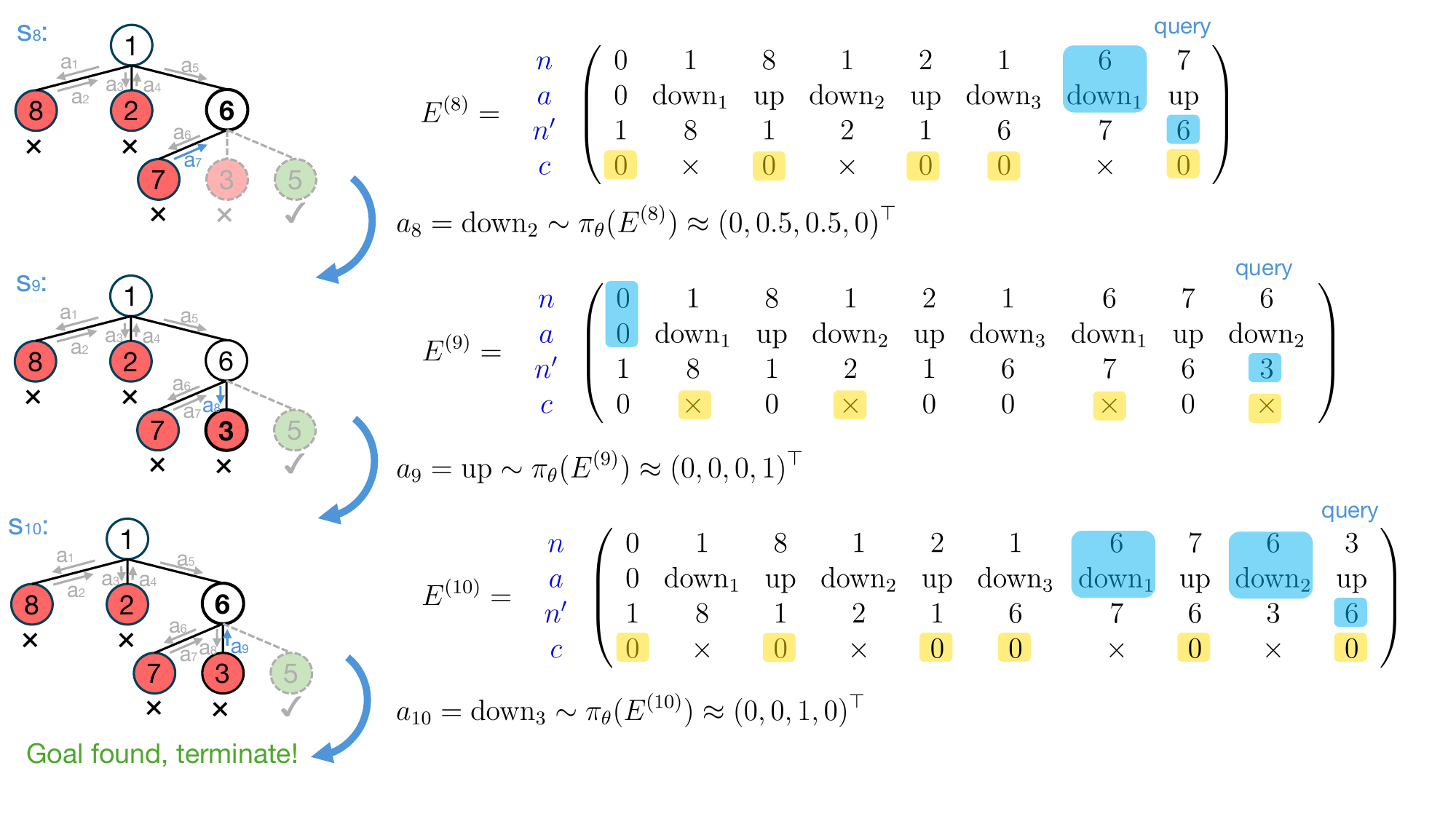}
  \caption{Illustration of the transformer mechanism implementing random DFS on a full 3-ary tree. Blue and yellow shading indicate the attention associations of heads~1 and~2, respectively. 
  }
  \label{fig:construction_balanced}
\end{figure}

Under our construction, the agent performs \emph{random DFS}: it selects randomly among unvisited children, and goes up only at a wrong leaf or after all children of the current node have been explored.
Figure~\ref{fig:construction_balanced} illustrates how the two heads specialize and cooperate to perform random DFS. 
\begin{itemize}
\item Head~2, associated with matrices $C$ and $Q$, uses label history to promote action $\up$ at non-goal leaves and suppress $\up$ at internal nodes. 
\item Head~1, associated with matrices $B$ and $P$, retrieves action history and suppresses tried children from being selected again. 
\end{itemize}
When all children of an internal node have been tried, every logit associated with $\down_i$ is suppressed by Head 1, then the logic associated with $\up$ becomes largest despite Head~2's internal-node bias, pushing agent to backtrack from an unsuccessful branch.  

\section{Training Dynamics and Generalization of Transformers}\label{sec:training}
 
 We begin with specifying the environment distribution that is used for policy gradient training, followed by the training dynamic and generalization performance of our transformer policy.

 \subsection{Training Environment and Policy Gradient}

\paragraph{Training distribution.} 
To describe the training distribution $\gP_l$ for any fixed depth $l\in[L]$, we define the training tree set $\gD_{\text{train},\text{tree}}^l$ as
\begin{align}\label{eq:training_tree_set}
  \gD_{\text{train},\text{tree}}^l \coloneqq \left\{\gT \mid \gT \text{ is a perfect $k$-ary tree with distinct nodes in $[N]$ and depth $l$} \right\},
\end{align}
and define $p_{\text{train,tree}}^l=\gU(\gD_{\text{train},\text{tree}}^l)$ as the uniform distribution on $\gD_{\text{train},\text{tree}}^l$. 
We assume, for any given tree $\gT$ with any depth $l\in[L]$, the environment generates the goal node $g(\gT)$ according to a latent \emph{goal distribution} 
$$p^a=(p^a_1, \cdots, p^a_k)\in\Delta(\gA_{\texttt{down}})$$ 
that is unknown to the agent. Specifically, after a tree $\gT$ is generated from $p_{\text{train,tree}}^l$, the environment takes action  $a^{\text{env}}\overset{i.i.d}\sim p^a$ at each node -- starting from the root -- until a leaf is reached and assigned as a goal. Let $p^a(\cdot\mid\gT)$ denote the goal distribution induced by $p^a$ given tree $\gT$. We define the training distribution $\gP_l$ of trees with depth $l\in[L]$ as
\begin{align}\label{eq:rho_l}
    \gP_l(\gT,g) = p_{\text{train,tree}}^l(\gT) p^a(g\mid\gT).
\end{align}
Here, we focus on the case where the goal distribution $p^a$ is balanced, i.e., we assume
\begin{align}\label{eq:balanced_goal_distribution}
 \text{(balanced goal:)}   \quad\quad  p^a_1 = p^a_2 = \cdots = p^a_k = \frac{1}{k}.
\end{align}
In Section~\ref{sec:extension_imbalanced}, we also consider the case where the goal distribution $p^a$ is imbalanced.

\paragraph{Training with policy gradient.} The construction in Theorem~\ref{thm:transformer_construction} suggests that the last row of $P$ can be fixed to 0, so for simplicity and consistency with our construction, we let
\begin{align}\label{eq:W_P}
  P=\begin{pmatrix}
    \overline P\\
    0_{1\times d_2}
  \end{pmatrix}
\end{align}
where $\overline P\in\R^{k\times d_2}$, and set trainable parameters to be $\theta = \{B, C, \overline{P}, Q\}$. We train the policy of the form \eqref{eq:pi_theta_simplified} with policy gradient. Specifically, we let $\theta^{(t)} = \{B^{(t)}, C^{(t)}, \overline{P}^{(t)}, Q^{(t)}\}$ denote the parameters at iteration $t\in\NN$, and initialize all matrices to be zero matrices, i.e.,  
\begin{align}\label{eq:initialization}
  B^{(0)} = 0_{d_1\times d_1}, \quad C^{(0)} = 0_{d_3\times d_3}, \quad \overline{P}^{(0)} = 0_{k\times d_2}, \quad Q^{(0)} = 0_{(k+1)\times d_3}.
\end{align} 
We write $J(\theta)$ as the shorthand of $J^{\pi_\theta}(\gP)$ when the training distribution $\gP$ is clear from the context.
Then under our setting, the policy gradient update at iteration $t$ is given by 
\begin{align}\label{eq:policy_gradient_update}
  \theta^{(t+1)} &= \theta^{(t)} + \eta \nabla_{\theta} J(\theta^{(t)})\notag\\
  &= \theta^{(t)} + \eta \EE_{\Phi\sim\gP,\pi_\theta}\left[\sum_{h=1}^H \nabla_{\theta} \log \pi_\theta(a_h\mid E^{(h)}) \sum_{i=h}^H r_{i+1}\right],
\end{align}
where the second equality follows from the policy gradient theorem~\citep{williams1992simple,sutton1999policy}, and $\eta>0$ is the learning rate.

\subsection{Depth-wise Curriculum Training }

  We aim to demonstrate that after training on trees of depth 1 and 2 well enough, the agent learns the DFS algorithm and can perform well on deeper trees. Specifically, we use a two-stage training with curriculum schedulers $\delta_1,\delta_2\in(0,1)$: 
\begin{enumerate}
\item[(i)] depth-1 curriculum training (training step $t\in[T_1]$): we first train on depth-1 trees $\gT\sim \gP_1$; 
\item[(ii)] depth-2 curriculum training (training step $T_1+1\leq t\leq T_2$): as soon as the success rate $R(\pi;\gP_1)$ reaches $1-\delta_1$, we 
start training on depth-2 trees $\gT\sim \gP_2$ until $R(\pi;\gP_2)$ reaches $1-\delta_2$. 
\end{enumerate} 
 
 To simplify the analysis, we make the following orthonormality assumption on the embedding matrices defined in \eqref{eq:U_V_Z}.
\begin{asmp}[orthonormal embeddings]\label{asmp:token_embedding}
The matrices $U,V,Z$ each have orthonormal columns.
\end{asmp}

\paragraph{Theoretical guarantees.} We now show that after training with proper choices of $\delta_1,\delta_2$ for sufficiently long, the trained policy can succeed 
on an arbitrary full $k$-ary tree with depth $l\in[L]$ that goes significantly beyond the training depth.  The formal versions (for both Theorem~\ref{thm:convergence_phase_1_balanced} and Theorem~\ref{thm:convergence_phase_2_balanced}) and their proofs are in Appendix~\ref{sec:proof_convergence}.

\begin{thm}[Depth-1 curriculum training, informal]\label{thm:convergence_phase_1_balanced}
  Suppose the goal node is balanced (cf.~\eqref{eq:balanced_goal_distribution}), Assumption~\ref{asmp:token_embedding} holds,  
  $k,N\in\NN_+$ are sufficiently large, and $\epsilon,\eta>0$ are sufficiently small, $\delta_1=\epsilon^{1.05}$. For depth-1 curriculum training, the success rate $R(\pi;\gP_1)$ reaches $1-\delta_1$ in $T_1=O(1/(\eta\epsilon^{1.1}))$ iterations. After the training, we have 
\begin{align}\label{eq:success_rate_generalization_balanced_phase_1}
  R(\pi^{(T_1)};\gP_l)= O\left(1/k^{l-1}\right), \quad \forall 2\leq l\leq L.
\end{align}
\end{thm}

Theorem~\ref{thm:convergence_phase_1_balanced} shows that after training on depth-1 trees, the transformer can achieve arbitrary accuracy on depth-1 trees, yet its generalization performance on deeper trees still degenerates exponentially. Examining the training dynamics, by time \(T_1\), the matrices $A_B$, $A_C$, and $A_Q$ are close to their respective constructions in Theorem~\ref{thm:transformer_construction}, but generalization still fails because $A_Q$ remains inadequately trained: after exhausting the children of a non-root internal node, the agent does not move up with high probability. 
We therefore freeze the weight matrices $B$, $C$, $Q$ and train only $\overline P$ on depth-2 trees to learn the remaining DFS behavior. The outcome of this stage is summarized in the following theorem.
\begin{thm}[Depth-2 curriculum training and depth generalization, informal]\label{thm:convergence_phase_2_balanced}
  Suppose the same assumptions in Theorem~\ref{thm:convergence_phase_1_balanced} hold, and fix $\delta_2=\epsilon^{1.02}/k^L$. For depth-2 curriculum training, the success rate $R(\pi;\gP_2)$ reaches $1-\delta_2$ in $T_2-T_1=O\left(k^L/(\eta\epsilon^{1.1})\right)$ iterations.
After the training, given any full $k$-ary tree $\gT$ with depth $l\in[L]$ and any goal node $g(\gT)$, we have
\begin{align}\label{eq:success_rate_generalization_balanced_phase_2}
  R(\pi^{(T_2)};(\gT,g(\gT)))\geq 1-\epsilon.
\end{align}
\end{thm}
Theorem~\ref{thm:convergence_phase_2_balanced} establishes that after training on depth-2 trees, the learned policy generalize to trees with arbitrary depth and goal node by  implementing DFS, where all weight matrices converge to their construction in Theorem~\ref{thm:transformer_construction}. While the training distribution is confined to perfect $k$-ary trees to exploit symmetry, the learned policy is applicable to any full $k$-ary trees.   Compared with existing analyses on prompt-revealed or fixed graphs~\citep{saparov2024transformers,wang2024alpine,wang2025benefits}, or RL/RLVR over generated traces~\citep{bu2025provable,huang2026learning}, our hidden-environment agentic setting requires tracking a multi-phase interaction among the induced matrices and return-to-go estimates; see Appendix~\ref{sec_app:main_statements_proof_convergence} for discussions.

\section{Extension to Imbalanced Goal Distribution}\label{sec:extension_imbalanced}

So far we have focused on balanced goal distributions (c.f.~\eqref{eq:balanced_goal_distribution}). In practice, goals can be imbalanced: for instance, in web search, desired information is usually more likely to appear among top-ranked results. This suggests that it is beneficial to prioritize the search over more promising branches according to some unknown non-uniform prior~\citep{koh2024tree}. We formalize the additional prior information via an averaged full-tree environment distribution $\gP$. For any $l\in[L]$, let
\begin{align}
  \gD_{\text{env},\text{tree}}^l \coloneqq \left\{\gT \mid \gT \text{ is a full $k$-ary tree with distinct nodes in $[N]$ and depth } l \right\},
\end{align}
and let $p_{\text{env},\text{tree}}^l=\gU(\gD_{\text{env},\text{tree}}^l)$ be the uniform distribution on $\gD_{\text{env},\text{tree}}^l$. Similar to the balanced setting, given a sampled tree $\gT$, the environment generates $g(\gT)$ by sampling latent actions independently from 
$$p^a=(p^a_1,\ldots,p^a_k)$$ 
from the root until a leaf is reached, where we index the actions by decreasing likelihood: 
\begin{align}\label{eq:imbalanced_goal_distribution}
\text{(imbalanced goal:)} \quad\quad  p^a_1 > p^a_2 > \cdots > p^a_k > 0.
\end{align}
Denoting the induced distribution by $p^a(\cdot\mid\gT)$, define
  \begin{align}\label{eq:env_distribution}
    \gP := \gP(\gT,g) = p_{\text{env},\text{tree}}(\gT)\, p^a(g\mid\gT),\,\,\text{ where }\,\, p_{\text{env},\text{tree}}(\gT)\coloneqq\sum_{l=1}^L \omega_l \,p_{\text{env},\text{tree}}^l(\gT)
\end{align}
as the environment distribution, where the weights $\omega_l>0$ and $\sum_{l=1}^L \omega_l=1$.

\paragraph{Benefit of discounting.} Define
\begin{align}
  \Pi_\text{success}\coloneqq\Pi_\text{success}(\gP)
  \coloneqq
  \left\{\pi:R(\pi;\gP)=1\right\},
\end{align}
which is the set of policies that find the goal almost surely under $\gP$. 
Because $\Pi_\text{success}$ contains infinitely many policies, which cannot be distinguished by the undiscounted objective, we consider maximizing the \emph{discounted} expected return, i.e.,
\begin{align}\label{eq:objective_discount}
  J_\gamma^{\pi}(\gP) &\coloneqq \EE_{s_1\sim\rho,  a_h\sim \pi(\cdot\mid s_h), \atop s_{h+1}\sim\PP(\cdot\mid s_h,a_h), h\in[H]}\left[\sum_{h=1}^H \gamma^{h-1} r_{h+1}\right]= \EE_{\Phi\sim\gP,\pi}\left[\sum_{h=1}^H \gamma^{h-1} r_{h+1}\right],
\end{align}
where $\gamma\in(0,1]$ is the discount factor. We further define the expected path length of policy $\pi$ as $H^\pi(\gP)$, which provides a quantitative means to distinguish the policies in $\Pi_\text{success}$. Since we only provide a terminal reward, shorter path lengths result in higher discounted expected returns. Proposition~\ref{prop:optimal_policy_balanced} shows that in the balanced case, every policy in $\Pi_\text{success}$ has the same expected path length, so any successful DFS policy, including random DFS, is optimal.

\begin{prop}[optimal policy under balanced goal distribution]\label{prop:optimal_policy_balanced}
  Under balanced goal distribution~\eqref{eq:balanced_goal_distribution}, any policy $\pi\in\Pi_\text{success}$ has the same expected path length $H^\pi(\gP)$.
\end{prop}

In contrast, the imbalanced case has a unique \emph{ranked DFS} policy that minimizes the expected path length among successful policies (up to behavior on unreachable histories): 
\begin{definition}[Ranked DFS]
 Under the imbalanced goal distribution~\eqref{eq:imbalanced_goal_distribution},  at any reachable, non-terminal state, if the current node has unvisited children, the \emph{ranked DFS} policy $\pi_{\mathsf{ranked}}^\star$ explores them in decreasing order of $p^a_j$; otherwise, it takes action $\up$.
\end{definition}
Moreover, for a suitable $\gamma<1$, maximizing $J_\gamma^{\pi}(\gP)$ exactly recovers this policy, formalized in Proposition~\ref{prop:optimal_policy_imbalanced}.
\begin{prop}[optimal policy under imbalanced goal distribution]\label{prop:optimal_policy_imbalanced}
Under the imbalanced goal distribution~\eqref{eq:imbalanced_goal_distribution}, we have $\pi^\star \in\Pi_\textnormal{success}$, and
  \begin{enumerate}[leftmargin=1.25em,labelsep=0.45em]
  \item  $H^{\pi^\star}(\gP)<H^\pi(\gP)$ for any $\pi\in\Pi_\text{success}$ whose reachable behavior differs from $\pi_{\mathsf{ranked}}^\star$.
  \item There exists an absolute constant $C>0$ such that for any $\gamma\in\Big(1-C\left(p^a_k\right)^{L},1\Big)$, $\pi^\star=\argmax_{\pi}J_\gamma^{\pi}(\gP)$ up to reachable behavior.
  \end{enumerate}
\end{prop}
The proof of both propositions is postponed to Appendix~\ref{sec:proof_optimal_policy}. We also call $\pi^\star$ the \textit{optimal policy} for the imbalanced case.

\paragraph{Transformer construction.} The ranked DFS policy can also be implemented by our two-head shallow transformer.
Theorem~\ref{thm:transformer_construction_imbalanced} modifies the induced matrices in~\eqref{eq:A_B_C_P_Q} by replacing the uniform row weights in $A_Q$ with a priority vector $\lambda\in\R^k$, yielding ranked DFS $\pi^\star$. The proof is provided in Appendix~\ref{sec:proof_construction_imbalanced}.

\begin{thm}[Transformer construction for ranked DFS]\label{thm:transformer_construction_imbalanced}
  Assume $U,V,Z$ all have full column rank. Then there exists $\theta=\{B, C, P, Q\}$ such that 
  \begin{align}\label{eq:A_B_C_P_Q_imbalanced}
    A_B &= \begin{pmatrix}
      0 & a_{b,0}1_{1\times N} \\
      0_{N\times 1} & a_{b,1}I_{N}
    \end{pmatrix},\quad\quad 
    A_C = \begin{pmatrix}
      a_{c,0} & -a_{c,1} & 0 \\
      -a_{c,0} & a_{c,1} & 0 \\
      0 & 0 & 0
    \end{pmatrix},\notag\\
    A_{P} &= \begin{pmatrix}
      0_{k\times 1} & -a_{p,1}I_{k} & 0_{k\times 1} \\
      0 & 0_{1\times k} & 0
    \end{pmatrix},\quad
    A_Q = \begin{pmatrix}
      \lambda & 0_{k\times 1} & 0_{k\times 1} \\
      -a_{q,0} & a_{q,\times} & 0
    \end{pmatrix}
  \end{align}
  for any $a_{b,0}, a_{b,1}, a_{c,0}, a_{c,1}, a_{q,0}, a_{q,\times},a_{p,1}\in\R$, any $\lambda=(\lambda_1,\ldots,\lambda_k)^\top\in\R^k$. For any $k\geq 2$, $L\in\NN_+$, $N\geq \frac{k^{L+1}-1}{k-1}$ and precision $\epsilon>0$, there exists an absolute constant $C>0$ such that when 
  \begin{align}\label{eq:constraints_imbalanced}
      e^{a_{c,0}}&\geq \ln k, \quad e^{a_{c,1}}\geq \sqrt{Nk}\ln k, \quad e^{a_{b,0}}\geq Nk\ln k, \quad e^{a_{b,1}}\geq \ln k e^{a_{b,0}}, \notag\\
      \lambda_{i}-\lambda_{i+1}&\asymp a_{q,0}\asymp a_{q,\times}\gtrsim \lambda_k\geq 0,\quad \lambda_{i}-\lambda_{i+1},\, a_{q,0},\, a_{q,\times}\geq C\ln\left(N/\epsilon\right),\,\,\forall i\in[k-1],\notag\\
      \frac{a_{p,1}}{k}&\asymp \lambda_1,\quad \frac{a_{p,1}}{k}\geq C\left(\ln\left(N/\epsilon\right)+a_{q,0}+\lambda_1\right),
  \end{align}
for any $\Phi=(\gT,g(\gT))$ where $\gT$ is a full $k$-ary tree with depth $l\in[L]$, and $g(\gT)$ is an arbitrary leaf, and any reachable non-terminal state $s_h$, we have 
  \begin{align}\label{eq:R_pi_Phi_imbalanced}
    R(\pi_\theta;\Phi)\geq 1-\epsilon,\quad \norm{\pi_\theta(\cdot\mid s_h)-\pi^\star(\cdot\mid s_h)}_\infty\leq \frac{\epsilon}{2N}.
  \end{align}
\end{thm} 

Our experiments in Section~\ref{sec:experiments} show that discounted-return training recovers $\pi^\star$ in the imbalanced setting and learns induced matrices similar to the construction above.

\section{Numerical Experiments}
\label{sec:experiments}

\paragraph{Setup.} To validate our training analysis, we conduct simulations using the transformer defined in \eqref{eq:pi_theta_simplified} with zero initialization of the parameters. We set $k=3$, $N=\frac{k^{L+1}-1}{k-1}+10$, $d_1=N+1$, $d_2=k+2$, $d_3=3$, and choose $u_i\in\R^{d_1}$, $v_j\in\R^{d_2}$, $z_p\in\R^{d_3}$ to be the one-hot vectors with respect to the index $i$, $j$, and $p$ respectively (assuming the index starts from 0 here). For both balanced and imbalanced cases, we follow the two-depth curriculum training procedure, and use stochastic policy gradient with batch size $b=256$ to train the model, i.e., we have
\begin{align}\label{eq:policy_gradient_sample}
  \theta^{(t+1)} &= \theta^{(t)} + \frac{\eta}{b} \sum_{j=1}^b \sum_{h=1}^H \nabla_{\theta} \log \pi_\theta\left(a_h^{(j)}\mid E^{(h,j)}\right) \sum_{i=h}^H \gamma^{i-1} r_{i+1}^{(j)},
\end{align}
where $E^{(h,j)}$ ($j\in[b]$) is the state embedding of the state $s_h^{(j)}$ at the $h$-th step in the $j$-th trajectory in the batch. 
Each trajectory $(s_1^{(j)}, a_1^{(j)}, \dots, s_H^{(j)}, a_H^{(j)})$ is obtained by first sampling a tree $\gT\overset{\text{iid}}{\sim} p_{\text{train,tree}}^l$ ($l=1,2$) from the training distribution and then executing $\pi_\theta$ starting from the root node, until the episode ends. 
We set the learning rate $\eta=10$. 
\begin{itemize}
\item For the balanced case, we set $L=4$, $\gamma=1$, and train $\theta=\{B,C,\overline P, Q\}$ at the depth-1 stage for 1000 iterations, and train $\theta=\{B, \overline P\}$ at the depth-2 stage for 49600 iterations.\footnote{For faster convergence, we also train $B$ at depth-2 stage in the experiments.} 
\item For the imbalanced case, we set $L=3$, $\gamma=0.9$, $p^a_{i+1}=0.9p^a_i, i\in[k-1]$, and train $\theta=\{B,C,P,Q\}$ at both depths for 1100 and 50000 iterations, respectively.
\end{itemize}
For every 50 iterations, we also evaluate on $L$ test sets, each of size $128$, consisting of i.i.d.\ random full $k$-ary trees with depth $l=1,\cdots,L$ respectively. 
 
\begin{figure}[ht]
  \centering
  \subfigure[Depth-1 curriculum training]{\includegraphics[width=0.45\textwidth]{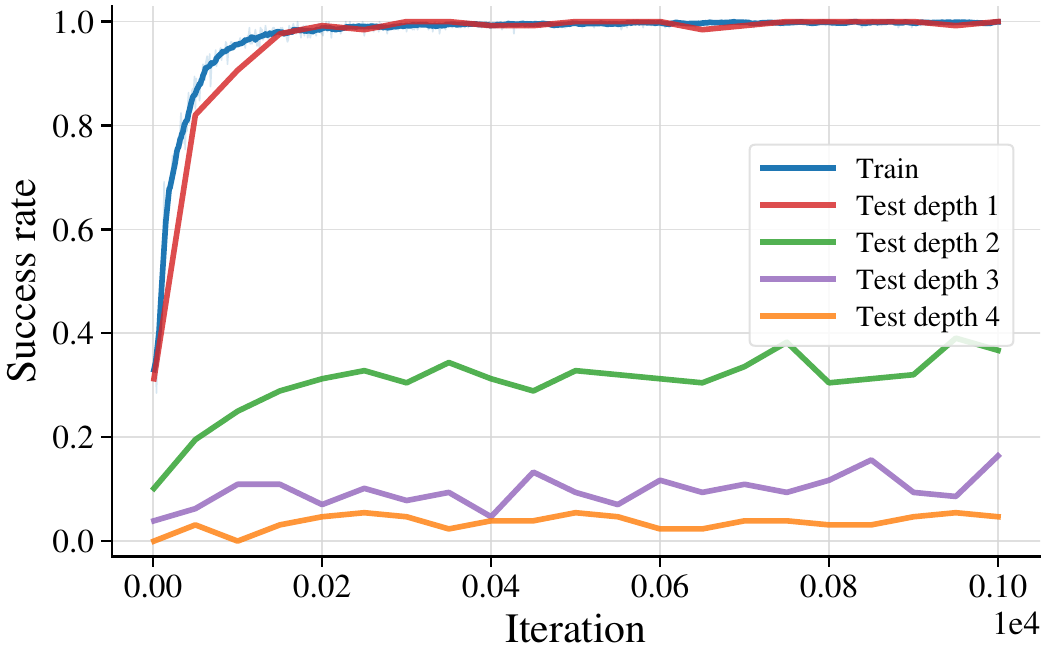}}
  \hspace{0.02\textwidth}
  \subfigure[Depth-2 curriculum training]{\includegraphics[width=0.45\textwidth]{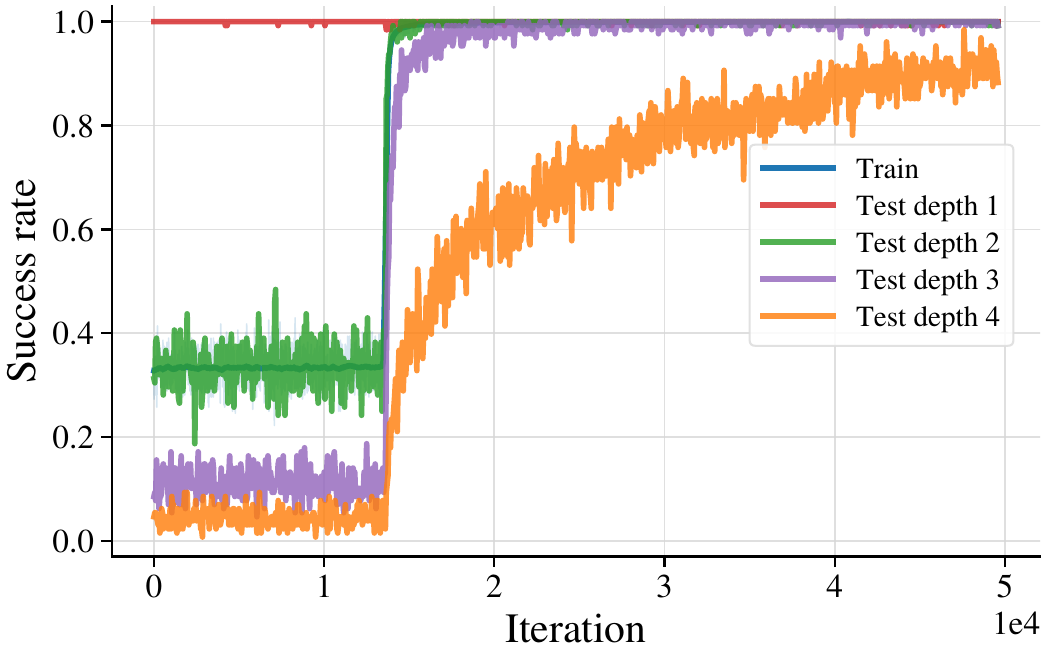}}
  \vspace{-0.05in}
  \caption{Train and test 
  rates under balanced goal distribution for depth-1 (left) and depth-2 curriculum training (right), trained on perfect 3-ary trees with stochastic policy gradient, and tested on randomly generated full 3-ary trees with depth 1 to 4.}
  \label{fig:reward_balanced}
\end{figure} 

\begin{figure}[ht]
  \centering
  \subfigure[Depth-1 curriculum training]{\includegraphics[width=0.45\textwidth]{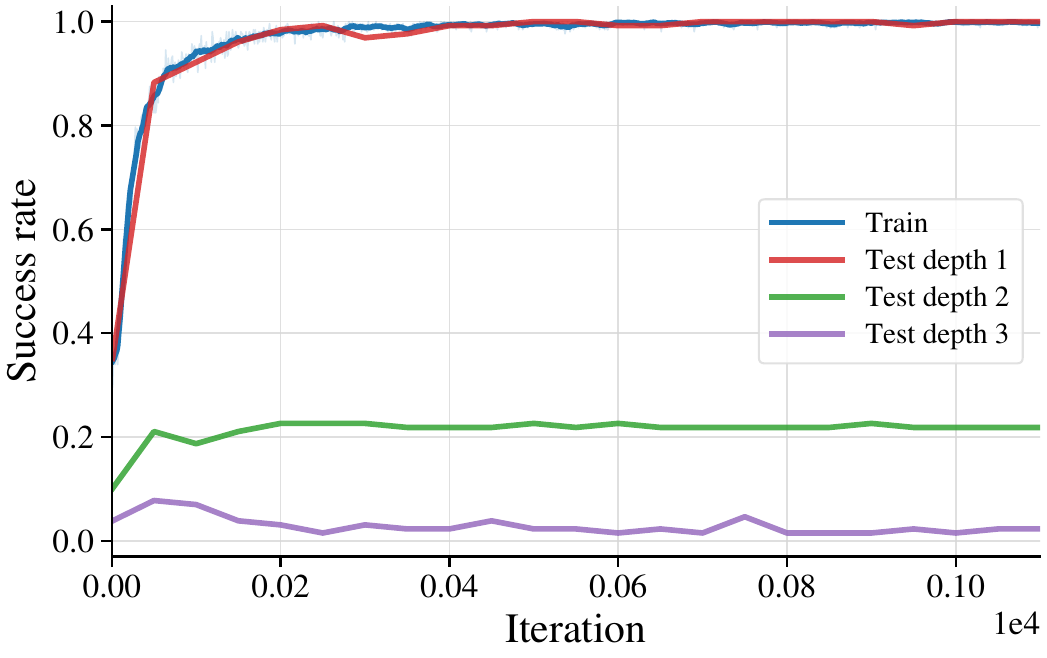}}
  \hspace{0.02\textwidth}
  \subfigure[Depth-2 curriculum training]{\includegraphics[width=0.45\textwidth]{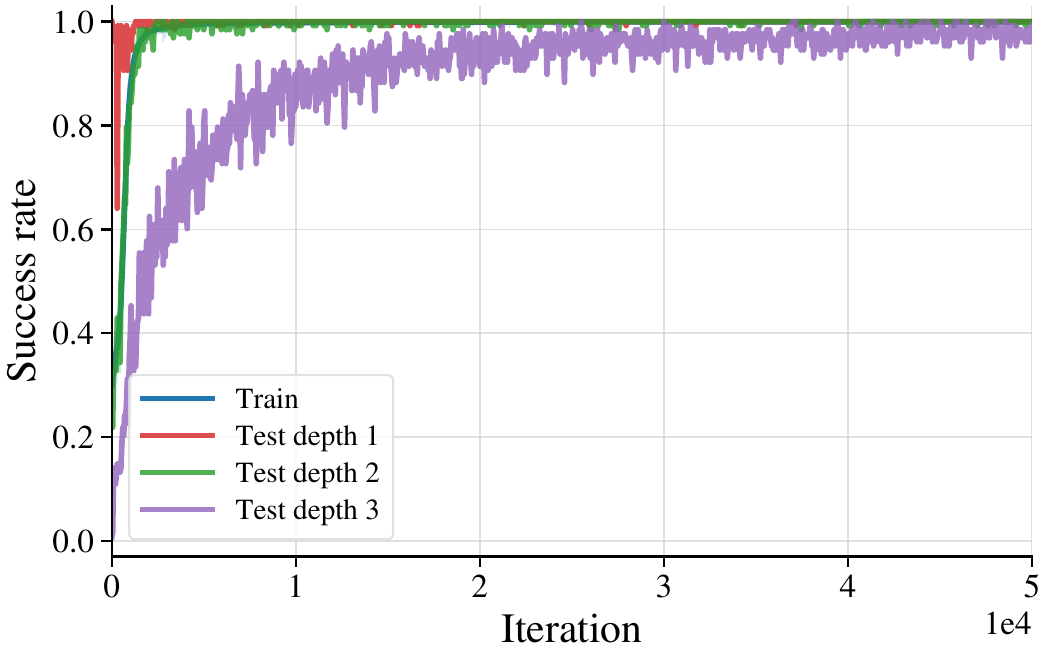}}
  \caption{Train and test 
  rates under imbalanced goal distribution for depth-1 (left) and depth-2 curriculum training (right), trained on perfect 3-ary trees with stochastic policy gradient, and tested on randomly generated full 3-ary trees with depth 1 to 3.}
  \label{fig:imbalanced_curves}
\end{figure}

\begin{figure}[htbp]
  \centering
 \subfigure[$A_C$]{\includegraphics[width=0.23\textwidth]{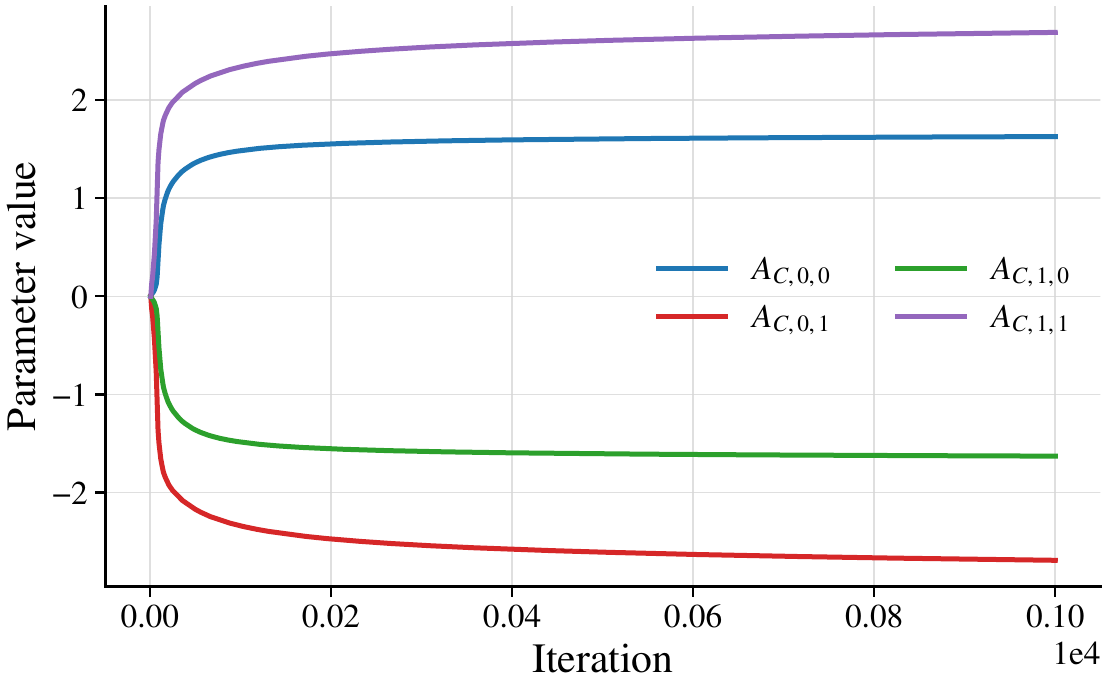}} 
   \hspace{0.01\textwidth}
  \subfigure[$A_Q$]{\includegraphics[width=0.23\textwidth]{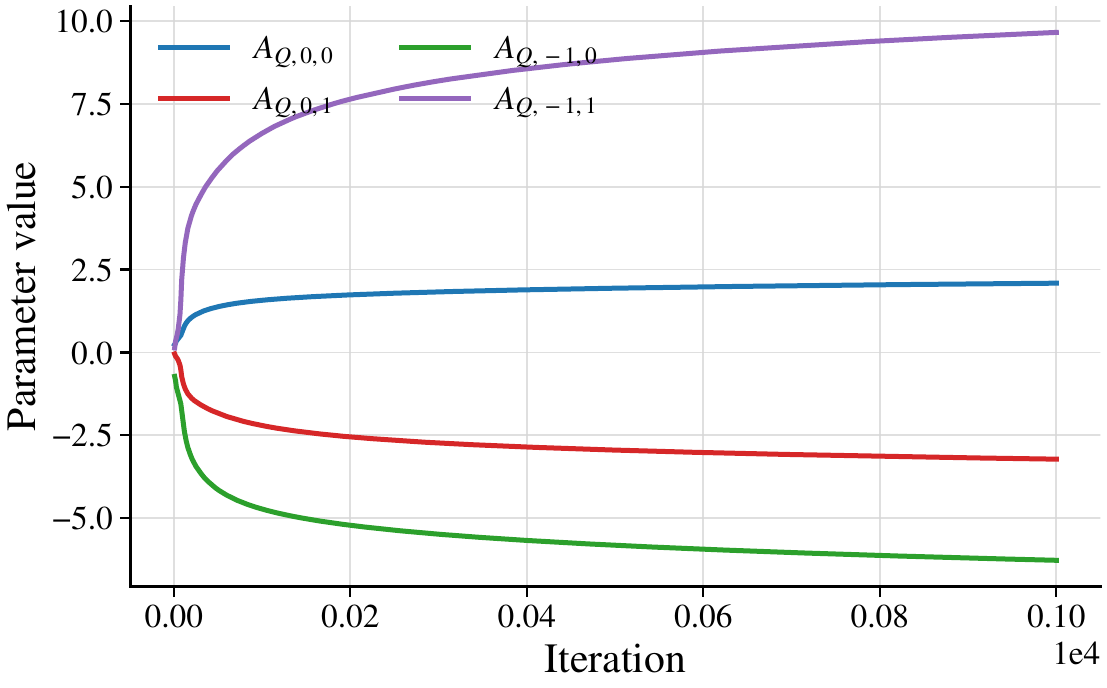}}
    \hspace{0.01\textwidth}
    \subfigure[$A_B$]{\includegraphics[width=0.23\textwidth]{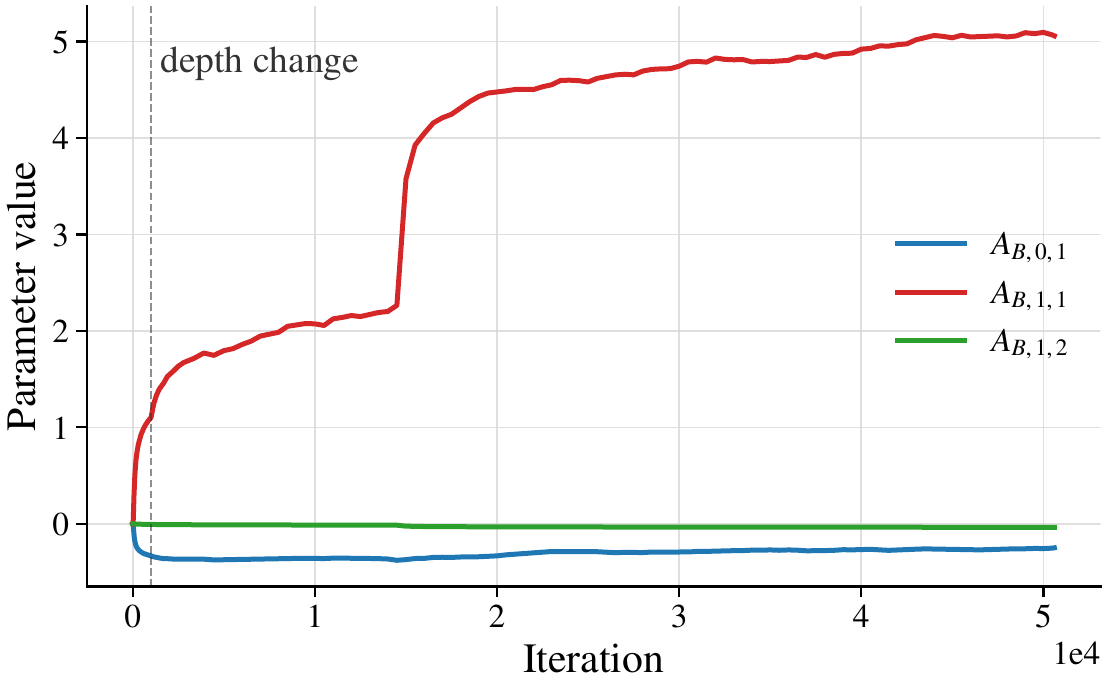}}
   \hspace{0.01\textwidth}
  \subfigure[$A_P$]{\includegraphics[width=0.23\textwidth]{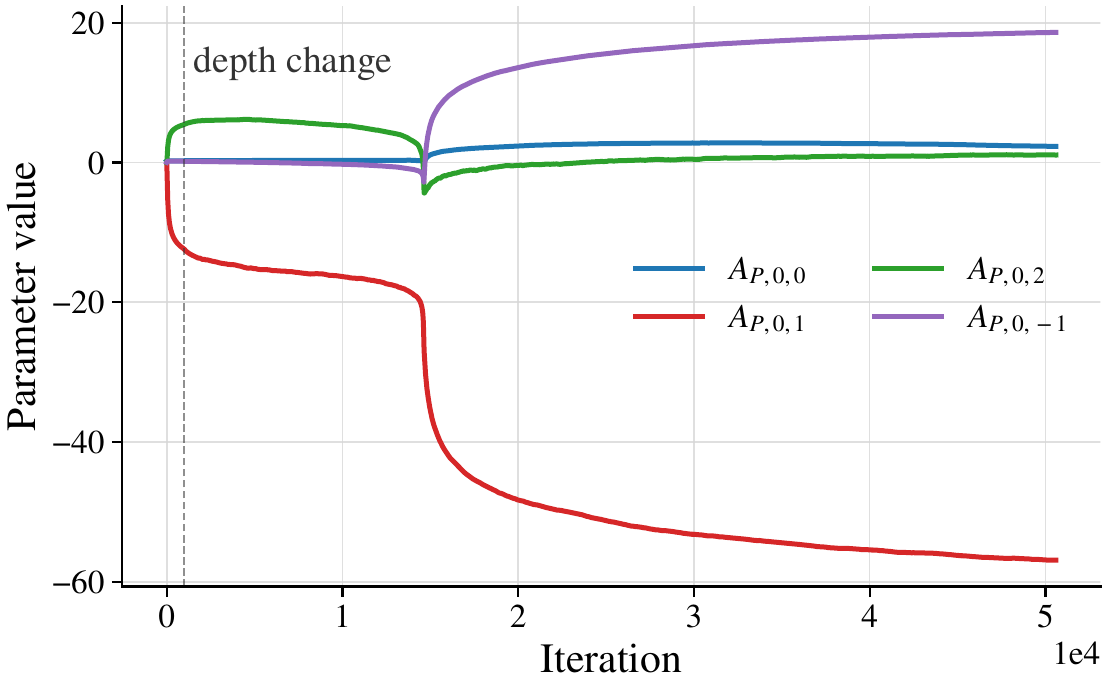}} 
  \caption{Training dynamics of selected entries of $A_C$,  $A_Q$, $A_B$, $A_P$ under balanced goal distribution: $A_B$ and $A_P$ (associated with head 1) through depth-1 and depth-2 training; $A_C$ and $A_Q$ (associated with head 2) through depth-1 training.}
  \label{fig:entries_balanced}
\end{figure}

\begin{figure}[ht]
  \centering
  \subfigure[$A_B$]{\includegraphics[width=0.23\textwidth]{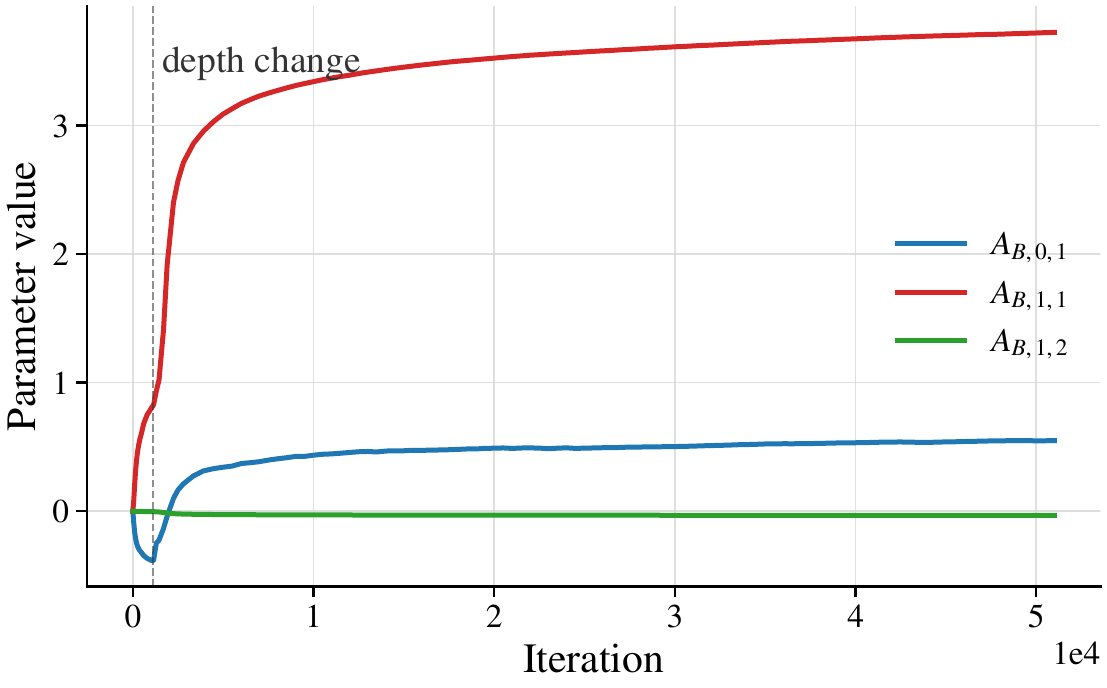}}
  \hspace{0.01\textwidth}
  \subfigure[$A_C$]{\includegraphics[width=0.23\textwidth]{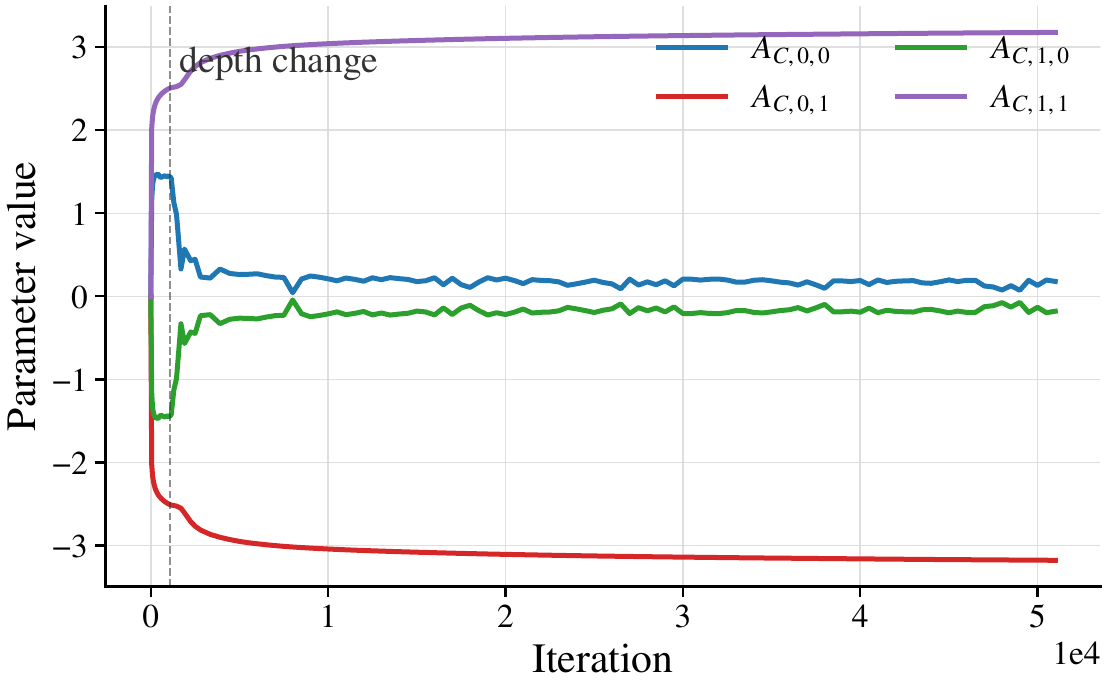}} 
  \hspace{0.01\textwidth}
    \subfigure[$A_P$]{\includegraphics[width=0.23\textwidth]{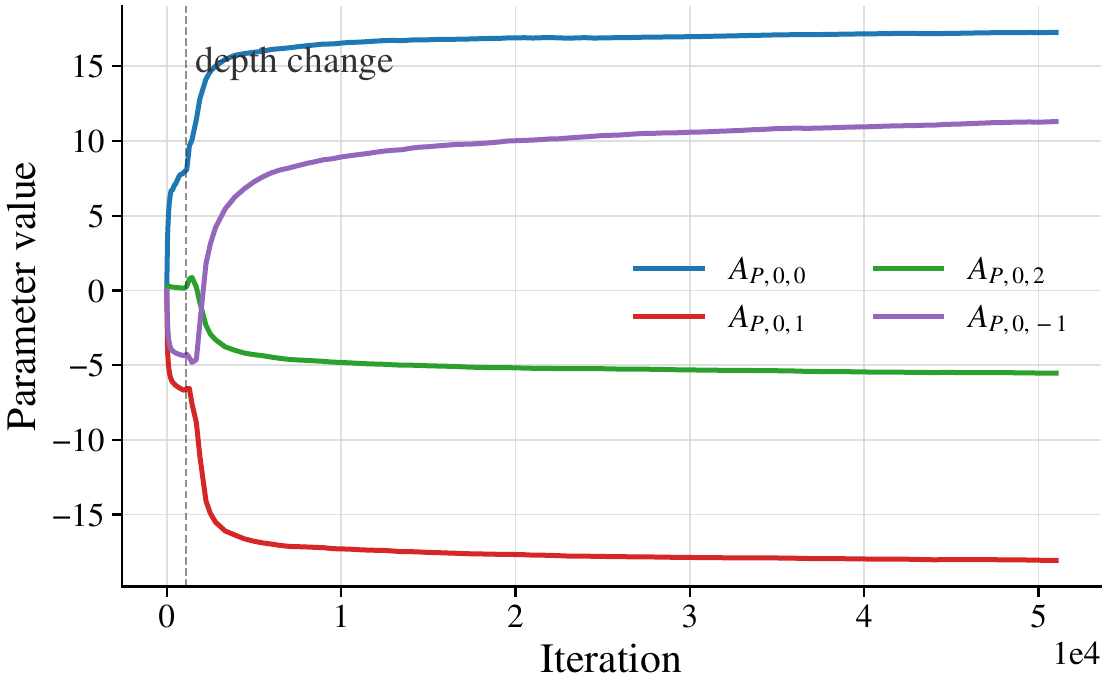}}
  \hspace{0.01\textwidth}
  \subfigure[$A_Q$]{\includegraphics[width=0.23\textwidth]{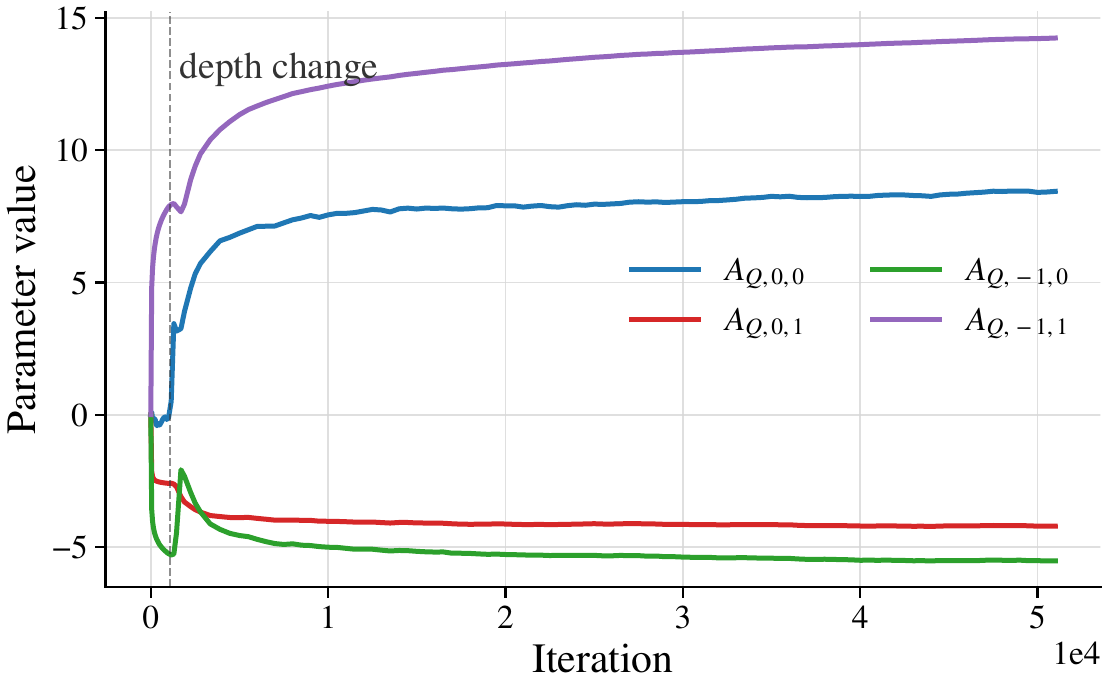}}
  \caption{Selected entries of $A_B$, $A_C$, $A_P$, and $A_Q$ under imbalanced training through depth-1 and depth-2 curriculum training.}
  \label{fig:entries_imbalanced}
\end{figure}

\begin{figure}[ht]
  \centering
  \subfigure[$A_C$]{\includegraphics[height=0.18\textwidth]{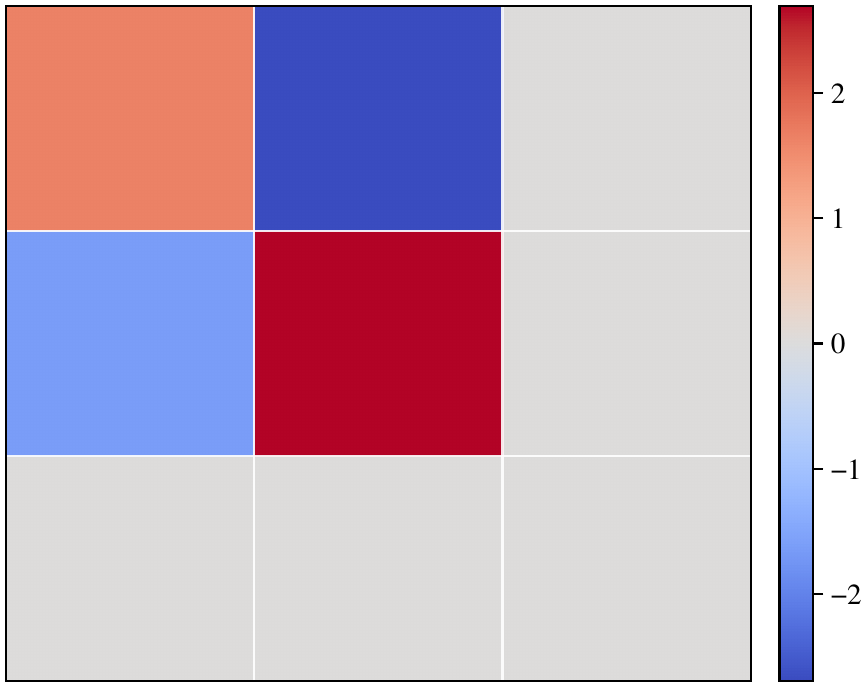}}  
  \hspace{0.01\textwidth}
  \subfigure[$A_Q$]{\includegraphics[height=0.18\textwidth]{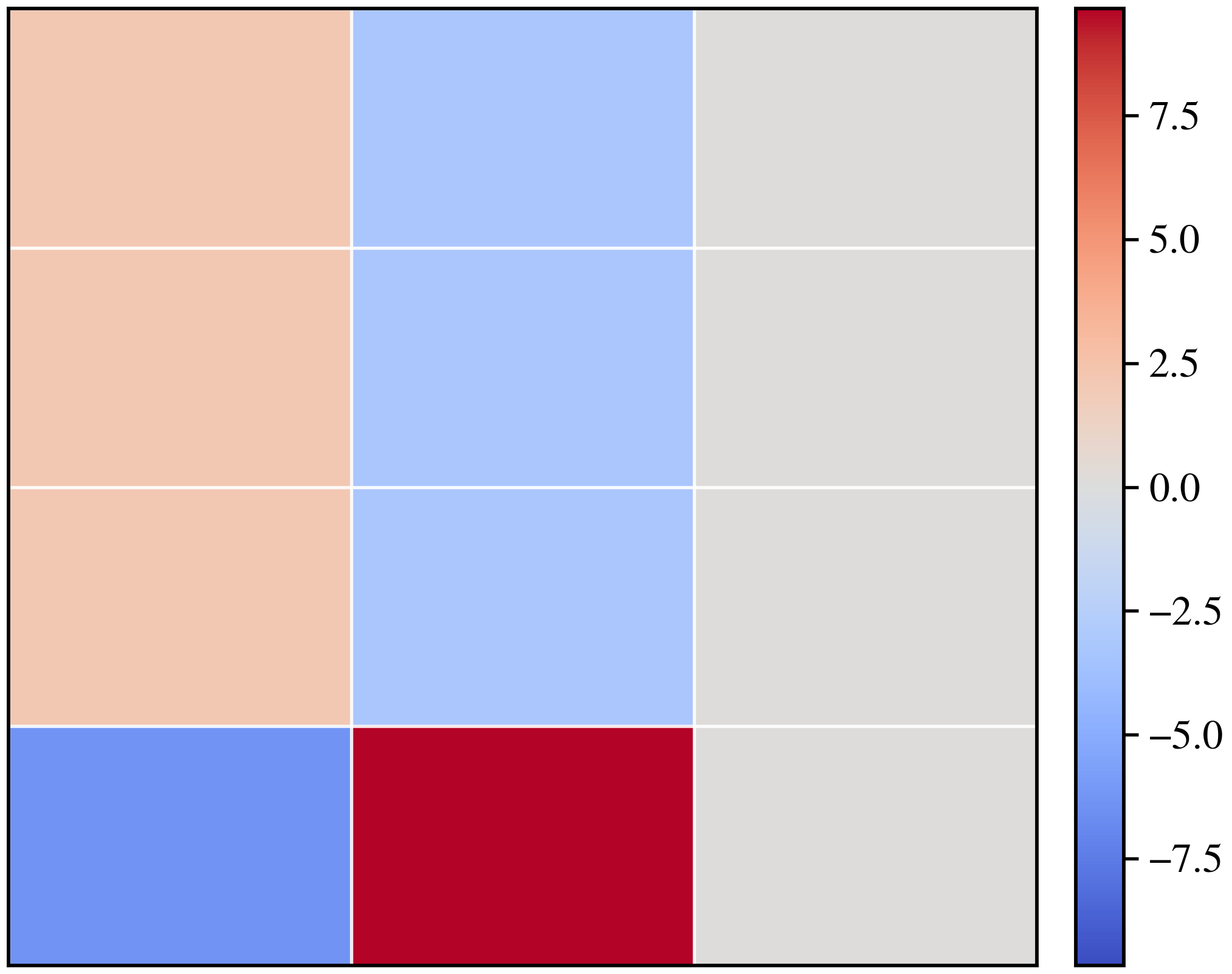}}
  \hspace{0.01\textwidth}
  \subfigure[{$A_B[:20,:20]$}]{\includegraphics[height=0.18\textwidth]{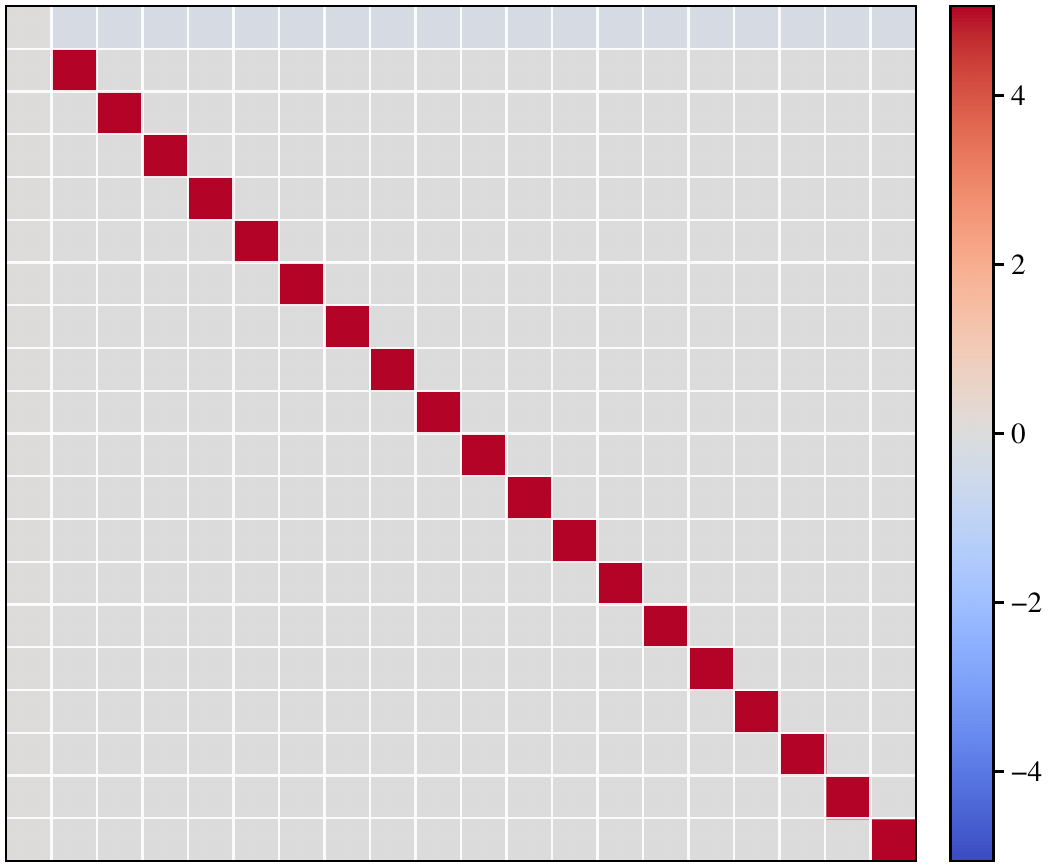}}
  \hspace{0.01\textwidth}
  \subfigure[$A_P$]{\includegraphics[height=0.18\textwidth]{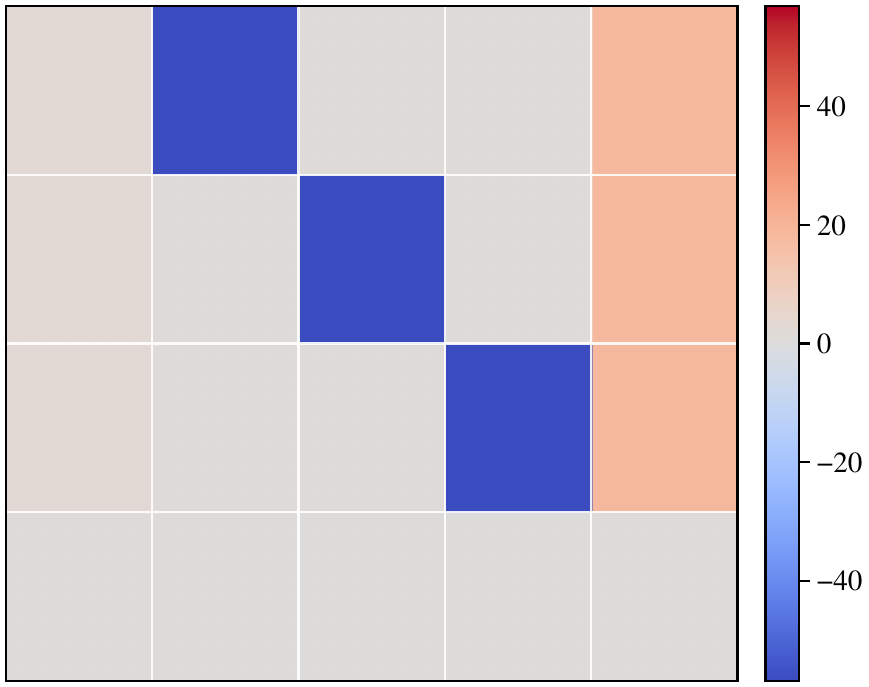}} \\
  \subfigure[$A_C$]{\includegraphics[height=0.18\textwidth]{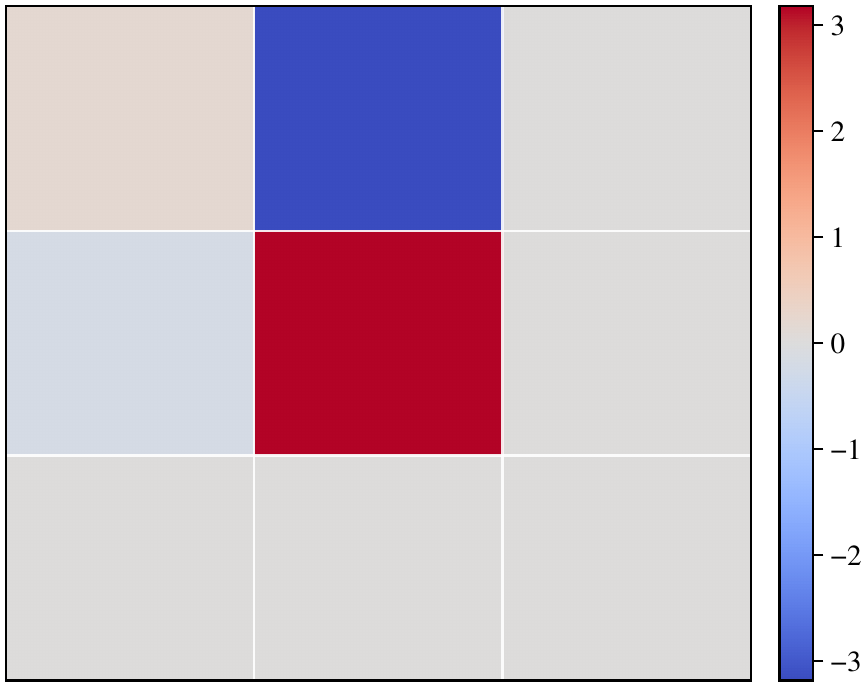}}
    \hspace{0.01\textwidth}
  \subfigure[$A_Q$]{\includegraphics[height=0.18\textwidth]{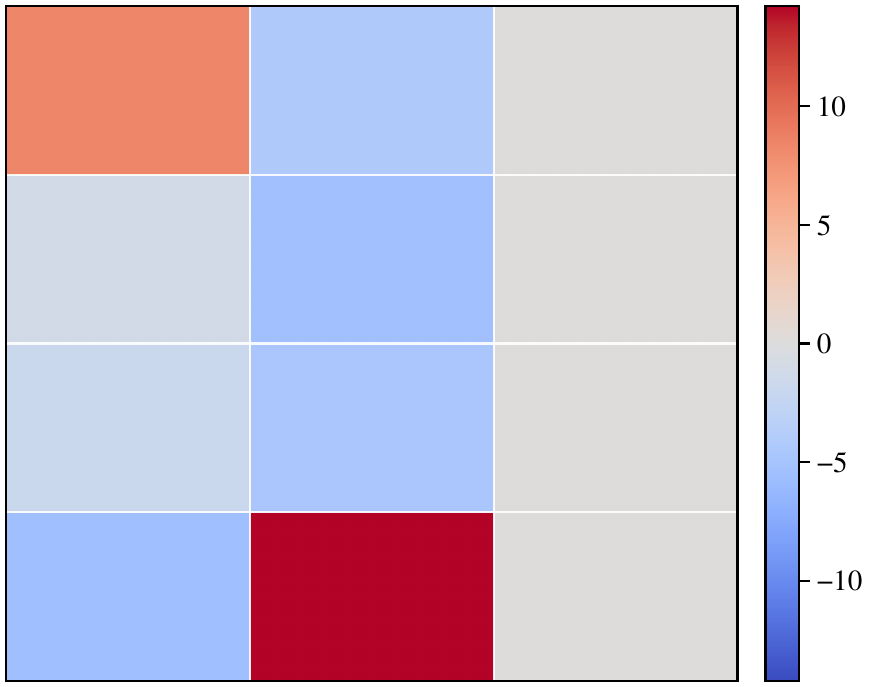}}
  \hspace{0.01\textwidth}
      \subfigure[$A_B$]{\includegraphics[height=0.18\textwidth]{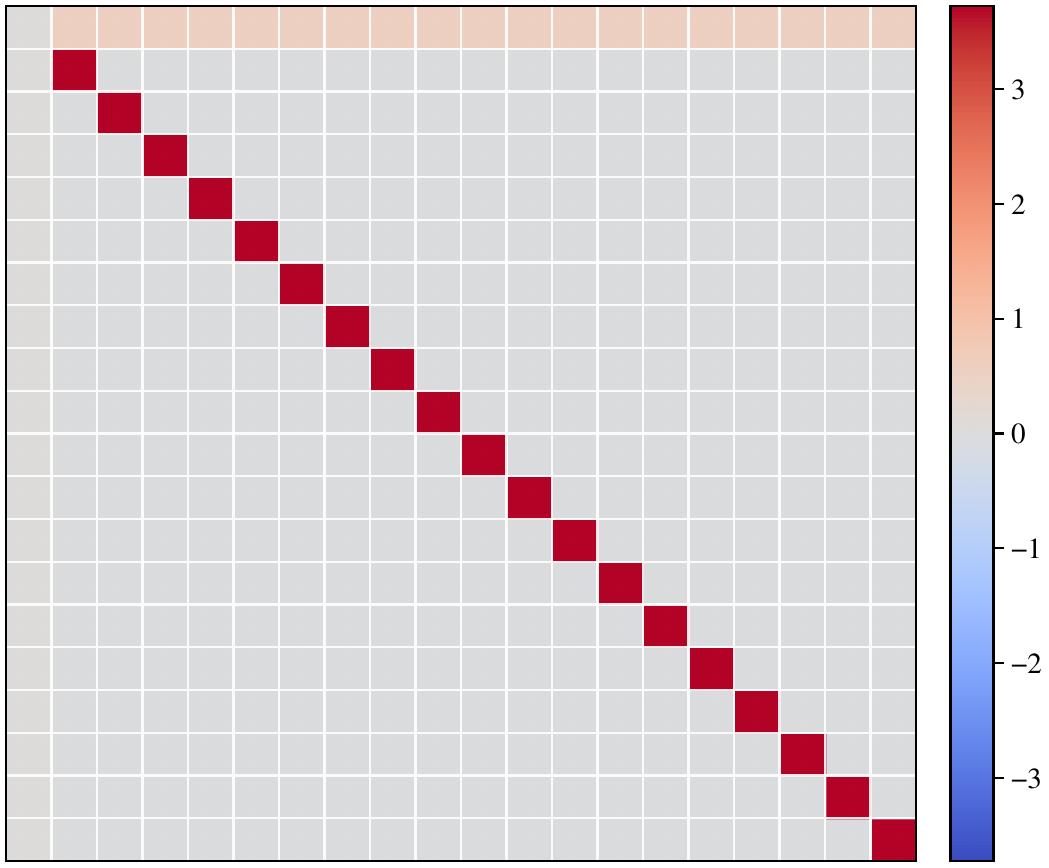}}
  \hspace{0.01\textwidth}
  \subfigure[$A_P$]{\includegraphics[height=0.18\textwidth]{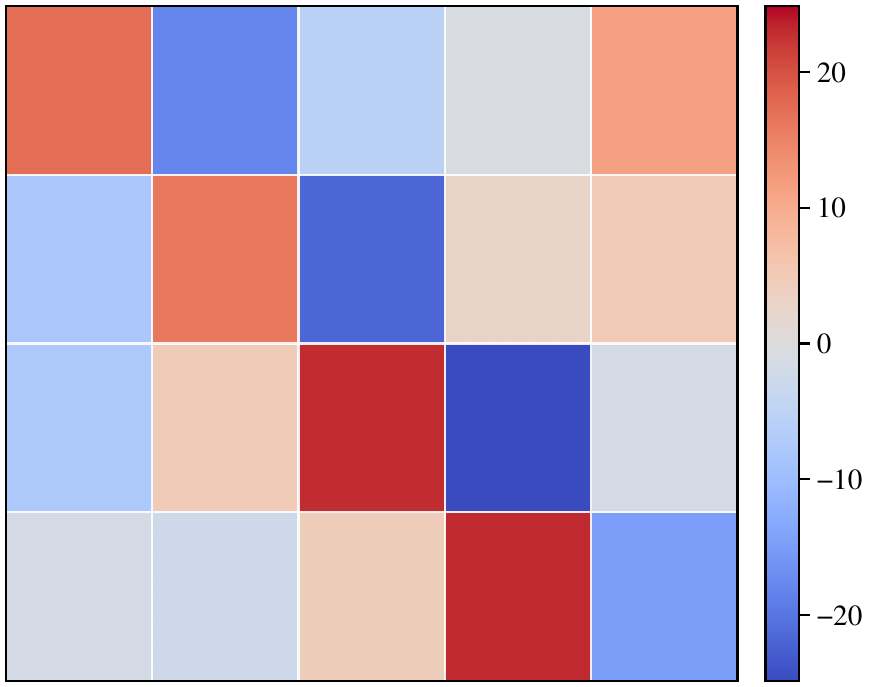}}
  \caption{Heat maps of $A_B[:20,:20]$, $A_C$, $A_P$, and $A_Q$ under the balanced (first row) and imbalanced (second row) goal distribution at the end of training.}
  \label{fig:matrix_evolution}
\end{figure}

\paragraph{Results.} Figures~\ref{fig:reward_balanced} and Figure~\ref{fig:imbalanced_curves} report the success rates at depths 1 and 2, together with the generalization performance for both balanced and imbalanced cases. Figure~\ref{fig:entries_balanced} and Figure~\ref{fig:entries_imbalanced} show trajectories of selected matrix entries in $A_B$, $A_C$, $A_P$, and $A_Q$, showing their dynamics converge to their constructed patterns. Figure~\ref{fig:matrix_evolution} shows heat maps of $A_C$, $A_P$, $A_Q$ and the first 20 rows and 20 columns of $A_B$ under the balanced and imbalanced goal distributions. We can see that the experiment results overall agree with our theoretical analysis.

\section{Conclusion}\label{sec:conclusion}

We have shown that an interactive search algorithm can emerge from RL training in a transformer agent, even when supervision comes solely from sparse terminal reward. In our analysis, policy gradient updates shape two attention heads into complementary memory-retrieval mechanisms: one suppresses previously tried actions, while the other detects failed outcomes and triggers backtracking. Their composition yields a DFS policy that generalizes from shallow curriculum training to deep trees, and under imbalanced goal distributions, discounting further biases the learned policy toward ranked search. Our work is a first step toward understanding complex agentic behavior in stochastic RL environments, and provides a tractable testbed for future directions including principled memory management, context pruning, and scalable RL algorithms for long-horizon interactive reasoning.

\bibliographystyle{abbrvnat}
\bibliography{biblio}


\appendix


\section{Proof of Convergence and Generalization}\label{sec:proof_convergence}
\subsection{Formal Statements of Theorems~\ref{thm:convergence_phase_1_balanced} and Theorem~\ref{thm:convergence_phase_2_balanced}}\label{sec_app:main_statements_proof_convergence}
Below we provide the formal statements of Theorems~\ref{thm:convergence_phase_1_balanced} and Theorem~\ref{thm:convergence_phase_2_balanced} in a merged version, where we use $\poly(k)$ to denote a polynomial function of $k$.
\begin{thm}[convergence and generalization results under balanced goal distribution, formal]\label{thm:convergence_balanced}
    Suppose \eqref{eq:balanced_goal_distribution}, Assumption~\ref{asmp:token_embedding} hold, $k\in\NN_+$ is larger than some absolute constant, $N=\poly(k)\geq \max\left\{k^5,\frac{k^{L+1}-1}{k-1}\right\}$, $L$ is treated as a constant, and $\eta>0$ is smaller than some absolute constant, $\epsilon>0$ satisfies $\ln (1/\epsilon)\geq e^k$. Let $\delta_1=\epsilon^{1.05}$, $\delta_2=\frac{\epsilon^{1.02}}{k^L}$. Under our training setting, for depth-1 curriculum training, the success rate $R(\pi;\gP_1)$ reaches $1-\delta_1$ in $T_1=O\left(\frac{1}{\eta\epsilon^{1.1}}\right)$ iterations. At the end of depth-1 curriculum training, we have 
\begin{align}\label{eq:success_rate_generalization_balanced_phase_1_app}
  R(\pi^{(T_1)};\gP_l)= O\left(\frac{1}{k^{l-1}}\right), \quad \forall 2\leq l\leq L.
\end{align}
    For depth-2 curriculum training, the success rate $R(\pi;\gP_2)$ reaches $1-\delta_2$ in $T_2-T_1=O\left(\frac{k^L}{\eta\epsilon^{1.1}}\right)$ iterations.
  After depth-2 curriculum training, given any full $k$-ary tree $\gT$ with depth $l\in[L]$ and any goal node $g(\gT)$, we have
  \begin{align}\label{eq:success_rate_generalization_balanced_phase_2_app}
    R(\pi^{(T_2)};(\gT,g(\gT)))\geq 1-\epsilon.
  \end{align}
  \end{thm}
The proof of Theorem~\ref{thm:convergence_balanced} is provided in Appendix~\ref{sec:proof_convergence_balanced}. Note that in Theorem~\ref{thm:convergence_balanced}, 
without loss of generality, we assume (the hardest and the most extreme case where) $\ln(1/\epsilon)\geq e^k$. If $\epsilon$ is larger, the convergence bound in Theorem~\ref{thm:convergence_balanced} also holds: the proof still goes through and possibly becomes simpler with a bit of extra discussions, as depth-1 curriculum training may end before $T_{1,i}$ for some $i\in[5]$ defined in Appendix~\ref{sec:proof_convergence_balanced} and thus we don't need to track some sub-phases.

\paragraph{Comparison with related training-dynamics analyses.}
As far as we know, ours is the first optimization-level analysis of policy-gradient training in which coupled transformer parameters ($B,C,\overline P,Q$) coevolve into a two-head memory circuit for agentic search, yielding DFS and depth generalization from depth-1/2 curriculum training to deeper unseen trees. Prior analyses simplify the setting along different axes. Graph-search works typically study static instances whose graph structure is revealed in the prompt \citep{saparov2024transformers,wang2024alpine,yang2025multi}. RL planning analyses operate on fixed graph abstractions, where generation amounts to extending a path by next-node tokens and process rewards may expose adjacency or target information \citep{wang2025benefits}. RL/RLVR reasoning analyses, while using outcome or verifier rewards, largely treat RL as post-training over self-generated reasoning traces rather than interaction with a partially observed environment~\citep{bu2025provable,huang2026learning}; in these settings the evolving state is essentially the generated prefix, or the analysis assumes/fixes atomic operations needed for reasoning. 

By contrast, our transformer is an agent in a hidden stochastic tree: it must infer the search state from trajectory history and sparse terminal reward, while learning to remember failed branches, avoid repeated actions, and backtrack from both failed leaves and exhausted internal nodes. This richer interactive setting also makes the training-dynamics analysis substantially more delicate: the proof must track a multi-phase evolution of interacting attention and output matrices, while resolving the relevant return-to-go terms under both a stochastic policy and posterior uncertainty over the hidden environment.

\subsection{Proof of Theorem~\ref{thm:convergence_balanced}}\label{sec:proof_convergence_balanced}

\textbf{Notation.} Throughout the proof, we use $o(1)$ or e.g., $O\left(\frac{1}{k}\right)$ to represent small quantities that may also be negative. We sometimes abuse the notation to use $[t_1,t_2]$ to denote $\{t_1,t_1+1,\ldots,t_2\}$ for integers $t_1\leq t_2$.
  Though we fix the last row of $P$ to be 0 and don't update it (cf.~\eqref{eq:W_P}),
its gradient is still useful and we'll compute it. Note that
\begin{align}
    \delta M^{(t)}&\coloneqq \EE_{\Phi\sim\gP,\pi_{\theta^{(t)}}}\left[\sum_{h=1}^H \nabla_{M} \log \pi_\theta(a_h\mid E^{(h)})\Big|_{M=M^{(t)}} \sum_{i=h}^H 
     r_{i+1}\right]\,\,\text{ for } M\in\{B,C, P,Q\}.
\end{align}
Then according to update rule \eqref{eq:policy_gradient_update}, we have
\begin{align}\label{eq:matrices_update}
    M^{(t+1)}=M^{(t)}+\eta\delta M^{(t)}\,\,\text{ for } M\in\{B,C, \overline{P},Q\}.
\end{align}
More generally, for any quantity $\{X^{(t)}\}_{t\in\NN}$ (except for the last row of $P$), we define 
\begin{align}\label{eq:delta_X_t}
    \delta X^{(t)}\coloneqq \frac{X^{(t+1)}-X^{(t)}}{\eta}.
\end{align}
We let
\begin{align}\label{eq:U_V}
    U\coloneqq (u_0,\cdots,u_N)\in\R^{d_1\times (N+1)},\,\, V\coloneqq (v_0,\cdots,v_{k+1})\in\R^{d_2\times (k+2)},\,\, Z\coloneqq (z_0,z_1,z_2)\in\R^{d_3\times 3}
\end{align}
denote the node, action, and label embedding matrices respectively.
We define the induced matrix $A_M$ $(M\in\{B,C,P,Q\})$ as 
\begin{align}\label{eq:induced_matrices}
    A_B\coloneqq U^\top B U\in\R^{(N+1)\times (N+1)},
    \quad A_C\coloneqq Z^\top CZ\in\R^{3\times 3},\notag\\
    A_P\coloneqq P  V\in\R^{(k+1)\times (k+2)},
    \quad A_Q\coloneqq Q Z\in\R^{(k+1)\times 3},
\end{align}
and let $A_M^{(t)}\coloneqq A_{M^{(t)}}$  and $\delta A_M^{(t)}\coloneqq \delta A_{M^{(t)}}$ for $M\in\{B,C,P,Q\}$.
Then by \eqref{eq:matrices_update}, we have
\begin{align}\label{eq:induced_matrices_update}
    A_M^{(t+1)}=A_M^{(t)}+\eta\delta A_M^{(t)}\,\,\text{ for } M\in\{B,C, \overline{P} ,Q\}.
\end{align}

We define 
\begin{align}\label{eq:pi_G_Q_h}
    \pi_h\coloneqq \pi_\theta(E^{(h)}),\quad \pi_h^{(t)}\coloneqq \pi_{\theta^{(t)}}(E^{(h)}),\quad \widehat{G}_h\coloneqq \sum_{i=h}^H r_{i+1},\notag\\
    Q_h^{(t)}(a)\coloneqq\EE_{\pi^{(t)}}[\widehat{G}_h|\gF_h,a_h=a],\quad G_h^{(t)}\coloneqq \EE_{\pi^{(t)}}[\widehat{G}_h|\gF_h]=\sum_{a\in\gA}\pi_h^{(t)}(a)Q_h^{(t)}(a),
\end{align}
where $\gF_h$ denotes the $\sigma$-field generated by the history available before choosing $a_h$, equivalently by $E^{(h)}$. We also write $Q_h^{(t)}\in\R^{k+1}$ for the vector whose $a$-th coordinate is $Q_h^{(t)}(a)$. 
Define $N_h$ as the number of leaf nodes explored up to step $h$:
\begin{align}\label{eq:N_h_times}
    N_{h,\times}\coloneqq \sum_{i=1}^h \mathbbm{1}\{c_i=\times\}.
\end{align}
At any step $h$, the agent is at node $n_h$. 
We let
\begin{align}\label{eq:gA_h_times}
\gA_{h,\times}\coloneqq \begin{cases}
\{i\in[k]|i \text{ is taken at } n_h\}, \quad \text{if } c_h=0,\\
\emptyset,\quad \text{if } c_h=\times,\\
\end{cases}
\end{align}
denote the set of downward actions that are taken at node $n_h$. 
Then for any action $i\in\gA_{h,\times}$, $Q_h^{(t)}(i)=0$.
We also define 
\begin{align}\label{eq:N_h_i}
    \forall i\in[k+1]:\quad N_h(i)\coloneqq \sum_{j=1}^{h-1} \mathbbm{1}\{a_j=i\}
\end{align}
as the number of times action $i$ is taken up to step $h$. And we let $N_h(\up)\coloneqq N_h(k+1)$. 

\paragraph{Gradient computation.}
 We first derive the expressions of $\delta A_M^{(t)}$ and the policy $\pi_h$ in Lemma~\ref{lm:gradients}. The expectations at step $t$ are all taken under policy $\pi^{(t)}$. Note that histories used for gradient updates exclude the terminal observation, and thus terminal token $\checkmark$  never appears in $E_c^{(h)}$.
\begin{lm}[gradients and policy expression]\label{lm:gradients}
  Under balanced goal distribution~\eqref{eq:balanced_goal_distribution}, Assumption~\ref{asmp:token_embedding}, we have the following gradient expressions for any $t\in\NN$:
\begin{align}\label{eq:A_C_simplified}
    A_C^{(t)} = \begin{pmatrix}
        a_{c,0}^{(t)} & -a_{c,1}^{(t)} & 0 \\
        -a_{c,0}^{(t)} & a_{c,1}^{(t)} & 0 \\
        0 & 0 & 0
    \end{pmatrix},\quad 
    \delta A_C^{(t)} = \begin{pmatrix}
        \delta a_{c,0}^{(t)} & -\delta a_{c,1}^{(t)} & 0 \\
        -\delta a_{c,0}^{(t)} & \delta a_{c,1}^{(t)} & 0 \\
        0 & 0 & 0
    \end{pmatrix},
\end{align}
where 
\small
\begin{subequations}
\begin{align}
    \delta a_{c,0}^{(t)} &= \left(1+\frac{1}{k}\right)\EE\left[\sum_{h=1}^H \mathbbm{1}\{c_h=0\} \frac{N_{h,\times}(h-N_{h,\times})}{\left(N_{h,\times}e^{-a_{c,0}^{(t)}}+(h-N_{h,\times})e^{a_{c,0}^{(t)}}\right)^2}\left(a_{q,\times}^{(t)}+a_{q,0}^{(t)}\right)\pi_h^{(t)}(\up)\left(G_h^{(t)}-Q_h^{(t)}(\up)\right)\right],\label{eq:delta_a_c_0_simplified_balanced}
    \\
    \delta a_{c,1}^{(t)} &= \left(1+\frac{1}{k}\right)\EE\left[\sum_{h=1}^H \mathbbm{1}\{c_h=\times\} \frac{N_{h,\times}(h-N_{h,\times})}{\left(N_{h,\times}e^{a_{c,1}^{(t)}}+(h-N_{h,\times})e^{-a_{c,1}^{(t)}}\right)^2}\left(a_{q,\times}^{(t)}+a_{q,0}^{(t)}\right)\pi_h^{(t)}(\up)(1-\pi_h^{(t)}(\up))Q_h^{(t)}(\up)\right].\label{eq:delta_a_c_1_simplified_balanced}
\end{align}
\end{subequations}
\normalsize
\begin{align}\label{eq:delta_A_B_simplified}
    A_B^{(t)} = \begin{pmatrix}
        0 & a_{b,0}^{(t)} & a_{b,0}^{(t)} & \cdots & a_{b,0}^{(t)} \\
        0 & a_{b,1}^{(t)} & a_{b,2}^{(t)} & \cdots & a_{b,2}^{(t)} \\
        0 & a_{b,2}^{(t)} & a_{b,1}^{(t)} & \cdots & a_{b,2}^{(t)} \\
        \vdots & \vdots & \vdots & \ddots & \vdots \\
        0 & a_{b,2}^{(t)} & a_{b,2}^{(t)} & \cdots & a_{b,1}^{(t)}
    \end{pmatrix},\quad
    \delta A_B^{(t)} = \begin{pmatrix}
        0 & \delta a_{b,0}^{(t)} & \delta a_{b,0}^{(t)} & \cdots & \delta a_{b,0}^{(t)} \\
        0 & \delta a_{b,1}^{(t)} & \delta a_{b,2}^{(t)} & \cdots & \delta a_{b,2}^{(t)} \\
        0 & \delta a_{b,2}^{(t)} & \delta a_{b,1}^{(t)} & \cdots & \delta a_{b,2}^{(t)} \\
        \vdots & \vdots & \vdots & \ddots & \vdots \\
        0 & \delta a_{b,2}^{(t)} & \delta a_{b,2}^{(t)} & \cdots & \delta a_{b,1}^{(t)}
    \end{pmatrix},
\end{align}
where
\begin{subequations}
\begin{align}
    \delta a_{b,0}^{(t)}&=\frac{1}{N}\EE\Bigg[\sum_{h=1}^H\frac{e^{a_{b,0}^{(t)}}}{\left(e^{a_{b,0}^{(t)}}+|\gA_{h,\times}|e^{a_{b,1}^{(t)}}+(h-1-|\gA_{h,\times}|)e^{a_{b,2}^{(t)}}\right)^2}\notag\\
    &\qquad\qquad \cdot\left\{
        \left(|\gA_{h,\times}|e^{a_{b,1}^{(t)}}+(h-1-|\gA_{h,\times}|)e^{a_{b,2}^{(t)}}\right)\beta_{h,0}^{(t)}-e^{a_{b,1}^{(t)}}\beta_{h,1}^{(t)}-e^{a_{b,2}^{(t)}}\beta_{h,2}^{(t)}
    \right\}\Bigg]\label{eq:delta_a_b_0_simplified_balanced},\\
    \delta a_{b,1}^{(t)}&=\frac{1}{N}\EE\Bigg[\sum_{h=1}^H\frac{e^{a_{b,1}^{(t)}}}{\left(e^{a_{b,0}^{(t)}}+|\gA_{h,\times}|e^{a_{b,1}^{(t)}}+(h-1-|\gA_{h,\times}|)e^{a_{b,2}^{(t)}}\right)^2}\notag\\
    &\qquad\qquad \cdot\left\{
        \left(e^{a_{b,0}^{(t)}}+(h-1-|\gA_{h,\times}|)e^{a_{b,2}^{(t)}}\right)\beta_{h,1}^{(t)}-e^{a_{b,0}^{(t)}}|\gA_{h,\times}|\beta_{h,0}^{(t)}-e^{a_{b,2}^{(t)}}|\gA_{h,\times}|\beta_{h,2}^{(t)}
    \right\}\Bigg]\label{eq:delta_a_b_1_simplified_balanced},\\
    a_{b,2}^{(t)}&=\frac{-a_{b,0}^{(t)}-a_{b,1}^{(t)}}{N-1},\quad \delta a_{b,2}^{(t)}=\frac{-\delta a_{b,0}^{(t)}-\delta a_{b,1}^{(t)}}{N-1},\label{eq:delta_a_b_2_simplified_balanced}
\end{align}
\end{subequations}
with
\begin{subequations}
\begin{align}
    \beta_{h,0}^{(t)}&\coloneqq \frac{a_{p,0}^{(t)}}{k}\pi_h^{(t)}(\up)\left(Q_h^{(t)}(\up)-G_h^{(t)}\right),\label{eq:beta_h_0_balanced} \\
    \beta_{h,1}^{(t)}&\coloneqq (a_{p,1}^{(t)}+a_{p,2}^{(t)})\sum_{i\in\gA_{h,\times}}\pi_h^{(t)}(i)
    G_h^{(t)}-a_{p,2}^{(t)}|\gA_{h,\times}|\pi_h^{(t)}(\up)(Q_h^{(t)}(\up)-G_h^{(t)}),\label{eq:beta_h_1_balanced}\\
    \beta_{h,2}^{(t)}&\coloneqq -\left((h-1-N_h(\up)-|\gA_{h,\times}|)a_{p,2}^{(t)}+\frac{1}{k}N_h(\up)a_{p,\up}^{(t)}\right)\pi_h^{(t)}(\up)(Q_h^{(t)}(\up)-G_h^{(t)})\notag\\
    &\quad+(a_{p,1}^{(t)}+a_{p,2}^{(t)})\left(\sum_{i\in\gA_{h,\times}}(N_h(i)-1)\pi_h^{(t)}(i)G_h^{(t)}+\sum_{i\in[k]\setminus\gA_{h,\times}}N_h(i)\pi_h^{(t)}(i)(G_h^{(t)}-Q_h^{(t)}(i))\right).\label{eq:beta_h_2_balanced}
\end{align}
\end{subequations}

\small
\begin{align}\label{eq:A_P_simplified}
    A_{P}^{(t)}=\begin{pmatrix}
    -\frac{a_{p,0}^{(t)}}{k} & -a_{p,1}^{(t)} & a_{p,2}^{(t)} & \cdots & a_{p,2}^{(t)} & \frac{a_{p,\up}^{(t)}}{k} \\
    -\frac{a_{p,0}^{(t)}}{k} & a_{p,2}^{(t)} & -a_{p,1}^{(t)} & \cdots & a_{p,2}^{(t)} & \frac{a_{p,\up}^{(t)}}{k} \\
    \vdots & \vdots & \vdots & \ddots & \vdots & \vdots \\
    -\frac{a_{p,0}^{(t)}}{k} & a_{p,2}^{(t)} & a_{p,2}^{(t)} & \cdots & -a_{p,1}^{(t)} & \frac{a_{p,\up}^{(t)}}{k}\\
    0 & 0 & 0 & \cdots & 0 & 0
    \end{pmatrix},\;
    \delta A_P^{(t)}=\begin{pmatrix}
        -\frac{\delta a_{p,0}^{(t)}}{k} & -\delta a_{p,1}^{(t)} & \delta a_{p,2}^{(t)} & \cdots & \delta a_{p,2}^{(t)} & \frac{\delta a_{p,\up}^{(t)}}{k} \\
    -\frac{\delta a_{p,0}^{(t)}}{k} & \delta a_{p,2}^{(t)} & -\delta a_{p,1}^{(t)} & \cdots & \delta a_{p,2}^{(t)} & \frac{\delta a_{p,\up}^{(t)}}{k} \\
    \vdots & \vdots & \vdots & \ddots & \vdots & \vdots \\
    -\frac{\delta a_{p,0}^{(t)}}{k} & \delta a_{p,2}^{(t)} & \delta a_{p,2}^{(t)} & \cdots & -\delta a_{p,1}^{(t)} & \frac{\delta a_{p,\up}^{(t)}}{k}\\
    \delta a_{p,0}^{(t)} & \delta a_{p,3}^{(t)} & \delta a_{p,3}^{(t)} & \cdots & \delta a_{p,3}^{(t)} & -\delta a_{p,\up}^{(t)}
    \end{pmatrix},
\end{align}
\normalsize
where 
\begin{subequations}
\begin{align}
    \delta a_{p,3}^{(t)}&= \delta a_{p,1}^{(t)}-(k-1)\delta a_{p,2}^{(t)},\quad a_{p,3}^{(t)}\coloneqq\eta\sum_{t'=0}^{t-1}\delta a_{p,3}^{(t')}=a_{p,1}^{(t)}-(k-1)a_{p,2}^{(t)},\label{eq:a_p3}\\
    \delta a_{p,0}^{(t)} &=\EE\left[\sum_{h=1}^H \pi_h^{(t)}(\up)\frac{e^{a_{b,0}^{(t)}}}{e^{a_{b,0}^{(t)}}+|\gA_{h,\times}|e^{a_{b,1}^{(t)}}+(h-1-|\gA_{h,\times}|)e^{a_{b,2}^{(t)}}}\left(Q_h^{(t)}(\up)-G_h^{(t)}\right)\right],\label{eq:delta_a_p_0_simplified_balanced}\\
    \delta a_{p,1}^{(t)} &=\EE\Bigg[\sum_{h=1}^H\pi_h^{(t)}(j)\bigg(\mathbbm{1}\{j\in\gA_{h,\times}\}
    \frac{e^{a_{b,1}^{(t)}}+(N_h(j)-1)e^{a_{b,2}^{(t)}}}{e^{a_{b,0}^{(t)}}+|\gA_{h,\times}|e^{a_{b,1}^{(t)}}+(h-1-|\gA_{h,\times}|)e^{a_{b,2}^{(t)}}}G_h^{(t)}\notag\\
    &\qquad-\mathbbm{1}\{j\in[k]\setminus\gA_{h,\times}\}\frac{N_h(j)e^{a_{b,2}^{(t)}}}{e^{a_{b,0}^{(t)}}+|\gA_{h,\times}|e^{a_{b,1}^{(t)}}+(h-1-|\gA_{h,\times}|)e^{a_{b,2}^{(t)}}}\left(Q_h^{(t)}(j)-G_h^{(t)}\right)\bigg)\Bigg]
    ,\label{eq:delta_a_p_1_simplified_balanced}\\
    \delta a_{p,3}^{(t)} &=\EE\Bigg[\sum_{h=1}^H \pi_h^{(t)}(\up)\bigg(\mathbbm{1}\{j\in\gA_{h,\times}\}\frac{e^{a_{b,1}^{(t)}}+(N_h(j)-1)e^{a_{b,2}^{(t)}}}{e^{a_{b,0}^{(t)}}+|\gA_{h,\times}|e^{a_{b,1}^{(t)}}+(h-1-|\gA_{h,\times}|)e^{a_{b,2}^{(t)}}}\notag\\
    &\qquad+\mathbbm{1}\{j\in[k]\setminus\gA_{h,\times}\}\frac{N_h(j)e^{a_{b,2}^{(t)}}}{e^{a_{b,0}^{(t)}}+|\gA_{h,\times}|e^{a_{b,1}^{(t)}}+(h-1-|\gA_{h,\times}|)e^{a_{b,2}^{(t)}}}\bigg)(Q_h^{(t)}(\up)-G_h^{(t)})\Bigg],
    \label{eq:delta_a_p_3_simplified_balanced}\\
    \delta a_{p,\up}^{(t)} &=-\EE\left[\sum_{h=1}^H \pi_h^{(t)}(\up)\frac{N_h(\up)e^{a_{b,2}^{(t)}}}{e^{a_{b,0}^{(t)}}+|\gA_{h,\times}|e^{a_{b,1}^{(t)}}+(h-1-|\gA_{h,\times}|)e^{a_{b,2}^{(t)}}}\left(Q_h^{(t)}(\up)-G_h^{(t)}\right)\right],\label{eq:delta_a_p_up_simplified_balanced}
\end{align}
\end{subequations}
\begin{align}\label{eq:A_Q_simplified}
    A_Q^{(t)} = \begin{pmatrix}
        \frac{a_{q,0}^{(t)}}{k} & -\frac{a_{q,\times}^{(t)}}{k} &  0\\
        \vdots & \vdots  & \vdots \\
        \frac{a_{q,0}^{(t)}}{k} & -\frac{a_{q,\times}^{(t)}}{k} &  0\\
        -a_{q,0}^{(t)} & a_{q,\times}^{(t)} & 0
    \end{pmatrix},\quad
    \delta A_Q^{(t)} = \begin{pmatrix}
        \frac{\delta a_{q,0}^{(t)}}{k} & -\frac{\delta a_{q,\times}^{(t)}}{k} &  0\\
        \vdots & \vdots  & \vdots \\
        \frac{\delta a_{q,0}^{(t)}}{k} & -\frac{\delta a_{q,\times}^{(t)}}{k} &  0\\
        -\delta a_{q,0}^{(t)} & \delta a_{q,\times}^{(t)} & 0
    \end{pmatrix},
\end{align}
where
\begin{subequations}
\begin{align}
    \delta a_{q,0}^{(t)}&=-\EE\left[\sum_{h=1}^H \mathbbm{1}\{c_h=\times\} \frac{(h-N_{h,\times})e^{-a_{c,1}^{(t)}}}{N_{h,\times}e^{a_{c,1}^{(t)}}+(h-N_{h,\times})e^{-a_{c,1}^{(t)}}}\pi_h^{(t)}(\up)(1-\pi_h^{(t)}(\up))Q_h^{(t)}(\up)\right]\notag\\
    &\quad + \EE\left[\sum_{h=1}^H \mathbbm{1}\{c_h=0\} \frac{(h-N_{h,\times})e^{a_{c,0}^{(t)}}}{N_{h,\times}e^{-a_{c,0}^{(t)}}+(h-N_{h,\times})e^{a_{c,0}^{(t)}}}\pi_h^{(t)}(\up)\left(G_h^{(t)}-Q_h^{(t)}(\up)\right)\right],\label{eq:delta_a_q_0_simplified_balanced}\\
    \delta a_{q,\times}^{(t)}&=\EE\left[\sum_{h=1}^H \mathbbm{1}\{c_h=\times\} \frac{N_{h,\times}e^{a_{c,1}^{(t)}}}{N_{h,\times}e^{a_{c,1}^{(t)}}+(h-N_{h,\times})e^{-a_{c,1}^{(t)}}}\pi_h^{(t)}(\up)(1-\pi_h^{(t)}(\up))Q_h^{(t)}(\up)\right]\notag\\
    &\quad - \EE\left[\sum_{h=1}^H \mathbbm{1}\{c_h=0\} \frac{N_{h,\times}e^{-a_{c,0}^{(t)}}}{N_{h,\times}e^{-a_{c,0}^{(t)}}+(h-N_{h,\times})e^{a_{c,0}^{(t)}}}\pi_h^{(t)}(\up)\left(G_h^{(t)}-Q_h^{(t)}(\up)\right)\right].\label{eq:delta_a_q_times_simplified_balanced}
\end{align}
\end{subequations}
Furthermore, for all $t\in\NN$, $\pi_h^{(t)}$ can be expressed as follows:
    \begin{align}\label{eq:pi_h_simplified_balanced}
        \pi_h^{(t)}(\up)&=\frac{1}{Z_h^{(t)}}\exp\left(
            \left(1+\frac{1}{k}\right)\xi_h^{(t)}\right),\notag\\
        \forall i\in[k]:\quad \pi_h^{(t)}(i)&=\frac{1}{Z_h^{(t)}}\exp\left(\frac{\varphi_h^{(t)}(i)}{e^{a_{b,0}^{(t)}}+|\gA_{h,\times}|e^{a_{b,1}^{(t)}}+(h-1-|\gA_{h,\times}|)e^{a_{b,2}^{(t)}}}\right),
    \end{align}
    where
    \begin{align}\label{eq:varphi_h_i_balanced}
        \varphi_h^{(t)}(i)&= -\frac{a_{p,0}^{(t)}}{k}e^{a_{b,0}^{(t)}}+|\gA_{h,\times}|a_{p,2}^{(t)}e^{a_{b,1}^{(t)}}-\mathbbm{1}\{i\in\gA_{h,\times}\}\left((a_{p,1}^{(t)}+a_{p,2}^{(t)})(e^{a_{b,1}^{(t)}}-e^{a_{b,2}^{(t)}})\right)\notag\\
        &\quad+\left(N_h(\up)\frac{a_{p,\up}^{(t)}}{k}-N_h(i)a_{p,1}^{(t)}+(h-1-N_h(\up)-N_h(i)-|\gA_{h,\times}|)a_{p,2}^{(t)}\right)e^{a_{b,2}^{(t)}}
    \end{align}
    for all $i\in[k]$,
    \begin{align}\label{eq:xi_h_balanced}
        \xi_h^{(t)}&=\begin{cases}
            \frac{N_{h,\times}e^{-a_{c,0}^{(t)}}a_{q,\times}^{(t)}-(h-N_{h,\times})e^{a_{c,0}^{(t)}}a_{q,0}^{(t)}}{N_{h,\times}e^{-a_{c,0}^{(t)}}+(h-N_{h,\times})e^{a_{c,0}^{(t)}}}, & \text{if } c_h=0,\\
            \frac{N_{h,\times}e^{a_{c,1}^{(t)}}a_{q,\times}^{(t)}-(h-N_{h,\times})e^{-a_{c,1}^{(t)}}a_{q,0}^{(t)}}{N_{h,\times}e^{a_{c,1}^{(t)}}+(h-N_{h,\times})e^{-a_{c,1}^{(t)}}}, & \text{if } c_h=\times,
        \end{cases}
    \end{align}
    and $Z_h^{(t)}$ is the normalizing factor:
    \begin{align}\label{eq:Z_h_balanced}
        Z_h^{(t)}&\coloneqq \exp\left(\left(1+\frac{1}{k}\right)\xi_h^{(t)}\right)+\sum_{i=1}^{k} \exp\left(\frac{\varphi_h^{(t)}(i)}{e^{a_{b,0}^{(t)}}+|\gA_{h,\times}|e^{a_{b,1}^{(t)}}+(h-1-|\gA_{h,\times}|)e^{a_{b,2}^{(t)}}}\right).
    \end{align}
\end{lm}
The proof of Lemma~\ref{lm:gradients} is postponed to Appendix~\ref{sec:proof_gradients}.


\subsubsection{Depth-1 Curriculum Training}\label{sec:proof_depth_1_curriculum_training}

To analyze the convergence of Depth-1 curriculum training, we define the following time points:
\begin{align}
    T_{1/2}&\coloneqq \sup\left\{t\in \NN: a_{p,1}^{(t)}\leq 4.1k \ln k\right\}.\label{eq:T_1/2}
\end{align}
We divide depth-1 curriculum training into two sub-phases by $T_{1/2}$: Phase 1.1: $t\leq  T_{1/2}$ and Phase 1.2: $ T_{1/2}+1\leq t\leq T_1$.

\paragraph{Phase 1.1: $t\leq  T_{1/2}$.} Define
\begin{subequations}
\begin{align}
    T_{1,1}&\coloneqq \sup\left\{t\in \NN: a_{q,\times}^{(t)}\leq \ln k\right\},\label{eq:T_1_1}\\
    T_{1,2}&\coloneqq \sup\left\{t\in \NN: a_{p,1}^{(t)}\leq k\right\},\label{eq:T_1_2}
\end{align}
\end{subequations}
We will show the following dynamic. 
\begin{itemize}
\item Before $T_{1,1}$, matrix $A_P$ is very close to 0, and the agent cannot distinguish if an action is taken or not. During this time period, the agent learns how to go up at leaf nodes that are not goal nodes, and go down at the root. \item Between $T_{1,1}$ and $T_{1,2}$, $a_{p,1}$ rapidly increases to $k$, the agent has high confidence to go up at leaf nodes and go down at the root, but is still bad at distinguishing if an action is taken or not when the trajectory gets long. 
\item From $T_{1,2}$ to $ T_{1/2}$, $a_{p,1}$ increases to $4k \ln k$, the agent has high confidence to follow all rules at trees of depth 1, and by the end of this period, the agent has a success rate higher than $1-\frac{1}{k}$. 
\end{itemize}
During the whole Phase 1.1, matrix $A_B$ is very close to 0 and can be neglected.
We rigorously state the training dynamics of Phase 1.1 in the following lemma.
\begin{lm}\label{lm:phase_1.1}
 Under balanced goal distribution~\eqref{eq:balanced_goal_distribution}, Assumption~\ref{asmp:token_embedding}, for all $t\leq T_{1/2}$, for all $t'\leq t\leq T_{1/2}$, there exists an absolute constant $C>0$ such that
 \begin{subequations}
\begin{align}
    -\frac{\ln^2 k}{Nk}\lesssim a_{b,1}^{(t)} &\lesssim \frac{k^3\ln^2 k}{N}, \label{eq:delta_a_b_1_balanced_phase_1}\\
    -\frac{k^3\ln^2 k}{N}\lesssim a_{b,0}^{(t)} &\lesssim \frac{\ln^2 k}{N},\label{eq:delta_a_b_0_balanced_phase_1}\\
    \left|a_{b,2}^{(t)}\right| &\lesssim \frac{k^3\ln^2 k}{N^2},\label{eq:delta_a_b_2_balanced_phase_1}\\
    e^{2a^{(t)}_{c,0}}-e^{2a^{(t')}_{c,0}}&\geq 0 , \label{eq:delta_a_c_0_balanced_phase_1}\\
        e^{2a_{c,1}^{(t)}}-e^{2a_{c,1}^{(t')}}&\geq \max\left\{\left(a_{q,\times}^{(t')}+a_{q,0}^{(t')}\right)\left(a_{q,\times}^{(t)}-a_{q,\times}^{(t')}\right),0\right\},\label{eq:delta_a_c_1_balanced_phase_1}\\
        |a_{p,0}^{(t)}|&\lesssim \ln k,\quad |a_{p,\up}^{(t)}|\lesssim \ln k,\label{eq:a_p_0_up_bound_T_1}\\
         a_{p,1}^{(t)}-a_{p,1}^{(t')}&\geq0,\label{eq:a_p_1_bound_T_11}\\
        a_{q,0}^{(t)}&\lesssim a_{q,\times}^{(t)},\label{eq:a_q_0<=a_q_x_bound_T_11}\\
        a_{q,\times}^{(t)}+a_{q,0}^{(t)}&\asymp a_{q,\times}^{(t)},\label{eq:a_q_x+a_q_0_=a_q_x_T1}\\
        \left(a_{q,\times}^{(t)}+a_{q,0}^{(t)}\right)-\left(a_{q,\times}^{(t')}+a_{q,0}^{(t')}\right)&\geq 0, \label{eq:a_q_x+a_q_0_increase}\\
        -\left(a_{q,0}^{(t)}-a_{q,0}^{(t')}\right)&\lesssim \frac{1}{e^{2a_{c,1}^{(t')}}}\left(\left(a_{q,\times}^{(t)}+a_{q,0}^{(t)}\right)-\left(a_{q,\times}^{(t')}+a_{q,0}^{(t')}\right)\right),\label{eq:-a_q_0_bound_T_12}\\
        \xi_{1,2}^{(t)}&\lesssim a_{q,\times}^{(t)}.\label{eq:xi_1_2_bound_T_11}
    \end{align}
\end{subequations}
    Moreover, for all $t'\leq t\leq T_{1,1}$:
     \begin{subequations}
    \begin{align}
        a_{p,1}^{(t)}\lesssim \frac{\ln k}{k^2},\quad |a_{p,2}^{(t)}|&\lesssim \frac{\ln k}{k^2},\label{eq:a_p_1_2_tiny_T_11}\\
        a_{q,\times}^{(t)}-a_{q,\times}^{(t')}&\gtrsim \frac{\eta(t-t')}{k^2},\label{eq:a_q_x_bound_T_11}\\
        \left(a_{q,\times}^{(t)}+a_{q,0}^{(t)}\right)-\left(a_{q,\times}^{(t')}+a_{q,0}^{(t')}\right)&\asymp a_{q,\times}^{(t)}-a_{q,\times}^{(t')},\label{eq:a_q_x+a_q_0_=a_q_T11}\\
        a_{q,0}^{(t)}-a_{q,0}^{(t')}&\leq \frac{2}{3}\left(a_{q,\times}^{(t)}-a_{q,\times}^{(t')}\right)+C.\label{eq:a_q_0<=2/3a_q_x_T11}
    \end{align}
    \end{subequations}
For any $t',t$ such that $T_{1,1}+1\leq t'\leq t\leq T_{1/2}$, we have
\begin{align}
    a_{p,1}^{(t)}-a_{p,1}^{(t')}&\gtrsim \frac{\eta(t-t')}{k^5},\label{eq:a_p_1_bound_T_12}\\
    e^{2a_{c,1}^{(t)}}&\gtrsim \ln^2 k.\label{eq:e2a_c_1_bound_T_12}
\end{align}
For any $t',t$ such that $T_{1,1}+1\leq t'\leq t\leq T_{1,2}$, we have 
\begin{subequations}
    \begin{align}
        -\frac{\ln k}{k^2}\lesssim a_{p,2}^{(t)} &\leq 1+O\left(\frac{\ln k}{k^2}\right),\label{eq:a_p_2<=1_T_12}\\
        e^{2a_{c,0}^{(t)}}-e^{2a_{c,0}^{(t')}}&\gtrsim \left(a_{q,\times}^{(t')}+a_{q,0}^{(t')}\right)\left(a_{q,0}^{(t)}-a_{q,0}^{(t')}\right),\label{eq:delta_e2a_c_0_bound_T_12}\\
    0\leq a_{q,\times}^{(t)}-a_{q,0}^{(t)}&\leq 2.1\ln k,\label{eq:a_q_x-a_q_0_bound_T_12}\\
    a_{q,0}^{(t)}+a_{q,\times}^{(t)}&\asymp \ln k.\label{eq:a_q_x+a_q_0_=lnk_T_12}
    \end{align}
        \end{subequations}
For any $t',t$ such that $T_{1,2}+1\leq t'\leq t\leq T_{1/2}$, we have
\begin{subequations}
\begin{align}
    -C\leq a_{q,\times}^{(t)}-a_{q,0}^{(t)}&\leq 6.2\ln k,\label{eq:a_q_x-a_q_0_bound>T_12}\\
    a_{q,0}^{(t)}&\geq 1.9\ln k-1,\label{eq:a_q_0>=1.9lnk_T_12}\\
    a_{q,\times}^{(t)}-\frac{1}{2}a_{p,2}^{(t)}&\geq 2.425\ln k,\label{eq:a_q_x-a_p_2_bound_T_12}\\
    a_{p,2}^{(t)}&=\frac{1+o(1)}{k}a_{p,1}^{(t)},\label{eq:a_p_2_bound_T_12}\\
    e^{2a_{c,0}^{(t)}}&\asymp \ln^2 k,\label{eq:e2a_c_0>=lnk^2_T_12}\\
    5.9\ln k\leq a_{q,0}^{(t)}+a_{q,\times}^{(t)}&\leq 20.4\ln k.\label{eq:a_q_0+a_q_x=C_lnk_phase_1}
\end{align}
\end{subequations}
When $t=T_{1/2}$, we have
\begin{subequations}
\begin{align}
    a_{b,1}^{(T_{1/2})}&\asymp \frac{k^3\ln^2 k}{N},\label{eq:a_b_1_phase_1_end}\\
    -\frac{k^3\ln^2 k}{N}\lesssim a_{b,0}^{(T_{1/2})} &\lesssim -\frac{k^2\ln^2 k}{N},\label{eq:a_b_0_phase_1_end}\\
   \frac{1}{k^{3.05}}\lesssim 1-R\left(\pi^{(T_{1/2})};\gP_1\right)&\lesssim \frac{1}{k^{1.01}},\label{eq:1-R_pi_T_1_2_phase_1_end}\\
    \forall 2\leq l\leq L:\quad R\left(\pi^{(T_{1/2})};\gP_l\right)&\lesssim \frac{1}{k},\label{eq:R_pi_T_1_l_phase_1_end}\\
    a_{q,\times}^{(T_{1/2})}-\frac{1}{2}a_{p,2}^{(T_{1/2})}&\geq 3.9 \ln k.\label{eq:a_q_0-a_p_2_phase_1_end}
\end{align}
\end{subequations}
\end{lm}   
The proof of Lemma~\ref{lm:phase_1.1} is postponed to Appendix~\ref{sec_app:phase_1.1}. As a consequence of Lemma~\ref{lm:phase_1.1}, we have
\begin{align}\label{eq:T_1/2_bound}
    T_{1/2}\lesssim \frac{k^6\ln k}{\eta}.
\end{align}

\paragraph{Phase 1.2: $T_{1/2}+1\leq t\leq T_{1}$.} 
At Phase 1.2, the effect of matrix $A_B$ becomes non-negligible. Specifically, its main diagonal $a_{b,1}$ will monotonically increase, $a_{b,0}$ is negative at first, and then becomes positive if $\epsilon$ is small enough (smaller than $\frac{1}{k^{9/4}N\exp\left( k^{1.3}\sqrt{N}\right)}$), and $a_{b,2}$ stays small. Same as in Phase 1.1, $a_{c,0}$ and $a_{c,1}$ will monotonically increase, but $a_{c,0}$'s increase becomes much slower and $e^{2a_{c,0}}$ stays on the order $\ln^2 k$ after Phase 1.1. $a_{p,1}$ will monotonically increase, $a_{p,2}\approx \frac{a_{p,1}}{k-1}$ after Phase 1.1, and $a_{p,0},a_{p,\up}$ are no larger than $a_{p,2}$ in magnitude. $a_{q,0}$, $a_{q,\times}$ increase overall, and has the same magnitude as $a_{p,2}$.
 The failure probability will keep decreasing until it's bounded by $\epsilon^{1.05}$ at step $T_1$.

To formalize the above description, we define 
\begin{subequations}
\begin{align}
    T_{1,3}&\coloneqq \sup\left\{t\in \NN: a_{p,1}^{(t)}\leq 21k\ln k\right\},\label{eq:T_1_3}\\
    T_{1,4}&\coloneqq \inf\left\{t\in \NN: e^{a_{b,0}^{(t)}}\geq \frac{k}{\ln k}\right\},\label{eq:T_1_4}
\end{align}
\end{subequations}
and for all $h\in[k]$, define
\begin{subequations}
\begin{align}\label{eq:mu_h_phase1.2}
    \mu_{2h-1}^{(t)}&\coloneqq \frac{-\frac{a_{p,0}^{(t)}}{k}e^{a_{b,0}^{(t)}}+\frac{h-1}{k}a_{p,\up}^{(t)}e^{a_{b,2}^{(t)}}+(h-1)a_{p,2}^{(t)}e^{a_{b,1}^{(t)}}}{e^{a_{b,0}^{(t)}}+(h-1)e^{a_{b,1}^{(t)}}+(h-1)e^{a_{b,2}^{(t)}}}, \\
    \mu_{2h}^{(t)}&\coloneqq \frac{-\frac{a_{p,0}^{(t)}}{k}e^{a_{b,0}^{(t)}}+\frac{h-1}{k}a_{p,\up}^{(t)}e^{a_{b,2}^{(t)}}+ha_{p,2}^{(t)}e^{a_{b,2}^{(t)}}}{e^{a_{b,0}^{(t)}}+(2h-1)e^{a_{b,2}^{(t)}}},
\end{align}
\end{subequations}
and 
\begin{align}\label{eq:r_down_times_phase1.2}
    r_{\down_\times}^{(t)}\coloneqq \frac{e^{a_{b,1}^{(t)}}}{e^{a_{b,0}^{(t)}}+(k-1)e^{a_{b,1}^{(t)}}+(k-1)e^{a_{b,2}^{(t)}}}.
\end{align}
We state the training dynamics of Phase 1.2 in Lemma~\ref{lm:phase_1.2}. 
\begin{lm}\label{lm:phase_1.2}
    Under balanced goal distribution~\eqref{eq:balanced_goal_distribution}, Assumption~\ref{asmp:token_embedding}, for all $t,t'\in[T_{1/2},T_1]$ with $t'\leq t$, there exists an absolute constant $C>0$ such that we have
\begin{subequations}
\begin{align}
    e^{a_{b,1}^{(t)}}-e^{a_{b,1}^{(t')}}&\geq0,\label{eq:delta_ea_b_1_bound_phase_1.2}\\
    a_{q,0}^{(t)}&\asymp a_{q,\times}^{(t)}\asymp a_{p,2}^{(t)},\label{eq:a_q_x_q_0_p_2_asymp_phase_1.2}\\
    e^{2a^{(t)}_{c,0}}-e^{2a^{(t')}_{c,0}}&\geq 0, \quad e^{2a^{(t)}_{c,0}}\asymp \ln^2 k, \label{eq:e2a_c_0_balanced_phase_1.2}\\
    e^{2a_{c,1}^{(t)}}-e^{2a_{c,1}^{(t')}}&\geq \max\left\{(2+o(1))\left(a_{q,\times}^{(t')}+a_{q,0}^{(t')}\right)\left(a_{q,\times}^{(t)}-a_{q,\times}^{(t')}\right),0\right\},\label{eq:e2a_c_1_bound_phase_1.2}\\
    a_{p,1}^{(t)}-a_{p,1}^{(t')}&\geq 0,\label{eq:a_p_1_increase_phase_1.2}\\
    \left|a_{p,0}^{(t)}-a_{p,0}^{(t')}\right|&\lesssim \left(a_{q,\times}^{(t)}+a_{q,0}^{(t)}\right)-\left(a_{q,\times}^{(t')}+a_{q,0}^{(t')}\right),\label{eq:a_p_0_bound_phase_1.2}\\
    \left|a_{p,\up}^{(t)}-a_{p,\up}^{(t')}\right|&\lesssim \left(a_{q,\times}^{(t)}+a_{q,0}^{(t)}\right)-\left(a_{q,\times}^{(t')}+a_{q,0}^{(t')}\right),\label{eq:a_p_up_bound_phase_1.2}\\
    a_{p,2}^{(t)}&=\left(1+O\left(1/k^2\right)\right)\frac{a_{p,1}^{(t)}}{k-1}.\label{eq:a_p_2_bound_phase_1.2}
\end{align}
\end{subequations}
For $t,t'\in[T_{1/2}+1,T_{1,3}]$, $t'\leq t$, we have
\begin{subequations}
\begin{align}
    a_{q,\times}^{(t)}-\frac{1}{2}a_{p,2}^{(t)}&\geq 3.9 \ln k.\label{eq:a_q_x-a_p_2_bound_<T_1_3}\\
    a_{q,0}^{(t)}&\geq 1.9\ln k -2,\label{eq:a_q_0_bound_<T_1_3}\\
    a_{p,1}^{(t)}-a_{p,1}^{(t')}&\gtrsim \frac{\eta(t-t')}{k^{12.5}},\label{eq:a_p_1_bound_<T_1_3}\\
    a_{b,1}^{(t)}&\asymp \frac{k^3\ln^2 k}{N},\label{eq:a_b_1_bound_<T_1_3}\\
    -\frac{k^3\ln^2 k}{N}\lesssim a_{b,0}^{(t)} &\lesssim -\frac{k^2\ln^2 k}{N}.\label{eq:a_b_0_bound_<T_1_3}
\end{align}
\end{subequations}
For $t\in [T_{1,3}+1,T_{1,4}]$, $t'\leq t$, we have
\begin{subequations}
\begin{align}
    a_{p,1}^{(t)}\leq k^{9/4}\sqrt{{N}},\label{eq:a_p_1_ub_T_1_4}\\   
    \min\left\{a_{q,\times}^{(t)}-\mu_2^{(t)}-\ln k,a_{q,\times}^{(t)}-\mu_{2(k-1)}^{(t)}+\ln k\right\}+\frac{2}{k}a_{q,0}^{(t)}
    &\geq r_{\down_\times}^{(t)}\left(a_{p,1}^{(t)}+a_{p,2}^{(t)}\right)+\ln k,\label{eq:a_q_x>a_p_2_bound_T_13}\\
    r_{\down_\times}^{(t)}\left(a_{p,1}^{(t)}+a_{p,2}^{(t)}\right)+0.1\sqrt{a_{p,1}^{(t)}/k}+5\ln k &\geq \min\left\{a_{q,\times}^{(t)}-\mu_2^{(t)},a_{q,\times}^{(t)}-\mu_{2(k-1)}^{(t)}\right\},\label{eq:a_q_2>a_q_x_bound_T_13}\\
    r_{\down_\times}^{(t)}\left(a_{p,1}^{(t)}+a_{p,2}^{(t)}\right)+2.1\ln k &\geq \left(1+\frac{1}{k}\right)a_{q,0}^{(t)}-\frac{a_{p,0}^{(t)}}{k},\label{eq:a_p2>a_q_0_bound_T_13}\\
    \left(1+\frac{1}{k}\right)a_{q,0}^{(t)}-\frac{a_{p,0}^{(t)}}{k}+0.1\sqrt{a_{p,1}^{(t)}/k}+3\ln k&\geq  \min\left\{a_{q,\times}^{(t)}-\mu_2^{(t)},a_{q,\times}^{(t)}-\mu_{2(k-1)}^{(t)}\right\},\label{eq:a_q_0>a_q_x_bound_T_13}\\
    e^{a_{b,1}^{(t)}}-1&\gtrsim \frac{k\left(a_{p,1}^{(t)}\right)^2}{N}\gtrsim k^2\left(e^{a_{b,0}^{(t)}}-1\right),
    \label{eq:ea_b_2_bound_T_14}\\
    e^{a_{b,0}^{(t)}}&\gtrsim \frac{1}{\sqrt{k}},\label{eq:e2a_b_0_lb_T_14}\\
    e^{a_{b,1}^{(t)}}+\frac{1}{k^8 N}a_{q,\times}^{(t)}-\left(e^{a_{b,1}^{(t')}}+\frac{1}{k^8 N}a_{q,\times}^{(t')}\right)&\geq \frac{1+o(1)}{N}\frac{k a_{p,1}^{(t')}}{3}\left(a_{p,1}^{(t)}-a_{p,1}^{(t')}\right),\label{eq:ea_b_1_bound_T_14}\\
    e^{a_{b,1}^{(t)}}&\lesssim \frac{k^{6.5}}{\ln k},\label{eq:ea_b_1_ub_T_14}\\
    a_{p,1}^{(t)}-a_{p,1}^{(t')}&\gtrsim \frac{\eta(t-t')}{\exp\left(k^{1.3}\sqrt{N}\right)},\label{eq:a_p_1_increase_rate_T_13-T_14}
\end{align}
\end{subequations}
For $t\in [T_{1,4}+1,T_1]$, we have
\begin{subequations}
\begin{align}
    e^{a_{b,0}^{(t)}}&\gtrsim \frac{k}{\ln k},\label{eq:ea_b_0>=k/lnk_T_15}\\
    a_{p,1}^{(t)}&\gtrsim \frac{k^{2}\sqrt{N}}{\ln k},\label{eq:a_p_1_lb_T_1_5}\\
    e^{a_{b,1}^{(t)}}&\gtrsim \frac{k^{5}}{\ln^2 k},\label{eq:ea_b_1_bound_T_15}\\
    \min\left\{a_{q,\times}^{(t)}-\mu_2^{(t)}-\ln k,a_{q,\times}^{(t)}-\mu_{2(k-1)}^{(t)}+\ln k\right\}+\frac{2}{k}a_{q,0}^{(t)}
    &\geq r_{\down_\times}^{(t)}\left(a_{p,1}^{(t)}+a_{p,2}^{(t)}\right)-0.1\sqrt{a_{p,1}^{(t)}/k}+\ln k,\label{eq:a_q_x>a_p_2_bound_T_15}\\
    \frac{1}{k-1}\left(a_{p,1}^{(t)}+a_{p,2}^{(t)}\right)+2.1\ln k &\geq \left(1+\frac{1}{k}\right)a_{q,0}^{(t)}-\frac{a_{p,0}^{(t)}}{k},\label{eq:a_p2>a_q_0_bound_T_15}\\
    \left(1+\frac{1}{k}\right)a_{q,0}^{(t)}-\frac{a_{p,0}^{(t)}}{k}+0.1\sqrt{a_{p,1}^{(t)}/k}+3\ln k&\geq  \min\left\{a_{q,\times}^{(t)}-\mu_2^{(t)},a_{q,\times}^{(t)}-\mu_{2(k-1)}^{(t)}\right\},\label{eq:a_q_0>a_q_x_bound_T_15}\\
    e^{\frac{N}{N-1}a_{b,1}^{(t)}}&\gtrsim \frac{k\left(a_{p,1}^{(t)}\right)^2}{N}\gtrsim k^2e^{\frac{N}{N-1}a_{b,0}^{(t)}},\label{eq:ea_b_1>=ea_b_0_T_15}\\
    \text{If }a_{p,1}^{(t)}\geq Nk^3:\quad e^{a_{b,0}^{(t)}}&\gtrsim \frac{\left(a_{p,1}^{(t)}\right)^{1-\frac{6}{N-1}}}{N k},\label{eq:a_b_0>=k^2_T_15}\\
    (1+o(1))\left(a_{p,1}^{(t)}\right)^{-\frac{6}{N-1}}\leq e^{a_{b,2}^{(t)}}&<1.\label{eq:e_ab_2_lb>T14}
\end{align}
\end{subequations}
When $t=T_1$, we have 
\begin{subequations}
\begin{align}
    a_{p,1}^{(T_1)}&\asymp k\ln(1/\epsilon),\label{eq:a_p_1_bound_T_1}\\
    \left(k-1+o\left(\frac{1}{\ln(1/\epsilon)}\right)\right)\ln\left(\frac{1}{k\epsilon^{1.05}}-(k-1)\right)&\leq a_{p,1}^{(T_1)}+a_{p,2}^{(T_1)}\leq (k+C)\ln (k^{2}/\epsilon^{1.05}),\label{eq:a_p_1_2_bound_T_1}\\
    \left(1+\frac{1}{k}\right)a_{q,0}^{(T_1)}-\frac{a_{p,0}^{(T_1)}}{k}&\geq \ln\left(\frac{1}{k\epsilon^{1.05}}-\frac{1}{k}\right)\label{eq:a_q_0_bound_T_1}\\
    \left(1+\frac{1}{k}\right)a_{q,\times}^{(T_1)}+\frac{a_{p,0}^{(T_1)}}{k}&\geq \ln\left(\frac{(k-1)^2}{k\epsilon^{1.05}}-(k-1)\right)+\frac{C}{\ln(1/\epsilon)},\label{eq:a_q_x_bound_T_1}\\
    a_{c,1}^{(T_1)}&\gtrsim \ln^2\left(1/\epsilon\right)\label{eq:a_c_1_bound_T_1}\\
    e^{a_{b,0}^{(T_1)}}&\gtrsim \frac{1}{N}\left(\ln(1/\epsilon)\right)^{1-\frac{6}{N-1}},\label{eq:e_ab_0_lb_T_1}\\
    e^{a_{b,1}^{(T_1)}}\gtrsim \frac{k^2}{N^2}\left(\ln(1/\epsilon)\right)^{3-\frac{6}{N-1}}
    &\gtrsim \frac{k^2}{N}\left(\ln(1/\epsilon)\right)^{1-\frac{5}{N}}e^{a_{b,0}^{(T_1)}},\label{eq:ea_b_1_lb_T_1}\\
    (1+o(1))\left(\ln(1/\epsilon)\right)^{-\frac{6}{N-1}}\lesssim e^{a_{b,2}^{(T_1)}}&\leq \min\left\{1,(1+o(1))\left(\ln(1/\epsilon)\right)^{-\frac{4}{N-1}+\frac{12}{N(N-1)}}\right\},\label{eq:e_ab_2_lb_T_1}\\
    \forall 2\leq l\leq L:\quad R\left(\pi^{(T_1)};\gP_l\right)&\leq \frac{1+o(1)}{k^{l-1}},\label{eq:win_rate_l>1}
\end{align}
\end{subequations}
and
\begin{align}\label{eq:T_1_bound}
    T_1\lesssim \frac{1}{\eta\epsilon^{1.1}}.
\end{align}
\end{lm}
The proof of Lemma~\ref{lm:phase_1.2} is postponed to Appendix~\ref{sec_app:phase_1.2}. The statement tracks the most stringent precision regime in Theorem~\ref{thm:convergence_balanced}. For a larger target precision $\epsilon$, the depth-1 success criterion may already be met before the later Phase 1.2 thresholds, such as $T_{1,4}$, are reached. This is the regime reflected in Figure~\ref{fig:entries_balanced} and Figure~\ref{fig:matrix_evolution}: training can stop while $a_{b,0}$ is still slightly negative or close to zero, even though the analysis would make it positive after sufficiently long depth-1 training. The resulting policy still learns the same search mechanism. It simply realizes it with a slightly different finite-precision balance between $A_B$ and $A_P$; in particular, during depth-2 curriculum training, $A_P$ can converge to adjusted values, with $a_{p,\up}$ growing to the scale needed for backtracking in the search policy.

\subsubsection{Depth-2 curriculum training}\label{sec:phase_2} 
The main reason of poor depth generalization after depth-1 curriculum training is that $a_{p,2}$ is too large. From the proof of \eqref{eq:win_rate_l>1} in Lemma~\ref{lm:phase_1.2} we can see that this causes the agent not to go up with high probability at a non-leaf and non-root node, after all its children have been visited.
We'll show that at depth-2 curriculum training, $a_{p,1}$ keeps monotonically increasing, $a_{p,2}=\frac{a_{p,1}-a_{p,3}}{k-1}$ (c.f.~\eqref{eq:delta_a_p_3_simplified_balanced}) keeps decreasing until it has the right proportion to $a_{p,1}$. The changes of $a_{p,0}$ and $a_{p,\up}$ are relatively small during depth-2 curriculum training.
Below we formally state the training dynamics of depth-2 curriculum training in Lemma~\ref{lm:phase_2}. Recall we fix $B,C,Q$ to be $B^{(T_1)},C^{(T_1)},Q^{(T_1)}$ after depth-1 curriculum training and only train $\overline P$ starting from $\overline P^{(T_1)}$.
\begin{lm}\label{lm:phase_2} 
    Under balanced goal distribution~\eqref{eq:balanced_goal_distribution}, Assumption~\ref{asmp:token_embedding}, for all $t,t'\in[T_1,T_2]$ with $t'\leq t$, we have
    \begin{subequations}
    \begin{align}
        |a_{p,0}^{(t)}-a_{p,0}^{(T_1)}|& \lesssim \frac{a_{q,0}}{\left(\ln (1/\epsilon)\right)^{1+o(1)}},\label{eq:a_p_0_bound_phase2}\\
        |a_{p,\up}^{(t)}-a_{p,\up}^{(T_1)}| & \lesssim \frac{a_{q,0}}{\left(\ln (1/\epsilon)\right)^{3+o(1)}},\label{eq:a_p_up_bound_phase2}\\
        a_{p,1}^{(t)}-a_{p,1}^{(t')} & \gtrsim \frac{\epsilon^{1.1}}{k^L}(\eta(t-t')),\label{eq:a_p_1_increase_rate_phase2}\\
        -\frac{a_{q,0}}{k}\lesssim a_{p,3}^{(t)} &\lesssim k a_{q,0},\label{eq:a_p_3_vs_a_q_0_phase2}\\
        a_{p,1}^{(t)} & \asymp k a_{q,0}.\label{eq:a_p_1_vs_a_q_0_phase2}
    \end{align}
    \end{subequations}
And there exists an absolute constant $C\in \RR$ such that for 
\begin{align}\label{eq:T_2_1_definition}
    T_{2,1}\coloneqq \inf_{t}\left\{t\in[T_1+1,T_2]:a_{p,3}^{(t)}\geq\frac{a_{p,1}^{(t)}}{2}+k\left(1+\frac{C}{\ln^2 k}\right)a_{q,0}\right\},
\end{align}
for all $t,t'\in[T_{1},T_{2,1}]$ with $t'\leq t$:
\begin{align}\label{eq:a_p_3_increase_rate<T_2_1}
    a_{p,3}^{(t)}-a_{p,3}^{(t')}\geq (1+o(1))k \left(a_{p,1}^{(t)}-a_{p,1}^{(t')}\right),
\end{align}
and for all $t\in[T_{2,1}+1,T_2]$:
\begin{subequations}
\begin{align}\label{eq:a_p_3_bound_phase2}
    a_{p,3}^{(t)}=\frac{a_{p,1}^{(t)}}{2}+k\left(1+O\left(\frac{1}{\ln^2 k}\right)\right)a_{q,0}.
\end{align}
When $t=T_2$, there exists a constant $c_1,c_2,c_3,c_4\in \RR$ such that
\begin{align}
    \frac{a_{p,3}^{(T_2)}}{k}-\left(1-\frac{c_1}{\ln^2 k}\right)a_{q,0}&\geq \ln\left(\frac{k^{L}}{\epsilon^{1.02}}\right),\label{eq:a_p_3_bound_phase2_T1}\\
    \frac{a_{p,1}^{(T_2)}}{k}&\geq 2\ln\left(\frac{k^L}{\epsilon^{1.02}}\right)+\frac{c_2}{\ln^2 k}a_{q,0},\label{eq:a_p_1_bound_phase2_T1}\\
    \frac{a_{p,3}^{(T_2)}}{k}-\left(1-\frac{c_3}{\ln^2 k}\right)a_{q,0}&\leq \ln \left(\frac{k^L}{\epsilon^{1.02+o(1)}}\right),\label{eq:a_p_3_ub_T2}\\
    \frac{a_{p,1}^{(T_2)}}{k}&\leq 2\ln \left(\frac{k^L}{\epsilon^{1.02+o(1)}}\right)+\frac{c_4}{\ln^2 k}a_{q,0}.\label{eq:a_p_1_ub_T2}
\end{align}
\end{subequations}
Furthermore, we have 
\begin{align}
    T_2-T_1\lesssim \frac{k^L}{\eta\epsilon^{1.1}}.\label{eq:T_2_bound}
\end{align}
For any full $k$-ary tree $\gT$ (that may not be a perfect $k$-ary tree) with depth $l\in[L]$ and any goal node $g(\gT)$, 
\begin{align}\label{eq:success_rate_generalization_phase2}
    R(\pi^{(T_2)},(\gT,g(\gT)))\geq 1-\epsilon.
\end{align}
\end{lm}
The proof of Lemma~\ref{lm:phase_2} is postponed to Appendix~\ref{sec_app:phase_2}. Theorem~\ref{thm:convergence_balanced} follows directly from Lemma~\ref{lm:phase_1.1}, Lemma~\ref{lm:phase_1.2} and Lemma~\ref{lm:phase_2}.


\subsection{Proof of Key Lemmas}\label{sec:proof_convergence_key_lemmas}

\subsubsection{Proof of Lemma~\ref{lm:gradients}}\label{sec:proof_gradients}

In the proof, for notation simplicity, we may drop superscript $t$ when it's clear from the context.

\paragraph{Step 1: compute $\delta B$, $\delta C$, $\delta P$, $\delta Q$.}
We first compute $\delta B$, $\delta C$, $\delta P$, $\delta Q$ by chain rule.
Note that 
    $ d\phi_h = (dP)E_a^{(h)}\alpha_h^n$,
    and thus
    $$d\log\pi_h(a_h) = \delta_h^\top d\phi_h = \delta_h^\top (dP)E_a^{(h)}\alpha_h^n=\tr\left(\delta_h(dP) E_a^{(h)}\alpha_h^n\right)=\tr\left((dP) E_a^{(h)}\alpha_h^n \delta_h^\top\right).$$
    Therefore, we have
    \begin{align}
        \nabla_P \log\pi_h(a_h) = \delta_h \left(E_a^{(h)}\alpha_h^n\right)^\top,
    \end{align}
    and thus by \eqref{eq:policy_gradient_update}, we have
    \begin{align}\label{eq:delta_P}
        \delta P = \E\left[\sum_{h=1}^H \widehat{G}_h\delta_h \left(E_a^{(h)}\alpha_h^n\right)^\top\right],
    \end{align}
where $\widehat{G}_h$ is defined in \eqref{eq:pi_G_Q_h}. By the same argument, we have
    \begin{align}\label{eq:delta_Q}
        \delta Q = \E\left[\sum_{h=1}^H \widehat{G}_h\delta_h \left(E_c^{(h)}\alpha_h^c\right)^\top\right].
    \end{align}

Let the softmax Jacobians be
    \begin{align*}
        J_h^n\coloneqq \diag(\alpha_h^n)-\alpha_h^n(\alpha_h^n)^\top,\quad J_h^c\coloneqq \diag(\alpha_h^c)-\alpha_h^c(\alpha_h^c)^\top,
        \end{align*}
then we have
\begin{align*}
    d\log\pi_h(a_h)=\delta_h^\top d\phi_h=\delta_h^\top PE_a^{(h)}J_h^n\left(E_n^{(h)} \right)^\top (dB)\overline{n}_h=\tr\left(\overline{n}_h\delta_h^\top PE_a^{(h)}J_h^n\left(E_n^{(h)} \right)^\top (dB)\right).
\end{align*}
Therefore, we have
\begin{align}
    \nabla_B \log\pi_h(a_h) = E_n^{(h)} J_h^n\left(E_a^{(h)}\right)^\top P^\top \delta_h \overline{n}_h^\top,
\end{align}
and thus by \eqref{eq:policy_gradient_update}, we have
\begin{align}\label{eq:delta_B}
    \delta B = \E\left[\sum_{h=1}^H \widehat{G}_h E_n^{(h)} J_h^n\left(E_a^{(h)}\right)^\top P^\top \delta_h \overline{n}_h^\top\right].
\end{align}
By the same argument, we have
\begin{align}\label{eq:delta_C}
    \delta C = \E\left[\sum_{h=1}^H \widehat{G}_h E_c^{(h)} J_h^c\left(E_c^{(h)}\right)^\top Q^\top \delta_h \overline{c}_h^\top\right].
\end{align}

\paragraph{Step 2: compute $\delta A_B$, $\delta A_C$, $\delta A_P$, $\delta A_Q$.} 
We write the per-step selector factorizations
\begin{align}\label{eq:selector_factorizations}
    E_n^{(h)}=U S_n^{(h)},\quad E_a^{(h)}=V S_a^{(h)},\quad E_c^{(h)}=Z S_c^{(h)},
\end{align}
where each column of $S_n^{(h)}$, $S_a^{(h)}$, $S_c^{(h)}$ is a one-hot selector vector that has 1 at the position of the corresponding node, action, or label and 0 otherwise.
Then we have
\begin{align} 
    \delta A_P & =\delta PV \overset{\eqref{eq:delta_P}}= \E\left[\sum_{h=1}^H \widehat{G}_h\delta_h \left(E_a^{(h)}\alpha_h^n\right)^\top\right]V\overset{\eqref{eq:selector_factorizations}}= \E\left[\sum_{h=1}^H \widehat{G}_h\delta_h \left(\alpha_h^n\right)^\top \left(S_a^{(h)}\right)^\top\right],  \label{eq:delta_A_P} \\
    \delta A_Q & =\delta QZ \overset{\eqref{eq:delta_Q}}= \E\left[\sum_{h=1}^H \widehat{G}_h\delta_h \left(E_c^{(h)}\alpha_h^c\right)^\top\right]Z\overset{\eqref{eq:selector_factorizations}}= \E\left[\sum_{h=1}^H \widehat{G}_h\delta_h \left(\alpha_h^c\right)^\top \left(S_c^{(h)}\right)^\top\right], \label{eq:delta_A_Q}  \\
    \delta A_B & =U^\top \delta B U \overset{\eqref{eq:delta_B}}= U^\top \E\left[\sum_{h=1}^H \widehat{G}_h E_n^{(h)} J_h^n\left(E_a^{(h)}\right)^\top P^\top \delta_h \overline{n}_h^\top\right]U \overset{\eqref{eq:selector_factorizations}}= \E\left[\sum_{h=1}^H \widehat{G}_h S_n^{(h)}J_h^n \left(S_a^{(h)}\right)^\top A_P^\top \delta_h e_{n_h}^\top\right], \label{eq:delta_A_B} \\
    \delta A_C& =Z^\top \delta C Z \overset{\eqref{eq:delta_C}}= Z^\top \E\left[\sum_{h=1}^H \widehat{G}_h E_c^{(h)} J_h^c\left(E_c^{(h)}\right)^\top Q^\top \delta_h \overline{c}_h^\top\right]Z 
   \overset{\eqref{eq:selector_factorizations}}= \E\left[\sum_{h=1}^H \widehat{G}_h S_c^{(h)}J_h^c \left(S_c^{(h)}\right)^\top A_Q^\top \delta_h e_{c_h}^\top\right], \label{eq:delta_A_C}
\end{align}
where $e_{n_h}\in\R^{N+1}$ and $e_{c_h}\in\R^3$ are the one-hot vectors that have 1 at the position of $n_h$ and $c_h$ respectively and 0 otherwise.

We next simplify the matrix expressions and their gradients by exploiting the specific structure of the gradient matrices. To do so, we first introduce some notation. We define the (pre-softmax) logits as
\begin{align}\label{eq:logits_phi}
  \phi_h\coloneqq P E_a^{(h)}\alpha_h^n + Q E_c^{(h)} \alpha_h^c
\end{align}
with
\begin{align}\label{eq:alpha_h_n_c}
    \alpha_h^n\coloneqq \sm\left(\left(E_n^{(h)}\right)^\top B \overline{n}_h\right), \quad \alpha_h^c\coloneqq \sm\left(\left(E_c^{(h)}\right)^\top C \overline{c}_h\right).
\end{align}
Then by \eqref{eq:pi_theta_simplified}, we have
\begin{align}
\pi_h =\sm(\phi_h) = \sm\left(P E_a^{(h)}\alpha_h^n + Q E_c^{(h)} \alpha_h^c\right).
\end{align}
We also define 
\begin{align}
    \delta_h\coloneqq e_{a_h}-\pi_h,
\end{align}
where $e_{a_h}$ is the one-hot vector that has 1 at the position of $a_h$ and 0 otherwise.

To better align with the node/action/label numbering, we let $A_B,A_C$ be 0-indexed, $A_P,A_Q$'s column indices be 0-indexed, and $A_P,A_Q$'s row indices be 1-indexed, and we let $\times$ and $\checkmark$ denote index 1 and 2 respectively, e.g., $A_C(0,\times)=A_C(0,1)$, $e_\times=e_1$. We also let $\up$ denote index $k+1$. For each $h\in[H]$, we define
\begin{align}\label{eq:p_c_x,p_c_0}
    p_{h,\times}^c\coloneqq \sum_{i=1}^h\alpha_h^c(i)\mathbbm{1}\{c_i=\times\},\quad p_{h,0}^c\coloneqq \sum_{i=1}^h\alpha_h^c(i)\mathbbm{1}\{c_i=0\}.
\end{align}
Then since $c_h\in\{0, \times\}$ for all $h\in[H]$, we have
\begin{align}\label{eq:p_c}
    \forall h\in[H]:\quad p_{h,\times}^c + p_{h,0}^c = 1.
\end{align}
Analogously, we define
\begin{subequations}
\begin{align}\label{eq:p_n}
    p_{h,j}^a&\coloneqq \sum_{i=1}^h\alpha_h^n(i)\mathbbm{1}\{a_{i-1}=j\},\,\,\forall j\in\{0,1,\cdots,k+1\}, \\
    p_{h,j}^n&\coloneqq \sum_{i=1}^h\alpha_h^n(i)\mathbbm{1}\{n_{i-1}=j\},\,\,\forall j\in\{0,1,\cdots,N\},
\end{align}
\end{subequations}
which satisfies
\begin{align}\label{eq:p_n_sum}
    \forall h\in[H]:\quad \sum_{j=0}^{k+1} p_{h,j}^a = 1,\quad \sum_{j=0}^{N} p_{h,j}^n = 1.
\end{align}

First, since once goal is found, i.e., $\checkmark$ is observed, the episode terminates and the gradient matrices are not updated anymore. Therefore, the input matrix $E_c^{(h)}$ won't contain any $z_2$ (the embedding of label $\checkmark$) for any $h\in[H]$. Thus by \eqref{eq:delta_A_C} and our zero initialization, we have
\begin{align}\label{eq:A_C_0}
    \forall t\in\NN:\quad\delta A_C^{(t)}(\checkmark,:)=\delta A_C^{(t)}(:,\checkmark)=0\quad\text{and}\quad A_C^{(t)}(\checkmark,:)=A_C^{(t)}(:,\checkmark)=0.
\end{align}
For the same reason, by \eqref{eq:delta_A_Q}, we have
\begin{align}\label{eq:A_Q_0}
    \forall t\in\NN:\quad\delta A_Q^{(t)}(:,\checkmark)=0\quad\text{and}\quad A_Q^{(t)}(:,\checkmark)=0.
\end{align}
Further, since $n_h\neq 0$ for all $h\in[H]$,  by \eqref{eq:delta_A_B}, we have
\begin{align}\label{eq:A_B_0}
    \forall t\in\NN:\quad\delta A_B^{(t)}(:,0)=0\quad\text{and}\quad A_B^{(t)}(:,0)=0.
\end{align}

Moreover, by \eqref{eq:delta_A_C}, we have
\begin{align}\label{eq:delta_A_C_sum0}
    1^\top\delta A_C &=\E\left[\sum_{h=1}^H \widehat{G}_h 1^\top S_c^{(h)}J_h^c \left(S_c^{(h)}\right)^\top A_Q^\top \delta_h e_{c_h}^\top\right]\notag\\
    &= \E\left[\sum_{h=1}^H \widehat{G}_h 1^\top J_h^c \left(S_c^{(h)}\right)^\top A_Q^\top \delta_h e_{c_h}^\top\right]=0,
\end{align}
where the first equality uses the fact that $1^\top S_c^{(h)}=1^\top$ and the second equality uses the fact that $1^\top J_h^c=0$. This combined with \eqref{eq:A_C_0} and our zero initialization suggests that
\begin{align}\label{eq:A_C_abs}
    \forall t\in\NN,j\in\{0,1\}: \quad\delta A_C^{(t)}(0,j)=-\delta A_C^{(t)}(1,j)\quad\text{and}\quad A_C^{(t)}(0,j)=-A_C^{(t)}(1,j).
\end{align}
The equations \eqref{eq:A_C_0} and \eqref{eq:A_C_abs} allow us to express $A_C^{(t)}$ and $\delta A_C^{(t)}$ as in \eqref{eq:A_C_simplified}. 
Similarly, we can further compute that each column of other 3 matrices also sum up to 0 using \eqref{eq:delta_A_B}, \eqref{eq:delta_A_P}, \eqref{eq:delta_A_Q}, i.e.,
\begin{align}\label{eq:delta_A_M_sum0}
    1^\top\delta A_M =0\quad\text{for } M\in\{B,P,Q\}.
\end{align}
This and \eqref{eq:delta_A_C_sum0} combined with our zero initialization yields
\begin{align}\label{eq:A_M_sum0}
    \forall t\in\NN: \quad 1^\top \delta A_M^{(t)}=0, \quad\text{for } M\in\{B,C,P,Q\};\quad 1^\top  A_{M}^{(t)}=0, \quad\text{for } M\in\{B,C,Q\}.
\end{align}

By exploiting the symmetry of the data distribution and using \eqref{eq:A_M_sum0}, we could represent $A_P$ and $A_Q$ as in \eqref{eq:A_P_simplified} and \eqref{eq:A_Q_simplified}, and we can also obtain \eqref{eq:a_p3}.

We next compute $\delta a_{c,0}$ and $\delta a_{c,1}$ in \eqref{eq:A_C_simplified}.
By \eqref{eq:delta_A_C}, we have
\begin{align}\label{eq:delta_a_c_1_intermediate}
    \delta a_{c,1}     &= e_1^\top \delta A_C e_1 \notag\\
    &=\E\left[\sum_{h=1}^H \left((\mathbbm{1}\{c_1=\times\},\cdots,\mathbbm{1}\{c_h=\times\})\right)\left(\diag\{\alpha_h^c\}-\alpha_h^c(\alpha_h^c)^\top\right)\left(S_c^{(h)}\right)^\top A_Q^\top (e_{a_h}-\pi_h)\mathbbm{1}\{c_h=\times\} \widehat{G}_h\right]\notag\\
    &=\E\Bigg[\sum_{h=1}^H \Bigg\{\left(\sum_{i=1}^h\alpha_h^c(i)\mathbbm{1}\{c_i=\times\}\right) 
A_Q(:,\times)^\top -\left(\sum_{i=1}^h\alpha_h^c(i)\mathbbm{1}\{c_i=\times\}\right)\notag\\
&\qquad\qquad\cdot\left(\sum_{i=1}^h\alpha_h^c(i)\mathbbm{1}\{c_i=\times\}A_Q(:,\times)^\top+\sum_{i=1}^h\alpha_h^c(i)\mathbbm{1}\{c_i=0\}A_Q(:,0)^\top\right)\Bigg\}(e_{a_h}-\pi_h)\mathbbm{1}\{c_h=\times\} \widehat{G}_h\Bigg]\notag\\
&=\E\left[\sum_{h=1}^H \left(\sum_{i=1}^h\alpha_h^c(i)\mathbbm{1}\{c_i=\times\}\right) \left(\sum_{i=1}^h\alpha_h^c(i)\mathbbm{1}\{c_i=0\}\right) \left(A_Q(:,\times)-A_Q(:,0)\right)^\top (e_{a_h}-\pi_h)\mathbbm{1}\{c_h=\times\} \widehat{G}_h\right].
\end{align}
Combining \eqref{eq:p_c_x,p_c_0}, we have
\begin{align}\label{eq:delta_a_c_1_intermediate_2}
    \delta a_{c,1} &= \E\left[\sum_{h=1}^H \mathbbm{1}\{c_h=\times\} p_{h,\times}^c p_{h,0}^c \left(A_Q(:,\times)-A_Q(:,0)\right)^\top (e_{a_h}-\pi_h)\widehat{G}_h\right]\notag\\
    &= \E\left[\sum_{h=1}^H \mathbbm{1}\{c_h=\times\} p_{h,\times}^c p_{h,0}^c \left(A_Q(:,\times)-A_Q(:,0)\right)^\top \left(\pi_h\odot\left(Q_h-G_h \vone\right)\right)\right].
\end{align}
Similarly, we can compute that 
\begin{align}\label{eq:delta_a_c_0_intermediate}
    &\delta a_{c,0}\notag\\
&=\E\left[\sum_{h=1}^H \left(\sum_{i=1}^h\alpha_h^c(i)\mathbbm{1}\{c_i=0\}\right) \left(\sum_{i=1}^h\alpha_h^c(i)\mathbbm{1}\{c_i=\times\}\right) \left(A_Q(:,0)-A_Q(:,\times)\right)^\top (e_{a_h}-\pi_h)\mathbbm{1}\{c_h=0\} \widehat{G}_h\right]\notag\\
&=\E\left[\sum_{h=1}^H \mathbbm{1}\{c_h=0\} p_{h,0}^c p_{h,\times}^c \left(A_Q(:,0)-A_Q(:,\times)\right)^\top (e_{a_h}-\pi_h)\widehat{G}_h\right]\notag\\
&=\E\left[\sum_{h=1}^H \mathbbm{1}\{c_h=0\} p_{h,0}^c p_{h,\times}^c \left(A_Q(:,0)-A_Q(:,\times)\right)^\top \left(\pi_h\odot\left(Q_h-G_h \vone\right)\right)\right].
\end{align}
From \eqref{eq:delta_a_c_1_intermediate_2}, \eqref{eq:delta_a_c_0_intermediate} and \eqref{eq:A_Q_simplified} we further deduce
\begin{align}
    &\delta a_{c,0}\notag\\
    &= \E\left[\sum_{h=1}^H \mathbbm{1}\{c_h=0\} p_{h,0}^c p_{h,\times}^c \left(A_Q(:,0)-A_Q(:,\times)\right)^\top \left(\pi_h\odot\left(Q_h-G_h \vone\right)\right)\right]\notag\\
    &=\EE\left[\sum_{h=1}^H \mathbbm{1}\{c_h=0\} p_{h,0}^c p_{h,\times}^c \left(\frac{a_{q,\times}+a_{q,0}}{k}\sum_{i=1}^k \pi_h(i)(Q_h(i)-G_h)-(a_{q,\times}+a_{q,0})\pi_h(\up)(Q_h(\up)-G_h)\right)\right]\notag\\
    &=\left(1+\frac{1}{k}\right)\EE\left[\sum_{h=1}^H \mathbbm{1}\{c_h=0\} p_{h,0}^c p_{h,\times}^c \left(a_{q,\times}+a_{q,0}\right)\pi_h(\up)(G_h-Q_h(\up))\right],
\end{align}
and
\begin{align}
    \delta a_{c,1}&=\E\left[\sum_{h=1}^H \mathbbm{1}\{c_h=\times\} p_{h,\times}^c p_{h,0}^c \left(A_Q(:,\times)-A_Q(:,0)\right)^\top \left(\pi_h\odot\left(Q_h-G_h \vone\right)\right)\right]\notag\\
    &=\left(1+\frac{1}{k}\right)\EE\left[\sum_{h=1}^H \mathbbm{1}\{c_h=\times\} p_{h,0}^c p_{h,\times}^c \left(a_{q,\times}+a_{q,0}\right)\pi_h(\up)(Q_h(\up)-G_h)\right]\notag\\
    &=\left(1+\frac{1}{k}\right)\EE\left[\sum_{h=1}^H \mathbbm{1}\{c_h=\times\} p_{h,0}^c p_{h,\times}^c \left(a_{q,\times}+a_{q,0}\right)\pi_h(\up)(1-\pi_h(\up))Q_h(\up)\right],
\end{align}
where the last relation uses the fact that $Q_h(\down_i)=0$ when $c_h=\times$.
Further note that 
\begin{align}\label{eq:p_c_simplified}
    p_{h,\times}^c=\begin{cases}
    \frac{N_{h,\times}e^{a_{c,1}}}{N_{h,\times}e^{a_{c,1}}+(h-N_{h,\times})e^{-a_{c,1}}},\,\,\text{if }c_h=\times,\\
    \frac{N_{h,\times}e^{-a_{c,0}}}{N_{h,\times}e^{-a_{c,0}}+(h-N_{h,\times})e^{a_{c,0}}},\,\,\text{if }c_h=0,
    \end{cases},\quad p_{h,0}^c=\begin{cases}
        \frac{(h-N_{h,\times})e^{-a_{c,1}}}{N_{h,\times}e^{a_{c,1}}+(h-N_{h,\times})e^{-a_{c,1}}},\,\,\text{if }c_h=\times,\\
        \frac{(h-N_{h,\times})e^{a_{c,0}}}{N_{h,\times}e^{-a_{c,0}}+(h-N_{h,\times})e^{a_{c,0}}},\,\,\text{if }c_h=0.
    \end{cases}
\end{align}
Plugging \eqref{eq:p_c_simplified} into the above two equations about $\delta a_{c,0}$ and $\delta a_{c,1}$, we get \eqref{eq:delta_a_c_0_simplified_balanced} and \eqref{eq:delta_a_c_1_simplified_balanced}.

Analogously, we can compute that for any $i,j\in\{0,\cdots,N\}$,
\begin{align}
    &\delta A_B(i,j)\notag\\
    &=\E\left[\sum_{h=1}^H \mathbbm{1}\{n_h=j\} \sum_{m=1}^h \alpha_h^n(m)\left(\mathbbm{1}\{n_{m-1}=i\}-\sum_{l=1}^h\alpha_h^n(l)\mathbbm{1}\{n_{l-1}=i\}\right) A_P(:,a_{m-1})^\top (e_{a_h}-\pi_h)\widehat G_h\right].
\end{align}
By the above expression, the symmetry of the node distribution, our zero initialization, \eqref{eq:delta_A_M_sum0}, \eqref{eq:A_B_0} and \eqref{eq:p_n}, we immediately obtain \eqref{eq:delta_A_B_simplified}, in which 
\begin{subequations}
\begin{align}
    \delta a_{b,0} &= \E\left[\sum_{h=1}^H \mathbbm{1}\{n_h=j\} \sum_{m=1}^h \alpha_h^n(m)\left(\mathbbm{1}\{n_{m-1}=0\}-p_{h,0}^n\right) A_P(:,a_{m-1})^\top \left(\pi_h\odot\left(Q_h-G_h \vone\right)\right)\right],\label{eq:delta_a_b_0_intermediate}\\
    \delta a_{b,1} &= \E\left[\sum_{h=1}^H \mathbbm{1}\{n_h=j\} \sum_{m=1}^h \alpha_h^n(m)\left(\mathbbm{1}\{n_{m-1}=j\}-p_{h,j}^n\right) A_P(:,a_{m-1})^\top \left(\pi_h\odot\left(Q_h-G_h \vone\right)\right)\right],\label{eq:delta_a_b_1_intermediate}\\
    \delta a_{b,2} &= \E\left[\sum_{h=1}^H \mathbbm{1}\{n_h=j\} \sum_{m=1}^h \alpha_h^n(m)\left(\mathbbm{1}\{n_{m-1}=i\}-p_{h,i}^n\right) A_P(:,a_{m-1})^\top \left(\pi_h\odot\left(Q_h-G_h \vone\right)\right)\right],\notag\\
    &=\frac{-\delta a_{b,0}-\delta a_{b,1}}{N-1}\label{eq:delta_a_b_2_intermediate}
\end{align}
\end{subequations}
for any $i,j\in[N]$, $i\neq j$. Combining \eqref{eq:delta_a_b_2_intermediate} and \eqref{eq:A_M_sum0} yield \eqref{eq:delta_a_b_2_simplified_balanced}. We can further simplify $\delta a_{b,0}$ as follows: 
\small
\begin{align}
    &\delta a_{b,0} \notag\\
    &\overset{\eqref{eq:delta_a_b_0_intermediate}}=\frac{1}{N}\E\Bigg[\sum_{h=1}^H \bigg\{\frac{e^{a_{b,0}}}{S_h}\left(1-\frac{e^{a_{b,0}}}{S_h}\right)\underbrace{A_P(:,0)^\top \left(\pi_h\odot\left(Q_h-G_h \vone\right)\right)}_{=:\beta_{h,0}}\notag\\
    &\quad-\frac{e^{a_{b,0}}}{S_h}\frac{e^{a_{b,1}}}{S_h}\underbrace{\sum_{i\in\gA_{h,\times}}A_P(:,i)^\top \left(\pi_h\odot\left(Q_h-G_h \vone\right)\right)}_{=:\beta_{h,1}}\notag\\
    &\quad-\frac{e^{a_{b,0}}}{S_h}\frac{e^{a_{b,2}}}{S_h}\underbrace{\left(\sum_{i=1}^{k}\left((N_h(i)-\mathbbm{1}\{i\in\gA_{h,\times}\})A_P(:,i)^\top \left(\pi_h\odot\left(Q_h-G_h \vone\right)\right)\right)+N_h(\up)A_P(:,\up)^\top \left(\pi_h\odot\left(Q_h-G_h \vone\right)\right)\right)}_{=:\beta_{h,2}}
    \bigg\}\Bigg],
\end{align}
\normalsize
where 
$$S_h\coloneqq e^{a_{b,0}}+|\gA_{h,\times}|e^{a_{b,1}}+(h-1-|\gA_{h,\times}|)e^{a_{b,2}}.$$
Similarly, we can further simplify $\delta a_{b,1}$ as follows:
\begin{align}
    \delta a_{b,1} &\overset{\eqref{eq:delta_a_b_1_intermediate}}=\frac{1}{N}\E\Bigg[\sum_{h=1}^H \bigg\{\frac{e^{a_{b,1}}}{S_h}\left(1-\frac{|\gA_{h,\times}|e^{a_{b,1}}}{S_h}\right)\beta_{h,1}-\frac{|\gA_{h,\times}|e^{a_{b,1}}}{S_h}\frac{e^{a_{b,0}}}{S_h}\beta_{h,0}-\frac{|\gA_{h,\times}|e^{a_{b,1}}}{S_h}\frac{e^{a_{b,2}}}{S_h}\beta_{h,2}\bigg\}\Bigg].
\end{align}
By \eqref{eq:A_P_simplified} we have
\begin{subequations}
\begin{align}
    A_P(:,0)^\top \left(\pi_h\odot\left(Q_h-G_h \vone\right)\right)&=\frac{1}{k}a_{p,0}\pi_h(\up)\left(Q_h(\up)-G_h\right),\\
    A_P(:,\up)^\top \left(\pi_h\odot\left(Q_h-G_h \vone\right)\right)&=-\frac{1}{k}a_{p,\up}\pi_h(\up)\left(Q_h(\up)-G_h\right),\\
   \forall i\in[k]:\,\, A_P(:,i)^\top \left(\pi_h\odot\left(Q_h-G_h \vone\right)\right)&=-\left(a_{p,1}+a_{p,2}\right)\pi_h(i)\left(Q_h(i)-G_h\right)-a_{p,2}\pi_h(\up)(Q_h(\up)-G_h).
\end{align}
\end{subequations}
Plugging them into the expressions of $\beta_{h,0}$, $\beta_{h,1}$ and $\beta_{h,2}$ and organizing the terms, we get \eqref{eq:delta_a_b_0_simplified_balanced}, \eqref{eq:delta_a_b_1_simplified_balanced}, \eqref{eq:beta_h_0_balanced}, \eqref{eq:beta_h_1_balanced} and \eqref{eq:beta_h_2_balanced}.

We next compute $\delta a_{p,i}$ for $i\in\{0,1,3,\up\}$ and $\delta a_{q,j}$ for $j\in\{0,\times\}$. First, by \eqref{eq:delta_A_P}, we have for all $i\in[k+1],\,\,j\in\{0,1,\cdots,k+1\}$:
\begin{align}
    \delta A_P(i,j)&=\E\left[\sum_{h=1}^H \left(\mathbbm{1}\{a_h=i\}-\pi_h(i)\right) \left(\sum_{m=1}^h \alpha_h^n(m) \mathbbm{1}\{a_{m-1}=j\}\right) \widehat{G}_h\right]\notag\\
    &=\E\left[\sum_{h=1}^H  \left(\sum_{m=1}^h \alpha_h^n(m) \mathbbm{1}\{a_{m-1}=j\}\right) \left(\pi_h(i)\left(1-\pi_h(i)\right)Q_h(i)-\sum_{a\neq i}\pi_h(a)\pi_h(i)Q_h(a)\right)\right]\notag\\
    &=\E\left[\sum_{h=1}^H \pi_h(i) \left(\sum_{m=1}^h \alpha_h^n(m) \mathbbm{1}\{a_{m-1}=j\}\right) \left(Q_h(i)-G_h\right)\right].
\end{align}
This together with \eqref{eq:p_n} gives 
\begin{align}\label{eq:delta_A_P_simplified_entries}
    \delta A_P(:,j)=\E\left[\sum_{h=1}^H p_{h,j}^a \left(\pi_h\odot\left(Q_h-G_h \vone\right)\right)\right],\,\,\forall j\in\{0,1,\cdots,k+1\}.
\end{align}
By a similar argument, we can compute that for any $i\in[k+1]$, $j\in\{0,\times\}$,
\begin{align}\label{eq:delta_A_Q_simplified_entries_app}
    \delta A_Q(i,j)&=\E\left[\sum_{h=1}^H \left(\mathbbm{1}\{a_h=i\}-\pi_h(i)\right) \left(\sum_{m=1}^h \alpha_h^c(m) \mathbbm{1}\{c_m=j\}\right) \widehat{G}_h\right]\notag\\
    &=\E\left[\sum_{h=1}^H\pi_h(i)\left(\sum_{m=1}^h \alpha_h^c(m) \mathbbm{1}\{c_m=j\}\right) \left(Q_h(i)-G_h\right)\right]\notag\\
    &\overset{\eqref{eq:p_c}}= \E\left[\sum_{h=1}^H \pi_h(i) p_{h,j}^c \left(Q_h(i)-G_h\right)\right].
\end{align}
This together with \eqref{eq:A_Q_0} gives 
\begin{align}\label{eq:delta_A_Q_simplified_entries}
    \delta A_Q(:,j)=\begin{cases}
        \E\left[\sum_{h=1}^H  p_{h,j}^c \left(\pi_h\odot\left(Q_h-G_h \vone\right)\right)\right], & j\in\{0,\times\},\\
        0, & \text{otherwise}.
    \end{cases}
\end{align}
\begin{subequations}
By \eqref{eq:delta_A_P_simplified_entries} and \eqref{eq:A_P_simplified} we have
\begin{align}
    \delta a_{p,0}=\delta A_P(\up,0)=\EE\left[\sum_{h=1}^H \pi_h(\up)\frac{e^{a_{b,0}}}{e^{a_{b,0}}+|\gA_{h,\times}|e^{a_{b,1}}+(h-1-|\gA_{h,\times}|)e^{a_{b,2}}}\left(Q_h(\up)-G_h\right)\right],
\end{align}
which yields \eqref{eq:delta_a_p_0_simplified_balanced},
\begin{align}
    \delta a_{p,\up}=-\delta A_P(\up,\up)=-\EE\left[\sum_{h=1}^H \pi_h(\up)\frac{N_h(\up)e^{a_{b,2}}}{e^{a_{b,0}}+|\gA_{h,\times}|e^{a_{b,1}}+(h-1-|\gA_{h,\times}|)e^{a_{b,2}}}\left(Q_h(\up)-G_h\right)\right],
\end{align}
which yields \eqref{eq:delta_a_p_up_simplified_balanced},
\begin{align}
    \delta a_{p,1}=-\delta A_P(j,j)&=-\EE\Bigg[\sum_{h=1}^H\pi_h(j)\bigg(\mathbbm{1}\{j\in\gA_{h,\times}\}
\frac{e^{a_{b,1}}+(N_h(j)-1)e^{a_{b,2}}}{e^{a_{b,0}}+|\gA_{h,\times}|e^{a_{b,1}}+(h-1-|\gA_{h,\times}|)e^{a_{b,2}}}(-G_h)\notag\\
&\qquad+\mathbbm{1}\{j\in[k]\setminus\gA_{h,\times}\}\frac{N_h(j)e^{a_{b,2}}}{e^{a_{b,0}}+|\gA_{h,\times}|e^{a_{b,1}}+(h-1-|\gA_{h,\times}|)e^{a_{b,2}}}\left(Q_h(j)-G_h\right)\bigg)\Bigg],
\end{align}
where $j\in[k]$. This gives \eqref{eq:delta_a_p_1_simplified_balanced}. And similarly, we have
\begin{align}
    \delta a_{p,3}=\delta A_P(\up,j)&=\EE\Bigg[\sum_{h=1}^H \pi_h(\up)\bigg(\mathbbm{1}\{j\in\gA_{h,\times}\}\frac{e^{a_{b,1}}+(N_h(j)-1)e^{a_{b,2}}}{e^{a_{b,0}}+|\gA_{h,\times}|e^{a_{b,1}}+(h-1-|\gA_{h,\times}|)e^{a_{b,2}}}\notag\\
&\qquad+\mathbbm{1}\{j\in[k]\setminus\gA_{h,\times}\}\frac{N_h(j)e^{a_{b,2}}}{e^{a_{b,0}}+|\gA_{h,\times}|e^{a_{b,1}}+(h-1-|\gA_{h,\times}|)e^{a_{b,2}}}\bigg)(Q_h(\up)-G_h)\Bigg],
\end{align}
\end{subequations}
where $j\in[k]$. This gives \eqref{eq:delta_a_p_3_simplified_balanced}. 
Lastly, combining \eqref{eq:delta_A_Q_simplified_entries} and \eqref{eq:p_c_simplified}, we have \eqref{eq:delta_a_q_0_simplified_balanced} and \eqref{eq:delta_a_q_times_simplified_balanced}.

\paragraph{Policy computation.}
We let 
\begin{align}
\varphi_h&\coloneqq A_P(:,0)e^{a_{b,0}}+\sum_{j\in\gA_{h,\times}}A_P(:,j)e^{a_{b,1}}\notag\\
&\qquad+\left(N_h(\up)A_P(:,\up)+\sum_{i=1}^k\left(N_h(i)-\mathbbm{1}\{i\in\gA_{h,\times}\}\right)A_P(:,i)\right)e^{a_{b,2}}.
\end{align}
Then by \eqref{eq:pi_theta_simplified}, under Assumption~\ref{asmp:token_embedding}, we have
\small
\begin{align}\label{eq:pi_h_simplified_balanced_vector}
    \pi_h&=\softmax\Bigg(
        \frac{\varphi_h}{e^{a_{b,0}}+|\gA_{h,\times}|e^{a_{b,1}}+(h-1-|\gA_{h,\times}|)e^{a_{b,2}}}+\begin{cases}
        \frac{N_{h,\times}e^{-a_{c,0}}A_Q(:,\times)+(h-N_{h,\times})e^{a_{c,0}}A_Q(:,0)}{N_{h,\times}e^{-a_{c,0}}+(h-N_{h,\times})e^{a_{c,0}}}, & \text{if } c_h=0\\
        \frac{N_{h,\times}e^{a_{c,1}}A_Q(:,\times)+(h-N_{h,\times})e^{-a_{c,1}}A_Q(:,0)}{N_{h,\times}e^{a_{c,1}}+(h-N_{h,\times})e^{-a_{c,1}}}, & \text{if } c_h=\times
    \end{cases}\Bigg).
\end{align}
\normalsize
By \eqref{eq:pi_h_simplified_balanced_vector} and \eqref{eq:A_P_simplified}, we have
\begin{subequations}
\begin{align}
    \varphi_h(\up)&=0,  \\
    \forall m\in[k]:\quad\varphi_h(m)&= -\frac{a_{p,0}}{k}e^{a_{b,0}}+|\gA_{h,\times}|a_{p,2}e^{a_{b,1}}-\mathbbm{1}\{m\in\gA_{h,\times}\}\left((a_{p,1}+a_{p,2})(e^{a_{b,1}}-e^{a_{b,2}})\right)\notag\\
    &\quad+\left(N_h(\up)\frac{a_{p,\up}}{k}-N_h(m)a_{p,1}+(h-1-N_h(\up)-N_h(m)-|\gA_{h,\times}|)a_{p,2}\right)e^{a_{b,2}}.
\end{align}
\end{subequations}
The above two expressions give \eqref{eq:varphi_h_i_balanced}. \eqref{eq:A_P_simplified} together with \eqref{eq:pi_h_simplified_balanced_vector} gives \eqref{eq:xi_h_balanced}.




\subsubsection{Proof of Lemma~\ref{lm:phase_1.1}}\label{sec_app:phase_1.1}
We show by induction.
First, by our zero initialization, Lemma~\ref{lm:phase_1.1} holds for $t=0$. We make the following induction hypothesis:
\begin{center}
\textbf{Induction Hypothesis I:} For all $t\leq s-1$ ($s\in [T_{1/2}]$), the relations in Lemma~\ref{lm:phase_1.1} hold. 
\end{center}

Below we show Lemma~\ref{lm:phase_1.1} holds for $t=s$ under Induction Hypothesis I. To do so, we first simplify the gradients for $l=1$.
We define events
\begin{align}\label{eq:gE_1_h}
    \gE_{1,h}\coloneqq \left\{\text{the environment has not terminated at step } h, \text{ and the tree depth  } l=1\right\}.
\end{align}
Recall we define $\widehat{G}_h$ in~\eqref{eq:pi_G_Q_h}. We let $V_{1,h}^\pi$ denote the expectation of $\widehat{G}_h$ conditioned on $\gE_{1,h}$ under policy $\pi$:
\begin{align}\label{eq:V_1_h}
    V_{1,h}^\pi\coloneqq \EE^\pi[\widehat{G}_h|\gE_{1,h}].
\end{align}
We let $V_{1,h}^{(t)}\coloneqq V_{1,h}^{\pi^{(t)}}$. Note that $V_{1,h}^{(t)}$ is the same for any possible $s_{h}$ conditioned on $\gE_{1,h}$ and $\pi_h^{(t)}$ by symmetry under the balanced case. We let $\pi_{1,h}^{(t)}(i)$ denote $\pi_h^{(t)}(i)$ conditioned on $\gE_{1,h}$ and $\pi_h^{(t)}$ for any $i\in[k+1]$.
Then by symmetry, for all odd steps $2h-1$ ($h\in[k]$) (where the agent is at the root node when $l=1$), we have $\pi_{1,2h-1}^{(t)}(i)$ are equal for any $i\in[k]\setminus\gA_{2h-1,\times}$ for any $t\in \NN$. 
We let $\pi_{1,2h-1}^{(t)}(\down)\coloneqq \pi_{1,2h-1}^{(t)}(i)$ for any $i\in[k]\setminus\gA_{2h-1,\times}$. Similarly, $\pi_{1,2h-1}^{(t)}(i)$ are also equal for any $i\in\gA_{2h-1,\times}$, and we let $\pi_{1,2h-1}^{(t)}(\down_\times)\coloneqq \pi_{1,2h-1}^{(t)}(i)$ for any $i\in\gA_{2h-1,\times}$. We also let 
$$\pi_{1,2h}^{(t)}(\down_\times^0)\coloneqq \pi_{1,2h}^{(t)}(i) \qquad \mbox{for any}~{i\in[k]\setminus\left(\gA_{2h-1,\times}\cup\{a_{2h-1}\}\right)}$$ 
and 
$$\pi_{1,2h}^{(t)}(\down_\times^1)\coloneqq \pi_{1,2h}^{(t)}(i)\qquad \mbox{for any}~{i\in\gA_{2h-1,\times}\cup\{a_{2h-1}\}},$$ 
whose values are also independent of $s_{2h}$ conditioned on $\gE_{1,2h}$. We let $\PP^{\pi}(\gE_{1,h})$ denote the probability of event $\gE_{1,h}$ happening under policy $\pi$. We let $\PP^{(t)}(\gE_{1,h})\coloneqq \PP^{\pi^{(t)}}(\gE_{1,h})$ for notation simplicity.

\paragraph{Gradients/policy simplification for Phase 1.1.} By Induction Hypothesis I, using the above notation we define, we could further simplify the gradients and policy expressions in Lemma~\ref{lm:gradients} when $l=1$. We summarize the results in the following lemma.
\begin{lm}\label{lm:gradients_simplified_phase1.1_balanced}
    Under Induction Hypothesis I, we have for any $0\leq t\leq s-1$:
    \begin{subequations}
    \begin{align}
       \delta a_{q,0}^{(t)}&=\pi_{1,1}^{(t)}(\up)V_{1,1}^{(t)}\notag\\
       &\quad+\sum_{h=1}^{k-1} \PP^{(t)}\left(\gE_{1,2h}\right)\pi_{1,2h}^{(t)}(\up)\bigg(\frac{1}{1+\frac{h}{h+1}e^{-2a_{c,0}^{(t)}}}\pi_{1,2h+1}^{(t)}(\up)-\frac{1}{1+e^{2a_{c,1}^{(t)}}}\left(1-\pi_{1,2h}^{(t)}(\up)\right)\bigg)V_{1,2h+1}^{(t)}, \label{eq:delta_a_q_0_simplified_balanced_phase_1} \\
    \delta a_{q,\times}^{(t)}&=\sum_{h=1}^{k-1} \PP^{(t)}\left(\gE_{1,2h}\right)\pi_{1,2h}^{(t)}(\up)\bigg(\frac{1}{1+e^{-2a_{c,1}^{(t)}}}\left(1-\pi_{1,2h}^{(t)}(\up)\right)-\frac{1}{1+\frac{h+1}{h}e^{2a_{c,0}^{(t)}}}\pi_{1,2h+1}^{(t)}(\up)\bigg)V_{1,2h+1}^{(t)}, \label{eq:delta_a_q_x_simplified_balanced_phase_1}  \\
        \delta a_{c,0}^{(t)}&=\left(1+\frac{1}{k}\right)\sum_{h=1}^{k-1} \PP^{(t)}\left(\gE_{1,2h}\right)\pi_{1,2h}^{(t)}(\up)\frac{h(h+1)}{\left(he^{-a_{c,0}^{(t)}}+(h+1)e^{a_{c,0}^{(t)}}\right)^2}\left(a_{q,\times}^{(t)}+a_{q,0}^{(t)}\right)\pi_{1,2h+1}^{(t)}(\up)V_{1,2h+1}^{(t)}, \label{eq:delta_a_c_0_simplified_balanced_phase_1} \\
        \delta a_{c,1}^{(t)}&=\left(1+\frac{1}{k}\right)\sum_{h=1}^{k-1} \PP^{(t)}\left(\gE_{1,2h}\right)\pi_{1,2h}^{(t)}(\up)\frac{1}{\left(e^{a_{c,1}^{(t)}}+e^{-a_{c,1}^{(t)}}\right)^2}\left(a_{q,\times}^{(t)}+a_{q,0}^{(t)}\right)(1-\pi_{1,2h}^{(t)}(\up))V_{1,2h+1}^{(t)}, \label{eq:delta_a_c_1_simplified_balanced_phase_1} \\
        \delta a_{p,0}^{(t)} &=(1+o(1))\left(-\pi_{1,1}^{(t)}(\up)V_{1,1}^{(t)}+\sum_{h=1}^{k-1} \PP^{(t)}\left(\gE_{1,2h}\right)\pi_{1,2h}^{(t)}(\up)\left(\frac{1-\pi_{1,2h}^{(t)}(\up)}{2h}-\frac{\pi_{1,2h+1}^{(t)}(\up)}{2h+1}\right)V_{1,2h+1}^{(t)}\right), \label{eq:delta_a_p_0_simplified_balanced_phase_1} \\
    \delta a_{p,1}^{(t)} &=(1+o(1))\frac{1}{k}\sum_{h=1}^{k-1}\PP^{(t)}\left(\gE_{1,2h}\right)\pi_{1,2h}^{(t)}(\up)\left(\frac{1}{2}\pi_{1,2h}^{(t)}(\down_\times^1)+\frac{h}{2h+1}\pi_{1,2h+1}^{(t)}(\down_\times)\right)V_{1,2h+1}^{(t)}, \label{eq:delta_a_p_1_simplified_balanced_phase_1} \\
    \delta a_{p,3}^{(t)} &=(1+o(1))\frac{1}{k}\sum_{h=1}^{k-1}\PP^{(t)}\left(\gE_{1,2h}\right)\pi_{1,2h}^{(t)}(\up)\left(\frac{1}{2}\left(1-\pi_{1,2h}^{(t)}(\up)\right)-\frac{h}{2h+1}\pi_{1,2h+1}^{(t)}(\up)\right)V_{1,2h+1}^{(t)}, \label{eq:delta_a_p_3_simplified_balanced_phase_1} \\
    \delta a_{p,\up}^{(t)}&=-(1+o(1))\sum_{h=1}^{k-1} \PP^{(t)}\left(\gE_{1,2h}\right)\pi_{1,2h}^{(t)}(\up)\left(\frac{h-1}{2h}\left(1-\pi_{1,2h}^{(t)}(\up)\right)-\frac{h}{2h+1}\pi_{1,2h+1}^{(t)}(\up)\right)V_{1,2h+1}^{(t)}, \label{eq:delta_a_p_up_simplified_balanced_phase_1} \\
    \delta a_{b,0}^{(t)}& =\frac{1+o(1)}{N}\sum_{h=1}^{k-1}\PP^{(t)}\left(\gE_{1,2h}\right)\pi_{1,2h}^{(t)}(\up)I_{h,0}^{(t)}V_{1,2h+1}^{(t)}, \label{eq:delta_a_b_0_simplified_balanced_phase_1} \\
    \delta a_{b,1}^{(t)}&=\frac{1+o(1)}{N}\sum_{h=1}^{k-1}\PP^{(t)}\left(\gE_{1,2h}\right)\pi_{1,2h}^{(t)}(\up)I_{h,1}^{(t)}V_{1,2h+1}^{(t)}, \label{eq:delta_a_b_1_simplified_balanced_phase_1}
\end{align}
\end{subequations}
where 
\begin{subequations}
\begin{align}
    I_{h,0}^{(t)}&\coloneqq \frac{1}{4h^2}\left(\left(\frac{2h-1}{k}a_{p,0}^{(t)}+ha_{p,2}^{(t)}+\frac{h-1}{k}a_{p,\up}^{(t)}\right)\left(1-\pi_{1,2h}^{(t)}(\up)\right)-\left(a_{p,1}^{(t)}+a_{p,2}^{(t)}\right)h\pi_{1,2h}^{(t)}(\down_\times^1)\right)\notag\\
    &\quad \,-\frac{h}{(2h+1)^2}\left(\left(\frac{1}{k}\left(2a_{p,0}^{(t)}+a_{p,\up}^{(t)}\right)+a_{p,2}^{(t)}\right)\pi_{1,2h+1}^{(t)}(\up)+\left(a_{p,1}^{(t)}+a_{p,2}^{(t)}\right)\pi_{1,2h+1}^{(t)}(\down_\times)\right), \label{eq:I_h_0_balanced} \\
    I_{h,1}^{(t)}&\coloneqq \frac{h}{(2h+1)^2}\bigg[(h+1)\left(a_{p,1}^{(t)}+a_{p,2}^{(t)}\right)\pi_{1,2h+1}^{(t)}(\down_\times)+\left(\frac{1}{k}\left(a_{p,0}^{(t)}-ha_{p,\up}^{(t)}\right)+(h+1)a_{p,2}^{(t)}\right)\pi_{1,2h+1}^{(t)}(\up)\bigg]. \label{eq:I_h_1_balanced}
\end{align}
\end{subequations}
For all $\forall h\in[k]$, $i\in[k]$, we have
\begin{subequations}
\begin{align}
        \pi_{1,2h-1}^{(t)}(\up)&=\left(1+o(1)\right)\frac{\exp\left(\xi_{2h-1}^{(t)}-\frac{h-1}{2h-1}a_{p,2}^{(t)}\right)}{(h-1)\exp\left(-\frac{a_{p,1}^{(t)}+a_{p,2}^{(t)}}{2h-1}\right)+k-h+1+\exp\left(\xi_{2h-1}^{(t)}-\frac{h-1}{2h-1}a_{p,2}^{(t)}\right)},\label{eq:pi_2h-1_up_phase1}\\
        \pi_{1,2h}^{(t)}(\up)&=\left(1+o(1)\right)\frac{\exp\left(\xi_{2h}^{(t)}-\frac{1}{2}a_{p,2}^{(t)}\right)}{h\exp\left(-\frac{a_{p,1}^{(t)}+a_{p,2}^{(t)}}{2h}\right)+k-h+\exp\left(\xi_{2h}^{(t)}-\frac{1}{2}a_{p,2}^{(t)}\right)},\label{eq:pi_2h_up_phase1}\\
        \pi_{1,2h-1}^{(t)}(i)&=
            \left(1+o(1)\right)\frac{\exp\left(-\frac{N_{2h-1}(i)}{2h-1}\left(a_{p,1}^{(t)}+a_{p,2}^{(t)}\right)\right)}{(h-1)\exp\left(-\frac{a_{p,1}^{(t)}+a_{p,2}^{(t)}}{2h-1}\right)+k-h+1+\exp\left(\xi_{2h-1}^{(t)}-\frac{h-1}{2h-1}a_{p,2}^{(t)}\right)},\label{eq:pi_1_2h-1_i_phase1}\\
        \pi_{1,2h}^{(t)}(i)&=
            \left(1+o(1)\right)\frac{\exp\left(-\frac{N_{2h}(i)}{2h}\left(a_{p,1}^{(t)}+a_{p,2}^{(t)}\right)\right)}{h\exp\left(-\frac{a_{p,1}^{(t)}+a_{p,2}^{(t)}}{2h}\right)+k-h+\exp\left(\xi_{2h}^{(t)}-\frac{1}{2}a_{p,2}^{(t)}\right)},\label{eq:pi_1_2h_i_phase1}
\end{align}
\end{subequations}
where
\begin{align}\label{eq:xi_h_phase1}
    \xi_{2h-1}^{(t)}\coloneqq -\frac{a_{q,0}^{(t)}-\frac{h-1}{h}e^{-2a_{c,0}^{(t)}}a_{q,\times}^{(t)}}{1+\frac{h-1}{h}e^{-2a_{c,0}^{(t)}}},\quad \xi_{2h}^{(t)}\coloneqq \frac{a_{q,\times}^{(t)}-e^{-2a_{c,1}^{(t)}}a_{q,0}^{(t)}}{1+e^{-2a_{c,1}^{(t)}}}.
\end{align}
\end{lm}
The proof of Lemma~\ref{lm:gradients_simplified_phase1.1_balanced} is postponed to Appendix~\ref{sec:proof_gradients_simplified_phase1_balanced}.

Define the ratios 
\begin{align}\label{eq:rho_ratio}
    \nu_h^{(t)}\coloneqq \frac{\pi_{1,2h-1}^{(t)}(\up)}{\pi_{1,2h}^{(t)}(\up)},\quad r_h^{(t)}\coloneqq \frac{\exp\left(\xi_{2h-1}^{(t)}-\frac{h-1}{2h-1}a_{p,2}^{(t)}\right)}{\exp\left(\xi_{2h}^{(t)}-\frac{1}{2}a_{p,2}^{(t)}\right)}.
\end{align}
We next divide Phase 1.1 into two cases: (i) $s-1\leq T_{1,1}$ and (ii) $T_{1,1}+1\leq s-1\leq T_{1/2}$.


\paragraph{Case 1: $s-1\leq T_{1,1}$.}
\eqref{eq:delta_a_p_1_simplified_balanced_phase_1} and Induction Hypothesis I imply that \eqref{eq:a_p_1_bound_T_11} holds at step $s-1$.
By Induction Hypothesis I, we can simplify the policy expressions \eqref{eq:pi_2h-1_up_phase1}, \eqref{eq:pi_2h_up_phase1}, \eqref{eq:pi_1_2h-1_i_phase1}, \eqref{eq:pi_1_2h_i_phase1} when $s-1\leq T_{1,1}$ as follows:
\begin{subequations}
\begin{align}
    \forall h\in[k],i\in[k]:\quad\pi_{1,2h-1}^{(s-1)}(\up)&=\left(1+o(1)\right)\frac{\exp\left(\xi_{2h-1}^{(s-1)}\right)}{k+\exp\left(\xi_{2h-1}^{(s-1)}\right)},\label{eq:pi_2h-1_up_phase1.1}\\
        \pi_{1,2h}^{(s-1)}(\up)&=\left(1+o(1)\right)\frac{\exp\left(\xi_{2h}^{(s-1)}\right)}{k+\exp\left(\xi_{2h}^{(s-1)}\right)},\label{eq:pi_2h_up_phase1.1}\\
        \pi_{1,2h-1}^{(s-1)}(i)&=\left(1+o(1)\right)\frac{1}{k+\exp\left(\xi_{2h-1}^{(s-1)}\right)},\label{eq:pi_2h-1_down_times_phase1.1}\\
        \pi_{1,2h}^{(s-1)}(i)&=\left(1+o(1)\right)\frac{1}{k+\exp\left(\xi_{2h}^{(s-1)}\right)}.\label{eq:pi_2h_down_times_phase1.1}
\end{align}
\end{subequations}
Then by \eqref{eq:pi_2h-1_up_phase1.1}, \eqref{eq:pi_2h_up_phase1.1}, \eqref{eq:a_q_x+a_q_0_=a_q_T11} and \eqref{eq:a_p_1_2_tiny_T_11} in Induction Hypothesis I, we have
\begin{align}\label{eq:rho_ratio_T_11_expression}
    r_h^{(s-1)}&=(1+o(1))\exp\Bigg(\underbrace{-\frac{1-\frac{h-1}{h}e^{-2a_{c,0}^{(s-1)}}e^{-2a_{c,1}^{(s-1)}}}{\left(1+\frac{h-1}{h}e^{-2a_{c,0}^{(s-1)}}\right)\left(1+e^{-2a_{c,1}^{(s-1)}}\right)}\left(a_{q,\times}^{(s-1)}+a_{q,0}^{(s-1)}\right)}_{\leq 0}\Bigg)\leq 1+o(1).
\end{align}
Note that if $a_{q,\times}^{(s-1)}\gtrsim \ln k$, there exists $s'<s$ such that
\begin{align*}
    \frac{1}{3}a_{q,\times}^{(s-1)}\leq a_{q,\times}^{(s')} \leq\frac{1}{2}a_{q,\times}^{(s-1)},
\end{align*}
and by \eqref{eq:delta_a_c_1_balanced_phase_1}, \eqref{eq:a_q_x+a_q_0_=a_q_x_T1}, we have
\begin{align}\label{eq:e2a_c_1>a_q_x^2}
    e^{2a_{c,1}^{(s-1)}}\geq e^{2a_{c,1}^{(s-1)}}-e^{2a_{c,1}^{(s')}}\gtrsim \left(a_{q,\times}^{(s-1)}\right)^2.
\end{align}
Thus by \eqref{eq:a_q_0<=a_q_x_bound_T_11} in Induction Hypothesis I, we have
\begin{align}\label{eq:xi_2h=a_q_x}
   \text{If } a_{q,\times}^{(s-1)}\gtrsim \ln k:\quad \xi_{2h}^{(s-1)}=\left(1+O\left(\frac{1}{\ln k}\right)\right)a_{q,\times}^{(s-1)}.
\end{align}
Thus by the definition of $T_{1,1}$ in \eqref{eq:T_1_1} and \eqref{eq:xi_2h=a_q_x}, we have
\begin{align}\label{eq:pi_1_2<=1/2}
    \forall s-1\leq T_{1,1}:\quad \pi_{1,2}^{(s-1)}(\up)\leq \frac{1+o(1)}{2}.
\end{align}

By \eqref{eq:pi_2h_up_phase1.1} and \eqref{eq:rho_ratio_T_11_expression}, we have when $s-1\leq T_{1,1}$:
\begin{align}\label{eq:pi_1_2h_up_bound_T_11}
   \forall h\in[k]:\quad \pi_{1,2h-1}^{(s-1)}(\up)\leq(1+o(1))\pi_{1,2h}^{(s-1)}(\up)=(1+o(1))\pi_{1,2}^{(s-1)}(\up).
\end{align}

If $\pi_{1,2}^{(s-1)}(\up)\lesssim \frac{1}{\sqrt{k}}$, then by \eqref{eq:pi_1_2h_up_bound_T_11}, we have
\begin{align}\label{eq:pi_1_2h_up<=1/sqrt(k)}
    \forall h\in[k]:\quad \pi_{1,2h}^{(s-1)}(\up)\lesssim \frac{1}{\sqrt{k}},\quad \pi_{1,2h-1}^{(s-1)}(\up)\lesssim \frac{1}{\sqrt{k}}.
\end{align}
Thus by \eqref{eq:delta_a_q_x_simplified_balanced_phase_1}, we have
\begin{align}\label{eq:delta_a_q_x_T_11}
    \text{when $\pi_{1,2}^{(s-1)}(\up)\lesssim \frac{1}{\sqrt{k}}$:} \quad\delta a_{q,\times}^{(s-1)}&=\frac{1+o(1)}{1+e^{-2a_{c,1}^{(s-1)}}}\sum_{h=1}^{k-1} \PP^{(s-1)}\left(\gE_{1,2h}\right)\pi_{1,2h}^{(s-1)}(\up)V_{1,2h+1}^{(s-1)}.
\end{align}
If $\pi_{1,2}^{(s-1)}(\up)\gtrsim \frac{1}{\sqrt{k}}$, since $\pi_{1,2}^{(s-1)}(\up)\leq\frac{1+o(1)}{2}$ (c.f. \eqref{eq:pi_1_2<=1/2}),   
we have
\begin{align*}
    \xi_2^{(s-1)}\overset{\eqref{eq:xi_h_phase1}}=\frac{a_{q,\times}^{(s-1)}-e^{-2a_{c,1}^{(s-1)}}a_{q,0}^{(s-1)}}{1+e^{-2a_{c,1}^{(s-1)}}}\asymp \ln k.
\end{align*}
By \eqref{eq:a_q_x+a_q_0_=a_q_T11} and \eqref{eq:xi_1_2_bound_T_11} Induction Hypothesis I, we have
\begin{align}\label{eq:a_q_x+a_q_0=xi_2}
    a_{q,\times}^{(s-1)}\asymp a_{q,\times}^{(s-1)}+a_{q,0}^{(s-1)}\gtrsim \xi_2^{(s-1)}\asymp \ln k.
\end{align}
Note that when $a_{q,0}^{(s-1)}+a_{q,\times}^{(s-1)}\gtrsim 1$, by \eqref{eq:delta_a_c_1_balanced_phase_1} in Induction Hypothesis I, we have
\begin{align}\label{eq:e2a_c_1>C}
    e^{2a_{c,1}^{(s-1)}}\geq C \text{ for some constant } C>1.
\end{align}
Thus by \eqref{eq:a_q_x+a_q_0=xi_2}, \eqref{eq:e2a_c_1>C}, \eqref{eq:rho_ratio_T_11_expression}, we have when $\pi_{1,2}^{(s-1)}(\up)\gtrsim \frac{1}{\sqrt{k}}$,
\begin{align}\label{eq:rho_h_up_bound_T11}
    \nu_h^{(s-1)}\leq \frac{1}{k^{c_1}} \text{ for some constant } c_1>0.
\end{align}
By \eqref{eq:pi_1_2h_up<=1/sqrt(k)}, \eqref{eq:rho_h_up_bound_T11} and \eqref{eq:pi_1_2h_up_bound_T_11}, we know that 
\begin{align}\label{eq:pi_1_2h-1_up=o(1)}
    \forall h\in[k]:\quad \pi_{1,2h-1}^{(s-1)}(\up)\leq \frac{1}{k^c} \text{ for some constant } c>0.
\end{align}
Combining \eqref{eq:pi_1_2h-1_up=o(1)} with \eqref{eq:delta_a_q_x_T_11}, we have for any $s-1\leq T_{1,1}$:
\begin{align}\label{eq:delta_a_q_x_T_11_bound_2}
    \delta a_{q,\times}^{(s-1)}&=\frac{1+o(1)}{1+e^{-2a_{c,1}^{(s-1)}}}\sum_{h=1}^{k-1} \PP^{(s-1)}\left(\gE_{1,2h}\right)\pi_{1,2h}^{(s-1)}(\up)\left(1-\pi_{1,2h}^{(s-1)}(\up)\right)V_{1,2h+1}^{(s-1)}.
\end{align}

By comparing \eqref{eq:delta_a_p_1_simplified_balanced_phase_1} and \eqref{eq:delta_a_q_x_T_11_bound_2}, we have
\begin{align}\label{eq:delta_a_p_1_small_T11}
    \forall t\leq s-1:\quad 0\leq \delta a_{p,1}^{(t)}\lesssim \frac{1}{k^2}\delta a_{q,\times}^{(t)},
\end{align}
and thus 
\begin{align}\label{eq:a_p_1_small_case_1}
    a_{p,1}^{(s)}\lesssim \frac{1}{k^2} a_{q,\times}^{(s)}\lesssim \frac{\ln k}{k^2}.
\end{align}
Similarly, by comparing \eqref{eq:delta_a_p_3_simplified_balanced_phase_1} and \eqref{eq:delta_a_q_x_T_11_bound_2}, we have
\begin{align}\label{eq:delta_a_p_3_small_T11}
    \forall t\leq s-1:\quad |\delta a_{p,3}^{(t)}|\lesssim \frac{1}{k}\delta a_{q,\times}^{(t)},
\end{align}
and thus by \eqref{eq:a_p3} we have
\begin{align*}
    |\delta a_{p,2}^{(s)}|\lesssim  \left|\frac{\delta a_{p,1}^{(s)}-\delta a_{p,3}^{(s)}}{k-1}\right|\lesssim \frac{1}{k^2}\delta a_{q,\times}^{(s)},
\end{align*}
which yields
\begin{align}\label{eq:a_p_2_small_case_1}
    |a_{p,2}^{(s)}|\lesssim \frac{\ln k}{k^2}.
\end{align}
\eqref{eq:a_p_2_small_case_1} and \eqref{eq:a_p_1_small_case_1} together show that \eqref{eq:a_p_1_2_tiny_T_11} holds at step $s$.
Similarly, by comparing the gradient expressions of $a_{p,0}^{(t)}$ in \eqref{eq:delta_a_p_0_simplified_balanced_phase_1} and $a_{p,\up}^{(t)}$ in \eqref{eq:delta_a_p_up_simplified_balanced_phase_1} with \eqref{eq:delta_a_q_x_T_11_bound_2} and by our choice of $T_{1,1}$, we immediately know that \eqref{eq:a_p_0_up_bound_T_1} holds at step $s$.

By \eqref{eq:delta_a_q_0_simplified_balanced_phase_1}, we have
\begin{align}\label{eq:delta_a_q_0_upper_bound_T11}
    \delta a_{q,0}^{(s-1)}&\leq\pi_{1,1}^{(s-1)}(\up)V_{1,1}^{(s-1)}+\frac{1}{1+\frac{h}{h+1}e^{-2a_{c,0}^{(s-1)}}}\sum_{h=1}^{k-1} \PP^{(s-1)}\left(\gE_{1,2h}\right)\pi_{1,2h}^{(s-1)}(\up)\pi_{1,2h+1}^{(s-1)}(\up)V_{1,2h+1}^{(s-1)}\notag\\
    &\leq \nu^{(s-1)}_1\pi_{1,2}^{(s-1)}(\up)V_{1,1}^{(s-1)}+\frac{1}{k^c}\frac{1}{1+\frac{h}{h+1}e^{-2a_{c,0}^{(s-1)}}}\sum_{h=1}^{k-1} \PP^{(s-1)}\left(\gE_{1,2h}\right)\pi_{1,2h}^{(s-1)}(\up)V_{1,2h+1}^{(s-1)}.
 \end{align}

We introduce the following lemma which reveals the ralationships of values at different states.
\begin{lm}\label{lm:G_2h_1_bound}
    Under Induction Hypothesis I, when $l=1$, $s-1\leq T_{1,1}$, then for any $t\leq s-1$ we have
    \begin{align}\label{eq:G_2h_1_bound_asymp}
        \forall h\in[k]:\quad \frac{V_{1,2h-1}^{(t)}}{V_{1,1}^{(t)}}\asymp 1.
    \end{align}
    
    Moreover, if we also have $\pi_{1,2}^{(t)}(\up)\lesssim \frac{1}{\ln k}$, then
    \begin{align}\label{eq:G_2h_1_bound}
        \forall h\in[k]:\quad \frac{V_{1,2h-1}^{(t)}}{V_{1,1}^{(t)}}=1+o(1).
    \end{align}
\end{lm}
The proof of Lemma~\ref{lm:G_2h_1_bound} is postponed to Appendix~\ref{sec:proof_G_2h_1_bound}. By \eqref{eq:delta_a_q_x_T_11_bound_2}, \eqref{eq:delta_a_q_0_upper_bound_T11}, Lemma~\ref{lm:G_2h_1_bound} and \eqref{eq:rho_ratio_T_11_expression} we immediately know that 
\begin{align}\label{eq:delta_a_q_0<=delta_a_q_x_1}
    \delta a_{q,0}^{(s-1)}\lesssim \delta a_{q,\times}^{(s-1)}.
\end{align}
This suggests \eqref{eq:a_q_0<=a_q_x_bound_T_11} holds at step $s$.
By Lemma~\ref{lm:G_2h_1_bound} and \eqref{eq:delta_a_q_0_upper_bound_T11} we also know that when $\pi_{1,2}^{(s-1)}(\up)\lesssim \frac{1}{\sqrt{k}}$, we have
\begin{align}\label{eq:delta_a_q_0_upper_bound2_T11}
    \delta a_{q,0}^{(s-1)}\leq (1+o(1))
    \nu^{(s-1)}_1\pi_{1,2}^{(s-1)}(\up)V_{1,3}^{(s-1)}+\frac{1}{k^c}\frac{1}{1+\frac{h}{h+1}e^{-2a_{c,0}^{(s-1)}}}\sum_{h=1}^{k-1} \PP^{(s-1)}\left(\gE_{1,2h}\right)\pi_{1,2h}^{(s-1)}(\up)V_{1,2h+1}^{(s-1)}.
\end{align}
If $
\nu^{(s-1)}_1\geq \frac{1}{4}$, by \eqref{eq:rho_ratio_T_11_expression}, we have $a_{q,\times}^{(s-1)}+a_{q,0}^{(s-1)}\lesssim 1$, and by \eqref{eq:a_q_x+a_q_0_=a_q_x_T1} and \eqref{eq:a_q_0<=a_q_x_bound_T_11} in Induction Hypothesis I we have $a_{q,0}^{(s-1)}\leq C$ for some constant $C>0$. And from the first two relations in \eqref{eq:a_q_x+a_q_0=xi_2}, we know that $\xi_2^{(s-1)}\lesssim 1$, and thus by \eqref{eq:pi_1_2h_up_bound_T_11} we have
\begin{align}
    \text{When } 
    \nu^{(s-1)}_1\geq \frac{1}{4}:\quad \pi_{1,2h}^{(s-1)}(\up)\gtrsim \frac{1}{k}.
\end{align}
If $
\nu^{(s-1)}_1\leq \frac{1}{4}$, $\pi_{1,2h}^{(s-1)}(\up)\lesssim \frac{1}{\sqrt{k}}$, by comparing \eqref{eq:delta_a_q_0_upper_bound2_T11} and \eqref{eq:delta_a_q_x_T_11}, we have
\begin{align}
    \delta a_{q,0}^{(s-1)}\leq \frac{2}{3}\delta a_{q,\times}^{(s-1)}.
\end{align}
When $\pi_{1,2}^{(s-1)}(\up)\gtrsim \frac{1}{\sqrt{k}}$, $\xi_2^{(s-1)}\asymp\ln k$,  by \eqref{eq:rho_h_up_bound_T11}, Lemma~\ref{lm:G_2h_1_bound} and \eqref{eq:delta_a_q_0_upper_bound2_T11}, \eqref{eq:delta_a_q_x_T_11}, there exists constant $c_2>0$ such that
\begin{align}
    \delta a_{q,0}^{(s-1)}\leq \frac{1}{k^{c_2}}\delta a_{q,\times}^{(s-1)}\leq \frac{2}{3}\delta a_{q,\times}^{(s-1)}.
\end{align}
The above two inequalities suggest \eqref{eq:a_q_0<=2/3a_q_x_T11} in Induction Hypothesis I holds at step $s$. 

On the other hand, for any $t\leq s-1$, we have
\begin{align}\label{eq:-delta_a_q_0_bound_T_11}
    -\delta a_{q,0}^{(t)}&\overset{\eqref{eq:delta_a_q_0_simplified_balanced_phase_1}}\leq \frac{1}{1+e^{2a_{c,1}^{(t)}}}\sum_{h=1}^{k-1} \PP^{(t)}\left(\gE_{1,2h}\right)\pi_{1,2h}^{(t)}(\up)\left(1-\pi_{1,2h}^{(t)}(\up)\right)V_{1,2h+1}^{(t)}\notag\\
    &\overset{\eqref{eq:delta_a_q_x_T_11_bound_2}}\leq (1+o(1))\frac{1+e^{-2a_{c,1}^{(t)}}}{1+e^{2a_{c,1}^{(t)}}}\delta a_{q,\times}^{(t)}.
\end{align}
This indicates that
\begin{align}
    -\left(a_{q,0}^{(t)}-a_{q,0}^{(t')}\right)\lesssim \frac{1}{e^{2a_{c,1}^{(t')}}}\left(a_{q,\times}^{(t)}-a_{q,\times}^{(t')}\right).
\end{align}
By \eqref{eq:a_q_x+a_q_0_=a_q_T11} and the above inequality, we know that \eqref{eq:-a_q_0_bound_T_12} in Induction Hypothesis I holds at step $s$.



By \eqref{eq:a_q_0<=2/3a_q_x_T11}, \eqref{eq:xi_h_phase1} and \eqref{eq:pi_1_2h_up_bound_T_11}, we have for any $s-1\leq T_{1,1}$:
\begin{align}\label{eq:pi_2h_up_bound_T11}
    \forall h\in[k-1]:\quad \pi_{1,2h}^{(s-1)}(\up)\gtrsim \frac{1}{k}.
\end{align}
By \eqref{eq:pi_1_2h-1_i_phase1} and Induction Hypothesis I, we have
\begin{align}\label{eq:pi_1_2h-1_down_bound_T_11}
    \forall h\in[k]:\quad \pi_{1,2h-1}^{(s)}(\down)=(1+o(1))\frac{1-\pi_{1,2h-1}^{(s)}(\up)}{k}\overset{\eqref{eq:pi_1_2h-1_up=o(1)}}=\frac{1+o(1)}{k},
\end{align}
and thus 
\begin{align}\label{eq:V_lower_bound_T11}
    \forall h\in[k]:\quad 
    V_{1,2h-1}^{(t)}\geq \pi_{1,2h-1}^{(t)}(\down)=(1+o(1))\frac{1}{k}.
\end{align}
Plugging \eqref{eq:V_lower_bound_T11} and \eqref{eq:pi_2h_up_bound_T11} into \eqref{eq:delta_a_q_x_T_11_bound_2}, we have
\begin{align}\label{eq:delta_a_q_x_T_11_bound}
    \delta a_{q,\times}^{(s-1)}\gtrsim \frac{1}{k^2}.
\end{align}
This combined with \eqref{eq:a_q_x_bound_T_11} in Induction Hypothesis I yields
\eqref{eq:a_q_x_bound_T_11} holds at step $s$.

By \eqref{eq:delta_a_q_0_simplified_balanced_phase_1} and \eqref{eq:delta_a_q_x_simplified_balanced_phase_1}, we have
\begin{align}\label{eq:delta_a_q_0+delta_a_q_x_T_11_expression}
    \delta a_{q,0}^{(s-1)}+\delta a_{q,\times}^{(s-1)}&=\pi_{1,1}^{(s-1)}(\up)V_{1,1}^{(s-1)}+ \sum_{h=1}^{k-1} \PP^{(s-1)}\left(\gE_{1,2h}\right)\pi_{1,2h}^{(s-1)}(\up)\bigg(\frac{1-\frac{h}{h+1}e^{-2a_{c,0}^{(s-1)}}}{1+\frac{h}{h+1}e^{-2a_{c,0}^{(s-1)}}}\pi_{1,2h+1}^{(s-1)}(\up)\notag\\
    &\qquad +\frac{1-e^{-2a_{c,1}^{(s-1)}}}{1+e^{-2a_{c,1}^{(s-1)}}}\left(1-\pi_{1,2h}^{(s-1)}(\up)\right)\bigg)V_{1,2h+1}^{(s-1)}>0.
\end{align}
This indicates \eqref{eq:a_q_x+a_q_0_increase} holds at step $s$.
Note that by \eqref{eq:rho_ratio_T_11_expression} and \eqref{eq:pi_2h_up_bound_T11},
 if $a_{q,0}^{(s-1)}+a_{q,\times}^{(s-1)}\lesssim 1$, by \eqref{eq:a_q_x+a_q_0=xi_2} and \eqref{eq:rho_ratio_T_11_expression}, we have
\begin{align}\label{eq:pi_1>=1/k}
    \pi_{1,1}^{(s-1)}(\up)\asymp \frac{1}{k},\quad \pi_{1,2h}^{(s-1)}(\up)\asymp \frac{1}{k},
\end{align}
and thus
\begin{align}
    \PP^{(s-1)}\left(\gE_{1,2h}\right)\leq \prod_{i=1}^{h-1}\pi_{1,2i}^{(s-1)}(\up)\leq \left(\frac{C_1}{k}\right)^{h-1}\text{ for some constant } C_1>0,
\end{align}
which, combine with \eqref{eq:delta_a_q_0+delta_a_q_x_T_11_expression} and Lemma~\ref{lm:G_2h_1_bound}, yields
\begin{align}\label{eq:delta_a_q_0+delta_a_q_x_T_11_bound_2_expression}
    \delta a_{q,0}^{(s-1)}+\delta a_{q,\times}^{(s-1)}&=\pi_{1,1}^{(s-1)}(\up)V_{1,1}^{(s-1)}+\left(1+o(1)\right)\frac{1-e^{-2a_{c,1}^{(s-1)}}}{1+e^{-2a_{c,1}^{(s-1)}}}\pi_{1,2}^{(s-1)}(\up)V_{1,3}^{(s-1)}.
\end{align}
For the same reason, if $a_{q,0}^{(s-1)}+a_{q,\times}^{(s-1)}\lesssim 1$, by \eqref{eq:delta_a_q_x_T_11_bound_2}, we have
\begin{align}\label{eq:delta_a_q_x_T_11_bound_2_expression}
    \delta a_{q,\times}^{(s-1)}=\frac{1+o(1)}{1+e^{-2a_{c,1}^{(s-1)}}}\pi_{1,2}^{(s-1)}(\up)V_{1,3}^{(s-1)}.
\end{align}
By \eqref{eq:pi_1>=1/k} and Lemma~\ref{lm:G_2h_1_bound}, \eqref{eq:delta_a_q_0+delta_a_q_x_T_11_bound_2_expression} and \eqref{eq:delta_a_q_x_T_11_bound_2_expression}, we have
\begin{align}\label{eq:delta_a_q_0+delta_a_q_x=delta_a_q_x_T_11_1}
    \text{If } \delta a_{q,0}^{(s-1)}+a_{q,\times}^{(s-1)}\lesssim 1:\quad \delta a_{q,0}^{(s-1)}+\delta a_{q,\times}^{(s-1)}\asymp \delta a_{q,\times}^{(s-1)}.
\end{align}
By \eqref{eq:e2a_c_1>C}, if $a_{q,0}^{(s-1)}+a_{q,\times}^{(s-1)}\gtrsim 1$, we have
\begin{align}
    \frac{1-e^{-2a_{c,1}^{(s-1)}}}{1+e^{-2a_{c,1}^{(s-1)}}}\geq C_2 \text{ for some constant } C_2>0.
\end{align}
Therefore, by Lemma~\ref{lm:G_2h_1_bound}, Induction Hypothesis I, \eqref{eq:delta_a_q_0+delta_a_q_x_T_11_expression} and \eqref{eq:delta_a_q_x_T_11_bound_2}, we have \eqref{eq:delta_a_q_0+delta_a_q_x=delta_a_q_x_T_11_1} also holds when $a_{q,0}^{(s-1)}+a_{q,\times}^{(s-1)}\gtrsim 1$, i.e., for any $s-1\leq T_{1,1}$:
\begin{align}
    \delta a_{q,0}^{(s-1)}+\delta a_{q,\times}^{(s-1)}\asymp \delta a_{q,\times}^{(s-1)}.
\end{align}
This shows \eqref{eq:a_q_x+a_q_0_=a_q_T11} and \eqref{eq:a_q_x+a_q_0_=a_q_x_T1} in Induction Hypothesis I holds at step $s$. By a similar argument, we can show \eqref{eq:xi_1_2_bound_T_11} in Induction Hypothesis I holds at step $s$.

Now we bound $\delta a_{b,1}^{(s-1)}$ to show \eqref{eq:delta_a_b_1_balanced_phase_1} holds at step $s$. We decompose $I_{h,1}$ defined in \eqref{eq:I_h_1_balanced} into two parts:
\begin{align}\label{eq:I_h_1_balanced_decomposition}
    I_{h,1}^{(t)}&\coloneqq \underbrace{\frac{h(h+1)}{(2h+1)^2}\left(a_{p,1}^{(t)}+a_{p,2}^{(t)}\right)\pi_{1,2h+1}^{(t)}(\down_\times)}_{\coloneqq I_{h,1,A}^{(t)}}\notag\\
    &\quad + \underbrace{\frac{h}{(2h+1)^2}\left(\frac{1}{k}\left(a_{p,0}^{(t)}-ha_{p,\up}^{(t)}\right)+(h+1)a_{p,2}^{(t)}\right)\pi_{1,2h+1}^{(t)}(\up)}_{\coloneqq I_{h,1,B}^{(t)}},
\end{align}
and thus by \eqref{eq:delta_a_b_1_simplified_balanced_phase_1} we have
\begin{align}\label{eq:delta_a_b_1_balanced_decomposition}
    \delta a_{b,1}^{(s-1)}&=\underbrace{\frac{1+o(1)}{N}\sum_{h=1}^{k-1}\PP^{(s-1)}\left(\gE_{2h}\right)\pi_{1,2h}^{(s-1)}(\up)I_{h,1,A}^{(s-1)}V_{1,2h+1}^{(s-1)}}_{\coloneqq \delta a_{b,1,A}^{(s-1)}}\notag\\
    &\quad + \underbrace{\frac{1+o(1)}{N}\sum_{h=1}^{k-1}\PP^{(s-1)}\left(\gE_{2h}\right)\pi_{1,2h}^{(s-1)}(\up)I_{h,1,B}^{(s-1)}V_{1,2h+1}^{(s-1)}}_{\coloneqq \delta a_{b,1,B}^{(s-1)}}.
\end{align}
From \eqref{eq:pi_2h-1_down_times_phase1.1}, \eqref{eq:pi_2h_down_times_phase1.1} and \eqref{eq:rho_ratio_T_11_expression} we deduce 
\begin{align}\label{eq:pi_2h_down_times>pi_2h_down_times^1}
    \pi_{1,2h+1}^{(t)}(\down_\times)\gtrsim \pi_{1,2h}^{(t)}(\down_\times^1),
\end{align}
and thus by \eqref{eq:a_p_1_2_tiny_T_11}, \eqref{eq:I_h_1_balanced_decomposition}, \eqref{eq:delta_a_b_1_balanced_decomposition} and \eqref{eq:delta_a_p_1_simplified_balanced_phase_1} we have
\begin{align}\label{eq:delta_a_b_1_A_bound_T_11}
    |\delta a_{b,1,A}^{(s-1)}|\lesssim \frac{\ln k}{Nk}\delta a_{p,1}^{(s-1)}.
\end{align}
From \eqref{eq:a_p_0_up_bound_T_1}, \eqref{eq:a_p_1_2_tiny_T_11}, \eqref{eq:delta_a_q_x_T_11_bound_2}, \eqref{eq:I_h_1_balanced_decomposition}, \eqref{eq:delta_a_b_1_balanced_decomposition} and \eqref{eq:pi_1_2h-1_up=o(1)} we have
\begin{align}\label{eq:delta_a_b_1_B_bound_T_11}
    |\delta a_{b,1,B}^{(s-1)}|\lesssim \frac{\ln k}{Nk^{1+c}}\delta a_{q,\times}^{(s-1)}.
\end{align}
The above two expressions together suggest \eqref{eq:delta_a_b_1_balanced_phase_1} holds at step $s$.

Similarly, to bound \eqref{eq:delta_a_b_0_balanced_phase_1}, we decompose $I_{h,0}$ defined in \eqref{eq:I_h_0_balanced} into two parts:
\small
\begin{align}\label{eq:I_h_0_balanced_decomposition}
    I_{h,0}^{(t)}&\coloneqq \underbrace{\frac{1}{4h^2}\left(\frac{2h-1}{k}a_{p,0}^{(t)}+ha_{p,2}^{(t)}+\frac{h-1}{k}a_{p,\up}^{(t)}\right)\left(1-\pi_{1,2h}^{(t)}(\up)\right)-\frac{h}{(2h+1)^2}\left(\frac{2}{k}\left(a_{p,0}^{(t)}+a_{p,\up}^{(t)}\right)+a_{p,2}^{(t)}\right)\pi_{1,2h+1}^{(t)}(\up)}_{\coloneqq I_{h,0,A}^{(t)}}\notag\\
    &\quad -\underbrace{\left(\frac{1}{4h}\left(a_{p,1}^{(t)}+a_{p,2}^{(t)}\right)\pi_{1,2h}^{(t)}(\down_\times^1)+\frac{h}{(2h+1)^2}\left(a_{p,1}^{(t)}+a_{p,2}^{(t)}\right)\pi_{1,2h+1}^{(t)}(\down_\times)\right)}_{\coloneqq I_{h,0,B}^{(t)}},
\end{align}
\normalsize
and thus by \eqref{eq:delta_a_b_0_simplified_balanced_phase_1} we have
\begin{align}\label{eq:delta_a_b_0_balanced_decomposition}
    \delta a_{b,0}^{(s-1)}&=\underbrace{\frac{1+o(1)}{N}\sum_{h=1}^{k-1}\PP^{(s-1)}\left(\gE_{2h}\right)\pi_{1,2h}^{(t)}(\up)I_{h,0,A}^{(s-1)}V_{1,2h+1}^{(s-1)}}_{\coloneqq \delta a_{b,0,A}^{(s-1)}}-\underbrace{\frac{1+o(1)}{N}\sum_{h=1}^{k-1}\PP^{(s-1)}\left(\gE_{2h}\right)\pi_{1,2h}^{(t)}(\up)I_{h,0,B}^{(s-1)}V_{1,2h+1}^{(s-1)}}_{\coloneqq \delta a_{b,0,B}^{(s-1)}}.
\end{align}
Then analogous to how we bound $\delta a_{b,1,B}^{(s-1)}$ in \eqref{eq:delta_a_b_1_B_bound_T_11}, we can bound $\delta a_{b,0,A}^{(s-1)}$ as
\begin{align*}
    |\delta a_{b,0,A}^{(s-1)}|\lesssim \frac{\ln k}{Nk}\delta a_{q,\times}^{(s-1)}.
\end{align*}
And analogous to how we bound $\delta a_{b,1,A}^{(s-1)}$ in \eqref{eq:delta_a_b_1_A_bound_T_11}, we can bound $\delta a_{b,0,B}^{(s-1)}$ as
\begin{align*}
    |\delta a_{b,0,B}^{(s-1)}|\lesssim \frac{\ln k}{Nk}\delta a_{p,1}^{(s-1)}.
\end{align*}
The above two expressions together suggest \eqref{eq:delta_a_b_0_balanced_phase_1} holds at step $s$. 

By \eqref{eq:delta_a_b_2_balanced_phase_1} we know that
\begin{align}\label{eq:a_b_2=-a_b_0-a_b_1/N-1}
    \forall t\in \NN:\quad a_{b,2}^{(t)}=\frac{-a_{b,0}^{(t)}-a_{b,1}^{(t)}}{N-1}.
\end{align}
Combining \eqref{eq:a_b_2=-a_b_0-a_b_1/N-1} with the fact that \eqref{eq:delta_a_b_0_balanced_phase_1} and \eqref{eq:delta_a_b_1_balanced_phase_1} hold at step $s$ we just proved, we immediately know that \eqref{eq:delta_a_b_2_balanced_phase_1} holds at step $s$.

Define
\begin{align}\label{eq:delta_e2a_c_i}
    \delta e^{2a_{c,i}^{(t)}}\coloneqq \frac{e^{2a_{c,i}^{(t+1)}}-e^{2a_{c,i}^{(t)}}}{\eta}=\frac{e^{2a_{c,i}^{(t)}}}{\eta}\left(e^{2\eta\delta a_{c,i}^{(t)}}-1\right)\geq \frac{e^{2a_{c,i}^{(t)}}}{\eta}\cdot 2\eta\delta a_{c,i}^{(t)}=2e^{2a_{c,i}^{(t)}}\delta a_{c,i}^{(t)}
\end{align}
for $i=0,1$.
Then by \eqref{eq:delta_a_c_1_simplified_balanced_phase_1} we have for any $t\leq T_{1,1}$:
\begin{align}\label{eq:delta_e2a_c_1_bound_T_11}
    \delta e^{2a_{c,1}^{(t)}}&\geq\sum_{h=1}^{k-1} \PP^{(t)}\left(\gE_{1,2h}\right)\pi_{1,2h}^{(t)}(\up)\frac{2}{\left(1+e^{-2a_{c,1}^{(t)}}\right)^2}\left(a_{q,\times}^{(t)}+a_{q,0}^{(t)}\right)(1-\pi_{1,2h}^{(t)}(\up))V_{1,2h+1}^{(t)}\notag\\
    &\overset{\eqref{eq:delta_a_q_x_simplified_balanced_phase_1}}\geq\max\left\{\left(a_{q,\times}^{(t)}+a_{q,0}^{(t)}\right)\delta a_{q,\times}^{(t)},0\right\}.
\end{align}
Thus by \eqref{eq:delta_a_c_1_balanced_phase_1} and \eqref{eq:a_q_x+a_q_0_=a_q_T11} in Induction Hypothesis I, we have
\begin{align}
    e^{2a_{c,1}^{(s)}}-e^{2a_{c,1}^{(s')}}\geq \left(a_{q,\times}^{(s')}+a_{q,0}^{(s')}\right)\left(a_{q,\times}^{(s)}-a_{q,\times}^{(s')}\right),
\end{align}
holds for any $s'\leq s$, i.e., \eqref{eq:delta_a_c_1_balanced_phase_1} in Induction Hypothesis I holds at step $s$. Also,
\begin{align}
    \delta e^{2a_{c,0}^{(t)}}&\overset{\eqref{eq:delta_a_c_0_simplified_balanced_phase_1}}>0,
\end{align}
and thus \eqref{eq:delta_a_c_0_balanced_phase_1} in Induction Hypothesis I holds at step $s$.

\paragraph{At $t=T_{1,1}$.} If $t=T_{1,1}$, by \eqref{eq:T_1_1} and \eqref{eq:xi_2h=a_q_x} we have
\begin{align}\label{eq:pi_1_2=1/2_T11}
    \pi_{1,2}^{(T_{1,1})}(\up)=\frac{1+o(1)}{2},
\end{align}
and by \eqref{eq:T_1_1} and \eqref{eq:a_q_x_bound_T_11}, we have
\begin{align}\label{eq:T_1_1_bound}
    T_{1,1}\lesssim 
    \frac{k^2\ln k}{\eta}.
\end{align}
Also note that by our data distribution assumption, the probability of finding the goal within $2h-1$ steps is less than $\frac{h}{k}$. And 
$$\PP\left(\gE_{1,2h}^{(T_{1,1})}\right)\leq \left(\frac{1+o(1)}{2}\right)^{h-1}\leq \left(\frac{2}{3}\right)^{h-1}$$
where we use \eqref{eq:pi_1_2h_up_bound_T_11} and \eqref{eq:pi_1_2=1/2_T11}.
Thus letting $h=\floor{\frac{\ln k}{\ln (3/2)}}$, we have the success rate
\begin{align}\label{eq:success_rate_T_11}
    R(\pi^{(T_{1,1})};\gP_1)\lesssim \frac{\ln k}{k}.
\end{align}
By \eqref{eq:e2a_c_1>a_q_x^2} we know that 
\begin{align}\label{eq:e2a_c_1>lnk^2_T_11_end}
    e^{2a_{c,1}^{(T_{1,1})}}\gtrsim \ln^2 k.
\end{align}

\paragraph{Case 2: $T_{1,1}+1\leq s-1\leq T_{1/2}$.} By \eqref{eq:delta_a_c_0_simplified_balanced_phase_1} and \eqref{eq:delta_a_p_1_simplified_balanced_phase_1} we know that \eqref{eq:delta_a_c_0_balanced_phase_1} and \eqref{eq:a_p_1_bound_T_11} hold at step $s$, respectively.
Note that \eqref{eq:delta_e2a_c_1_bound_T_11} still holds when $s-1>T_{1,1}$, and thus for Case 2, \eqref{eq:delta_a_c_1_balanced_phase_1} and \eqref{eq:e2a_c_1_bound_T_12} in Induction Hypothesis I holds at step $s$. 
Also note that \eqref{eq:e2a_c_1>a_q_x^2} still holds here, by which and Induction Hypothesis I we know that when $s-1\geq T_{1,1}+1$:
\begin{align}\label{eq:xi_2h_bound_phase_1.2}
    \forall h\in[k]:\quad \xi_{2h}^{(s-1)}=a_{q,\times}^{(s-1)}+O\left(\frac{1}{\ln k}\right),
\end{align}
and for all $h\in[k]$:
\begin{align}\label{eq:r_h_bound_phase_1.2}
    r_h^{(s-1)}=\left(1+o(1)\right)\exp\left(-\frac{1}{1+\frac{h-1}{h}e^{-2a_{c,0}^{(s-1)}}}\left(a_{q,\times}^{(s-1)}+a_{q,0}^{(s-1)}\right)+\frac{1}{2(2h-1)}a_{p,2}^{(s)}\right).
\end{align}
\eqref{eq:xi_2h_bound_phase_1.2} suggests \eqref{eq:xi_1_2_bound_T_11} in Induction Hypothesis I holds at step $s$;
\eqref{eq:r_h_bound_phase_1.2} combined with \eqref{eq:a_p_2<=1_T_12}, \eqref{eq:a_q_x+a_q_0_=lnk_T_12} (for $s-1\in [T_{1,1}+1,T_{1,2}]$) or \eqref{eq:a_q_0>=1.9lnk_T_12}, \eqref{eq:a_q_x-a_p_2_bound_T_12} (for $s-1\in [T_{1,2}+1,T_1]$) suggests \eqref{eq:pi_1_2h-1_up=o(1)} still holds for Case 2, i.e.,
\begin{align}\label{eq:pi_1_2h-1_up=o(1)_case2}
    \forall h\in[k]:\quad \pi_{1,2h-1}^{(s-1)}(\up)\leq \frac{1}{k^c} \text{ for some constant } c>0.
\end{align}

Note that for any $s-1\in[T_{1,1}+1,T_1]$, we have
\begin{align*}
    \pi_{1,1}(\down)\leq V_{1,1}^{(s-1)}\leq \pi_{1,1}(\down)+V_{1,3}^{(s-1)}= \frac{1-\pi_{1,1}^{(s-1)}(\up)}{k}+V_{1,3}^{(s-1)}\overset{\eqref{eq:pi_1_2h-1_up=o(1)_case2}}=\frac{1+o(1)}{k}+V_{1,3}^{(s-1)},
\end{align*}
and by Induction Hypothesis I and \eqref{eq:pi_1_2h-1_i_phase1}, we have for any $s-1\in[T_{1,1}+1,T_1]$,
\begin{align}\label{eq:V_lower_bound_1_over_k}
    \forall h\in[k]:\quad V_{1,2h-1}^{(s-1)}\geq 
    \pi_{1,2h-1}^{(s-1)}(\down)\gtrsim \frac{1}{k}.
\end{align}
The above two relations give
\begin{align}\label{eq:V_1_1_upper_bound_case2}
    V_{1,1}^{(s-1)}\lesssim V_{1,3}^{(s-1)}.
\end{align}
Furthermore, by \eqref{eq:pi_2h-1_up_phase1}, \eqref{eq:a_p_2<=1_T_12} we know that 
\begin{align}
    \text{If } T_{1,1}+1\leq s-1\leq T_{1,2}:\quad \pi_{1,3}^{(s-1)}(\up)\gtrsim \pi_{1,1}^{(s-1)}(\up).
\end{align}
Thus by \eqref{eq:V_1_1_upper_bound_case2}, we have
\begin{align}\label{eq:term1<=term3}
    \text{If } T_{1,1}+1\leq s-1\leq T_{1,2}:\quad \pi_{1,1}^{(s-1)}(\up)V_{1,1}^{(s-1)}\lesssim \PP\left(\gE_{1,2}\right)\pi_{1,3}^{(s-1)}(\up)V_{1,3}^{(s-1)},
\end{align}
where we use the fact that $\PP\left(\gE_{1,2}\right)=1+o(1)$ guaranteed by \eqref{eq:pi_1_2h-1_up=o(1)_case2}. Thus by \eqref{eq:term1<=term3}, \eqref{eq:delta_e2a_c_i} and \eqref{eq:delta_a_c_0_simplified_balanced_phase_1} and \eqref{eq:delta_a_q_0_simplified_balanced_phase_1}, we obtain
\begin{align}
    \text{If } T_{1,1}+1\leq s-1\leq T_{1,2}:\quad \delta e^{2a_{c,0}^{(s-1)}}&\gtrsim (a_{q,\times}^{(s-1)}+a_{q,0}^{(s-1)})\delta a_{q,0}^{(s-1)}.
\end{align}
This suggests \eqref{eq:delta_e2a_c_0_bound_T_12} in Induction Hypothesis I holds at step $s$.

Moreover, if $s-1\in[T_{1,2}+1,T_{1/2}]$, then by \eqref{eq:a_q_0>=1.9lnk_T_12} in Induction Hypothesis I we deduce 
\begin{align*}
    a_{q,0}^{(T_{1,2})}\geq 1.9\ln k-1,
\end{align*}
From \eqref{eq:a_q_x+a_q_0_=lnk_T_12} we know that
\begin{align*}
    a_{q,0}^{(T_{1,1})}+a_{q,\times}^{(T_{1,1})}\asymp \ln k.
\end{align*}
and by \eqref{eq:a_q_0<=2/3a_q_x_T11} and our choice of $T_{1,1}$ we know that
\begin{align*}
    a_{q,0}^{(T_{1,1})}\leq 2/3 \ln k +C.
\end{align*}
Combining the above three inequalities with \eqref{eq:delta_e2a_c_0_bound_T_12} in Induction Hypothesis I we have
\begin{align}\label{eq:e2a_c_0_bound_T_12_T_12}
    e^{2a_{c,0}^{(T_{1,2})}}\geq e^{2a_{c,0}^{(T_{1,2})}}-e^{2a_{c,0}^{(T_{1,1})}}\gtrsim \ln^2 k.
\end{align}
Thus by \eqref{eq:delta_a_c_0_balanced_phase_1} we know that for any $T_{1,2}+1\leq s-1\leq T_1$, we have 
\begin{align}\label{eq:e2a_c_0_lb_phase1}
    e^{2a_{c,0}^{(s)}}\gtrsim \ln^2 k.
\end{align}
On the other hand, by \eqref{eq:delta_a_c_0_simplified_balanced_phase_1}, \eqref{eq:delta_a_q_0_simplified_balanced_phase_1}, \eqref{eq:delta_a_q_x_simplified_balanced_phase_1} and Induction Hypothesis I, we have
\begin{align}
    \forall t\leq s-1:\quad \delta e^{2a_{c,0}^{(t)}}\lesssim \left(a_{q,\times}^{(t)}+a_{q,0}^{(t)}\right)\left(\delta a_{q,0}^{(t)}+\delta a_{q,\times}^{(t)}\right),
\end{align}
which together with \eqref{eq:a_q_0+a_q_x=C_lnk_phase_1} suggests 
\begin{align}\label{eq:e2a_c_0_ub_phase1}
    e^{2a_{c,0}^{(s)}}\lesssim \ln^2 k.
\end{align}
Thus by \eqref{eq:e2a_c_0_lb_phase1} and \eqref{eq:e2a_c_0_ub_phase1} we know that \eqref{eq:e2a_c_0>=lnk^2_T_12} holds at step $s$.

By \eqref{eq:delta_a_q_0+delta_a_q_x_T_11_expression} (which holds for any $s-1\in[T_{1,1}+1,T_{1/2}]$) 
we have
\eqref{eq:a_q_x+a_q_0_increase} holds at step $s$.
And by comparing \eqref{eq:delta_a_q_0+delta_a_q_x_T_11_expression} and \eqref{eq:delta_a_q_0_simplified_balanced_phase_1} we know that for any $s-1\in[T_{1,1}+1,T_{1/2}-1]$,
\begin{align}
    -\delta a_{q,0}^{(s-1)}\lesssim \frac{1}{e^{2a_{c,1}^{(s-1)}}}\left(\delta a_{q,\times}^{(s-1)}+\delta a_{q,0}^{(s-1)}\right).
\end{align}
This suggests \eqref{eq:-a_q_0_bound_T_12} in Induction Hypothesis I holds at step $s$. Similarly, by comparing the gradient expressions of $a_{p,1}^{(t)}$ in \eqref{eq:delta_a_p_1_simplified_balanced_phase_1} and $a_{p,\up}^{(t)}$ in \eqref{eq:delta_a_p_up_simplified_balanced_phase_1} with \eqref{eq:delta_a_q_0+delta_a_q_x_T_11_expression} and by \eqref{eq:a_q_x+a_q_0_=lnk_T_12}, \eqref{eq:a_q_0+a_q_x=C_lnk_phase_1}, we know that \eqref{eq:a_p_0_up_bound_T_1} holds at step s for any $s-1\in[T_{1,1}+1,T_{1/2}]$. By comparing \eqref{eq:delta_a_p_3_simplified_balanced_phase_1} and \eqref{eq:delta_a_q_0+delta_a_q_x_T_11_expression}, we have
\begin{align}\label{eq:delta_a_p_3_small_T12}
    \forall t\in[T_{1,1}+1,T_{1/2}]:\quad |\delta a_{p,3}^{(t)}|\lesssim \frac{1}{k}\left(\delta a_{q,\times}^{(t)}+\delta a_{q,0}^{(t)}\right),
\end{align}
and thus by \eqref{eq:delta_a_p_3_small_T11}, \eqref{eq:delta_a_p_3_small_T12}, \eqref{eq:a_q_0+a_q_x=C_lnk_phase_1} and  \eqref{eq:a_q_x+a_q_0_=a_q_x_T1} we know that 
\begin{align}\label{eq:a_p_3}
    -\frac{\ln k}{k}\lesssim a_{p,3}^{(s)}\coloneqq \eta\sum_{t=0}^{s-1}\delta a_{p,3}^{(t)}\lesssim \frac{\ln k}{k}.
\end{align}
and by \eqref{eq:a_p3} we have 
\begin{align}\label{eq:a_p3_1}
    a_{p,2}^{(s)}=\frac{a_{p,1}^{(s)}-a_{p,3}^{(s)}}{k-1}.
\end{align}
From the above two expressions and our choice of $T_{1,2}$ we deduce that when $T_{1,1}+1\leq s-1\leq T_{1,2}$, \eqref{eq:a_p_2<=1_T_12} holds at step $s$; when $T_{1,2}+1\leq s-1\leq T_{1/2}-1$, \eqref{eq:a_p_2_bound_T_12} holds at step $s$.

Now we show \eqref{eq:delta_a_b_1_balanced_phase_1} in Induction Hypothesis I holds at step $s$ for Case 2. Note by \eqref{eq:pi_1_2h_i_phase1}, \eqref{eq:pi_1_2h-1_i_phase1} and Induction Hypothesis I, \eqref{eq:pi_2h_down_times>pi_2h_down_times^1} still holds for any $s-1\in[T_{1,1}+1,T_{1/2}-1]$. By our choice of $T_{1/2}$, \eqref{eq:a_p_1_bound_T_11}, \eqref{eq:a_p_2<=1_T_12} or \eqref{eq:a_p_2_bound_T_12} we have
\begin{align*}
    \forall t\in[T_{1,1}+1,s-1]:\quad -\frac{\ln k}{k^2}\lesssim a_{p,1}^{(t)}+a_{p,2}^{(t)}\lesssim k\ln k,
\end{align*}
and 
\begin{align*}
    a_{p,1}^{(t)}+a_{p,2}^{(t)}>0
\end{align*}
after $a_{p,1}^{(t)}\geq 2$.
By the above two relations and \eqref{eq:I_h_1_balanced_decomposition}, \eqref{eq:delta_a_b_1_balanced_decomposition}, \eqref{eq:pi_2h_down_times>pi_2h_down_times^1} and \eqref{eq:delta_a_p_1_simplified_balanced_phase_1} we have
\begin{align}
    \forall t\in[T_{1,1}+1,s-1]:\quad -\frac{\ln k}{Nk}\delta a_{p,1}^{(t)}\lesssim \delta a_{b,1,A}^{(t)}\lesssim \frac{k^2 \ln k}{N}\delta a_{p,1}^{(t)},
\end{align}
and 
\begin{align}
    \delta a_{b,1,A}^{(t)}>0
\end{align}
after $a_{p,1}^{(t)}\geq 2$.
Thus by the above two relations and our choice of $T_{1/2}$ we have
\begin{align}\label{eq:a_b_1_A_bound_case_2}
    -\frac{\ln k}{Nk}\lesssim a_{b,1,A}^{(s)}-a_{b,1,A}^{(T_{1,1})}\lesssim \frac{k^3 \ln^2 k}{N}.
\end{align}
By \eqref{eq:a_p_0_up_bound_T_1}, \eqref{eq:a_p_2<=1_T_12} or \eqref{eq:a_p_2_bound_T_12} and our choice of $T_{1/2}$ we have
\begin{align}\label{eq:I_h_1_B_bound_case_2}
    \forall t\in[T_{1,1}+1,s-1]:\quad -\frac{\ln k}{k}\lesssim \frac{h}{(2h+1)^2}\left(\frac{1}{k}\left(a_{p,0}^{(t)}-ha_{p,\up}^{(t)}\right)+(h+1)a_{p,2}^{(t)}\right)\lesssim \ln k,
\end{align}
Note that summing up \eqref{eq:delta_a_q_0+delta_a_q_x_T_11_expression} and \eqref{eq:delta_a_q_0_simplified_balanced_phase_1} gives us
\begin{align}
    2\delta a_{q,0}^{(t)}+\delta a_{q,\times}^{(t)}&=2\pi_{1,1}^{(t)}(\up)V_{1,1}^{(t)}+ \sum_{h=1}^{k-1} \PP^{(t)}\left(\gE_{1,2h}\right)\pi_{1,2h}^{(t)}(\up)\bigg(\frac{2-\frac{h}{h+1}e^{-2a_{c,0}^{(t)}}}{1+\frac{h}{h+1}e^{-2a_{c,0}^{(t)}}}\pi_{1,2h+1}^{(t)}(\up)\notag\\
    &\qquad +\frac{1-2e^{-2a_{c,1}^{(t)}}}{1+e^{-2a_{c,1}^{(t)}}}\left(1-\pi_{1,2h}^{(t)}(\up)\right)\bigg)V_{1,2h+1}^{(t)}\notag\\
    &\overset{\eqref{eq:e2a_c_1_bound_T_12}}\geq \frac{1}{2}\sum_{h=1}^{k-1} \PP^{(t)}\left(\gE_{1,2h}\right)\pi_{1,2h}^{(t)}(\up)\pi_{1,2h+1}^{(t)}(\up)V_{1,2h+1}^{(t)}>0
\end{align}
for any $t\in[T_{1,1}+1,s-1]$.
Thus by \eqref{eq:delta_a_b_1_balanced_decomposition}, \eqref{eq:I_h_1_B_bound_case_2} and the above inequality, we have
\begin{align}\label{eq:delta_a_b_1_B_bound_case_2}
    \forall t\in[T_{1,1}+1,s-1]:\quad -\frac{\ln k}{Nk}\left( 2\delta a_{q,0}^{(t)}+\delta a_{q,\times}^{(t)}\right)\lesssim \delta a_{b,1,B}^{(t)}\lesssim \frac{\ln k}{N}\left( 2\delta a_{q,0}^{(t)}+\delta a_{q,\times}^{(t)}\right),
\end{align}
which, combined with \eqref{eq:a_q_0<=a_q_x_bound_T_11} and \eqref{eq:a_q_x+a_q_0_=a_q_x_T1}, \eqref{eq:a_q_0+a_q_x=C_lnk_phase_1} or \eqref{eq:a_q_x+a_q_0_=lnk_T_12}, gives us
\begin{align}\label{eq:a_b_1_B_bound_case_2}
    -\frac{\ln^2 k}{Nk}\lesssim a_{b,1,B}^{(s)}-a_{b,1,B}^{(T_{1,1})}\lesssim \frac{\ln^2 k}{N}.
\end{align}
By \eqref{eq:a_b_1_A_bound_case_2} and \eqref{eq:a_b_1_B_bound_case_2} and that \eqref{eq:delta_a_b_1_balanced_phase_1} holds at step $T_{1,1}$ by Induction Hypothesis I, we know that \eqref{eq:delta_a_b_1_balanced_phase_1} holds at step $s$.

Analogous to how we bound $\delta a_{b,1,B}^{(s-1)}$ in \eqref{eq:delta_a_b_1_B_bound_case_2}, we can bound $\delta a_{b,0,A}^{(s-1)}$ as
\begin{align*}
   \forall t\in[T_{1,1}+1,s-1]:\quad |\delta a_{b,0,A}^{(t)}|\lesssim \frac{\ln k}{N}\left(\delta a_{q,\times}^{(t)}+\delta a_{q,0}^{(t)}\right),
\end{align*}
and thus by \eqref{eq:a_q_0+a_q_x=C_lnk_phase_1} or \eqref{eq:a_q_x+a_q_0_=lnk_T_12} we have
\begin{align*}
    |a_{b,0,A}^{(s)}-a_{b,0,A}^{(T_{1,1})}|\lesssim \frac{\ln^2 k}{N}.
\end{align*}
And analogous to how we bound $a_{b,1,A}^{(s-1)}$ in \eqref{eq:a_b_1_A_bound_case_2}, here we can bound $ a_{b,0,B}^{(s-1)}$ as
\begin{align*}
    -\frac{\ln k}{Nk}\lesssim a_{b,0,B}^{(s)}-a_{b,0,B}^{(T_{1,1})}\lesssim \frac{k^3 \ln^2 k}{N}.
\end{align*}
The above two expressions together with \eqref{eq:delta_a_b_0_balanced_decomposition} and that 
\eqref{eq:delta_a_b_0_balanced_phase_1} holds at step $T_{1,1}$ imply that \eqref{eq:delta_a_b_0_balanced_phase_1} holds at step $s$. Combining \eqref{eq:a_b_2=-a_b_0-a_b_1/N-1} with the fact that \eqref{eq:delta_a_b_0_balanced_phase_1} and \eqref{eq:delta_a_b_1_balanced_phase_1} hold at step $s$ we just proved, we immediately know that \eqref{eq:delta_a_b_2_balanced_phase_1} holds at step $s$.

Now we show \eqref{eq:a_p_1_bound_T_12} in Induction Hypothesis I holds at step $s$. Note that it implies \eqref{eq:a_p_1_bound_T_11} in Induction Hypothesis I holds at step $s$ for any $s-1\in[T_{1,1}+1,T_{1/2}]$. We consider two sub-cases: (i) $a_{p,1}^{(s-1)}\leq 5\ln k$ and (ii) $a_{p,1}^{(s-1)}\geq 5\ln k$.
\begin{enumerate}[label=(\roman*)]
    \item $a_{p,1}^{(s-1)}\leq 5\ln k$.
If $a_{p,1}^{(s-1)}\leq 5\ln k$, then by \eqref{eq:pi_1_2h-1_i_phase1}, \eqref{eq:a_p_2<=1_T_12} and \eqref{eq:a_p_2_bound_T_12} we have for any $s-1\in[T_{1,1}+1,T_{1/2}]$:
\begin{align}
    \pi_{1,5}^{(s-1)}(\down_\times)\geq \frac{1+o(1)}{k^2},
\end{align}
and thus by  \eqref{eq:delta_a_p_1_simplified_balanced_phase_1}, we have
\begin{align}\label{eq:delta_a_p_1_bound_a_p_1_(i)}
    \text{If } a_{p,1}^{(s-1)}\leq 5\ln k:\quad \delta a_{p,1}^{(s-1)}\gtrsim \frac{1}{k}\PP^{(s-1)}(\gE_{1,4})\pi_{1,4}^{(s-1)}(\up)\pi_{1,5}^{(s-1)}(\down_\times)V_{1,5}^{(s-1)}\gtrsim \frac{1}{k^4}.
\end{align}
where we use the fact that $\PP(\gE_{1,4})\gtrsim 1$, $\pi_{1,4}^{(s-1)}(\up)\gtrsim 1$, and $V_{1,5}^{(s-1)}\gtrsim \frac{1}{k}
$ guaranteed by \eqref{eq:V_lower_bound_1_over_k}, Induction Hypothesis I and our choice of $T_{1,1}$. \eqref{eq:delta_a_p_1_bound_a_p_1_(i)} suggests \eqref{eq:a_p_1_bound_T_12} holds at step $s$  when $a_{p,1}^{(s-1)}\leq 5\ln k$.

\item $a_{p,1}^{(s-1)}\geq 5\ln k$.
We first use contradiction to show that 
\begin{align}\label{eq:a_q_0+a_q_x>=4lnk_phase1}
    \text{If } a_{p,1}^{(s-1)}\geq 5\ln k:\quad a_{q,0}^{(s-1)}+a_{q,\times}^{(s-1)}\geq 4\ln k.
\end{align}
Suppose on the contrary that \eqref{eq:a_q_0+a_q_x>=4lnk_phase1} does not hold, then by \eqref{eq:a_q_x+a_q_0_increase} we know that for any $t\leq s-1$, we have 
\begin{align}\label{eq:a_q_0+a_q_x<4lnk}
    a_{q,0}^{(t)}+a_{q,\times}^{(t)}< 4\ln k.
\end{align}
Since $a_{p,1}^{(t)}\lesssim \frac{\ln k}{k^2}$ for any $t\leq T_{1,1}$ (c.f.~\eqref{eq:a_p_1_2_tiny_T_11}),
there exists $t\in [T_{1,1}+1,T_{1,2}]$ such that 
\begin{align}\label{eq:a_p_1_4lnk<=a_p_1<=5lnk}
    4\ln k\leq a_{p,1}^{(t)}\leq 5\ln k.
\end{align}
Then for any $t'\in[T_{1,1},t]$:
\begin{itemize}
    \item If $a_{q,\times}^{(t')}< 3 \ln k$, then by \eqref{eq:pi_2h_up_phase1}, \eqref{eq:e2a_c_1>a_q_x^2}, \eqref{eq:a_p_2<=1_T_12} and\eqref{eq:a_p_1_4lnk<=a_p_1<=5lnk} we know that we have
    \begin{align}
   1-\pi_{1,2h}^{(t')}(\up)\geq \frac{1+o(1)}{k^2},\,\,\forall h\in[k-1].
\end{align}
And by \eqref{eq:e2a_c_1>a_q_x^2} (which still holds here) we have
\begin{align}\label{eq:e2a_c_1_lnk^2}
    e^{2a_{c,1}^{(t')}}\gtrsim \ln^2 k.
\end{align}
Thus by \eqref{eq:delta_a_q_0+delta_a_q_x_T_11_expression} and the above two expressions, we have for all $t'\in[T_{1,1},t]$:
\begin{align}\label{eq:a_q_x<3lnk_delta_a_q_0+delta_a_q_x}
    \text{If } a_{q,\times}^{(t')}< 3 \ln k:\quad \delta a_{q,0}^{(t')}+\delta a_{q,\times}^{(t')}\geq \frac{1+o(1)}{k^2}\sum_{h=1}^{k-1}\PP^{(t')}\left(\gE_{1,2h}\right)\pi_{1,2h}^{(t')}(\up)V_{1,2h+1}^{(t')}.
\end{align}
    \item If $a_{q,\times}^{(t')}\geq 3\ln k$, then by \eqref{eq:a_q_x-a_q_0_bound_T_12} in Induction Hypothesis I and \eqref{eq:a_q_0+a_q_x<4lnk}, we have
    \begin{align}\label{eq:a_q_0=ln k}
        0.9\ln k\leq a_{q,0}^{(t')}< \ln k.
    \end{align}
    Thus by \eqref{eq:pi_2h-1_up_phase1}, \eqref{eq:a_q_0=ln k}, \eqref{eq:a_p_2<=1_T_12} and \eqref{eq:a_p_1_4lnk<=a_p_1<=5lnk} we know that 
    \begin{align}\label{eq:pi_1_2h-1_up<=1/k^2}
        \pi_{1,2h-1}^{(t')}(\up)\geq \frac{1+o(1)}{k^2},\,\,\forall h\in[k].
     \end{align}
     Additionally, by \eqref{eq:delta_e2a_c_0_bound_T_12} in Induction Hypothesis I, \eqref{eq:a_q_0=ln k} and \eqref{eq:e2a_c_0>=lnk^2_T_12}, we have
     \begin{align}\label{eq:e2a_c_0_lnk^2}
        e^{2a_{c,0}^{(t')}}\gtrsim \ln^2 k.
     \end{align}
     Thus by \eqref{eq:delta_a_q_0+delta_a_q_x_T_11_expression} and the above two expressions, we have for all $t'\in[T_{1,1},t]$:
     \begin{align}\label{eq:a_q_x>=3lnk_delta_a_q_0+delta_a_q_x}
        \text{If } a_{q,\times}^{(t')}\geq 3\ln k:\quad \delta a_{q,0}^{(t')}+\delta a_{q,\times}^{(t')}\geq \frac{1+o(1)}{k^2}\sum_{h=1}^{k-1}\PP^{(t')}\left(\gE_{1,2h}\right)\pi_{1,2h}^{(t')}(\up)V_{1,2h+1}^{(t')}.
     \end{align}
\end{itemize}

On the other hand, by \eqref{eq:delta_a_p_1_simplified_balanced_phase_1}, we have for all $t'\in[T_{1,1},t]$:
\begin{align}\label{eq:delta_a_p_1<delta_a_q_0+delta_a_q_x}
    \delta a_{p,1}^{(t')} 
    &\leq (1+o(1))\frac{1}{k}\sum_{h=1}^{k-1}\PP^{(t')}\left(\gE_{1,2h}\right)\pi_{1,2h}^{(t')}(\up)\left(\frac{1}{2}\cdot\frac{1-\pi_{1,2h}^{(t')}(\up)}{k}+\frac{h}{2h+1}\cdot\frac{1-\pi_{1,2h+1}^{(t')}(\up)}{k}\right)V_{1,2h+1}^{(t')}\notag\\
    &\leq (1+o(1))\frac{1}{k}\sum_{h=1}^{k-1}\PP^{(t')}\left(\gE_{1,2h}\right)\pi_{1,2h}^{(t')}(\up)\left(\frac{1+o(1)}{4}\cdot\frac{1}{k}+\frac{h}{2h+1}\cdot\frac{1+o(1)}{k}\right)V_{1,2h+1}^{(t')}\notag\\
    &\leq \frac{3}{4}\frac{1+o(1)}{k^2}\sum_{h=1}^{k-1}\PP^{(t')}\left(\gE_{1,2h}\right)\pi_{1,2h}^{(t')}(\up)V_{1,2h+1}^{(t')}\notag\\
    &\leq \frac{3+o(1)}{4}\left(\delta a_{q,0}^{(t')}+\delta a_{q,\times}^{(t')}\right),
\end{align}
where the second inequality uses \eqref{eq:pi_1_2h-1_up=o(1)_case2}, $a_{p,1}^{(t')}\geq 0$ guaranteed by \eqref{eq:a_p_1_bound_T_11} in Induction Hypothesis I, and the fact that 
\begin{align}\label{eq:pi_1_2h_up_asymp_1}
    \pi_{1,2h}^{(t')}(\up)\geq \frac{1+o(1)}{2}
\end{align}
guaranteed by our choice of $T_{1,1}$ and \eqref{eq:xi_2h_bound_phase_1.2}, and the last inequality follows from \eqref{eq:a_q_x<3lnk_delta_a_q_0+delta_a_q_x}, \eqref{eq:a_q_x>=3lnk_delta_a_q_0+delta_a_q_x}.
\eqref{eq:delta_a_p_1<delta_a_q_0+delta_a_q_x} 
indicates that 
\begin{align*}
    a_{p,1}^{(t)}-a_{p,1}^{(T_{1,1})}\leq \frac{3(1+o(1))}{4}\left(\left(a_{q,0}^{(t)}+a_{q,\times}^{(t)}\right)-\left(a_{q,0}^{(T_{1,1})}+a_{q,\times}^{(T_{1,1})}\right)\right).
\end{align*}
By \eqref{eq:a_p_1_2_tiny_T_11} and \eqref{eq:a_q_x+a_q_0_=lnk_T_12} in Induction Hypothesis I, we have
\begin{align*}
a_{p,1}^{(T_{1,1})}\lesssim \frac{1}{k^2}\left(a_{q,0}^{(T_{1,1})}+a_{q,\times}^{(T_{1,1})}\right).
\end{align*}
Combining the above two expressions, we have
\begin{align*}
    a_{q,0}^{(t)}+a_{q,\times}^{(t)}>a_{p,1}^{(t)}\overset{\eqref{eq:a_p_1_4lnk<=a_p_1<=5lnk}}{\geq}4\ln k.
\end{align*}
This contradicts \eqref{eq:a_q_0+a_q_x<4lnk}. Therefore, \eqref{eq:a_q_0+a_q_x>=4lnk_phase1} must hold.

Now we show \eqref{eq:a_p_1_bound_T_12} in Induction Hypothesis I holds at step $s$ under \eqref{eq:a_q_0+a_q_x>=4lnk_phase1}. 
First note that for all $h\in[k-1]$:
\begin{align}\label{eq:PP_gE_1_2h_expression_phase_1}
    \PP\left(\gE_{1,2h}\right)=\underbrace{\frac{k-h}{k}}_{\text{(a)}}\underbrace{\prod_{i=1}^{h-1}\pi_{1,2i}^{(s-1)}(\up)}_{\text{(b)}}\underbrace{\prod_{i=1}^h \left(1-(i-1)\pi_{1,2i-1}^{(s-1)}(\down_\times)-\pi_{1,2i-1}^{(s-1)}(\up)\right)}_{\text{(c)}},
\end{align}
where (a) stands for the probability of the event that the goal is not among the first $h$ children visited. By \eqref{eq:a_q_0+a_q_x>=4lnk_phase1}, \eqref{eq:a_q_x-a_q_0_bound_T_12}, \eqref{eq:a_p_2<=1_T_12} (if $s-1\in[T_{1,1}+1,T_{1,2}-1]$) or \eqref{eq:a_q_x-a_p_2_bound_T_12}, \eqref{eq:a_p_2_bound_T_12} (if $s-1\in[T_{1,2}+1,T_{1/2}-1]$) we have that if $a_{p,1}^{(s-1)}\geq 5\ln k$, then
\begin{align}
    a_{q,\times}^{(s-1)}-\frac{1}{2}a_{p,2}^{(s-1)}\geq 2\ln k-2,
\end{align}
and hence by \eqref{eq:pi_2h_up_phase1} and \eqref{eq:xi_2h_bound_phase_1.2} we have
\begin{align}\label{eq:1-pi_1_2h_up<=1/k}
    \forall h\in[k-1]:\quad 1-\pi_{1,2h}^{(s-1)}(\up)\lesssim \frac{1}{k}.
\end{align}
From the above inequality we know that (b) in \eqref{eq:PP_gE_1_2h_expression_phase_1} satisfies
\begin{align}\label{eq:PP_gE_1_2h_expression_phase_1_b=1}
    \forall h\in[k]:\quad (b)\asymp 1.
\end{align}
Note that if $T_{1,1}+1\leq s-1\leq T_{1,2}$, then by \eqref{eq:a_q_0+a_q_x>=4lnk_phase1} and \eqref{eq:a_q_x-a_q_0_bound_T_12} in Induction Hypothesis I, we have
\begin{align}\label{eq:a_q_0_asymp_a_q_x_and_lnk}
    a_{q,0}^{(s-1)}\geq 0.95 \ln k,\quad a_{q,0}^{(s-1)}\asymp a_{q,\times}^{(s-1)},
\end{align}
and if $s-1\in[T_{1,1}+1,T_{1,2}]$, by \eqref{eq:a_q_0_asymp_a_q_x_and_lnk}, \eqref{eq:delta_e2a_c_0_bound_T_12}, \eqref{eq:a_q_0<=2/3a_q_x_T11} and \eqref{eq:a_q_x+a_q_0_=a_q_T11} and a similar argument as how we prove \eqref{eq:e2a_c_0_bound_T_12_T_12}, we have 
\begin{align}\label{eq:e2a_c_0_bound_T_12}
    e^{2a_{c,0}^{(s-1)}}
    \gtrsim \ln^2 k.
\end{align}
If $s-1\in [T_{1,2}+1,T_{1/2}-1]$, by \eqref{eq:e2a_c_0>=lnk^2_T_12} we know that the above inequality \eqref{eq:e2a_c_0_bound_T_12} still holds. And by Induction Hypothesis I, we have
\eqref{eq:a_q_0_asymp_a_q_x_and_lnk} holds for any $t$ where $a_{p,1}^{(t)}\in [5\ln k, k]$. Thus by 
 \eqref{eq:-a_q_0_bound_T_12}, \eqref{eq:e2a_c_1_bound_T_12} and \eqref{eq:a_q_x+a_q_0_=lnk_T_12}, we have if $s-1\in [T_{1,2}+1,T_{1/2}-1]$:
 \begin{align}\label{eq:a_q_0_asymp_a_q_x_and_lnk_C}
    a_{q,0}^{(s-1)}\geq 0.95 \ln k-C,\quad a_{q,0}^{(s-1)}\asymp a_{q,\times}^{(s-1)}
 \end{align}
 for some constant $C>0$.
Combining \eqref{eq:pi_2h-1_up_phase1}, \eqref{eq:a_q_0_asymp_a_q_x_and_lnk} or \eqref{eq:a_q_0_asymp_a_q_x_and_lnk_C} and \eqref{eq:e2a_c_0_bound_T_12} (or \eqref{eq:e2a_c_0>=lnk^2_T_12}), \eqref{eq:a_p_2<=1_T_12} or \eqref{eq:a_p_2_bound_T_12}, we obtain
\begin{align}\label{eq:pi_1_2h-1_up<=k^-0.95}
    \text{If } k-h\gtrsim k:\quad \pi_{1,2h-1}^{(s-1)}(\up)\lesssim k^{-1.95}.
\end{align}
Thus if 
\begin{align}\label{eq:pi_1_2i-1_leq_1_over_h2}
    \forall i\leq h-1:\quad \pi_{1,2i-1}^{(s-1)}(\down_\times)\leq \frac{4}{h^2},
\end{align}
then (c) in \eqref{eq:PP_gE_1_2h_expression_phase_1} satisfies
\begin{align}\label{eq:PP_gE_1_2h_expression_phase_1_c=1}
    \text{If } k-h\gtrsim k:\quad (c)\asymp 1.
\end{align}
Thus by \eqref{eq:PP_gE_1_2h_expression_phase_1},\eqref{eq:PP_gE_1_2h_expression_phase_1_b=1} and \eqref{eq:PP_gE_1_2h_expression_phase_1_c=1}, when $a_{p,1}^{(s-1)}\geq 5\ln k$, we have
\begin{align}\label{eq:when_PP_gE_1_2h_asymp_1}
    \text{If \eqref{eq:pi_1_2i-1_leq_1_over_h2} is satisfied}:\quad \PP\left(\gE_{1,2h}\right)\asymp 1,\,\,\forall h\in\left[\ceil{\frac{3k}{4}}\right].
\end{align}

By \eqref{eq:pi_1_2h-1_i_phase1} and the definition of $\pi_{1,2h-1}^{(t)}(\down_\times)$ at the beginning of the proof, we have for all $h\in[k]$:
\begin{align}\label{eq:pi_1_2h-1_down_times_bound_T_1}
    \pi_{1,2h-1}^{(s-1)}(\down_\times)
    &=\left(1+o(1)\right)\frac{\exp\left(-\frac{a_{p,1}^{(t)}+a_{p,2}^{(t)}}{2h-1}\right)}{(h-1)\exp\left(-\frac{a_{p,1}^{(t)}+a_{p,2}^{(t)}}{2h-1}\right)+k-h+1+\exp\left(-a_{q,0}^{(t)}-\frac{h-1}{2h-1}a_{p,2}^{(t)}\right)}\notag\\
    &\overset{\eqref{eq:a_q_0_asymp_a_q_x_and_lnk}}=\left(1+o(1)\right)\frac{\exp\left(-\frac{a_{p,1}^{(t)}+a_{p,2}^{(t)}}{2h-1}\right)}{(h-1)\exp\left(-\frac{a_{p,1}^{(t)}+a_{p,2}^{(t)}}{2h-1}\right)+k-h+1}\overset{\eqref{eq:a_p_1_bound_T_11}}\leq \frac{1+o(1)}{k},
\end{align}
which implies \eqref{eq:pi_1_2i-1_leq_1_over_h2} holds for $h=\ceil{\sqrt{k}}$. Thus we have
\begin{align}\label{eq:delta_a_p_1_bound_h=sqrt(k)}
    \delta a_{p,1}^{(s-1)}&\overset{\eqref{eq:delta_a_p_1_simplified_balanced_phase_1}}\geq \frac{1+o(1)}{k}\underbrace{\PP^{(s-1)}\left(\gE_{1,2h}\right)}_{\asymp 1 \text{ by \eqref{eq:when_PP_gE_1_2h_asymp_1}}}\underbrace{\pi_{1,2h}^{(s-1)}(\up)}_{\asymp 1 \text{ by \eqref{eq:pi_1_2h_up_asymp_1}} }\frac{h}{2h+1}\pi_{1,2h+1}^{(s-1)}(\down_\times)\underbrace{V_{1,2h+1}^{(s-1)}}_{\gtrsim 
    \frac{1}{k}
    \text{ by  \eqref{eq:V_lower_bound_1_over_k}}}\notag\\
    &\gtrsim \frac{1}{k^2} 
    \pi_{1,2h+1}^{(s-1)}(\down_\times),\quad\text{for } h=\ceil{\sqrt{k}}.
\end{align}
Then by \eqref{eq:pi_1_2h-1_down_times_bound_T_1}, \eqref{eq:delta_a_p_1_bound_h=sqrt(k)} and \eqref{eq:a_p_2<=1_T_12}, we have
\begin{align}\label{eq:delta_a_p_1_bound_a_p_1_initial}
    \text{When } a_{p,1}^{(s-1)}\leq 4\ceil{\sqrt{k}}\ln k:\quad \delta a_{p,1}^{(s-1)}\gtrsim \frac{1}{k^5}. 
\end{align}
Note that from \eqref{eq:pi_1_2h-1_down_times_bound_T_1} we also know that \eqref{eq:pi_1_2i-1_leq_1_over_h2} holds for all $i\leq h\leq \ceil{\frac{3k}{4}}$ if
\begin{align}\label{eq:sufficient_condition_for_pi_1_2i-1_leq_1_over_h2}
    a_{p,1}^{(s-1)}\geq (2h-1)\ln k.
\end{align}
Therefore, when $a_{p,1}^{(s-1)}\geq 4\ceil{\sqrt{k}}\ln k$, by \eqref{eq:sufficient_condition_for_pi_1_2i-1_leq_1_over_h2} we have \eqref{eq:pi_1_2i-1_leq_1_over_h2} holds for $h=2\ceil{\sqrt{k}}$, and thus by a similar argument as we proved \eqref{eq:delta_a_p_1_bound_a_p_1_initial}, we can show that 
\begin{align}
    \text{When } 4\ceil{\sqrt{k}}\ln k\leq a_{p,1}^{(s-1)}\leq 8\ceil{\sqrt{k}}\ln k:\quad \delta a_{p,1}^{(s-1)}\gtrsim \frac{1}{k^5}. 
\end{align}
We set $m\in\NN_+$ such that 
\begin{align}
    2^{m-1}\ceil{\sqrt{k}}\leq\ceil{\frac{3k}{2}}< 2^m\ceil{\sqrt{k}}.
\end{align}
Then by repeating the above argument, we can show that for all $m'\in\{3,\cdots,m\}:$
\begin{align}\label{eq:delta_a_p_1_bound_a_p_1_case2_general}
    \text{When } 2^{m'-1}\ceil{\sqrt{k}}\ln k\leq a_{p,1}^{(s-1)}\leq 2^{m'}\ceil{\sqrt{k}}\ln k:\quad \delta a_{p,1}^{(s-1)}\gtrsim \frac{\gamma^{2^{m'-1}\sqrt{k}}}{k^5}.
\end{align}
Finally, when $a_{p,1}^{(s-1)}\geq 2^{m}\ceil{\sqrt{k}}\ln k\geq\ceil{\frac{3k}{2}}\ln k$, then by \eqref{eq:sufficient_condition_for_pi_1_2i-1_leq_1_over_h2} and \eqref{eq:when_PP_gE_1_2h_asymp_1}, we have 
$$\PP\left(\gE_{1,2h}\right)\asymp 1 \text{ when } h=\ceil{\frac{3k}{4}}.$$
Thus similar as \eqref{eq:delta_a_p_1_bound_h=sqrt(k)}, we have
\begin{align}\label{eq:delta_a_p_1_bound_a_p_1_2sqrt(k)lnk_case2}
    \delta a_{p,1}^{(s-1)}\gtrsim \frac{1}{k}\pi_{1,2\ceil{\frac{3k}{4}}+1}^{(s-1)}(\down_\times)V_{1,2\ceil{\frac{3k}{4}}+1}^{(s-1)}. 
\end{align}
Note that when $\ceil{\frac{3k}{2}}\ln k\leq a_{p,1}^{(s-1)}\leq 4.1k\ln k$, by \eqref{eq:a_p_2_bound_T_12} we have
\begin{align}\label{eq:pi_1_2i+1_down_times>=1/k^(11/3)}
    \pi_{1,2\ceil{\frac{3k}{4}}+1}^{(s-1)}(\down_\times)\overset{\eqref{eq:pi_1_2h-1_down_times_bound_T_1}}\gtrsim \frac{1}{k^{11.2/3}},
\end{align}
and for all $i\in\left[\ceil{\frac{3k}{4}},\ceil{\frac{3k}{4}}+\ceil{k^{2.2/3}}\right]$:
\begin{align}\label{eq:pi_1_2i+1_down<=1/k^2}
    \pi_{1,2i+1}^{(s-1)}(\down_\times)\overset{\eqref{eq:pi_1_2h-1_down_times_bound_T_1}}\lesssim \frac{1}{k^2}.
\end{align}
which gives 
\begin{align}\label{eq:V_1_2i+1_bound_2}
    V_{1,2\ceil{\frac{3k}{4}}+1}^{(s-1)}&\geq\sum_{j=\ceil{\frac{3k}{4}}}^{\ceil{\frac{3k}{4}}+\ceil{k^{2.2/3}}}
    \frac{1}{k} 
    \underbrace{\prod_{i=\ceil{\frac{3k}{4}}}^{j}\left(1-(k-i)\pi_{1,2i+1}^{(s-1)}(\down_\times)-\pi_{1,2i+1}^{(s-1)}(\up)\right)}_{\asymp 1 \text{ by \eqref{eq:pi_1_2i+1_down<=1/k^2} and \eqref{eq:pi_1_2h-1_up<=k^-0.95}}}\underbrace{\prod_{i=\ceil{\frac{3k}{4}}+1}^{j}\pi_{1,2i}^{(s-1)}(\up)}_{\asymp 1 \text{ by \eqref{eq:1-pi_1_2h_up<=1/k}}}  \notag \\
    & \gtrsim \frac{1}{k^{0.8/3}}. 
\end{align}
Plugging \eqref{eq:pi_1_2i+1_down_times>=1/k^(11/3)}, \eqref{eq:V_1_2i+1_bound_2} into \eqref{eq:delta_a_p_1_bound_a_p_1_2sqrt(k)lnk_case2}, we have
\begin{align}\label{eq:delta_a_p_1_bound_a_p_1_case2_final}
    \delta a_{p,1}^{(s-1)}\gtrsim \frac{1}{k^5}. 
\end{align}
By \eqref{eq:delta_a_p_1_bound_a_p_1_initial}, \eqref{eq:delta_a_p_1_bound_a_p_1_case2_general} and \eqref{eq:delta_a_p_1_bound_a_p_1_case2_final}, we know that \eqref{eq:a_p_1_bound_T_12} holds at step $s$ also for subcase (ii) (i.e., when $a_{p,1}^{(s-1)}\geq 5\ln k$).
\end{enumerate}

When $T_{1,2}+1\leq s-1\leq T_{1/2}-1$, if \eqref{eq:a_q_x-a_p_2_bound_T_12}, \eqref{eq:a_p_2_bound_T_12} and \eqref{eq:a_q_0+a_q_x=C_lnk_phase_1} hold at step $s$, then \eqref{eq:a_q_0<=a_q_x_bound_T_11} and \eqref{eq:a_q_x+a_q_0_=a_q_x_T1} hold at step $s$.


Note that when $T_{1,1}+1\leq s-1\leq T_{1,2}$, 
the first relation in \eqref{eq:a_q_x-a_q_0_bound_T_12} can indicate \eqref{eq:a_q_0<=a_q_x_bound_T_11} and \eqref{eq:a_q_x+a_q_0_=a_q_x_T1} in Case 2. Thus to show they hold at step $s$, we only need to show \eqref{eq:a_q_x-a_q_0_bound_T_12} holds at step $s$. To do so, 
we consider two sub-cases: (i) $1-\pi_{1,2}^{(s-1)}(\up)\geq \frac{1}{k^{C'}}$ and (ii) $1-\pi_{1,2}^{(s-1)}(\up)\leq \frac{1}{k^{C'}}$, where we choose constant (recall $c$ is the constant in \eqref{eq:pi_1_2h-1_up=o(1)_case2})
\begin{align}\label{eq:C'}
    C'\in (0,\min\{c,1/2\}).
\end{align}
\begin{enumerate}[label=(\roman*)]
    \item $1-\pi_{1,2}^{(s-1)}(\up)\geq \frac{1}{k^{C'}}$ ($T_{1,1}+1\leq s-1\leq T_{1,2}$). By comparing \eqref{eq:delta_a_p_1_simplified_balanced_phase_1} and \eqref{eq:delta_a_q_x_simplified_balanced_phase_1} we know that 
\begin{align}
    \forall t\leq s-1:\quad \delta a_{p,1}^{(s-1)}\lesssim \frac{1}{k^{2-C'}}\delta a_{q,\times}^{(s-1)},
\end{align}
which indicates
\begin{align}
    a_{p,1}^{(s-1)}\lesssim \frac{1}{k^{2-C'}} a_{q,\times}^{(s-1)}\lesssim \frac{\ln k}{k^{2-C'}}.
\end{align}
Therefore, \eqref{eq:pi_2h-1_up_phase1.1} and \eqref{eq:pi_1_2h_up_bound_T_11} still hold, and thus when $1-\pi_{1,2}^{(s-1)}(\up)\geq \frac{1}{k^{C'}}$, \eqref{eq:delta_a_q_x_T_11_bound_2} still holds. 
By comparing \eqref{eq:delta_a_q_0_simplified_balanced_phase_1} and \eqref{eq:delta_a_q_x_T_11_bound_2}, we also have
\begin{align}
    \text{If } 1-\pi_{1,2}^{(s-1)}(\up)\geq \frac{1}{k^{C'}}:\quad \delta a_{q,0}^{(s-1)}\leq \delta a_{q,\times}^{(s-1)}.
\end{align}
This together with \eqref{eq:a_q_0<=2/3a_q_x_T11} indicates 
the first relation in \eqref{eq:a_q_x-a_q_0_bound_T_12} in Induction Hypothesis I holds at step $s$ when $1-\pi_{1,2}^{(s-1)}(\up)\geq \frac{1}{k^{C'}}$. 

To show the second relation in \eqref{eq:a_q_x-a_q_0_bound_T_12} in Induction Hypothesis I holds at step $s$, we first show
\begin{align}\label{eq:a_q_0<=lnk/2_case2}
    -a_{q,0}^{(s-1)}< \frac{\ln k}{3}
\end{align}
by contradiction. Suppose  
\begin{align}\label{eq:a_q_0>=lnk/2_contradiction}
    -a_{q,0}^{(s-1)}\geq \frac{\ln k}{3},
\end{align}
then there exists some $s'\leq s-1$ such that 
\begin{align*}
    \frac{\ln k}{5}\leq -a_{q,0}^{(s')} \leq \frac{\ln k}{4}.
\end{align*}
Then by \eqref{eq:a_q_0<=a_q_x_bound_T_11} we have
\begin{align*}
    a_{q,\times}^{(s')}\gtrsim \ln k.
\end{align*}
And by \eqref{eq:e2a_c_1>a_q_x^2} (which still holds in Case 2) and \eqref{eq:-a_q_0_bound_T_12} we know that 
\begin{align*}
    -a_{q,0}^{(s-1)}+a_{q,0}^{(s')}\lesssim \frac{1}{a_{q,\times}^{(s')}}\lesssim \frac{1}{\ln k},
\end{align*}
and thus
\begin{align*}
    -a_{q,0}^{(s-1)}\leq -a_{q,0}^{(s')}+O\left(\frac{1}{\ln k}\right)\leq \frac{\ln k}{4}+O\left(\frac{1}{\ln k}\right)< \frac{\ln k}{3},
\end{align*}
this contradicts \eqref{eq:a_q_0>=lnk/2_contradiction}. Thus \eqref{eq:a_q_0<=lnk/2_case2} holds. 

From our choice of $C'$ in \eqref{eq:C'} and \eqref{eq:pi_2h_up_phase1.1}, \eqref{eq:xi_2h_bound_phase_1.2} and Induction Hypothesis I, we have
\begin{align}\label{eq:a_q_x<=3/2lnk_case2}
    a_{q,\times}^{(s-1)}\leq (3/2+o(1))\ln k.
\end{align}
Combining \eqref{eq:a_q_x<=3/2lnk_case2} and \eqref{eq:a_q_0<=lnk/2_case2}, we have
\begin{align*}
    a_{q,\times}^{(s-1)}-a_{q,0}^{(s-1)}\leq \left(\frac{11}{6}+o(1)\right)\ln k,
\end{align*}
and by our gradient updates we know that
\begin{align*}
    a_{q,\times}^{(s)}-a_{q,0}^{(s)}\leq 2\ln k.
\end{align*}
This shows the second relation in \eqref{eq:a_q_x-a_q_0_bound_T_12} in Induction Hypothesis I holds at step $s$.


    \item $1-\pi_{1,2}^{(s-1)}(\up)\leq \frac{1}{k^{C'}}$ ($T_{1,1}+1\leq s-1\leq T_{1,2}$). Define ratio
\begin{align}\label{eq:kappa_h_ratio}
    \forall h\in[k-1]:\quad \kappa_h^{(t)}\coloneqq \frac{\pi_{1,2h+1}^{(t)}(\up)}{1-\pi_{1,2h}^{(t)}(\up)}
\end{align}
By \eqref{eq:a_p_2<=1_T_12}, \eqref{eq:pi_2h_up_phase1}, \eqref{eq:pi_2h-1_up_phase1}, \eqref{eq:xi_2h_bound_phase_1.2} and \eqref{eq:pi_1_2h-1_up=o(1)_case2}, we have
\begin{align}\label{eq:kappa_h_ratio_expression}
    \kappa_h^{(s-1)}=(1+o(1))\frac{\exp\left(\zeta_h^{(s-1)}\right)}{\left(h\exp\left(-\frac{a_{p,1}^{(s-1)}}{2h+1}\right)+k-h\right)\left(h\exp\left(-\frac{a_{p,1}^{(s-1)}}{2h}\right)+k-h\right)},
\end{align}
where
\begin{align}\label{eq:zeta}
    \zeta_h^{(s-1)}&\coloneqq \left(1+\frac{\frac{h}{h+1}e^{-2a_{c,0}^{(s-1)}}}{\frac{h}{h+1}e^{-2a_{c,0}^{(s-1)}}+1}\right)a_{q,\times}^{(s-1)}-\frac{1}{1+\frac{h}{h+1}e^{-2a_{c,0}^{(s-1)}}}a_{q,0}^{(s-1)}-\frac{4h+1}{4h+2}a_{p,2}^{(s-1)}\notag\\
    &=a_{q,\times}^{(s-1)}-a_{q,0}^{(s-1)}+\frac{\frac{h}{h+1}e^{-2a_{c,0}^{(s-1)}}}{1+\frac{h}{h+1}e^{-2a_{c,0}^{(s-1)}}}\left(a_{q,\times}^{(s-1)}+a_{q,0}^{(s-1)}\right)-\frac{4h+1}{4h+2}a_{p,2}^{(s-1)}.
\end{align}
Below we first show by contradiction that the second relation in \eqref{eq:a_q_x-a_q_0_bound_T_12} in Induction Hypothesis I holds at step $s$. Suppose otherwise, i.e.,
\begin{align}\label{eq:a_q_x-a_q_0_bound_T_12_contradiction}
    a_{q,\times}^{(s)}-a_{q,0}^{(s)}> 2.1\ln k,
\end{align}
Then by our gradient updates we have
\begin{align*}
    a_{q,\times}^{(s-1)}-a_{q,0}^{(s-1)}> 2.1\ln k-\frac{1}{2}.
\end{align*}
This combined with the fact that $a_{q,\times}^{(t-1)}+a_{q,0}^{(t-1)}>0$ guaranteed by Induction Hypothesis I and \eqref{eq:a_p_2<=1_T_12} gives
\begin{align*}
    \forall h\in[k-1]:\quad \zeta_h^{(s-1)}> 2.1\ln k-2,
\end{align*}
Therefore, we have
\begin{align*}
    \forall h\in[k-1]:\quad \kappa_h^{(s-1)}\overset{\eqref{eq:a_p_1_bound_T_11}}\geq (1+o(1))\frac{\exp(\zeta_h^{(s-1)})}{k^2}\gtrsim k^{0.1},
\end{align*}
which indicates
\begin{align*}
    \forall h\in[k-1]:\quad \pi_{1,2h}^{(s-1)}(\up)=o\left(\pi_{1,2h+1}^{(s-1)}(\up)\right).
\end{align*}
Thus by comparing \eqref{eq:delta_a_q_0_simplified_balanced_phase_1} and \eqref{eq:delta_a_q_x_simplified_balanced_phase_1}, and using \eqref{eq:term1<=term3}, we know that 
\begin{align*}
    \delta a_{q,\times}^{(s-1)}-\delta a_{q,0}^{(s-1)}<0,
\end{align*}
which suggests 
\begin{align*}
    a_{q,\times}^{(s)}-a_{q,0}^{(s)}< a_{q,\times}^{(s-1)}-a_{q,0}^{(s-1)}\leq 2.1\ln k,
\end{align*}
where the last inequality is guaranteed by Induction Hypothesis I. This contradicts \eqref{eq:a_q_x-a_q_0_bound_T_12_contradiction}. Thus the second relation in \eqref{eq:a_q_x-a_q_0_bound_T_12} in Induction Hypothesis I holds at step $s$.

We now show the first relation in \eqref{eq:a_q_x-a_q_0_bound_T_12} in Induction Hypothesis I holds at step $s$ also by contradiction. Suppose otherwise, i.e.,
\begin{align}\label{eq:a_q_x-a_q_0_bound_T_12_contradiction_1}
    a_{q,\times}^{(s)}-a_{q,0}^{(s)}< 0,
\end{align}
Then by our gradient updates we have
\begin{align}\label{eq:a_q_x-a_q_0_bound_T_12_contradiction_2}
    a_{q,\times}^{(s-1)}-a_{q,0}^{(s-1)}< \frac{1}{2}.
\end{align}
\eqref{eq:a_q_x-a_q_0_bound_T_12_contradiction_2} and our choice of $T_{1,1}$ imply
\begin{align*}
    a_{q,0}^{(s-1)}\geq a_{q,\times}^{(s-1)}\geq \ln k-\frac{1}{2}.
\end{align*}
Note that by \eqref{eq:a_q_0<=2/3a_q_x_T11} we know that there exists $s'\in[T_{1,1}+1,s-1]$ and constants $0<c_1<c_2<1$ such that
\begin{align*}
    c_1 a_{q,\times}^{(s-1)}\leq a_{q,0}^{(s')}\leq c_2 a_{q,\times}^{(s-1)}.
\end{align*}
Thus by \eqref{eq:delta_e2a_c_0_bound_T_12}, \eqref{eq:a_q_x+a_q_0_=a_q_x_T1} and the above two inequalities we have
\begin{align}\label{eq:e2a_c_0_is_large}
    e^{2a_{c,0}^{(s-1)}}\geq e^{2a_{c,0}^{(s-1)}}-e^{2a_{c,0}^{(s')}}\gtrsim a_{q,\times}^{(s-1)} \left(a_{q,0}^{(s')}+a_{q,\times}^{(s')}\right)\gtrsim \left(a_{q,\times}^{(s-1)}+a_{q,0}^{(s-1)}\right)\ln k.
\end{align}
This combined with \eqref{eq:zeta} and \eqref{eq:a_p_2<=1_T_12} gives
\begin{align*}
    \zeta_h^{(s-1)}\leq a_{q,\times}^{(s-1)}-a_{q,0}^{(s-1)}+O\left(\frac{1}{\ln k}\right)\overset{\eqref{eq:a_q_x-a_q_0_bound_T_12_contradiction_2}}<1.
\end{align*}
Plugging this into \eqref{eq:kappa_h_ratio_expression}, and note that $a_{p,1}\leq k$ by our choice of $T_{1,2}$ (c.f.~\eqref{eq:T_1_2}), thus we have
\begin{align*}
    \forall h\in[k-1]:\quad \kappa_h^{(s-1)} &\lesssim \frac{1}{k^2}, \\
    \forall h\in k: \quad \pi_{1,2h-1}^{(s-1)}(\up)& \lesssim \frac{1}{k^2}.
\end{align*}
Thus by the above two expressions, \eqref{eq:delta_a_q_0_simplified_balanced_phase_1} and \eqref{eq:delta_a_q_x_simplified_balanced_phase_1} we have
\begin{align*}
    \delta a_{q,\times}^{(s-1)}-\delta a_{q,0}^{(s-1)}>0,
\end{align*}
which suggests
\begin{align*}
    a_{q,\times}^{(s)}-a_{q,0}^{(s)}> a_{q,\times}^{(s-1)}-a_{q,0}^{(s-1)}\geq 0,
\end{align*}
where the last inequality is guaranteed by Induction Hypothesis I. This contradicts \eqref{eq:a_q_x-a_q_0_bound_T_12_contradiction_1}. Thus the first relation in \eqref{eq:a_q_x-a_q_0_bound_T_12} in Induction Hypothesis I holds at step $s$.
\end{enumerate}

By \eqref{eq:a_p_2_bound_T_12} and our choice of $T_{1/2}$ we know that 
\begin{align}
    a_{p,1}^{(s-1)}\leq (4.1+o(1))\ln k.
\end{align}
Thus by a similar argument as we proved \eqref{eq:a_q_x-a_q_0_bound_T_12} above, we can show by contradiction that \eqref{eq:a_q_x-a_q_0_bound>T_12} holds at step $s$ for some constant $C>0$.

We next show by contradiction that the second relation in \eqref{eq:a_q_0+a_q_x=C_lnk_phase_1} holds at step $s$. Suppose otherwise, i.e.,
\begin{align}\label{eq:a_q_0+a_q_x_bound_T_12_contradiction}
    a_{q,0}^{(s)}+a_{q,\times}^{(s)}> 20.4\ln k,
\end{align}
Then there exists $s'<s$ such that 
\begin{align}\label{eq:a_q_0+a_q_x_bound_T_12_contradiction_2}
    20.3\ln k\leq a_{q,0}^{(s')}+a_{q,\times}^{(s')}\leq 20.35\ln k.
\end{align}
Then by \eqref{eq:a_q_x-a_q_0_bound>T_12} we have
\begin{align}
    a_{q,\times}^{(s')}\geq 10.05\ln k,\quad \text{and}\quad a_{q,0}^{(s')}\geq 7.05\ln k.
\end{align}

Then by \eqref{eq:pi_2h_up_phase1}, \eqref{eq:pi_2h-1_up_phase1}, \eqref{eq:a_p_2_bound_T_12} and \eqref{eq:e2a_c_0_is_large} (which still applies here by the above expression) we have for any $t\in[s',s-1]$:
\begin{align}
    \forall h\in[k]:\quad \pi_{1,2h-1}^{(t)}(\up)\lesssim \frac{1}{k^{7.05}},\quad \text{and}\quad \forall h\in [k-1]:\quad \pi_{1,2h}^{(t)}(\up)\lesssim \frac{1}{k^{7.05}}.
\end{align}
Thus by \eqref{eq:delta_a_q_0+delta_a_q_x_T_11_expression} we have 
\begin{align}\label{eq:a_q_0+a_q_x_s'_cond}
    \forall t\in[s',s-1]:\quad \delta a_{q,0}^{(t)}+\delta a_{q,\times}^{(t)}\lesssim \frac{1}{k^{6.05}}\overset{\eqref{eq:a_p_1_bound_T_12}}\lesssim \frac{1}{k^{1.05}}\delta a_{p,1}^{(t)}.
\end{align}
Thus 
\begin{align}
    \left(a_{q,0}^{(s)}+a_{q,\times}^{(s)}\right)-\left(a_{q,0}^{(s')}+a_{q,\times}^{(s')}\right)\lesssim \frac{1}{k^{1.05}}\left(a_{p,1}^{(s)}-a_{p,1}^{(s')}\right)\overset{\eqref{eq:T_1/2}}\lesssim \frac{k\ln k}{k^{1.05}}=o(1).
\end{align}
Then by \eqref{eq:a_q_0+a_q_x_bound_T_12_contradiction_2} and the above relation, we have
\begin{align}
    a_{q,0}^{(s)}+a_{q,\times}^{(s)}<20.4\ln k.
\end{align}
This contradicts \eqref{eq:a_q_0+a_q_x_bound_T_12_contradiction}. Thus the second relation in \eqref{eq:a_q_0+a_q_x=C_lnk_phase_1} holds at step $s$.
Note that by \eqref{eq:a_q_x+a_q_0_increase}, \eqref{eq:a_q_x+a_q_0_=a_q_x_T1} and our choice of $T_{1,1}$ we know that $a_{q,0}^{(t)}+a_{q,\times}^{(t)}\gtrsim \ln k$ after $t\geq T_{1,1}+1$. This together with the second relation in \eqref{eq:a_q_0+a_q_x=C_lnk_phase_1} indicates \eqref{eq:a_q_x+a_q_0_=lnk_T_12} holds at step $s$.

We next show the first relation in \eqref{eq:a_q_0+a_q_x=C_lnk_phase_1} holds at step $s$ also by contradiction ($s-1\in[T_{1,2}+1,T_{1/2}-1]$). Suppose otherwise, i.e.,
\begin{align}\label{eq:a_q_0+a_q_x_T_12_contradiction}
    a_{q,0}^{(s)}+a_{q,\times}^{(s)}< 5.9\ln k,
\end{align}
Then since by \eqref{eq:a_q_x+a_q_0_increase},  $a_{q,0}^{(t)}+a_{q,\times}^{(t)}$ monotonically increases for $t\leq s-1$, thus we have
\begin{align}\label{eq:a_q_0+a_q_x_T_12_contradiction_2}
    \forall t\in[T_{1,1}+1,T_{1,2}]:\quad a_{q,0}^{(t)}+a_{q,\times}^{(t)}< 5.9\ln k.
\end{align}
Then we have either $a_{q,0}^{(t)}\leq 1.95\ln k$ or $a_{q,\times}^{(t)}\geq 3.95\ln k$ for any $t\in[T_{1,1}+1,T_{1,2}]$. In either case, by \eqref{eq:pi_2h_up_phase1}, \eqref{eq:pi_2h-1_up_phase1}, Induction Hypothesis I and our choice of $T_{1,2}$ we have for any $t\in[T_{1,1}+1,T_{1,2}]$:
\begin{align*}
    \forall h\in[k-1]:\quad \pi_{1,2h}^{(t)}(\up)+\pi_{1,2h+1}^{(t)}(\up)\gtrsim \frac{1}{k^{2.95}}.
\end{align*}
By the above expression and the second line of \eqref{eq:delta_a_p_1<delta_a_q_0+delta_a_q_x}, we have
\begin{align}\label{eq:delta_a_q_0+delta_a_q_x_T_12_inequality}
    \forall t\in[T_{1,1}+1,T_{1,2}]:\quad \delta a_{q,0}^{(t)}+\delta a_{q,\times}^{(t)}\gtrsim \frac{1}{k^{0.95}}\delta a_{p,1}^{(t)}.
\end{align}
Let $t_0\in[T_{1,1}+1,T_{1,2}]$ be the smallest $t$ such that $a_{p,1}^{(t)}\geq 5\ln k$. Then by \eqref{eq:a_q_0+a_q_x>=4lnk_phase1} we have
\begin{align}\label{eq:a_q_0+a_q_x_t0>=4lnk}
    a_{q,0}^{(t_0)}+a_{q,\times}^{(t_0)}\geq 4\ln k.
\end{align}
Thus by \eqref{eq:delta_a_q_0+delta_a_q_x_T_12_inequality} we have
\begin{align*}
    \left(a_{q,0}^{(T_{1,2})}+a_{q,\times}^{(T_{1,2})}\right)-\left(a_{q,0}^{(t_0)}+a_{q,\times}^{(t_0)}\right)\gtrsim \frac{1}{k^{0.95}}\left(a_{p,1}^{(T_{1,2})}-a_{p,1}^{(t_0)}\right)\gtrsim k^{0.05}\geq 2\ln k.
\end{align*}
Thus by \eqref{eq:a_q_0+a_q_x_t0>=4lnk} and the above relation, we have
\begin{align*}
    a_{q,0}^{(T_{1,2})}+a_{q,\times}^{(T_{1,2})}\geq 6\ln k.
\end{align*}
This contradicts \eqref{eq:a_q_0+a_q_x_T_12_contradiction_2}. Thus the first relation in \eqref{eq:a_q_0+a_q_x=C_lnk_phase_1} holds at step $s$.

By the above argument we also know that
\begin{align}\label{eq:a_q_0+a_q_x_T_12_intermediate}
    a_{q,0}^{(T_{1,2})}+a_{q,\times}^{(T_{1,2})}\geq 5.9\ln k.
\end{align}
This together with \eqref{eq:a_q_x-a_q_0_bound_T_12} gives 
\begin{align}\label{eq:a_q_0>=1.9lnk_T_12_intermediate}
    a_{q,0}^{(T_{1,2})}\geq 1.9\ln k.
\end{align}
By \eqref{eq:e2a_c_1>lnk^2_T_11_end}, \eqref{eq:-a_q_0_bound_T_12}, \eqref{eq:a_q_0+a_q_x=C_lnk_phase_1} we know that 
\begin{align}\label{eq:a_q_0_bound_T_12_intermediate}
    -\left(a_{q,0}^{(s-1)}-a_{q,0}^{(T_{1,2})}\right)\lesssim \frac{1}{\ln k},
\end{align}
From \eqref{eq:a_q_0+a_q_x_T_12_intermediate}, \eqref{eq:a_q_0_bound_T_12_intermediate} and our update rule, we know that 
\eqref{eq:a_q_0>=1.9lnk_T_12} holds at step $s$.

We next show \eqref{eq:a_q_x-a_p_2_bound_T_12} holds at step $s$ for $s-1\in[T_{1,2}+1,T_{1/2}-1]$. By \eqref{eq:a_p_2<=1_T_12}, \eqref{eq:a_q_x-a_q_0_bound_T_12} and \eqref{eq:a_q_0+a_q_x_T_12_intermediate} we know that
\begin{align}
    a_{q,\times}^{(T_{1,2})}-\frac{1}{2}a_{p,2}^{(T_{1,2})}\geq 2.95\ln k-2,
\end{align}
and thus \eqref{eq:a_q_x-a_p_2_bound_T_12} holds at step $T_{1,2}$. Suppose \eqref{eq:a_q_x-a_p_2_bound_T_12} does not hold at step $s$ ($s-1\in[T_{1,2}+1,T_{1/2}-1]$), i.e.,
\begin{align}\label{eq:a_q_x-a_p_2_T_12_contradiction_2}
    a_{q,\times}^{(s)}-\frac{1}{2}a_{p,2}^{(s)}< 2.425\ln k,
\end{align}
then by our update rule, we know that
\begin{align}\label{eq:a_q_x-a_p_2_T_12_contradiction}
    a_{q,\times}^{(s-1)}-\frac{1}{2}a_{p,2}^{(s-1)}\leq 2.425\ln k+1,
\end{align}
By \eqref{eq:a_q_0+a_q_x_T_12_intermediate}, \eqref{eq:a_q_x-a_q_0_bound_T_12} and a similar argument as we proved \eqref{eq:a_q_0>=1.9lnk_T_12} holds at step $s$ above, by using the gradient expression of $a_{q,\times}^{(t)}$ in \eqref{eq:delta_a_q_x_simplified_balanced_phase_1} and \eqref{eq:e2a_c_0>=lnk^2_T_12} we can show that
\begin{align*}
    a_{q,\times}^{(s-1)}\geq 2.95\ln k-1,
\end{align*}
By the above two inequalities, we have
\begin{align*}
    \frac{1}{2}a_{p,2}^{(s-1)}\geq 0.525\ln k-2.
\end{align*}
This together with \eqref{eq:a_q_0>=1.9lnk_T_12} and \eqref{eq:pi_2h-1_up_phase1} gives
\begin{align*}
    \forall h\in[k]:\quad \pi_{1,2h-1}^{(s-1)}(\up)\lesssim \frac{1}{k^{2.425}}.
\end{align*}
By \eqref{eq:a_q_x-a_p_2_T_12_contradiction} and \eqref{eq:pi_2h_up_phase1} we have
\begin{align*}
    \forall h\in[k-1]:\quad 1-\pi_{1,2h}^{(s-1)}(\up)\gtrsim \frac{1}{k^{2.425}}.
\end{align*}
Plugging the above two relations into \eqref{eq:delta_a_q_x_simplified_balanced_phase_1}, and by \eqref{eq:e2a_c_1_bound_T_12}, \eqref{eq:e2a_c_0>=lnk^2_T_12} we have
\begin{align*}
    \delta a_{q,\times}^{(s-1)}\lesssim\frac{1}{2k^{2.425}}\sum_{h=1}^{k-1} \PP^{(s-1)}\left(\gE_{1,2h}\right)\pi_{1,2h}^{(s-1)}(\up)V_{1,2h+1}^{(s-1)}.
\end{align*}
Comparing the above expression with \eqref{eq:delta_a_p_1_simplified_balanced_phase_1} and \eqref{eq:delta_a_p_3_simplified_balanced_phase_1} and by Induction Hypothesis I, we deduce
\begin{align*}
    \delta a_{p,1}^{(s-1)}\lesssim k^{0.425}\delta a_{q,\times}^{(s-1)},\quad \delta a_{p,3}^{(s-1)}\lesssim \frac{1}{k}\delta a_{q,\times}^{(s-1)}.
\end{align*}
Therefore, by \eqref{eq:a_p3} we have
\begin{align*}
    \delta a_{p,2}^{(s-1)}=\frac{\delta a_{p,1}^{(s-1)}-\delta a_{p,3}^{(s-1)}}{k-1}\lesssim k^{-0.575}\delta a_{q,\times}^{(s-1)},
\end{align*}
which suggests
\begin{align*}
    \delta a_{q,\times}^{(s-1)}-\frac{1}{2}\delta a_{p,2}^{(s-1)}>0,
\end{align*}
and thus 
\begin{align*}
    a_{q,\times}^{(s)}-\frac{1}{2}a_{p,2}^{(s)}>a_{q,\times}^{(s-1)}-\frac{1}{2}a_{p,2}^{(s-1)}\overset{\eqref{eq:a_q_x-a_p_2_bound_T_12}}\geq 2.425\ln k,
\end{align*}
which contradicts \eqref{eq:a_q_x-a_p_2_T_12_contradiction_2}. Thus \eqref{eq:a_q_x-a_p_2_bound_T_12} holds at step $s$.

\paragraph{When $s=T_{1/2}$.} By our choice of $T_{1/2}$, our update rule and \eqref{eq:a_p_2_bound_T_12}, we have
\begin{align}
    a_{p,1}^{(T_{1/2})}&\geq 4.1k\ln k-1, \label{eq:a_p_1_bound_T_12_end} \\
    a_{p,2}^{(T_{1/2})}& \geq 4.1\ln k-1. \label{eq:a_p_2_bound_T_12_end}
\end{align}
We first show \eqref{eq:a_q_0-a_p_2_phase_1_end} holds. 
Define 
\begin{align}\label{eq:t_1'}
    t_1'=\inf\{t\in[T_{1,2}+1,T_{1/2}-1]:\quad a_{p,2}^{(t)}\geq 4\ln k\}.
\end{align}
We first show by contradiction that there exists $t_1\in[t_1',T_{1/2}-1]$ such that 
\begin{align}\label{eq:a_q_x-a_p_2_end_contradiction}
    a_{q,\times}^{(t_1)}-\frac{1}{2}a_{p,2}^{(t_1)}\geq 3.9 \ln k.
\end{align}
Suppose otherwise, i.e.,
\begin{align*}
    \forall t\in[t_1',T_{1/2}-1]:\quad a_{q,\times}^{(t)}-\frac{1}{2}a_{p,2}^{(t)}< 3.9 \ln k,
\end{align*}
then for all such $t$, by \eqref{eq:pi_2h_up_phase1} we have
\begin{align*}
    \forall h\in[k-1]:\quad 1-\pi_{1,2h}^{(t)}(\up)\lesssim \frac{1}{k^{3.9}},
\end{align*}
and by \eqref{eq:a_q_0>=1.9lnk_T_12}, \eqref{eq:pi_2h-1_up_phase1} and our choice of $t_1'$ we have
\begin{align*}
    \forall h\in[k-1]:\quad \pi_{1,2h+1}^{(t)}(\up)\gtrsim \frac{1}{k^{3.9}},
\end{align*}
Plugging the above two relations into \eqref{eq:delta_a_q_x_T_11}, we have
\begin{align}\label{eq:delta_a_q_x_T_11_cond}
    \delta a_{q,\times}^{(t)}\lesssim\frac{1}{k^{3.9}}\sum_{h=1}^{k-1} \PP^{(t)}\left(\gE_{1,2h}\right)\pi_{1,2h}^{(t)}(\up)V_{1,2h+1}^{(t)}.
\end{align}
By \eqref{eq:pi_1_2h_i_phase1}, \eqref{eq:pi_1_2h-1_i_phase1}, \eqref{eq:a_p_2_bound_T_12} and our choice of $t_1'$ we have 
\begin{align*}
   \forall h\in[k-1]:\quad \pi_{1,2h}^{(t)}(\down_\times^1)=o\left(\pi_{1,2h+1}^{(t)}(\down_\times)\right),\quad \pi_{1,2h+1}^{(t)}(\down_\times)\lesssim \frac{1}{k^2},
\end{align*}
Thus by the above expression, \eqref{eq:delta_a_q_x_T_11_cond}, \eqref{eq:delta_a_p_1_simplified_balanced_phase_1}, \eqref{eq:delta_a_p_3_simplified_balanced_phase_1} and \eqref{eq:a_p3} we have
\begin{align}
    \delta a_{p,2}^{(t)}\lesssim \frac{1}{k^{4}}\sum_{h=1}^{k-1} \PP^{(t)}\left(\gE_{1,2h}\right)\pi_{1,2h}^{(t)}(\up)V_{1,2h+1}^{(t)} & \lesssim \frac{1}{k^{0.1}}\delta a_{q,\times}^{(t)}, \label{eq:delta_a_p_2_vs_delta_a_q_x_contradiction} \\
    a_{q,\times}^{(T_{1/2})}-a_{q,\times}^{(t_1')} & \gtrsim k^{0.1}\ln k,
\end{align}
and thus
\begin{align*}
    a_{q,\times}^{(T_{1/2})}-\frac{1}{2}a_{p,2}^{(T_{1/2})}\gtrsim k^{0.1}\ln k\geq 3.9 \ln k.
\end{align*} 
This contradicts \eqref{eq:a_q_0-a_p_2_phase_1_end}. Therefore, such $t_1$ exists. Then we show
\begin{align}\label{eq:a_q_x-a_p_2_end_t_1_to_end}
    \forall t\in[t_1,T_{1/2}]:\quad a_{q,\times}^{(t)}-\frac{1}{2}a_{p,2}^{(t)}\geq 3.9 \ln k
\end{align}
again by contradiction. Suppose otherwise, then there exists the first $t_2\in[t_1,T_{1/2}]$ such that
\begin{align}
    a_{q,\times}^{(t_2)}-\frac{1}{2}a_{p,2}^{(t_2)}< 3.9 \ln k, \label{eq:a_q_x-a_p_2_end_t_1_to_end_contradiction} \\
    a_{q,\times}^{(t_2-1)}-\frac{1}{2}a_{p,2}^{(t_2-1)}\geq 3.9 \ln k. \label{eq:a_q_x-a_p_2_end_t_1_to_end_contradiction_2}
\end{align}
By our update rule, we have
\begin{align*}
    a_{q,\times}^{(t_2-1)}-\frac{1}{2}a_{p,2}^{(t_2-1)}\leq 3.9 \ln k+1.
\end{align*}
Then by the same argument as above, we can show that 
\begin{align*}
    \delta a_{p,\times}^{(t_2-1)}\gtrsim k^{0.1}\delta a_{p,2}^{(t_2-1)},
\end{align*}
which indicates
\begin{align*}
    \delta a_{q,\times}^{(t_2-1)}-\frac{1}{2}\delta a_{p,2}^{(t_2-1)}>0,
\end{align*}
and thus
\begin{align*}
    a_{q,\times}^{(t_2)}-\frac{1}{2}a_{p,2}^{(t_2)}>a_{q,\times}^{(t_2-1)}-\frac{1}{2}a_{p,2}^{(t_2-1)}\overset{\eqref{eq:a_q_x-a_p_2_end_t_1_to_end_contradiction_2}}\geq 3.9 \ln k.
\end{align*}
This contradicts \eqref{eq:a_q_x-a_p_2_end_t_1_to_end_contradiction}. Therefore, \eqref{eq:a_q_x-a_p_2_end_t_1_to_end}, and hence \eqref{eq:a_q_0-a_p_2_phase_1_end} holds.

We now prove \eqref{eq:1-R_pi_T_1_2_phase_1_end}.
\begin{align}\label{eq:pi_2h-1_down_phase1_end}
    \forall h\in[k]:\quad\pi_{1,2h-1}^{(T_{1/2})}(\down_\times)\geq (1+o(1))\frac{k^{-2.05}}{k-h+1}.
\end{align}
By \eqref{eq:a_p_2_bound_T_12_end}, \eqref{eq:a_q_0>=1.9lnk_T_12},
\eqref{eq:a_q_x-a_q_0_bound>T_12}, and \eqref{eq:pi_2h-1_up_phase1}, we have
\begin{align}\label{eq:pi_2h-1_up_phase1_end}
    \forall h\in[k]:\quad
    \pi_{1,2h-1}^{(T_{1/2})}(\up)\lesssim \frac{1}{k^{2.9}}.
\end{align}

 By \eqref{eq:a_q_0-a_p_2_phase_1_end}, \eqref{eq:pi_2h_up_phase1} we have
\begin{align}\label{eq:pi_2h_up_phase1_end}
    \forall h\in[k-1]:\quad 1-\pi_{1,2h}^{(T_{1/2})}(\up)\lesssim \frac{1}{k^{2.9}}.
\end{align}


Note that for any $t\in\NN$:
\begin{align}\label{eq:R_pi_s_1_expression_phase_1}
    R(\pi^{(t)};\gP_1)=\underbracket{\frac{1}{k}}_{\text{(a)}}\underbracket{\left(1-\pi_{1,1}^{(t)}(\up)\right)}_{(b)}+&\sum_{h=1}^{k-1}\underbracket{\frac{1}{k}}_{\text{(c)}}\bigg(\underbracket{\prod_{i=1}^{h-1}\pi_{1,2i}^{(t)}(\up)\prod_{i=1}^h\left(1-(i-1)\pi_{1,2i-1}^{(t)}(\down_\times)-\pi_{1,2i-1}^{(t)}(\up)\right)}_{(d)}\notag\\
    &\qquad\cdot\underbracket{\pi_{1,2h}^{(t)}(\up)\left(1-\pi_{1,2h+1}^{(t)}(\up)-h\pi_{1,2h+1}^{(t)}(\down_\times)\right)}_{(e)}\bigg),
\end{align}
where (a) stands for the probability of the event that the goal is the first leaf visited by the agent, and (b) stands for the probability of the agent finding the goal conditioned on this event; (c) stands for the probability of the event that the goal is the $h+1$-th child visited, (d) stands for the probability of event $\gE_{1,2h}$ conditioned on this event, and (e) stands for the probability of the agent finding the goal conditioned on this event and $\gE_{1,2h+1}$.

By \eqref{eq:pi_2h-1_down_phase1_end}, \eqref{eq:pi_2h-1_up_phase1_end}, \eqref{eq:pi_2h_up_phase1_end} we know that
\begin{align}\label{eq:(b),(e)_contradiction}
    \text{(b)}\geq 1-o\left(\frac{1}{k^{1.05}}\right),\quad \text{(e)}\geq 1-\frac{1}{k^{1.05}}.
\end{align}
For (d), 
\begin{align}
    \text{(d)}=\underbrace{\prod_{i=1}^{h-1}\pi_{1,2i}^{(t)}(\up)}_{\text{(d1)}}\underbrace{\prod_{i=1}^h\left(1-(i-1)\pi_{1,2i-1}^{(t)}(\down_\times)-\pi_{1,2i-1}^{(t)}(\up)\right)}_{\text{(d2)}}.
\end{align}
For (d1), when $t=T_{1/2}$, we have
\begin{align}\label{eq:(d1)_contradiction}
    \text{(d1)}\geq 1-o\left(\frac{1}{k^{1.05}}\right).
\end{align}
For (d2), when $t=T_{1/2}$, we have
\begin{align}\label{eq:(d2)_contradiction}
    \text{(d2)}&\geq \prod_{i=1}^{k-1}\left(1-(i-1)\pi_{1,2i-1}^{(T_{1/2})}(\down_\times)-\pi_{1,2i-1}^{(T_{1/2})}(\up)\right)\notag\\
    &\geq \prod_{i=1}^{k-1}\left(1-\frac{2k^{-1.05}}{k-i+1}\right)\notag\\
    &=\prod_{i=2}^{k}\left(1-\frac{2k^{-1.05}}{i}\right)\notag\\
    & \geq \prod_{i=2}^{\floor{k^{0.04}}}\left(1-2k^{-1.05}\right)\prod_{i=\floor{k^{0.04}}+1}^{\floor{k^{0.08}}}\left(1-2k^{-1.09}\right)\cdots\prod_{i=\floor{k^{0.96}}+1}^{k}\left(1-2k^{-2.01}\right)\notag\\
    &\geq \left(1-2k^{-1.05}\right)^{k^{0.04}}\left(1-2k^{-1.09}\right)^{k^{0.08}}\cdots\left(1-2k^{-2.01}\right)^{k}.\notag\\
    &\geq 1-\frac{c}{k^{1.01}}
\end{align}
for some constant $c>0$.
Plugging \eqref{eq:(b),(e)_contradiction}, \eqref{eq:(d1)_contradiction}, \eqref{eq:(d2)_contradiction} into \eqref{eq:R_pi_s_1_expression_phase_1}, we have
\begin{align}
    R(\pi^{(T_{1/2})};\gP_1)\geq 1-\frac{c'}{k^{1.01}}
\end{align}
where $c'>0$ is some constant. This gives the second relation in \eqref{eq:1-R_pi_T_1_2_phase_1_end}. 


On the other hand, note that $1-R\left(\pi^{(T_{1/2})};\gP_1\right)$---the failure probability of $\pi^{(T_{1/2})}$ on trees from $P_1$ is larger than the probability of the goal is at the $k$-th child the agent visits, but at step $2k-1$, the agent chooses some $\down_i$ that's already been taken, i.e., we have
\begin{align}
    1-R\left(\pi^{(T_{1/2})};\gP_1\right)\geq \frac{1}{k}\pi_{1,2k-1}(\down_\times)\overset{\eqref{eq:pi_2h-1_down_phase1_end}}\geq \frac{1+o(1)}{k^{3.05}}.
\end{align}
This gives the first relation in \eqref{eq:1-R_pi_T_1_2_phase_1_end}. 

When $s=T_{1/2}$, similar as how we show \eqref{eq:delta_a_b_1_balanced_phase_1} holds for both cases, by Induction Hypothesis I and comparing  \eqref{eq:delta_a_b_1_simplified_balanced_phase_1} and \eqref{eq:delta_a_p_1_simplified_balanced_phase_1}, we have
\begin{align}
    \forall t\leq T_{1/2}-1,\,\, \text{if } a_{p,1}^{(t)}\geq k\ln k:\quad \delta a_{b,1}^{(t)}\gtrsim \frac{k^2\ln k}{N}\delta a_{p,1}^{(t)}.
\end{align}
Thus by our choice of $T_{1/2}$ and \eqref{eq:delta_a_b_1_balanced_phase_1} we have \eqref{eq:a_b_1_phase_1_end} holds at step $s$.
Similarly, by noting that
\begin{align}
    \forall  t\leq T_{1/2}-1,\,\, \text{if } a_{p,1}^{(t)}\geq k\ln k:\quad \frac{k\ln k}{N}\delta a_{p,1}^{(t)}&\lesssim \delta a_{b,0,B}^{(t)}\lesssim \frac{k^2\ln k}{N}\delta a_{p,1}^{(t)},\notag\\
     \delta a_{b,0,A}^{(t)}&\lesssim \frac{\ln k}{N}\left(\delta a_{q,0}^{(t)}+\delta a_{q,\times}^{(t)}\right),
\end{align}
we have \eqref{eq:a_b_0_phase_1_end} holds at step $s$.

Below we prove \eqref{eq:R_pi_T_1_l_phase_1_end}. All probabilities here are under policy $\pi^{(T_{1/2})}$ with a training distribution $\gP_l$ that satisfies \eqref{eq:balanced_goal_distribution} for a fixed $l\in \{2,3,\cdots,L\}$. We define events
\begin{align}
    \gB_1&\coloneqq \left\{\text{goal is not at the first $k-l+1$ leaves $\pi^{(T_{1/2})}$ visits, and }a_{i}\in[k], a_i\neq a_j, \forall i,j\in[l], i\neq j\right\},\label{eq:event_B_1}\\
    \gB_2&\coloneqq \left\{a_{l+1}=\up\right\},\label{eq:event_B_2}
\end{align}
and for all $i\in\{2,3,\cdots, k-l+1\}$,
\begin{align}
    \gB_{2i-1}&\coloneqq \left\{a_{l+2(i-1)}\in[k]\setminus\{a_1,a_2,\cdots,a_{l},a_{l+2},\cdots,a_{l+2(i-2)}\}\right\},\label{eq:event_B_2i-1}\\
    \gB_{2i}&\coloneqq \left\{a_{l+2i-1}=\up\right\}.\label{eq:event_B_2i}
\end{align}
Then we have
\small
\begin{align}\label{eq:R_pi_T_1_l_phase_1_end_proof}
    &1-R\left(\pi^{(T_{1/2})};\gP_l\right)\notag\\
    &\geq \PP(\gB_1)\Big(\PP(\underbrace{a_{l+1}\neq \up|\gB_1}_{=1-\PP(\gB_2|\gB_1)})+\PP(\gB_2|\gB_1)\PP(\{\text{$\pi^{(T_{1/2})}$ fails at some step $h>l+1$}\}|\gB_1\gB_2)\Big)\notag\\
    &\geq \PP(\gB_1)\PP(\{\text{$\pi^{(T_{1/2})}$ fails at some step $h>l+1$}\}|\gB_1\gB_2)\notag\\
    &\geq \PP(\gB_1)\Big(\PP(a_{l+2}\in \gA_{l+2,\times}|\gB_1\gB_2)+\PP(\gB_3|\gB_1\gB_2)\PP\Big(\{\text{$\pi^{(T_{1/2})}$ fails at some step $h>l+2$}\}|\bigcap_{j=1}^3\gB_j\Big)\Big)\notag\\
    &\geq \PP(\gB_1)\PP(a_{l+2}\notin \{a_1,a_2,\cdots,a_{l-1},\up\}|\gB_1\gB_2)\PP\Big(\{\text{$\pi^{(T_{1/2})}$ fails at some step $h>l+2$}\}|\bigcap_{j=1}^3\gB_j\Big)\notag\\
    &\geq \cdots\notag\\
    &\geq \PP(\gB_1)\left(\prod_{i=1}^{k-l}\PP\Big(a_{l+2i}\notin \{a_1,a_2,\cdots,a_{l-1},\up\}|\bigcap_{j=1}^{2i}\gB_{j}\Big)\right)\notag\\
    &\quad\cdot\left(\PP\Big(a_{2k-l+1}\neq \up\Big|\bigcap_{j=1}^{2(k-l)+1}\gB_j\Big)+\PP\Big(\gB_{2(k-l+1)}\Big|\bigcap_{j=1}^{2(k-l)+1}\gB_j\Big)\PP\Big(a_{2k-l+2}\in \gA_{k-l+1,\times}\Big|\bigcap_{j=1}^{2(k-l+1)}\gB_{j}\Big)\right)\notag\\
    &\geq \PP(\gB_1)\left(\prod_{i=1}^{k-l}\PP\Big(a_{l+2i}\notin \{a_1,a_2,\cdots,a_{l-1},\up\}\Big|\bigcap_{j=1}^{2i}\gB_{j}\Big)\right)\PP\Big(a_{2k-l+2}\in \gA_{2k-l+2,\times}\Big|\bigcap_{j=1}^{2(k-l)+1}\gB_j\Big).
\end{align}
\normalsize
Note that conditioned on $\gB_1$, by Induction Hypothesis I and \eqref{eq:pi_h_simplified_balanced}, \eqref{eq:varphi_h_i_balanced}, \eqref{eq:xi_h_balanced}, \eqref{eq:Z_h_balanced}  in Lemma~\ref{lm:gradients}, we have  
\begin{align}
    \forall h\in[l]:\quad \varphi_h(m)&=-N_h(m)\left(a_{p,1}^{(T_{1/2})}+a_{p,2}^{(T_{1/2})}\right)+(h-1)a_{p,2}^{(T_{1/2})}+o(1),\notag\\
    &\xi_h=-a_{q,0}^{(T_{1/2})},
\end{align}
and thus conditioned on $\gB_1$, for all $h\in[l]$, we have for all $m\in[k]$:
\begin{align}
    \pi_h^{(T_{1/2})}(m)=(1+o(1))\frac{\exp\left(-N_h(m)\frac{a_{p,1}^{(T_{1/2})}+a_{p,2}^{(T_{1/2})}}{h}\right)}{\exp\left(-a_{q,0}^{(T_{1/2})}-\frac{h-1}{h}a_{p,2}^{(T_{1/2})}\right)+(h-1)\exp\left(-\frac{a_{p,1}^{(T_{1/2})}+a_{p,2}^{(T_{1/2})}}{h}\right)+k-h+1}.
\end{align}
From this, Induction Hypothesis I, and the fact that 
\begin{align}
    \PP\left(\text{goal is not at the first $k-l+1$ leaves $\pi^{(T_{1/2})}$ visits}\right)=1-\frac{k-l+1}{k^l},
\end{align}
 we deduce
 \small
\begin{align}\label{eq:PP_gB_1}
    &\PP(\gB_1)\notag\\
    &\geq \left(1-\frac{k-l+1}{k^l}\right)\prod_{h=1}^l\left(1-(1+o(1))\frac{\exp\left(-a_{q,0}^{(T_{1/2})}-\frac{h-1}{h}a_{p,2}^{(T_{1/2})}\right)+(h-1)\exp\left(-\frac{a_{p,1}^{(T_{1/2})}+a_{p,2}^{(T_{1/2})}}{h}\right)}{\exp\left(-a_{q,0}^{(T_{1/2})}-\frac{h-1}{h}a_{p,2}^{(T_{1/2})}\right)+(h-1)\exp\left(-\frac{a_{p,1}^{(T_{1/2})}+a_{p,2}^{(T_{1/2})}}{h}\right)+k-h+1}\right)\notag\\
    &\geq \left(1-\frac{k-l+1}{k^l}\right)\left(1-(1+o(1))\frac{\exp\left(-a_{q,0}^{(T_{1/2})}\right)}{k}\right)\notag\\
    &\overset{\eqref{eq:a_q_0>=1.9lnk_T_12}}\geq \left(1-\frac{c_1}{k^{2.9}}\right)\left(1-\frac{k-l+1}{k^l}\right)
\end{align}
\normalsize
for some constant $c_1>0$.

For all $i\in[k-l+1]$, conditioned on $\bigcap_{j=1}^{2i}\gB_{j}$, we have when $h=l+2i$:
\begin{align}
    |\gA_{l+2i,\times}|=N_{l+2i,\times}=N_{l+2i}(\up)=i,
\end{align}
and thus by Induction Hypothesis I and \eqref{eq:pi_h_simplified_balanced}, \eqref{eq:varphi_h_i_balanced}, \eqref{eq:xi_h_balanced}, \eqref{eq:Z_h_balanced} in Lemma~\ref{lm:gradients}, we have
\begin{align}
    \varphi_{l+2i}(m)&=-N_{l+2i}(m)\left(a_{p,1}^{(T_{1/2})}+a_{p,2}^{(T_{1/2})}\right)+(l+i-1)a_{p,2}^{(T_{1/2})}+o(1),\notag\\
    \xi_{l+2i}&=-a_{q,0}^{(T_{1/2})}+O\left(\frac{1}{\ln k}\right),
\end{align}
and thus by \eqref{eq:a_q_0>=1.9lnk_T_12}, \eqref{eq:a_p_2_bound_T_12} in Induction Hypothesis I and our choice of $T_{1/2}$, for all $i\in[k-l]$, we have
\begin{align}\label{eq:PP_a_l+2i_notin_a_1_a_2_cdots_a_l-1_up_gB_2i}
    &\PP\Big(a_{l+2i}\notin \{a_1,a_2,\cdots,a_{l-1},\up\}\Big|\bigcap_{j=1}^{2i}\gB_{j}\Big)\notag\\
    &=1-(1+o(1))\frac{\exp\left(-a_{q,0}^{(T_{1/2})}-\frac{l+i-1}{l+2i}a_{p,2}^{(T_{1/2})}\right)+(l-1)\exp\left(-\frac{a_{p,1}^{(T_{1/2})}+a_{p,2}^{(T_{1/2})}}{l+2i}\right)}{\exp\left(-a_{q,0}^{(T_{1/2})}-\frac{l+i-1}{l+2i}a_{p,2}^{(T_{1/2})}\right)+(l+i-1)\exp\left(-\frac{a_{p,1}^{(T_{1/2})}+a_{p,2}^{(T_{1/2})}}{l+2i}\right)+k-l-i+1}\notag\\
    &\geq 1-\frac{(1+o(1))l}{k^2},
\end{align}
and
\begin{align}\label{eq:PP_a_l+2i_in_gA_l+2i_times_gB_2i}
    &\PP\Big(a_{2k-l+2}\in \gA_{2k-l+2,\times}\Big|\bigcap_{j=1}^{2(k-l+1)}\gB_{j}\Big)\notag\\
    &=1-(1+o(1))\frac{(l-1)\exp\left(-\frac{a_{p,1}^{(T_{1/2})}}{l+2(k-l+1)}\right)+\exp\left(-a_{q,0}^{(T_{1/2})}-\frac{k}{l+2(k-l+1)}a_{p,2}^{(T_{1/2})}\right)}{k\exp\left(-\frac{a_{p,1}^{(T_{1/2})}}{l+2(k-l+1)}\right)+\exp\left(-a_{q,0}^{(T_{1/2})}-\frac{k}{l+2(k-l+1)}a_{p,2}^{(T_{1/2})}\right)}\notag\\
    &=1-(1+o(1))\frac{(l-1)\exp\left(-\frac{a_{p,1}^{(T_{1/2})}}{2k}\right)+\exp\left(-a_{q,0}^{(T_{1/2})}-\frac{1}{2}a_{p,2}^{(T_{1/2})}\right)}{k\exp\left(-\frac{a_{p,1}^{(T_{1/2})}}{2k}\right)+\exp\left(-a_{q,0}^{(T_{1/2})}-\frac{1}{2}a_{p,2}^{(T_{1/2})}\right)}\notag\\
    &\overset{\eqref{eq:a_p_2_bound_T_12}}=1-(1+o(1))\frac{l-1+\exp\left(-a_{q,0}^{(T_{1/2})}\right)}{k+\exp\left(-a_{q,0}^{(T_{1/2})}\right)}\notag\\
    &= 1-\frac{(1+o(1))l}{k}.
\end{align}
Plugging \eqref{eq:PP_gB_1}, \eqref{eq:PP_a_l+2i_notin_a_1_a_2_cdots_a_l-1_up_gB_2i} and \eqref{eq:PP_a_l+2i_in_gA_l+2i_times_gB_2i} into \eqref{eq:R_pi_T_1_l_phase_1_end_proof}, we deduce
\begin{align}
    1-R\left(\pi^{(T_{1/2})};\gP_l\right)\geq 1-\frac{c_2l}{k}
\end{align}
for some constant $c_2>0$. This gives \eqref{eq:R_pi_T_1_l_phase_1_end}.

\subsubsection{Proof of Lemma~\ref{lm:gradients_simplified_phase1.1_balanced}}\label{sec:proof_gradients_simplified_phase1_balanced}
\paragraph{Gradient computation.}
Note that when $l=1$, for all even steps, the agent is at a leaf node. If it's not a goal node, then the agent needs to go up, or the environment terminates. Thus for any $h\in[2k]$, conditioned on $\gE_{1,h}$, $N_{h,\times}=\floor{h/2}$. Therefore, we have
\begin{align}
    \delta a_{q,0}^{(t)}&\overset{\eqref{eq:delta_a_q_0_simplified_balanced}}=\-\EE\left[\sum_{h=1}^H \mathbbm{1}\{c_h=\times\} \frac{e^{-a_{c,1}^{(t)}}}{e^{a_{c,1}^{(t)}}+e^{-a_{c,1}^{(t)}}}\pi_h^{(t)}(\up)(1-\pi_h^{(t)}(\up))Q_h^{(t)}(\up)\bigg|l=1\right]\notag\\
    &\qquad +\EE\left[\sum_{h=1}^H \mathbbm{1}\{c_h=0\} \frac{(h+1)e^{a_{c,0}^{(t)}}}{(h-1)e^{-a_{c,0}^{(t)}}+(h+1)e^{a_{c,0}^{(t)}}}\pi_h^{(t)}(\up)G_h^{(t)}\bigg|l=1\right].
 \end{align}
With the notation we define at the beginning of Appendix~\ref{sec_app:phase_1.1}, by the above equation, we have
 \begin{align}
    &\delta a_{q,0}^{(t)}\notag\\
    &=\pi_{1,1}^{(t)}(\up)V_{1,1}^{(t)}+\sum_{h=1}^k \bigg(\PP^{(t)}\left(\gE_{1,2h+1}\right)\frac{1}{1+\frac{h}{h+1}e^{-2a_{c,0}^{(t)}}}\pi_{1,2h+1}^{(t)}(\up)\notag\\
    &\qquad -\PP^{(t)}\left(\gE_{1,2h}\right)\frac{1}{1+e^{2a_{c,1}^{(t)}}}\pi_{1,2h}^{(t)}(\up)\left(1-\pi_{1,2h}^{(t)}(\up)\right)\bigg)V_{1,2h+1}^{(t)}\notag\\
    &=\pi_{1,1}^{(t)}(\up)V_{1,1}^{(t)}+\sum_{h=1}^{k-1} \PP^{(t)}\left(\gE_{1,2h}\right)\pi_{1,2h}^{(t)}(\up)\bigg(\frac{1}{1+\frac{h}{h+1}e^{-2a_{c,0}^{(t)}}}\pi_{1,2h+1}^{(t)}(\up)-\frac{1}{1+e^{2a_{c,1}^{(t)}}}\left(1-\pi_{1,2h}^{(t)}(\up)\right)\bigg)V_{1,2h+1}^{(t)},
\end{align}
where the second relation uses 
\begin{align}
\EE^{\pi^{(t)}}[Q_{2h}^{(t)}(\up)|\gE_{1,2h}]=V_{1,2h+1}^{(t)}
\end{align}
by \eqref{eq:V_1_h}, and the third relation follows from the fact that 
\begin{align}\label{eq:PP_gE_1_2h_gE_1_2h+1_balanced}
    \PP^{(t)}\left(\gE_{1,2h+1}\right)=\pi_{1,2h}^{(t)}(\up)\PP^{(t)}\left(\gE_{1,2h}\right).
\end{align}
This gives \eqref{eq:delta_a_q_0_simplified_balanced_phase_1}.

Similarly, we have
\begin{align}
    \delta a_{q,\times}^{(t)}&\overset{\eqref{eq:delta_a_q_times_simplified_balanced}}=\EE\left[\sum_{h=1}^H \mathbbm{1}\{c_h=\times\} \frac{e^{a_{c,1}^{(t)}}}{e^{a_{c,1}^{(t)}}+e^{-a_{c,1}^{(t)}}}\pi_h^{(t)}(\up)(1-\pi_h^{(t)}(\up))Q_h^{(t)}(\up)\bigg|l=1\right]\notag\\
        &\qquad - \EE\left[\sum_{h=1}^H \mathbbm{1}\{c_h=0\} \frac{(h-1)e^{-a_{c,0}^{(t)}}}{(h-1)e^{-a_{c,0}^{(t)}}+(h+1)e^{a_{c,0}^{(t)}}}\pi_h^{(t)}(\up)G_h^{(t)}\bigg|l=1\right],
\end{align}
which gives \eqref{eq:delta_a_q_x_simplified_balanced_phase_1}.
\begin{align}
    \delta a_{c,0}^{(t)}&\overset{\eqref{eq:delta_a_c_0_simplified_balanced}}=\left(1+\frac{1}{k}\right)\EE\left[\sum_{h=1}^H \mathbbm{1}\{c_h=0\} \frac{(h-1)(h+1)}{\left((h-1)e^{-a_{c,0}^{(t)}}+(h+1)e^{a_{c,0}^{(t)}}\right)^2}\left(a_{q,\times}^{(t)}+a_{q,0}^{(t)}\right)\pi_h^{(t)}(\up)G_h^{(t)}\bigg|l=1\right],
\end{align}
which gives \eqref{eq:delta_a_c_0_simplified_balanced_phase_1}.
\begin{align}
    \delta a_{c,1}^{(t)}&\overset{\eqref{eq:delta_a_c_1_simplified_balanced}}=\left(1+\frac{1}{k}\right)\EE\left[\sum_{h=1}^H \mathbbm{1}\{c_h=\times\} \frac{1}{\left(e^{a_{c,1}^{(t)}}+e^{-a_{c,1}^{(t)}}\right)^2}\left(a_{q,\times}^{(t)}+a_{q,0}^{(t)}\right)\pi_h^{(t)}(\up)(1-\pi_h^{(t)}(\up))Q_h^{(t)}(\up)\bigg|l=1\right],
\end{align}
which gives \eqref{eq:delta_a_c_1_simplified_balanced_phase_1}.
\begin{align}
    \delta a_{p,0}^{(t)}&\overset{\eqref{eq:delta_a_p_0_simplified_balanced}}=(1+o(1))\EE\left[\sum_{h=1}^H \frac{\pi_h^{(t)}(\up)}{h}\left(Q_h^{(t)}(\up)-G_h^{(t)}\right)\bigg|l=1\right],
\end{align}
which gives \eqref{eq:delta_a_p_0_simplified_balanced_phase_1}. 
\begin{align}\label{eq:delta_a_p_1_intermediate_l=1}
    \delta a_{p,1}^{(t)} &\overset{\eqref{eq:delta_a_p_1_simplified_balanced}}=
    (1+o(1))\Bigg(\EE\left[\sum_{h=1}^H \mathbbm{1}\{c_h=0\}\frac{1}{k}\frac{\floor{h/2}}{h}\pi_h^{(t)}(j)G_h^{(t)}\bigg|j\in\gA_{h,\times},l=1\right]\notag\\
    &\quad\quad +\EE\left[\sum_{h=1}^H \mathbbm{1}\{c_h=\times\}\frac{N_h(j)}{h}\pi_h^{(t)}(j)\pi_h^{(t)}(\up)Q_h^{(t)}(\up)\bigg|j\in[k],l=1\right]\Bigg)\notag\\
    &=(1+o(1))\Bigg(\EE\left[\sum_{h=1}^{k-1}\frac{1}{k}\frac{h}{2h+1}\pi_{2h+1}^{(t)}(j)G_{2h+1}^{(t)}\bigg|j\in\gA_{2h+1,\times},l=1\right]\notag\\
    &\quad\quad +\EE\left[\sum_{h=1}^{k-1} \frac{1}{2k}\pi_{2h}^{(t)}(j)\pi_{2h}^{(t)}(\up)Q_{2h}^{(t)}(\up)\bigg|j\in\gA_{2h-1,\times}\cup\{a_{2h-1}\},l=1\right]\Bigg),
\end{align}
where for the even-step term we use
\[
\{N_{2h}(j)=1\}
=
\{j\in\gA_{2h-1,\times}\cup\{a_{2h-1}\}\},
\qquad
\PP\left(N_{2h}(j)=1\right)=\frac{h}{k}.
\]
On this event,
\[
\pi_{2h}^{(t)}(j)=\pi_{1,2h}^{(t)}(\down_\times^1).
\]
For the odd-step term, we use
\[
\PP\left(j\in\gA_{2h+1,\times}\right)=\frac{h}{k},
\qquad
\pi_{2h+1}^{(t)}(j)=\pi_{1,2h+1}^{(t)}(\down_\times).
\]
Therefore, \eqref{eq:delta_a_p_1_intermediate_l=1} gives \eqref{eq:delta_a_p_1_simplified_balanced_phase_1}. Similarly,
\begin{align}
    \delta a_{p,3}^{(t)} &\overset{\eqref{eq:delta_a_p_3_simplified_balanced}}=
    (1+o(1))\Bigg(\EE\left[\sum_{h=1}^H \mathbbm{1}\{c_h=0\} \frac{1}{k}\frac{\floor{h/2}}{h}\pi_h^{(t)}(\up)\left(Q_h^{(t)}(\up)-G_h^{(t)}\right)\bigg|j\in\gA_{h,\times},l=1\right]\notag\\
    &\quad\quad +\EE\left[\sum_{h=1}^H \mathbbm{1}\{c_h=\times\} \frac{1}{k}\frac{\floor{h/2}}{h}\pi_h^{(t)}(\up)\left(Q_h^{(t)}(\up)-G_h^{(t)}\right)\bigg|N_h(j)=1
    ,l=1\right]\Bigg)\notag\\
    &=(1+o(1))\Bigg(-\EE\left[\sum_{h=1}^{k-1}\frac{1}{k}\frac{h}{2h+1}\pi_{2h+1}^{(t)}(\up)G_{2h+1}^{(t)}\bigg|j\in\gA_{2h+1,\times},l=1\right]\notag\\
    &\quad \quad+\EE\left[\sum_{h=1}^{k-1} \frac{1}{2k}\pi_{2h}^{(t)}(\up)\left(1-\pi_{2h}^{(t)}(\up)\right)Q_{2h}^{(t)}(\up)\bigg|j\in\gA_{2h-1,\times}\cup\{a_{2h-1}\},l=1\right]\Bigg),
\end{align}
which gives \eqref{eq:delta_a_p_3_simplified_balanced_phase_1}.
\begin{align}
    \delta a_{p,\up}^{(t)} &\overset{\eqref{eq:delta_a_p_up_simplified_balanced}}=-(1+o(1))\EE\left[\sum_{h=1}^H \pi_h^{(t)}(\up)\frac{\floor{(h-1)/2}}{h}\left(Q_h^{(t)}(\up)-G_h^{(t)}\right)\bigg|l=1\right],
\end{align}
which gives \eqref{eq:delta_a_p_up_simplified_balanced_phase_1}.

To compute the simplified form of $\delta a_{b,i}^{(t)}$ for $i\in\{0,1,2\}$, we first simplify $\beta_{h,i}^{(t)}$ for $i\in\{0,1,2\}$ defined in \eqref{eq:beta_h_0_balanced}, \eqref{eq:beta_h_1_balanced} and \eqref{eq:beta_h_2_balanced} for $l=1$ as follows for all $h\in[k]$:
\begin{subequations}
\begin{align}
    \beta_{2h-1,0}^{(t)}&=-\frac{a_{p,0}^{(t)}}{k}\pi_{2h-1}^{(t)}(\up)G_{2h-1}^{(t)},\label{eq:beta_2h-1_0_l=1_balanced}\\
    \quad \beta_{2h,0}^{(t)}&=\frac{a_{p,0}^{(t)}}{k}\pi_{2h}^{(t)}(\up)\left(1-\pi_{2h}^{(t)}(\up)\right)Q_{2h}^{(t)}(\up),\label{eq:beta_2h_0_l=1_balanced}\\
    \beta_{2h-1,1}^{(t)}&=\left(\left(a_{p,1}^{(t)}+a_{p,2}^{(t)}\right)\pi_{2h-1}^{(t)}(\down_\times)+a_{p,2}^{(t)}\pi_{2h-1}^{(t)}(\up)\right)(h-1)G_{2h-1}^{(t)},\label{eq:beta_2h-1_1_l=1_balanced} \\
    \beta_{2h,1}^{(t)}&=0,\label{eq:beta_2h_1_l=1_balanced}\\
    \beta_{2h-1,2}^{(t)}&=\frac{h-1}{k}a_{p,\up}^{(t)}\pi_{2h-1}^{(t)}(\up)G_{2h-1}^{(t)},\label{eq:beta_2h-1_2_l=1_balanced}\\
    \beta_{2h,2}^{(t)}&=\left(-\left(ha_{p,2}^{(t)}+\frac{h-1}{k}a_{p,\up}^{(t)}\right)\left(1-\pi_{2h}^{(t)}(\up)\right)+\left(a_{p,1}^{(t)}+a_{p,2}^{(t)}\right)h\pi_{2h}^{(t)}(\down_\times^1)\right)\pi_{2h}^{(t)}(\up)Q_{2h}^{(t)}(\up).\label{eq:beta_2h_2_l=1_balanced}
\end{align}
\end{subequations}
By the above expressions and
\begin{align}
    \delta a_{b,0}^{(t)}&\overset{\eqref{eq:delta_a_b_0_simplified_balanced}}=\frac{1+o(1)}{N}\EE\left[\sum_{h=1}^H\frac{1}{h^2}\left((h-1)\beta_{h,0}^{(t)}-\beta_{h,1}^{(t)}-\beta_{h,2}^{(t)}\right)\right], \notag\\
    \delta a_{b,1}^{(t)}&\overset{\eqref{eq:delta_a_b_1_simplified_balanced}}=\frac{1+o(1)}{N}\EE\left[\sum_{h=1}^{k-1} \frac{1}{(2h+1)^2}\left((h+1)\beta_{2h+1,1}^{(t)}-h\beta_{2h+1,0}^{(t)}-h\beta_{2h+1,2}^{(t)}\right)\right],\notag
\end{align}
we obtain \eqref{eq:delta_a_b_0_simplified_balanced_phase_1} and \eqref{eq:delta_a_b_1_simplified_balanced_phase_1}.

\paragraph{Policy computation.} By 
Induction Hypothesis I, when $l=1$, \eqref{eq:varphi_h_i_balanced} can be simplified to (recall we assume $N\geq k^5$)
\begin{align}
    \forall i\in[k]:\quad\varphi_{2h-1}^{(t)}(i)&= -\frac{a_{p,0}^{(t)}}{k}+(h-1)\frac{a_{p,\up}^{(t)}}{k}+(h-1)a_{p,2}^{(t)}-N_{2h-1}(i)\left(a_{p,1}^{(t)}+a_{p,2}^{(t)}\right)+O\left(\frac{\ln^3 k}{k}\right),\label{eq:varphi_2h-1_i_l=1_balanced}\\
    \varphi_{2h}^{(t)}(i)&= -\frac{a_{p,0}^{(t)}}{k}+(h-1)\frac{a_{p,\up}^{(t)}}{k}+h a_{p,2}^{(t)}-N_{2h}(i)\left(a_{p,1}^{(t)}+a_{p,2}^{(t)}\right)+O\left(\frac{\ln^3 k}{k}\right),\label{eq:varphi_2h_i_l=1_balanced}
\end{align}
and \eqref{eq:xi_h_balanced} can be simplified to \eqref{eq:xi_h_phase1}.
By Induction Hypothesis I we know that 
\begin{align}\label{eq:small_quantities_in_policy_computation}
   \forall h\in[k]:\quad \left|\frac{1}{k}\xi_{h}^{(t)}\right|\lesssim \frac{\ln k}{k},\quad \left|\frac{a_{p,0}^{(t)}}{k}\right|\lesssim \frac{\ln k}{k},\quad \left|\frac{a_{p,\up}^{(t)}}{k}\right|\lesssim \frac{\ln k}{k}.
\end{align}
Combining  \eqref{eq:varphi_2h-1_i_l=1_balanced}, \eqref{eq:varphi_2h_i_l=1_balanced}, \eqref{eq:xi_h_phase1}, \eqref{eq:small_quantities_in_policy_computation} with \eqref{eq:pi_h_simplified_balanced}-\eqref{eq:Z_h_balanced} in Lemma~\ref{lm:gradients}, we have \eqref{eq:pi_2h-1_up_phase1}, \eqref{eq:pi_2h_up_phase1}, \eqref{eq:pi_1_2h-1_i_phase1} and \eqref{eq:pi_1_2h_i_phase1} hold.

\subsubsection{Proof of Lemma~\ref{lm:G_2h_1_bound}}\label{sec:proof_G_2h_1_bound}
We drop the superscript $(t)$ for simplicity.
    By the definition of $\widehat{G}_h$ in \eqref{eq:pi_G_Q_h} and the definition of $V_{1,h}$ in \eqref{eq:V_1_h}, we have the following recursion:
    \begin{align}\label{eq:G_h_recursion}
    V_{1,2k+1}=0,\quad \forall h\leq k:\,\,V_{1,2h-1}=\underbrace{\pi_{1,2h-1}(\down)}_{(i)}+\underbrace{(k-h)\pi_{1,2h-1}(\down)}_{(ii)}\pi_{1,2h}(\up)V_{1,2h+1}.
    \end{align}
where $(i)$ is the probability of the goal is reached after taking some down actions $i\in[k]\setminus\gA_{h,\times}$, and $(ii)$ is the probability of the goal is not reached after taking a down action $i\in[k]\setminus\gA_{h,\times}$ (note if $\up$ or $i\in\gA_{h,\times}$ is taken, the goal will not be reached and the environment terminates). 

To simplify notation, we define
\begin{align*}
    y_h\coloneqq V_{1,2h-1},\quad r_h=1+(k-h)\pi_{1,2h}(\up)V_{1,2h+1},\quad \beta_h=(k-h)\pi_{1,2h}(\up)\pi_{1,2h+1}(\down),
\end{align*}
then
\begin{align}\label{eq:y_h_ratio}
    \frac{y_h}{y_1}=\frac{\pi_{1,2h-1}(\down)r_h}{\pi_{1,1}(\down)r_1},
\end{align}
and
\begin{align*}
    r_k=1,\quad r_h=1+\beta_h r_{h+1}.
\end{align*}
If $\pi_{1,2}\lesssim \frac{1}{\ln k}$, then by \eqref{eq:pi_1_2h_up_bound_T_11}, 
\begin{align*}
    \forall h\in[k]:\quad \pi_{1,2h}(\up)\lesssim \frac{1}{\ln k}.
\end{align*}
This suggests $\beta_h\lesssim \frac{1}{\ln k}$ for any $h\in[k-1]$.
Thus we have
\begin{align*}
    r_1=1+\beta_1 r_2=\cdots=1+\beta_1 +\beta_1\beta_2 +\cdots+\beta_1\beta_2\cdots\beta_{k-1}=1+O\left(\frac{1}{\ln k}\right).
\end{align*}
Similarly, we have $r_h=1+O(\frac{1}{\ln k})$ for any $h\in[k]$. Thus $r_h/r_1= 1+O(\frac{1}{\ln k})$.
Moreover, when $s-1\leq T_{1,1}$, by Induction Hypothesis I we have
\begin{align*}
    \pi_{1,2h-1}(\down)=(1+o(1))\frac{1}{k},\quad \pi_{1,1}(\down)=(1+o(1))\frac{1}{k}.
\end{align*}
Thus $\pi_{1,2h-1}(\down)/\pi_{1,1}(\down)= 1+o(1)$.
Therefore, by \eqref{eq:y_h_ratio}, $y_h/y_1= 1+o(1)$. This gives \eqref{eq:G_2h_1_bound}.

If $\pi_{1,2}^{(s-1)}\gtrsim \frac{1}{\ln k}$, then since $s-1\leq T_{1,1}$, by \eqref{eq:pi_1_2h_up_bound_T_11}, we have
\begin{align*}
    \forall h\in[k]:\quad \pi_{1,2h}(\up)\leq \frac{1+o(1)}{2},
\end{align*}
which indicates $r_1\asymp 1$, $r_h\asymp 1$ for any $h\in[k]$. Thus by \eqref{eq:y_h_ratio}, \eqref{eq:G_2h_1_bound_asymp} holds.

\subsubsection{Proof of Lemma~\ref{lm:phase_1.2}}\label{sec_app:phase_1.2}
We show Lemma~\ref{lm:phase_1.2} by induction. First by Lemma~\ref{lm:phase_1.1} we know that Lemma~\ref{lm:phase_1.2} holds for $t=T_{1/2}$. We make the following induction hypothesis:
\begin{center}
\textbf{Induction Hypothesis II:} For all $t\leq s-1$ ($s\in [T_{1/2}+1,T_1]$), the relations in Lemma~\ref{lm:phase_1.2} hold.
\end{center}

Below we show Lemma~\ref{lm:phase_1.2} holds for $t=s$ under Induction Hypothesis II.  We still follow the notation defined at the beginning of Appendix~\ref{sec_app:phase_1.1}. We first simplify the gradients for Phase 1.2.

\paragraph{Gradients/policy simplification for phase 1.2.} Similar as Lemma~\ref{lm:gradients_simplified_phase1.1_balanced}, by Induction Hypothesis II, using the notation we define at the beginning of Appendix~\ref{sec_app:phase_1.1}, we could further simplify the gradients and policy expressions in Lemma~\ref{lm:gradients} when $l=1$. We summarize the results in the following lemma. 
\begin{lm}\label{lm:gradients_simplified_phase1.2_balanced}
    Under Induction Hypothesis II, we have for any $t\in[T_{1/2},s-1]$:
    \begin{subequations}
    \begin{align}
        \delta a_{q,0}^{(t)}&=\left(1+o\left(1\right)\right)\Bigg(\pi_{1,1}^{(t)}(\up)+\sum_{h=1}^{k-1} \frac{k-h}{k}\bigg(\frac{1}{1+\frac{h}{h+1}e^{-2a_{c,0}^{(t)}}}\pi_{1,2h+1}^{(t)}(\up)-\frac{1}{1+e^{2a_{c,1}^{(t)}}}\left(1-\pi_{1,2h}^{(t)}(\up)\right)\bigg)\Bigg), \label{eq:delta_a_q_0_simplified_balanced_phase_1.2} \\
     \delta a_{q,\times}^{(t)}&=\left(1+o\left(1\right)\right)\sum_{h=1}^{k-1} \frac{k-h}{k}\bigg(\frac{1}{1+e^{-2a_{c,1}^{(t)}}}\left(1-\pi_{1,2h}^{(t)}(\up)\right)-\frac{1}{1+\frac{h+1}{h}e^{2a_{c,0}^{(t)}}}\pi_{1,2h+1}^{(t)}(\up)\bigg), \label{eq:delta_a_q_x_simplified_balanced_phase_1.2} \\
         \delta a_{c,0}^{(t)}&=\left(1+o\left(1\right)\right)\sum_{h=1}^{k-1} \frac{k-h}{k}\frac{h(h+1)}{\left(he^{-a_{c,0}^{(t)}}+(h+1)e^{a_{c,0}^{(t)}}\right)^2}\left(a_{q,\times}^{(t)}+a_{q,0}^{(t)}\right)\pi_{1,2h+1}^{(t)}(\up), \label{eq:delta_a_c_0_simplified_balanced_phase_1.2} \\
         \delta a_{c,1}^{(t)}&=\left(1+o\left(1\right)\right)\sum_{h=1}^{k-1} \frac{k-h}{k}\frac{1}{\left(e^{a_{c,1}^{(t)}}+e^{-a_{c,1}^{(t)}}\right)^2}\left(a_{q,\times}^{(t)}+a_{q,0}^{(t)}\right)(1-\pi_{1,2h}^{(t)}(\up)), \label{eq:delta_a_c_1_simplified_balanced_phase_1.2} \\
         \delta a_{p,0}^{(t)} &=\left(1+o\left(1\right)\right)\Bigg(-\pi_{1,1}^{(t)}(\up)+\sum_{h=1}^{k-1} \frac{k-h}{k}\Bigg(\frac{e^{a_{b,0}^{(t)}}}{e^{a_{b,0}^{(t)}}+(2h-1)e^{a_{b,2}^{(t)}}}\left(1-\pi_{1,2h}^{(t)}(\up)\right)\notag\\
         &\hspace{6.3cm}-\frac{e^{a_{b,0}^{(t)}}}{e^{a_{b,0}^{(t)}}+he^{a_{b,1}^{(t)}}+he^{a_{b,2}^{(t)}}}\pi_{1,2h+1}^{(t)}(\up)\Bigg)\Bigg), \label{eq:delta_a_p_0_simplified_balanced_phase_1.2} \\
     \delta a_{p,1}^{(t)} &=\frac{1+o(1)}{k}\sum_{h=1}^{k-1}\frac{k-h}{k}\Bigg(\frac{he^{a_{b,2}^{(t)}}}{e^{a_{b,0}^{(t)}}+(2h-1)e^{a_{b,2}^{(t)}}}\pi_{1,2h}^{(t)}(\down_\times^1)+\frac{he^{a_{b,1}^{(t)}}}{e^{a_{b,0}^{(t)}}+he^{a_{b,1}^{(t)}}+he^{a_{b,2}^{(t)}}}\pi_{1,2h+1}^{(t)}(\down_\times)\Bigg), \label{eq:delta_a_p_1_simplified_balanced_phase_1.2} \\
     \delta a_{p,3}^{(t)} &=\frac{1+o(1)}{k}\sum_{h=1}^{k-1} \frac{k-h}{k}\Bigg(\frac{he^{a_{b,2}^{(t)}}}{e^{a_{b,0}^{(t)}}+(2h-1)e^{a_{b,2}^{(t)}}}\left(1-\pi_{1,2h}^{(t)}(\up)\right)-\frac{he^{a_{b,1}^{(t)}}}{e^{a_{b,0}^{(t)}}+he^{a_{b,1}^{(t)}}+he^{a_{b,2}^{(t)}}}\pi_{1,2h+1}^{(t)}(\up)\Bigg), \label{eq:delta_a_p_3_simplified_balanced_phase_1.2} \\
     \delta a_{p,\up}^{(t)}&=\left(1+o(1)\right)\sum_{h=1}^{k-1}\frac{k-h}{k}\Bigg(\frac{he^{a_{b,2}^{(t)}}}{e^{a_{b,0}^{(t)}}+he^{a_{b,1}^{(t)}}+he^{a_{b,2}^{(t)}}}\pi_{1,2h+1}^{(t)}(\up)-\frac{(h-1)e^{a_{b,2}^{(t)}}}{e^{a_{b,0}^{(t)}}+(2h-1)e^{a_{b,2}^{(t)}}}\left(1-\pi_{1,2h}^{(t)}(\up)\right)\Bigg), \label{eq:delta_a_p_up_simplified_balanced_phase_1.2} \\
     \delta a_{b,0}^{(t)}& =\frac{1+o(1)}{N}\sum_{h=1}^{k-1}\frac{k-h}{k}I_{h,0}^{(t)}, \label{eq:delta_a_b_0_simplified_balanced_phase_1.2} \\
          \delta a_{b,1}^{(t)}&=\frac{1+o(1)}{N}\sum_{h=1}^{k-1}\frac{k-h}{k}I_{h,1}^{(t)}, \label{eq:delta_a_b_1_simplified_balanced_phase_1.2}
 \end{align}
 \end{subequations}
 where 
 \small
 \begin{align}
     I_{h,0}^{(t)}&\coloneqq \frac{e^{a_{b,0}^{(t)}}e^{a_{b,2}^{(t)}}}{\left(e^{a_{b,0}^{(t)}}+(2h-1)e^{a_{b,2}^{(t)}}\right)^2}\left(\left(\frac{2h-1}{k}a_{p,0}^{(t)}+ha_{p,2}^{(t)}+\frac{h-1}{k}a_{p,\up}^{(t)}\right)\left(1-\pi_{1,2h}^{(t)}(\up)\right)-\left(a_{p,1}^{(t)}+a_{p,2}^{(t)}\right)h\pi_{1,2h}^{(t)}(\down_\times^1)\right)\notag\\
    &\quad -\frac{he^{a_{b,0}^{(t)}}}{\left(e^{a_{b,0}^{(t)}}+he^{a_{b,1}^{(t)}}+he^{a_{b,2}^{(t)}}\right)^2}\Bigg[\left(e^{a_{b,1}^{(t)}}\left(a_{p,2}^{(t)}+\frac{a_{p,0}^{(t)}}{k}\right)+e^{a_{b,2}^{(t)}}\left(\frac{a_{p,\up}^{(t)}}{k}+\frac{a_{p,0}^{(t)}}{k}\right)\right)\pi_{1,2h+1}^{(t)}(\up)\notag\\
    &\hspace{6.3cm}+e^{a_{b,1}^{(t)}}\left(a_{p,1}^{(t)}+a_{p,2}^{(t)}\right)\pi_{1,2h+1}^{(t)}(\down_\times)\Bigg], \label{eq:I_h_0_balanced_phase1.2} \\
     I_{h,1}^{(t)}&\coloneqq \frac{he^{a_{b,1}^{(t)}}}{\left(e^{a_{b,0}^{(t)}}+he^{a_{b,1}^{(t)}}+he^{a_{b,2}^{(t)}}\right)^2}\Bigg[\left(e^{a_{b,0}^{(t)}}+he^{a_{b,2}^{(t)}}\right)\left(a_{p,1}^{(t)}+a_{p,2}^{(t)}\right)\pi_{1,2h+1}^{(t)}(\down_\times)\notag\\
    &\qquad\qquad+\Bigg(e^{a_{b,0}^{(t)}}\Bigg(a_{p,2}^{(t)}+\frac{a_{p,0}^{(t)}}{k}\Bigg)+he^{a_{b,2}^{(t)}}\Bigg(a_{p,2}^{(t)}-\frac{a_{p,\up}^{(t)}}{k}\Bigg)\Bigg)\pi_{1,2h+1}^{(t)}(\up)\Bigg]. \label{eq:I_h_1_balanced_phase1.2}
 \end{align}
 \normalsize
 For all $\forall h\in[k]$, $i\in[k]$, we have 
 \begin{align}
        \pi_{1,h}^{(t)}\coloneqq \sm\left(\overline\phi_h^{(t)}\right),
 \end{align}
 where $\overline\phi_h^{(t)}\in\RR^{k+1}=\left(\overline\phi_{h,1}^{(t)},\cdots,\overline\phi_{h,k}^{(t)},\overline\phi_{h,\up}^{(t)}\right)^\top$ is defined as 
\begin{align}
    \forall i\in[k]:\quad\overline\phi_{2h-1,i}^{(t)}\coloneqq -\frac{N_{2h-1}(i)(a_{p,1}^{(t)}+a_{p,2}^{(t)})e^{a_{b,1}^{(t)}}}{e^{a_{b,0}^{(t)}}+(h-1)e^{a_{b,1}^{(t)}}+(h-1)e^{a_{b,2}^{(t)}}},
    \quad  \overline\phi_{2h,i}^{(t)} & \coloneqq -\frac{N_{2h}(i)(a_{p,1}^{(t)}+a_{p,2}^{(t)})e^{a_{b,2}^{(t)}}}{e^{a_{b,0}^{(t)}}+(2h-1)e^{a_{b,2}^{(t)}}}. \label{eq:overline_phi_h_i} \\
    \overline\phi_{h,\up}^{(t)}& \coloneqq \left(1+\frac{1}{k}\right)\xi_h^{(t)}-\mu_h^{(t)}, \label{eq:overline_phi_h_up}
\end{align}
where $\xi_h^{(t)}$ is defined in \eqref{eq:xi_h_phase1} and $\mu_h^{(t)}$ is defined in \eqref{eq:mu_h_phase1.2}.
\end{lm}
The proof of Lemma~\ref{lm:gradients_simplified_phase1.2_balanced} is postponed to Appendix~\ref{sec:proof_gradients_simplified_phase1.2_balanced}.

From \eqref{eq:delta_a_p_1_simplified_balanced_phase_1.2} and \eqref{eq:delta_a_b_1_simplified_balanced_phase_1.2} we immediately know that \eqref{eq:a_p_1_increase_phase_1.2} and \eqref{eq:delta_ea_b_1_bound_phase_1.2} hold at step $s$.
Also note that \eqref{eq:delta_e2a_c_i} still holds. And similar as \eqref{eq:delta_e2a_c_1_bound_T_11} for Phase 1.1, by comparing \eqref{eq:delta_a_c_1_simplified_balanced_phase_1.2} and \eqref{eq:delta_a_q_x_simplified_balanced_phase_1.2}, we have for any $t\leq s-1$:
\begin{align}
    \delta e^{2a_{c,1}^{(t)}}\geq \max\{(2+o(1))\left(a_{q,\times}^{(t)}+a_{q,0}^{(t)}\right)\delta a_{q,\times}^{(t)},0\}.
\end{align} 
This indicates \eqref{eq:e2a_c_1_bound_phase_1.2} holds at step $s$.
From \eqref{eq:delta_a_c_0_simplified_balanced_phase_1.2} we know that 
\begin{align}\label{eq:a_c_0_increase_phase1.2}
    \delta a_{c,0}^{(s-1)}>0,
\end{align}
and thus by Induction Hypothesis II and \eqref{eq:e2a_c_0>=lnk^2_T_12} in Lemma~\ref{lm:phase_1.1} we know that 
\begin{align}\label{eq:e2a_c_0_lb_phase1.2}
    e^{2a_{c,0}^{(s)}}\gtrsim \ln^2 k.
\end{align}

By \eqref{eq:a_q_x_q_0_p_2_asymp_phase_1.2} and \eqref{eq:a_q_x+a_q_0_=a_q_x_T1} in Lemma~\ref{lm:phase_1.1} we know that there exist constants $0<C_1\leq C_2$ such that for all $t\leq s-1$:
\begin{align}
    C_1\left( a_{q,0}^{(t)}+a_{q,\times}^{(t)}\right)\leq a_{q,\times}^{(t)}\leq C_2\left( a_{q,0}^{(t)}+a_{q,\times}^{(t)}\right),
\end{align}
and by 
\eqref{eq:e2a_c_1_bound_phase_1.2} and \eqref{eq:delta_a_c_1_balanced_phase_1} in Lemma~\ref{lm:phase_1.1} we have for any $t'\leq s-1$:
\begin{align}
    e^{2a_{c,1}^{(s-1)}}-e^{2a_{c,1}^{(t')}}\geq \left(a_{q,\times}^{(t')}+a_{q,0}^{(t')}\right)\left(a_{q,\times}^{(s-1)}-a_{q,\times}^{(t')}\right).
\end{align}
By our update rule, we can choose $t'\leq s-1$ such that 
\begin{align}
    a_{q,\times}^{(t')}+a_{q,0}^{(t')}\in\left[\frac{C_1}{2C_2}\left(a_{q,\times}^{(s-1)}+a_{q,0}^{(s-1)}\right),\frac{C_1}{2C_2}\left(a_{q,\times}^{(s-1)}+a_{q,0}^{(s-1)}\right)+1\right].
\end{align}
Combining the above three relations, we have
\begin{align}\label{eq:e^2ac1=(a_q_x+a_q_0)^2}
    e^{2a_{c,1}^{(s-1)}}&\geq \frac{C_1}{2C_2}\left(a_{q,\times}^{(s-1)}+a_{q,0}^{(s-1)}\right)\left(C_1\left(a_{q,\times}^{(s-1)}+a_{q,0}^{(s-1)}\right)-C_2\left(a_{q,\times}^{(t')}+a_{q,0}^{(t')}\right)\right)\notag\\
    &\geq \frac{C_1}{2C_2}\left(a_{q,\times}^{(s-1)}+a_{q,0}^{(s-1)}\right)\left(\frac{C_1}{2}\left(a_{q,\times}^{(s-1)}+a_{q,0}^{(s-1)}\right)-C_2\right)\notag\\
    &\asymp \left(a_{q,\times}^{(s-1)}+a_{q,0}^{(s-1)}\right)^2.
\end{align}
By \eqref{eq:e^2ac1=(a_q_x+a_q_0)^2} and \eqref{eq:xi_h_phase1} we have
\begin{align}\label{eq:xi_2h=a_q_x_phase1.2}
    \xi_{2h}^{(s-1)}=a_{q,\times}^{(s-1)}+O\left(\frac{1}{a_{q,\times}^{(s-1)}+a_{q,0}^{(s-1)}}\right)=a_{q,\times}^{(s-1)}+O\left(\frac{1}{\ln k}\right).
\end{align}
By \eqref{eq:e2a_c_0_balanced_phase_1.2} and \eqref{eq:e2a_c_0>=lnk^2_T_12} in Lemma~\ref{lm:phase_1.1} we have
\begin{align}
    e^{2a_{c,0}^{(s-1)}}\gtrsim \ln^2 k.
\end{align}
Thus by \eqref{eq:xi_h_phase1} and \eqref{eq:a_q_x_q_0_p_2_asymp_phase_1.2} we have
\begin{align}
    \xi_{2h-1}^{(s-1)}=-\left(1+O\left(\frac{1}{\ln^2 k}\right)\right)a_{q,0}^{(s-1)}.
\end{align}

For all $h\in[k-1]$, define
\begin{align}
    r_{\up,h,0}^{(t)}\coloneqq \frac{e^{a_{b,0}^{(t)}}}{e^{a_{b,0}^{(t)}}+(2h-1)e^{a_{b,2}^{(t)}}},\quad r_{\up,h,2}^{(t)}\coloneqq \frac{e^{a_{b,2}^{(t)}}}{e^{a_{b,0}^{(t)}}+(2h-1)e^{a_{b,2}^{(t)}}}, \label{eq:r_up_h_phase1.2} \\
    r_{\down,h,0}^{(t)}\coloneqq \frac{e^{a_{b,0}^{(t)}}}{e^{a_{b,0}^{(t)}}+he^{a_{b,1}^{(t)}}+he^{a_{b,2}^{(t)}}},\quad r_{\down,h,1}^{(t)}\coloneqq \frac{e^{a_{b,1}^{(t)}}}{e^{a_{b,0}^{(t)}}+he^{a_{b,1}^{(t)}}+he^{a_{b,2}^{(t)}}},\notag\\ 
    r_{\down,h,2}^{(t)}\coloneqq \frac{e^{a_{b,2}^{(t)}}}{e^{a_{b,0}^{(t)}}+he^{a_{b,1}^{(t)}}+he^{a_{b,2}^{(t)}}}. \label{eq:r_down_h_phase1.2} 
\end{align}
Then by \eqref{eq:r_down_times_phase1.2} we have for any $t$,
\begin{align*}
    r_{\down_\times}^{(t)}=r_{\down,k-1,1}^{(t)}.
\end{align*}
And by \eqref{eq:mu_h_phase1.2} and \eqref{eq:a_p_0_bound_phase_1.2}, \eqref{eq:a_p_up_bound_phase_1.2}, we have
\begin{align*}
    \mu_{1}^{(s-1)}=-\frac{a_{p,0}^{(s-1)}}{k},
\end{align*}
and for all $h\in[k-1]$:
\begin{align*}
    \mu_{2h+1}^{(s-1)}&=\left(1+O\left(\frac{1}{k}\right)\right)
    hr_{\down,h,1}^{(s-1)}a_{p,2}^{(s-1)}\geq \frac{h}{2h+1}a_{p,2}^{(s-1)},
\end{align*}
where the second inequality also uses the fact that $a_{b,1}^{(s-1)}$ is larger than $a_{b,0}^{(s-1)}$ and $a_{b,2}^{(s-1)}$ guaranteed by Induction Hypothesis II.
Thus by the gradient expressions in Lemma~\ref{lm:gradients_simplified_phase1.2_balanced} and the above three relations we have
\begin{align}
    \pi_{1,1}^{(s-1)}(\up)&=\frac{\exp\left(\overline\phi_{1,\up}^{(s-1)}\right)}{k+\exp\left(\overline\phi_{1,\up}^{(s-1)}\right)}\overset{\eqref{eq:overline_phi_h_up}}=\frac{\exp\left(-\left(1+\frac{1}{k}\right)a_{q,0}^{(s-1)}+\frac{a_{p,0}^{(s-1)}}{k}\right)}{k+\exp\left(-\left(1+\frac{1}{k}\right)a_{q,0}^{(s-1)}+\frac{a_{p,0}^{(s-1)}}{k}\right)}, \label{eq:pi_1_1_up_phase1.2} \\
    \forall h\in[k-1]:\quad \frac{k-h}{k}\pi_{1,2h+1}^{(s-1)}(\up)&\leq\frac{k-h}{k}\frac{\exp\left(-\left(1+O\left(\frac{1}{\ln^2 k}\right)\right)a_{q,0}^{(s-1)}-\frac{h}{2h+1}a_{p,2}^{(s-1)}\right)}{k-h}\notag\\
    &\leq\frac{\exp\left(-\left(1+\frac{1}{k}\right)a_{q,0}^{(s-1)}-\left(\frac{1}{3}+O\left(\frac{1}{\ln^2 k}\right)\right)a_{p,2}^{(s-1)}\right)}{k}\notag\\
    &\leq \frac{1}{k^{4/3}}\pi_{1,1}^{(s-1)}(\up), \label{eq:pi_1_up_dominate_phase1.2_intermediate}
\end{align}
where the second inequality uses \eqref{eq:a_q_x_q_0_p_2_asymp_phase_1.2}, the third inequality uses \eqref{eq:T_1/2}. From \eqref{eq:pi_1_up_dominate_phase1.2_intermediate} we know that 
\begin{align}\label{eq:pi_1_up_dominate_sum_phase1.2}
    \pi_{1,1}^{(s-1)}(\up)\geq k^{1/3}\sum_{h=1}^{k-1}\frac{k-h}{k}\pi_{1,2h+1}^{(s-1)}(\up).
\end{align}
And \eqref{eq:pi_1_1_up_phase1.2}, \eqref{eq:pi_1_up_dominate_phase1.2_intermediate}, \eqref{eq:a_q_x_q_0_p_2_asymp_phase_1.2}, \eqref{eq:a_p_0_bound_phase_1.2} also give the following inequality:
\begin{align}\label{eq:tiny_pi_1_2h+1_up_phase1.2}
    \sum_{h=1}^{k-1}\frac{k-h}{k}\pi_{1,2h+1}^{(s-1)}(\up)\leq \frac{\pi_{1,1}^{(s-1)}(\up)}{\exp\left(0.3a_{p,2}^{(s-1)}\right)}.
\end{align}


Similarly, from Lemma~\ref{lm:gradients_simplified_phase1.2_balanced} and Induction Hypothesis II we also know that
\begin{align}\label{eq:pi_1_2h+1_down_times_simplified_phase1.2}
    \forall h\in[k-1]:\quad \pi_{1,2h+1}^{(s-1)}(\down_\times)&=\left(1+o\left(\frac{1}{k}\right)\right)\frac{\exp\left(-r_{\down,h,1}^{(s-1)}(a_{p,1}^{(s-1)}+a_{p,2}^{(s-1)})\right)}{k-h},
\end{align}
which further indicates
\begin{align}\label{eq:sum_pi_1_2h+1_down_times_bound_phase1.2}
    \frac{1+o(1)}{k}\exp\left(-r_{\down_\times}^{(s-1)}\left(a_{p,1}^{(s-1)}+a_{p,2}^{(s-1)}\right)\right)&\leq \sum_{h=1}^{k-1}\frac{k-h}{k}\pi_{1,2h+1}^{(s-1)}(\down_\times)\notag\\
    &\leq \frac{1}{2}\exp\left(-r_{\down_\times}^{(s-1)}\left(a_{p,1}^{(s-1)}+a_{p,2}^{(s-1)}\right)\right).
\end{align}
Analogously, we have
\begin{align}\label{eq:1-pi_2h_bound_phase1.2}
    \forall h\in[k-1]:\quad (1+o(1))\frac{k-h}{\exp\left(a_{q,\times}^{(s-1)}-\mu_{2h}^{(s-1)}\right)}&\leq 1-\pi_{1,2h}^{(s-1)}(\up)\notag\\
    &\leq(1+o(1))\frac{k}{\exp\left(a_{q,\times}^{(s-1)}-\mu_{2h}^{(s-1)}\right)},
\end{align}
from which we deduce
\begin{align}\label{eq:sum_1-pi_2h_up_lb}
    \sum_{h=1}^{k-1}\frac{k-h}{k}\left(1-\pi_{1,2h}^{(s-1)}(\up)\right) 
    &\geq (1+o(1))\max\left\{\frac{k}{\exp\left(a_{q,\times}^{(s-1)}-\mu_{2}^{(s-1)}\right)},\frac{1}{k\exp\left(a_{q,\times}^{(s-1)}-\mu_{2(k-1)}^{(s-1)}\right)}\right\}\notag\\
    &\geq (1+o(1))\exp\left(-\min\left\{a_{q,\times}^{(s-1)}-\mu_{2}^{(s-1)}-\ln k,a_{q,\times}^{(s-1)}-\mu_{2(k-1)}^{(s-1)}+\ln k\right\}\right).
\end{align}
Note that by \eqref{eq:mu_h_phase1.2} we have
\begin{align}
    \min_{h\in[k-1]}\left\{\mu_{2h}^{(s-1)}\right\}=\min\{\mu_2^{(s-1)},\mu_{2(k-1)}^{(s-1)}\},\quad \max_{h\in[k-1]}\left\{\mu_{2h}^{(s-1)}\right\}=\max\{\mu_2^{(s-1)},\mu_{2(k-1)}^{(s-1)}\}.
\end{align}
Combining this with \eqref{eq:1-pi_2h_bound_phase1.2}, we have
\begin{align}\label{eq:sum_1-pi_2h_up_ub}
    \sum_{h=1}^{k-1}\frac{k-h}{k}\left(1-\pi_{1,2h}^{(s-1)}(\up)\right)&\leq (1+o(1))\sum_{h=1}^{k-1}\frac{k-h}{k}\frac{k}{\exp\left(a_{q,\times}^{(s-1)}-\max\left\{\mu_{2}^{(s-1)},\mu_{2(k-1)}^{(s-1)}\right\}\right)}\notag\\
    &\leq \frac{1+o(1)}{2}k^2\exp\left(-\min\left\{a_{q,\times}^{(s-1)}-\mu_{2}^{(s-1)},a_{q,\times}^{(s-1)}-\mu_{2(k-1)}^{(s-1)}\right\}\right).
\end{align}
By Lemma~\ref{lm:gradients_simplified_phase1.2_balanced} we also know that
\begin{align}\label{eq:pi_2h_down_1_bound_phase1.2}
    \forall h\in[k-1]:\,\, \pi_{1,2h}^{(s-1)}(\down_\times^1)&=\left(1-\pi_{1,2h}^{(s-1)}(\up)\right)\frac{\exp\left(-r_{\up,h,2}^{(s-1)}\left(a_{p,1}^{(s-1)}+a_{p,2}^{(s-1)}\right)\right)}{k-h+h\exp\left(-r_{\up,h,2}^{(s-1)}\left(a_{p,1}^{(s-1)}+a_{p,2}^{(s-1)}\right)\right)}\notag\\
    &\leq \frac{1-\pi_{1,2h}^{(s-1)}(\up)}{k},
\end{align}
and thus 
\begin{align}\label{eq:sum_pi_2h_down_1=1/k_sum_1-pi_2h_up}
    \sum_{h=1}^{k-1}\frac{k-h}{k}\pi_{1,2h}^{(s-1)}(\down_\times^1)\leq \frac{1}{k}\sum_{h=1}^{k-1}\frac{k-h}{k}\left(1-\pi_{1,2h}^{(s-1)}(\up)\right).
\end{align}

Summing up \eqref{eq:delta_a_q_0_simplified_balanced_phase_1.2} and \eqref{eq:delta_a_q_x_simplified_balanced_phase_1.2} and using \eqref{eq:pi_1_up_dominate_sum_phase1.2}, \eqref{eq:e^2ac1=(a_q_x+a_q_0)^2}, we have
\begin{align}\label{eq:delta_a_q_0+delta_a_q_x_phase1.2}
    \delta a_{q,0}^{(s-1)}+\delta a_{q,\times}^{(s-1)}=(1+o(1))\left(\pi_{1,1}^{(s-1)}(\up)+\sum_{h=1}^{k-1}\frac{k-h}{k}\left(1-\pi_{1,2h}^{(s-1)}(\up)\right)\right).
\end{align}
Comparing \eqref{eq:delta_a_q_0+delta_a_q_x_phase1.2} with \eqref{eq:delta_a_p_0_simplified_balanced_phase_1.2} and \eqref{eq:delta_a_p_up_simplified_balanced_phase_1.2}, we have
\begin{align}\label{eq:|delta_a_p_0_up|_bound_phase1.2}
    |\delta a_{p,0}^{(s-1)}|\lesssim \delta a_{q,0}^{(s-1)}+\delta a_{q,\times}^{(s-1)},\quad |\delta a_{p,\up}^{(s-1)}|\lesssim \delta a_{q,0}^{(s-1)}+\delta a_{q,\times}^{(s-1)}.
\end{align}
This implies that \eqref{eq:a_p_0_bound_phase_1.2} and \eqref{eq:a_p_up_bound_phase_1.2} hold at step $s$.
Similarly, we can also obtain (recall we define $a_{p,3}^{(s)}\coloneq\eta\sum_{t=0}^{s-1}\delta a_{p,3}^{(t)}$ in \eqref{eq:a_p_3})
\begin{align}\label{eq:delta_a_p_3_bound_phase1.2}
    \forall t\leq s:\quad |\delta a_{p,3}^{(t)}|\leq \frac{1+o(1)}{k}|\delta a_{q,0}^{(s-1)}+\delta a_{q,\times}^{(s-1)}|,\quad |a_{p,3}^{(t)}|\lesssim \frac{1}{k}\left(a_{q,0}^{(t)}+a_{q,\times}^{(t)}\right).
\end{align}
Thus by the above expresion, \eqref{eq:a_p3} and \eqref{eq:a_q_x_q_0_p_2_asymp_phase_1.2} we know that \eqref{eq:a_p_2_bound_phase_1.2} holds at step $s$.

Further, by comparing \eqref{eq:delta_a_q_0+delta_a_q_x_phase1.2}, \eqref{eq:delta_a_p_1_simplified_balanced_phase_1.2} and using \eqref{eq:sum_pi_2h_down_1=1/k_sum_1-pi_2h_up}, we know that 
\begin{align}
    \delta a_{p,1}^{(s-1)}\leq \frac{1+o(1)}{k}\sum_{h=1}^{k-1}\frac{k-h}{k}\pi_{1,2h}^{(s-1)}(\down_\times)+\frac{1+o(1)}{k^2}\left(\delta a_{q,0}^{(s-1)}+\delta a_{q,\times}^{(s-1)}\right),
\end{align}
and by \eqref{eq:delta_a_p_3_bound_phase1.2}, \eqref{eq:a_p3} we have
\begin{align}\label{eq:delta_a_p_2_ub_phase1.2}
    \left|\delta a_{p,2}^{(s-1)}-\frac{\delta a_{p,1}^{(s-1)}}{k-1}\right|=\left|\frac{\delta a_{p,3}^{(s-1)}}{k-1}\right|\leq \frac{1+o(1)}{k^2}\left(\delta a_{q,0}^{(s-1)}+\delta a_{q,\times}^{(s-1)}\right).
\end{align}

\paragraph{When $T_{1/2}+1\leq s\leq T_{1,3}$.} By the same argument as how we show \eqref{eq:a_q_0-a_p_2_phase_1_end}, \eqref{eq:a_q_0>=1.9lnk_T_12}, \eqref{eq:a_b_1_phase_1_end}, and \eqref{eq:a_b_0_phase_1_end} hold in the proof of Lemma~\ref{lm:phase_1.1}, we can show \eqref{eq:a_q_x-a_p_2_bound_<T_1_3}, \eqref{eq:a_q_0_bound_<T_1_3}, \eqref{eq:a_b_1_bound_<T_1_3}, \eqref{eq:a_b_0_bound_<T_1_3} and \eqref{eq:a_q_x_q_0_p_2_asymp_phase_1.2} hold at step $s$. And from \eqref{eq:delta_a_p_1_simplified_balanced_phase_1.2}, \eqref{eq:sum_pi_1_2h+1_down_times_bound_phase1.2} we deduce
\begin{align}
    \delta a_{p,1}^{(s-1)}\gtrsim \frac{\exp\left(-\frac{a_{p,1}^{(s-1)}}{2k}\right)}{k^2}\gtrsim \frac{1}{k^{12.5}}. 
\end{align}
This and Induction Hypothesis II imply that \eqref{eq:a_p_1_bound_<T_1_3} holds at step $s$. 

From \eqref{eq:delta_a_c_0_simplified_balanced_phase_1.2}, \eqref{eq:delta_a_q_0_simplified_balanced_phase_1.2} and \eqref{eq:delta_a_q_x_simplified_balanced_phase_1.2}, we have for any $t\in [T_{1,2}+1,s-1]$:
\begin{align}\label{eq:delta_e2a_c_0_simplified_balanced_ub_phase_1.2}
    \delta e^{2a_{c,0}^{(t)}}\lesssim \left(a_{q,\times}^{(t)}+a_{q,0}^{(t)}\right)\left(\delta a_{q,0}^{(t)}+\delta a_{q,\times}^{(t)}\right).
\end{align}
By \eqref{eq:delta_e2a_c_0_simplified_balanced_ub_phase_1.2}, our choice of $T_{1,3}$ and \eqref{eq:a_p_2_bound_phase_1.2}, \eqref{eq:a_q_x_q_0_p_2_asymp_phase_1.2} we know that if $s\leq T_{1,3}$, then 
\begin{align}\label{eq:e2ac_0<=ln^2k<=T_13}
    e^{2a_{c,0}^{(s)}}\lesssim \ln^2 k.
\end{align}
From this and \eqref{eq:a_c_0_increase_phase1.2}, \eqref{eq:e2a_c_0_lb_phase1.2} we know that
\eqref{eq:e2a_c_0_balanced_phase_1.2} holds at step $s$.

Let 
\begin{align}\label{eq:T_1_3'}
    T_{1,3}'\coloneqq  \sup\left\{t\in \NN: a_{p,1}^{(t)}\leq 20.1k\ln k\right\},
\end{align}
then $T_{1,3}'\in(T_{1/2},T_{1,3})$. 
We now prove \eqref{eq:a_q_0>a_q_x_bound_T_13} holds for all $t\in[T_{1,3}',T_{1,3}]$. We first show there exists $t_0\in[T_{1/2}+1,T_{1,3}']$ such that 
\begin{align}\label{eq:a_q_0>a_q_x_bound_T_13_tight}
    \left(1+\frac{1}{k}\right)a_{q,0}^{(t)}-\frac{a_{p,0}^{(t)}}{k}+0.05\sqrt{a_{p,1}^{(t)}/k}+3\ln k\geq  \min\left\{a_{q,\times}^{(t)}-\mu_2^{(t)},a_{q,\times}^{(t)}-\mu_{2(k-1)}^{(t)}\right\}
\end{align}
holds at step $t_0$. Suppose otherwise, i.e., for all $t\in[T_{1/2},T_{1,3}']$, we have
\begin{align}\label{eq:a_q_0<a_q_x_bound_T_13_contradiction}
    \left(1+\frac{1}{k}\right)a_{q,0}^{(t)}-\frac{a_{p,0}^{(t)}}{k}+0.05\sqrt{a_{p,1}^{(t)}/k}+3\ln k<  \min\left\{a_{q,\times}^{(t)}-\mu_2^{(t)},a_{q,\times}^{(t)}-\mu_{2(k-1)}^{(t)}\right\}
\end{align}
Then for all $t\in[T_{1/2},T_{1,3}']$: 
\begin{align}\label{eq:pi_1_1_up_T_13_contradiction}
    \pi_{1,1}^{(t)}(\up)&\overset{\eqref{eq:pi_1_1_up_phase1.2}}=(1+o(1))\frac{1}{k}\exp\left(-\left(1+\frac{1}{k}\right)a_{q,0}^{(t)}+\frac{a_{p,0}^{(t)}}{k}\right)\notag\\
    &\geq (1+o(1))k^2\exp\left(-\min\left\{a_{q,\times}^{(t)}-\mu_2^{(t)},a_{q,\times}^{(t)}-\mu_{2(k-1)}^{(t)}\right\}+0.05\sqrt{a_{p,1}^{(t)}/k}\right)\notag\\
    &\overset{\eqref{eq:sum_1-pi_2h_up_ub}}\geq 2(1+o(1))\exp\left(0.05\sqrt{a_{p,1}^{(t)}/k}\right)\sum_{h=1}^{k-1}\frac{k-h}{k}\left(1-\pi_{1,2h}^{(t)}(\up)\right).
\end{align}
Thus by \eqref{eq:pi_1_1_up_T_13_contradiction}, \eqref{eq:delta_a_q_0_simplified_balanced_phase_1.2} and \eqref{eq:delta_a_q_x_simplified_balanced_phase_1.2}, we have
\begin{align}\label{eq:delta_a_q_0_T_13_contradiction}
    \forall t\in[T_{1/2},T_{1,3}']:\quad \delta a_{q,0}^{(t)}\geq 2(1+o(1))\exp\left(0.05\sqrt{a_{p,1}^{(t)}/k}\right)\delta a_{q,\times}^{(t)}.
\end{align}
Note that from \eqref{eq:a_q_x-a_q_0_bound>T_12} and \eqref{eq:a_q_0+a_q_x=C_lnk_phase_1} we know that 
\begin{align}\label{eq:a_q_x_T_12_ub}
    a_{q,\times}^{(T_{1/2})}\leq 13.3\ln k,
\end{align}
and from \eqref{eq:a_q_x-a_p_2_bound_<T_1_3} and our choice of $T_{1,3},T_{1,3}'$ we know that 
\begin{align}\label{eq:a_q_x_T_13_lb}
    a_{q,\times}^{(T_{1,3})}\geq 14.4\ln k,\quad a_{q,\times}^{(T_{1,3}')}\geq 13.95\ln k.
\end{align}
By \eqref{eq:a_q_x_T_12_ub}, \eqref{eq:a_q_x_T_13_lb} and \eqref{eq:delta_a_q_0_T_13_contradiction}, we have
\begin{align}
    a_{q,0}^{(T_{1,3}')}>a_{q,0}^{(T_{1,3}')}-a_{q,0}^{(T_{1/2})}&\gtrsim \exp\left(0.05\sqrt{a_{p,1}^{(T_{1,3}')}/k}\right)\left(a_{q,\times}^{(T_{1,3}')}-a_{q,\times}^{(T_{1/2})}\right)\notag\\
    &\gtrsim \exp\left(0.05\sqrt{a_{p,1}^{(T_{1,3}')}/k}\right)\ln k\gg \ln k.
\end{align}
But by \eqref{eq:a_q_x_q_0_p_2_asymp_phase_1.2}, our choice of $T_{1,3}$ (c.f.~\eqref{eq:T_1_3}) and \eqref{eq:a_p_2_bound_phase_1.2}, we have
\begin{align}
    a_{q,0}^{(T_{1,3}')}\asymp a_{p,2}^{(T_{1,3}')}\asymp \ln k.
\end{align}
The above two relations contradict each other, and hence there must exist $t_0\in[T_{1/2}+1,T_{1,3}']$ such that \eqref{eq:a_q_0>a_q_x_bound_T_13_tight} holds at step $t_0$. We can choose $t_0\in[T_{1/2}+1,T_{1,3}']$ to be the biggest step such that \eqref{eq:a_q_0>a_q_x_bound_T_13_tight} holds, and show that \eqref{eq:a_q_0>a_q_x_bound_T_13} holds for all $t\in[t_0,T_{1,3}]$. Suppose otherwise, i.e., there exists $t_1\in[t_0+1,T_{1,3}']$ such that \eqref{eq:a_q_0>a_q_x_bound_T_13} does not hold at step $t_1$, then by Induction Hypothesis II we have
\begin{align}\label{eq:a_q_0<a_q_x_bound_contradiction_2_T_13}
    \left(1+\frac{1}{k}\right)a_{q,0}^{(t_1)}+0.1\sqrt{a_{p,1}^{(t_1)}/k}+3\ln k+o(1) 
    &=\left(1+\frac{1}{k}\right)a_{q,0}^{(t_1)}-\frac{a_{p,0}^{(t_1)}}{k}+0.1\sqrt{a_{p,1}^{(t_1)}/k}+3\ln k\notag\\
    &<  \min\left\{a_{q,\times}^{(t_1)}-\mu_2^{(t_1)},a_{q,\times}^{(t_1)}-\mu_{2(k-1)}^{(t_1)}\right\} \notag \\
    & =a_{q,\times}^{(t_1)}-\frac{a_{p,1}^{(t_1)}}{2k}+o(1).
\end{align}
By our choice of $t_0$, gradient update rule and Induction Hypothesis II, we have for all $t\in[t_0,t_1-1]$:
\begin{align}\label{eq:a_q_0<a_q_x_bound_contradiction_2_2_T_13}
    \left(1+\frac{1}{k}\right)a_{q,0}^{(t)}+0.05\sqrt{a_{p,1}^{(t)}/k}+3\ln k+o(1)
    &\leq a_{q,\times}^{(t)}-\frac{a_{p,1}^{(t)}}{2k}.
\end{align}
Then similar as how we obtain \eqref{eq:delta_a_q_0_T_13_contradiction}, from the above relation and \eqref{eq:delta_a_q_0_simplified_balanced_phase_1.2}, we can compute that
\begin{align}\label{eq:delta_a_q_0_t_1-1_contradiction_T_13}
    \forall t\in[t_0,t_1-1]:\quad \delta a_{q,0}^{(t)}\gtrsim \exp\left(0.05\sqrt{a_{p,1}^{(t)}/k}\right)\delta a_{q,\times}^{(t)}.
\end{align}
and thus we have
\begin{align}
    \left(1+\frac{1}{k}\right)a_{q,0}^{(t_1)}+0.05\sqrt{a_{p,1}^{(t_1)}/k}+3\ln k-\left(a_{q,\times}^{(t_1)}-\frac{a_{p,1}^{(t_1)}}{2k}\right) 
    &\overset{\eqref{eq:a_q_0>a_q_x_bound_T_13_tight}}\geq o(1)+\eta\sum_{t=t_0}^{t_1-1}\left(\delta a_{q,0}^{(t)}-\delta a_{q,\times}^{(t)}\right)\notag\\
    &\overset{\eqref{eq:delta_a_q_0_t_1-1_contradiction_T_13}}\gtrsim \eta\exp\left(0.05\sqrt{a_{p,1}^{(t_0)}/k}\right) 
    \sum_{t=t_0}^{t_1-1}\delta a_{q,\times}^{(t)}+o(1)>o(1),
\end{align}
which implies
\begin{align}
    \left(1+\frac{1}{k}\right)a_{q,0}^{(t_1)}+0.1\sqrt{a_{p,1}^{(t_1)}/k}+3\ln k\geq a_{q,\times}^{(t_1)}-\frac{a_{p,1}^{(t_1)}}{2k}+0.05\sqrt{a_{p,1}^{(t_1)}/k}+o(1).
\end{align}
This contradicts \eqref{eq:a_q_0<a_q_x_bound_contradiction_2_T_13}. Therefore, we have shown that \eqref{eq:a_q_0>a_q_x_bound_T_13} holds for all $t\in[T_{1,3}',T_{1,3}]$.

Then for all $t\in[T_{1,3}',s-1]$: 
\begin{align}\label{eq:sum_1-pi_2h_up>>sum_1-pi_2h+1_up}
    \sum_{h=1}^{k-1}\frac{k-h}{k}\left(1-\pi_{1,2h}^{(t)}(\up)\right)&\overset{\eqref{eq:sum_1-pi_2h_up_lb}}\geq \left(1+o(1)\right)\exp\left(-\min\left\{a_{q,\times}^{(t)}-\mu_{2}^{(t)}-\ln k,a_{q,\times}^{(t)}-\mu_{2(k-1)}^{(t)}+\ln k\right\}\right)\notag\\
    &\geq \frac{1+o(1)}{k}\exp\left(-\min\left\{a_{q,\times}^{(t)}-\mu_{2}^{(t)},a_{q,\times}^{(t)}-\mu_{2(k-1)}^{(t)}\right\}\right)\notag\\
    &\overset{\eqref{eq:a_q_0>a_q_x_bound_T_13}}\geq \frac{1+o(1)}{k^4}\exp\left(-\left(1+\frac{1}{k}\right)a_{q,0}^{(t)}+\frac{a_{p,0}^{(t)}}{k}-0.1\sqrt{a_{p,1}^{(t)}/k}\right)\notag\\
    &\overset{\eqref{eq:pi_1_1_up_phase1.2}}=\frac{1+o(1)}{k^3}\exp\left(-0.1\sqrt{a_{p,1}^{(t)}/k}\right)\pi_{1,1}^{(t)}(\up)\notag\\
    &\overset{\eqref{eq:tiny_pi_1_2h+1_up_phase1.2}}\geq \frac{1+o(1)}{k^3}\exp\left(0.3a_{p,2}^{(t)}-0.1\sqrt{a_{p,1}^{(t)}/k}\right)\sum_{h=1}^{k-1}\frac{k-h}{k}\pi_{1,2h+1}^{(t)}(\up)\notag\\
    &\geq k^3\sum_{h=1}^{k-1}\frac{k-h}{k}\pi_{1,2h+1}^{(t)}(\up),
\end{align}
where the last inequality follows from \eqref{eq:T_1_3} and \eqref{eq:a_p_2_bound_phase_1.2}, and the third line uses \eqref{eq:a_q_0>a_q_x_bound_T_13}, which is proven above to be hold also for all $t\in[T_{1,3}',T_{1,3}]$. Therefore, by \eqref{eq:delta_a_q_x_simplified_balanced_phase_1.2} and \eqref{eq:e^2ac1=(a_q_x+a_q_0)^2}, we have for all $t\in[T_{1,3}',s-1]$:
\begin{align}\label{eq:delta_a_q_x_simplified_T_13-T_14}
    \delta a_{q,\times}^{(t)}=(1+o(1))\sum_{h=1}^{k-1}\frac{k-h}{k}\left(1-\pi_{1,2h}^{(t)}(\up)\right)>0,
\end{align}
and by the penultimate line of \eqref{eq:sum_1-pi_2h_up>>sum_1-pi_2h+1_up}, \eqref{eq:delta_a_q_x_simplified_T_13-T_14} and our choice of $T_{1,3}'$ we know that 
\begin{align}\label{eq:sum_1-pi_2h+1_up_vs_sum_1-pi_2h_up}
    \forall t\in[T_{1,3}',s-1]:\quad \sum_{h=1}^{k-1}\frac{k-h}{k}\pi_{1,2h+1}^{(t)}(\up)\leq \exp\left(-0.1a_{p,2}^{(t)}\right)\delta a_{q,\times}^{(t)}.
\end{align}

We next show 
\eqref{eq:a_q_2>a_q_x_bound_T_13} holds at step $T_{1,3}$. To do so,
we first prove by contradiction that there exists $t_0\in[T_{1/2},T_{1,3}]$ such that 
\begin{align}\label{eq:a_q_2>a_q_x_bound_T_13_tight}
    r_{\down_\times}^{(t)}\left(a_{p,1}^{(t)}+a_{p,2}^{(t)}\right)+0.05\sqrt{a_{p,1}^{(t)}/k}+5\ln k &\geq \min\left\{a_{q,\times}^{(t)}-\mu_2^{(t)},a_{q,\times}^{(t)}-\mu_{2(k-1)}^{(t)}\right\}.
\end{align}
Suppose otherwise, i.e., for all $t\in[T_{1/2},T_{1,3}]$, we have
\begin{align}\label{eq:a_q_2>a_q_x_bound_T_13_contradiction}
    r_{\down_\times}^{(t)}\left(a_{p,1}^{(t)}+a_{p,2}^{(t)}\right)+0.05\sqrt{a_{p,1}^{(t)}/k}+5\ln k &<\min\left\{a_{q,\times}^{(t)}-\mu_2^{(t)},a_{q,\times}^{(t)}-\mu_{2(k-1)}^{(t)}\right\},
\end{align}
then 
\begin{align}
    \sum_{h=1}^{k-1}\frac{k-h}{k}\left(1-\pi_{1,2h}^{(t)}(\up)\right)&\overset{\eqref{eq:sum_1-pi_2h_up_ub}}\leq \frac{1+o(1)}{2}k^3\exp\left(-\min\left\{a_{q,\times}^{(s-1)}-\mu_{2}^{(s-1)},a_{q,\times}^{(s-1)}-\mu_{2(k-1)}^{(s-1)}\right\}\right)\notag\\
    &\overset{\eqref{eq:a_q_2>a_q_x_bound_T_13_contradiction}}\leq \frac{1+o(1)}{2k^2}\exp\left(-r_{\down_\times}^{(t)}\left(a_{p,1}^{(t)}+a_{p,2}^{(t)}\right)-0.05\sqrt{a_{p,1}^{(t)}/k}\right)\notag\\
    &\overset{\eqref{eq:sum_pi_1_2h+1_down_times_bound_phase1.2}}\leq \frac{1+o(1)}{2k^2}\exp\left(-0.05\sqrt{a_{p,1}^{(t)}/k}\right)\sum_{h=1}^{k-1}\frac{k-h}{k}\pi_{1,2h+1}^{(t)}(\down_\times),
\end{align}
which together with \eqref{eq:delta_a_p_1_simplified_balanced_phase_1.2}, \eqref{eq:delta_a_q_x_simplified_balanced_phase_1.2} gives
\begin{align}\label{eq:delta_a_p_1_contradiction_T_13}
    \forall t\in[T_{1/2},T_{1,3}]:\quad \delta a_{p,1}^{(t)}\geq 2(1+o(1))k\exp\left(0.05\sqrt{a_{p,1}^{(t)}/k}\right)\delta a_{q,\times}^{(t)}.
\end{align}
Combining \eqref{eq:delta_a_p_1_contradiction_T_13}, \eqref{eq:a_q_x_T_12_ub} and \eqref{eq:a_q_x_T_12_ub}, we can see that 
\begin{align}
    a_{p,1}^{(T_{1,3})}>a_{p,1}^{(T_{1,3})}-a_{p,1}^{(T_{1/2})}\geq 2(1+o(1))k\exp\left(0.05\sqrt{a_{p,1}^{(T_{1,3})}/k}\right)\left(a_{q,\times}^{(T_{1,3})}-a_{q,\times}^{(T_{1/2})}\right)\gg k\ln k.
\end{align}
But by \eqref{eq:a_q_x_q_0_p_2_asymp_phase_1.2} and \eqref{eq:a_p_2_bound_phase_1.2} we have
\begin{align}
    a_{p,1}^{(T_{1,3})}\lesssim ka_{q,\times}^{(T_{1,3})}\lesssim k\ln k.
\end{align}
The above two relations contradict each other. Thus there exists $t_0\in[T_{1/2},T_{1,3}]$ such that \eqref{eq:a_q_2>a_q_x_bound_T_13_tight} holds at step $t_0$. If $t_0=T_{1,3}$, then we are done. Otherwise, we can choose $t_0\in[T_{1/2},T_{1,3}]$ to be the biggest step such that \eqref{eq:a_q_2>a_q_x_bound_T_13_tight} holds.
We next show that \eqref{eq:a_q_2>a_q_x_bound_T_13} holds for all $t\in[t_0,T_{1,3}]$. Suppose otherwise, then let  $t_1\in[t_0+1,T_{1,3}]$ by the smallest step such that \eqref{eq:a_q_2>a_q_x_bound_T_13} does not hold at step $t_1$, then  by Induction Hypothesis II we have
\begin{align}\label{eq:a_q_2<a_q_x_t_1_contradiction_T_13}
    \frac{a_{p,1}^{(t_1)}}{2k}+0.1\sqrt{a_{p,1}^{(t_1)}/k}+5\ln k+o(1)&=r_{\down_\times}^{(t_1)}\left(a_{p,1}^{(t_1)}+a_{p,2}^{(t_1)}\right)+0.1\sqrt{a_{p,1}^{(t_1)}/k}+5\ln k\notag\\
    &<\min\left\{a_{q,\times}^{(t_1)}-\mu_2^{(t_1)},a_{q,\times}^{(t_1)}-\mu_{2(k-1)}^{(t_1)}\right\}=a_{q,\times}^{(t_1)}-\frac{a_{p,1}^{(t_1)}}{2k}+o(1).
\end{align}
By our choice of $t_0$, gradient update rule and Induction Hypothesis II, we have for all $t\in[t_0,t_1-1]$:
\begin{align}\label{eq:a_q_2<a_q_x_t_1-1_contradiction_T_13}
\frac{a_{p,1}^{(t)}}{2k}+0.05\sqrt{a_{p,1}^{(t)}/k}+5\ln k+o(1)\leq a_{q,\times}^{(t)}-\frac{a_{p,1}^{(t)}}{2k}.
\end{align}
and thus similar as \eqref{eq:delta_a_p_1_contradiction_T_13}, we can compute that
\begin{align}\label{eq:delta_a_p_1_t_1-1_contradiction_T_13}
    \forall t\in[t_0,t_1-1]:\quad \delta a_{p,1}^{(t)}\gtrsim k\exp\left(0.05\sqrt{a_{p,1}^{(t)}/k}\right)\delta a_{q,\times}^{(t)}.
\end{align}
Then
\begin{align}
\frac{a_{p,1}^{(t_1)}}{2k}+0.05\sqrt{a_{p,1}^{(t_1)}/k}+5\ln k
    -\left(a_{q,\times}^{(t_1)}-\frac{a_{p,1}^{(t_1)}}{2k}\right) 
    &\overset{\eqref{eq:a_q_2>a_q_x_bound_T_13_tight}}\geq
    o(1)+\eta\sum_{t=t_0}^{t_1-1}
    \left(\frac{1}{k}\delta a_{p,1}^{(t)}-\delta a_{q,\times}^{(t)}\right)\notag\\
    &\overset{\eqref{eq:delta_a_p_1_t_1-1_contradiction_T_13}}\gtrsim
    \eta\sum_{t=t_0}^{t_1-1}
    \left(
    \exp\left(0.05\sqrt{a_{p,1}^{(t)}/k}\right)-1
    \right)\delta a_{q,\times}^{(t)}+o(1)\notag\\
    &\gtrsim
    \eta\left(
    \exp\left(0.05\sqrt{a_{p,1}^{(t_0)}/k}\right)-1
    \right)
    \sum_{t=t_0}^{t_1-1}\delta a_{q,\times}^{(t)}
    +o(1)
    \geq o(1).
\end{align}
This together with our choice of $T_{1/2}$ suggests
\begin{align}
    \frac{a_{p,1}^{(t_1)}}{2k}+0.1\sqrt{a_{p,1}^{(t_1)}/k}+5\ln k\geq a_{q,\times}^{(t_1)}-\frac{a_{p,1}^{(t_1)}}{2k}+0.05\sqrt{a_{p,1}^{(t_1)}/k}+o(1),
\end{align}
which contradicts \eqref{eq:a_q_2<a_q_x_t_1_contradiction_T_13}. Therefore, we have shown that \eqref{eq:a_q_2>a_q_x_bound_T_13} holds for all $t\in[t_0,T_{1,3}]$.

We next show \eqref{eq:a_p2>a_q_0_bound_T_13} holds at step $T_{1,3}$. First, if $a_{q,0}^{(T_{1,3})}\leq 12\ln k$, then by our choice of $T_{1,3}$, \eqref{eq:a_p_2_bound_phase_1.2}, \eqref{eq:a_p_0_bound_phase_1.2} and the fact that $r_{\down_\times}^{(t)}\geq \frac{1}{2k}$ for all $t\in[T_{1/2},T_{1,3}]$ guaranteed by Induction Hypothesis II, we have \eqref{eq:a_p2>a_q_0_bound_T_13} holds at step $T_{1,3}$. Below we assume 
\begin{align}\label{eq:a_q0>12.5lnk_T_13}
    a_{q,0}^{(T_{1,3})}> 12\ln k.
\end{align}
We first show that there exists $t_0\in[T_{1/2},T_{1,3}]$ such that
\begin{align}\label{eq:a_p2>a_q_0_tight_T_13}
    r_{\down_\times}^{(t_0)}\left(a_{p,1}^{(t_0)}+a_{p,2}^{(t_0)}\right)+2.05\ln k \geq \left(1+\frac{1}{k}\right)a_{q,0}^{(t_0)}-\frac{a_{p,0}^{(t_0)}}{k}.
\end{align}
Suppose for all $t\in[T_{1/2},T_{1,3}]$, we have
\begin{align}\label{eq:a_p2>a_q_0_bound_T_13_contradiction_1}
    r_{\down_\times}^{(t)}\left(a_{p,1}^{(t)}+a_{p,2}^{(t)}\right)+2.05\ln k < \left(1+\frac{1}{k}\right)a_{q,0}^{(t)}-\frac{a_{p,0}^{(t)}}{k},
\end{align}
then for all $t\in[T_{1/2},T_{1,3}]$, we have
\begin{align}
    \sum_{h=1}^{k-1}\frac{k-h}{k}\pi_{1,2h+1}^{(t)}(\down_\times)&\overset{\eqref{eq:sum_pi_1_2h+1_down_times_bound_phase1.2}}\geq \frac{1+o(1)}{k}\exp\left(-r_{\down_\times}^{(t)}\left(a_{p,1}^{(t)}+a_{p,2}^{(t)}\right)\right)\notag\\
    &\geq \left(1+o(1)\right)k^{2.05}\frac{\exp\left(-\left(1+\frac{1}{k}\right)a_{q,0}^{(t)}+\frac{a_{p,0}^{(t)}}{k}\right)}{k} \overset{\eqref{eq:pi_1_1_up_phase1.2}}=k^{2.05}\pi_{1,1}^{(t)}(\up).
\end{align}
This combined with \eqref{eq:delta_a_p_1_simplified_balanced_phase_1.2} and \eqref{eq:delta_a_q_0_simplified_balanced_phase_1.2} suggests
\begin{align}\label{eq:delta_a_p_1_bound_T_13_contradiction_1}
    \delta a_{p,1}^{(t)}\geq  (1+o(1))k^{1.05}\pi_{1,1}^{(t)}(\up) \geq (1+o(1))k^{1.05}\delta a_{q,0}^{(t)}.
\end{align}
Note that by \eqref{eq:a_q_x-a_q_0_bound>T_12} and \eqref{eq:a_q_0+a_q_x=C_lnk_phase_1} we know that 
\begin{align}\label{eq:a_q_0_T_12_ub}
    a_{q,0}^{(T_{1/2})}\leq 10.2\ln k +C/2,
\end{align}
Therefore, by \eqref{eq:a_q_0_T_12_ub}, \eqref{eq:a_q0>12.5lnk_T_13} and \eqref{eq:delta_a_p_1_bound_T_13_contradiction_1}, we have
\begin{align}
    a_{p,1}^{(T_{1,3})}>a_{p,1}^{(T_{1,3})}-a_{p,1}^{(T_{1/2})}\gtrsim k^{1.05}\left(a_{q,0}^{(T_{1,3})}-a_{q,0}^{(T_{1/2})}\right)\gtrsim k^{1.05}\ln k,
\end{align}
which contradicts our choice of $T_{1,3}$ (c.f.~\eqref{eq:T_1_3}). Therefore, there exists $t_0\in[T_{1/2},T_{1,3}]$ such that \eqref{eq:a_p2>a_q_0_tight_T_13} holds. Then similar as how we show \eqref{eq:a_q_2>a_q_x_bound_T_13} or \eqref{eq:a_q_0>a_q_x_bound_T_13} holds at step $T_{1,3}$ above, building upon \eqref{eq:a_p2>a_q_0_tight_T_13}, we can show by contradiction that \eqref{eq:a_p2>a_q_0_bound_T_13} holds at step $T_{1,3}$.

We now prove that \eqref{eq:a_q_x>a_p_2_bound_T_13} holds at step $T_{1,3}$.
We first show that there exists $t_0\in[T_{1,3}',T_{1,3}]$ such that 
\begin{align}\label{eq:a_q_x>a_p_2_bound_T_13_tight}
        \min\left\{a_{q,\times}^{(t_0)}-\mu_2^{(t_0)}-\ln k,a_{q,\times}^{(t_0)}-\mu_{2(k-1)}^{(t_0)}+\ln k\right\}+\frac{2}{k}a_{q,0}^{(t_0)}\geq r_{\down_\times}^{(t_0)}\left(a_{p,1}^{(t_0)}+a_{p,2}^{(t_0)}\right)+1.95\ln k.
\end{align}
Suppose for all $t\in[T_{1,3}',T_{1,3}]$, we have
\begin{align}\label{eq:a_q_x>a_p_2_bound_T_13_contradiction_1}
    \min\left\{a_{q,\times}^{(t)}-\mu_2^{(t)}-\ln k,a_{q,\times}^{(t)}-\mu_{2(k-1)}^{(t)}+\ln k\right\}+\frac{2}{k}a_{q,0}^{(t)}< r_{\down_\times}^{(t)}\left(a_{p,1}^{(t)}+a_{p,2}^{(t)}\right)+1.95\ln k.
\end{align}
Then for all $t\in[T_{1,3}',T_{1,3}]$, we have
\begin{align}
    \sum_{h=1}^{k-1}\frac{k-h}{k}\left(1-\pi_{1,2h}^{(t)}(\up)\right) 
    &\overset{\eqref{eq:sum_1-pi_2h_up_lb}}\geq  (1+o(1))\exp\left(-\min\left\{a_{q,\times}^{(t)}-\mu_{2}^{(t)}-\ln k,a_{q,\times}^{(t)}-\mu_{2(k-1)}^{(t)}+\ln k\right\}\right)\notag\\
    &\geq (1+o(1))\exp\left(-r_{\down_\times}^{(t)}\left(a_{p,1}^{(t)}+a_{p,2}^{(t)}\right)-1.95\ln k+\frac{2}{k}a_{q,0}^{(t)}\right)\notag\\
    &\overset{\eqref{eq:sum_pi_1_2h+1_down_times_bound_phase1.2}}\geq \frac{2+o(1)}{k^{1.95}}\sum_{h=1}^{k-1}\frac{k-h}{k}\pi_{1,2h+1}^{(t)}(\down_\times),
\end{align}
and by \eqref{eq:pi_2h_down_1_bound_phase1.2} our choice of $T_{1,3}'$ we have
\begin{align}
    \sum_{h=1}^{k-1}\frac{k-h}{k}\pi_{1,2h}^{(t)}(\down_\times^1)=o\left(\frac{1}{k^{1.95}}\right)\sum_{h=1}^{k-1}\frac{k-h}{k}\left(1-\pi_{1,2h}^{(t)}(\up)\right).
\end{align}
the above two expressions and \eqref{eq:delta_a_p_1_simplified_balanced_phase_1.2}, \eqref{eq:delta_a_q_x_simplified_T_13-T_14} we have for all $t\in[T_{1,3}',T_{1,3}]$:
\begin{align}\label{eq:delta_a_p_1_bound_T_13_contradiction_2}
    \delta a_{p,1}^{(t)}\leq (2+o(1))k^{0.95}\delta a_{q,\times}^{(t)},
\end{align}
from which and our choice of $T_{1,3},T_{1,3}'$ we deduce
\begin{align}
    a_{q,\times}^{(T_{1,3})}>a_{q,\times}^{(T_{1,3})}-a_{q,\times}^{(T_{1,3}')}\gtrsim k^{-0.95}\left(a_{p,1}^{(T_{1,3})}-a_{p,1}^{(T_{1,3}')}\right)\gtrsim k^{0.05}\ln k,
\end{align}
which contradicts that fact that $a_{q,\times}^{(T_{1,3})}\asymp \ln k$ guaranteed by \eqref{eq:a_q_x_q_0_p_2_asymp_phase_1.2}, \eqref{eq:a_p_2_bound_phase_1.2} and our choice of $T_{1,3}$. Therefore, there exists $t_0\in[T_{1,3}',T_{1,3}]$ such that \eqref{eq:a_q_x>a_p_2_bound_T_13_tight} holds. Then similar as how we show \eqref{eq:a_q_2>a_q_x_bound_T_13} or \eqref{eq:a_q_0>a_q_x_bound_T_13} holds at step $T_{1,3}$ above, building upon \eqref{eq:a_q_x>a_p_2_bound_T_13_tight}, we can show by contradiction that \eqref{eq:a_q_x>a_p_2_bound_T_13} holds at step $T_{1,3}$.

\paragraph{When $T_{1,3}+1\leq s\leq T_{1,4}$.} We first show relations between $a_{b,0}^{(s)}$, $a_{b,1}^{(s)}$ and $a_{b,2}^{(s)}$. First note that \eqref{eq:sum_1-pi_2h_up>>sum_1-pi_2h+1_up}, \eqref{eq:delta_a_q_x_simplified_T_13-T_14} and \eqref{eq:sum_1-pi_2h+1_up_vs_sum_1-pi_2h_up} still hold here, and we'll frequently use them in the subsequent proof.

First, from \eqref{eq:sum_1-pi_2h+1_up_vs_sum_1-pi_2h_up} and  \eqref{eq:delta_a_c_0_simplified_balanced_phase_1.2} and \eqref{eq:a_q_x_q_0_p_2_asymp_phase_1.2} we know that for all $t\in [T_{1,3},s-1]$:
\begin{align}\label{eq:delta_e2a_c_0_simplified_balanced_ub_T_13-T_14}
    \delta e^{2a_{c,0}^{(t)}}\lesssim \frac{a_{p,2}^{(t)}}{\exp\left(0.1a_{p,2}^{(t)}\right)}\delta a_{q,\times}^{(t)}.
\end{align}
Then (i) if $a_{p,1}^{(s)}\leq k^2$, then by \eqref{eq:delta_e2a_c_0_simplified_balanced_ub_T_13-T_14}, \eqref{eq:a_p_0_bound_phase_1.2}, \eqref{eq:a_q_x_q_0_p_2_asymp_phase_1.2} and our choice of $T_{1,3}$ we have 
\begin{align}
    e^{2a_{c,0}^{(s)}}-e^{2a_{c,0}^{(T_{1,3})}}\lesssim \frac{21\ln k}{\exp\left(0.1\times 21\ln k\right)}k\lesssim \frac{\ln k}{k^{1.1}}.
\end{align}
And thus by \eqref{eq:e2ac_0<=ln^2k<=T_13} and the above relation we know that if $a_{p,1}^{(s)}\leq k^2$:
\begin{align}\label{eq:e2ac_0<=ln^2k_T_13-T_14_first_half}
    e^{2a_{c,0}^{(s)}}\lesssim \ln^2 k.
\end{align}
(ii) if $a_{p,1}^{(s)}>k^2$, then we can choose $t_0$ such that 
\begin{align}
    t_0\coloneqq \sup\left\{t\in [T_{1,3},s-1]: a_{p,1}^{(t)}= k^2\right\}.
\end{align}
Then above we have shown 
\begin{align}
    e^{2a_{c,0}^{(t_0)}}-e^{2a_{c,0}^{(T_{1,3})}}\lesssim \frac{\ln k}{k^{1.1}}.
\end{align}
From
\eqref{eq:delta_e2a_c_0_simplified_balanced_ub_T_13-T_14}, \eqref{eq:a_p_1_ub_T_1_4} and \eqref{eq:a_p_2_bound_phase_1.2}, \eqref{eq:a_q_x_q_0_p_2_asymp_phase_1.2} we deduce 
\begin{align}
    e^{2a_{c,0}^{(s)}}-e^{2a_{c,0}^{(t_0)}}\lesssim \frac{k}{\exp\left(0.1k\right)}k^{5/4}\sqrt{N}=o(1).
\end{align}
By the two relations above and \eqref{eq:e2ac_0<=ln^2k<=T_13} we know that 
\begin{align}\label{eq:e2ac_0<=ln^2k_T_13-T_14}
    e^{2a_{c,0}^{(s)}}\lesssim \ln^2 k.
\end{align}
Therefore, from \eqref{eq:e2ac_0<=ln^2k_T_13-T_14} and \eqref{eq:a_c_0_increase_phase1.2}, \eqref{eq:e2a_c_0_lb_phase1.2} we know that
\eqref{eq:e2a_c_0_balanced_phase_1.2} holds at step $s$.

By \eqref{eq:I_h_1_balanced_phase1.2} and Induction Hypothesis II we have for any $t\in[T_{1/2},s-1]$:
\begin{align}\label{eq:I_h_1_balanced_phase1.2_simplified}
    I_{h,1}^{(t)}
   &= \left(1+O\left(\frac{1}{k}\right)\right)\left(1-hr_{\down,h,1}^{(t)}\right)\notag\\
   &\quad\cdot\left(ka_{p,2}^{(t)}\frac{he^{a_{b,1}^{(t)}}}{e^{a_{b,0}^{(t)}}+he^{a_{b,1}^{(t)}}+he^{a_{b,2}^{(t)}}}\pi_{1,2h+1}^{(t)}(\down_\times)+hr_{\down,h,1}^{(t)}a_{p,2}^{(t)}\pi_{1,2h+1}^{(t)}(\up)\right).
\end{align}

From \eqref{eq:ea_b_1_bound_T_15} and \eqref{eq:delta_ea_b_1_bound_phase_1.2} in Induction Hypothesis II we know that
\begin{align}
    \forall t\in[T_{1/2},s-1]:\quad 0<a_{b,1}^{(t)}\lesssim \ln k.
\end{align}
From our choice of $T_{1,4}$ in \eqref{eq:T_1_4}, \eqref{eq:e2a_b_0_lb_T_14} and \eqref{eq:a_b_0_bound_<T_1_3} we know that
\begin{align}
    \forall t\in[T_{1/2},s-1]:\quad |a_{b,0}^{(t)}|\lesssim \ln k.
\end{align}
Therefore, by the above two expressions and \eqref{eq:A_M_sum0} we know that 
\begin{align}\label{eq:a_b_2_approx_0_t<=T_1_4}
    \forall t\in[T_{1/2},s-1]:\quad |a_{b,2}^{(t)}|\lesssim \frac{\ln k}{N}.
\end{align}

By our choice of $T_{1,4}$ in \eqref{eq:T_1_4}, \eqref{eq:ea_b_2_bound_T_14} in Induction Hypothesis II and \eqref{eq:a_b_2_approx_0_t<=T_1_4} we have
\begin{align}\label{eq:1-hr_down_h_1_bound_phase1.2}
    1-hr_{\down,h,1}^{(t)}\overset{\eqref{eq:r_down_h_phase1.2}}=\frac{e^{a_{b,0}^{(t)}}+he^{a_{b,2}^{(t)}}}{e^{a_{b,0}^{(t)}}+he^{a_{b,1}^{(t)}}+he^{a_{b,2}^{(t)}}}\lesssim  \frac{N}{\left(a_{p,1}^{(t)}\right)^2\ln k}.
\end{align}
Combining \eqref{eq:I_h_1_balanced_phase1.2_simplified}, \eqref{eq:1-hr_down_h_1_bound_phase1.2}, \eqref{eq:delta_a_p_1_simplified_balanced_phase_1.2} and \eqref{eq:sum_1-pi_2h+1_up_vs_sum_1-pi_2h_up} we obtain that for all $t\in[T_{1,3},s-1]$:
\begin{align}\label{eq:delta_a_b_1_bound_phase1.2}
    0< \delta a_{b,1}^{(t)}&\lesssim\frac{N}{\left(a_{p,1}^{(t)}\right)^2\ln k}\left(\frac{k^2 a_{p,2}^{(t)}}{N}\delta a_{p,1}^{(t)}+\frac{a_{p,2}^{(t)}}{N\exp\left(0.1a_{p,2}^{(t)}\right)}\delta a_{q,\times}^{(t)}\right)\notag\\
    &\overset{\eqref{eq:a_p_2_bound_phase_1.2}}\lesssim \frac{1}{a_{p,2}^{(t)}\ln k}\delta a_{p,1}^{(t)}+\frac{1}{k^2\ln ka_{p,2}^{(t)}\exp\left(0.1a_{p,2}^{(t)}\right)}\delta a_{q,\times}^{(t)}.
\end{align}

Similar as \eqref{eq:delta_e2a_c_i}, we define
\begin{align}\label{eq:delta_ea_b_i}
    \forall i\in \{0,1,2\}: \quad \delta e^{a_{b,i}^{(t)}}\coloneqq \frac{e^{a_{b,i}^{(t+1)}}-e^{a_{b,i}^{(t)}}}{\eta}=\frac{e^{a_{b,i}^{(t)}}}{\eta}\left(e^{\eta\delta a_{b,i}^{(t)}}-1\right)=(1+o(1))e^{a_{b,i}^{(t)}}\delta a_{b,i}^{(t)}.
\end{align}
Then by \eqref{eq:delta_a_b_0_simplified_balanced_phase_1.2}, \eqref{eq:I_h_0_balanced_phase1.2}, \eqref{eq:a_b_2_approx_0_t<=T_1_4}, \eqref{eq:a_p_2_bound_phase_1.2} and \eqref{eq:delta_a_q_x_simplified_T_13-T_14} we know that for all $t\in [T_{1/2}, s-1]$: 
\begin{align}
    \delta e^{a_{b,0}^{(t)}}\leq (1+o(1))\frac{a_{p,1}^{(t)}}{N}\delta a_{q,\times}^{(t)},
\end{align}
which suggests 
\begin{align}
    e^{a_{b,0}^{(s-1)}}-e^{a_{b,0}^{(T_{1,3})}}\leq \frac{1+o(1)}{N}a_{p,1}^{(s-1)}a_{q,\times}^{(s-1)}&\lesssim \frac{1}{Nk}\left(a_{p,1}^{(s-1)}\right)^2,
\end{align}
where the last inequality follows from \eqref{eq:a_p_2_bound_phase_1.2} and \eqref{eq:a_q_x_q_0_p_2_asymp_phase_1.2}. Thus by the above expression and \eqref{eq:a_b_0_bound_<T_1_3}, there exists constant $c>0$ such that 
\begin{align}\label{eq:e_a_b_0_bound_T_1_4}
    e^{a_{b,0}^{(s-1)}}\leq 1+\frac{c}{Nk}\left(a_{p,1}^{(s-1)}\right)^2.
\end{align}
And by \eqref{eq:I_h_1_balanced_phase1.2_simplified} we have for all $t\in[T_{1/2}, s-1]$ and $h\in[k-1]$:
\begin{align}\label{eq:e_ab1_I_h_1_lb_T_13-T_14}
    e^{a_{b,1}^{(t)}}I_{h,1}^{(t)}
    &\geq \left(1+O\left(\frac{1}{k}\right)\right)\frac{e^{a_{b,1}^{(t)}}\left(e^{a_{b,0}^{(t)}}+he^{a_{b,2}^{(t)}}\right)}{e^{a_{b,0}^{(t)}}+he^{a_{b,1}^{(t)}}+he^{a_{b,2}^{(t)}}}a_{p,1}^{(t)}\cdot\frac{he^{a_{b,1}^{(t)}}}{e^{a_{b,0}^{(t)}}+he^{a_{b,1}^{(t)}}+he^{a_{b,2}^{(t)}}}\pi_{1,2h+1}^{(t)}(\down_\times)\notag\\
    &\overset{\eqref{eq:a_b_2_approx_0_t<=T_1_4}}\geq \left(1+O\left(\frac{1}{k}\right)\right)\frac{he^{a_{b,1}^{(t)}}}{e^{a_{b,0}^{(t)}}+he^{a_{b,1}^{(t)}}+he^{a_{b,2}^{(t)}}}a_{p,1}^{(t)}\cdot\frac{he^{a_{b,1}^{(t)}}}{e^{a_{b,0}^{(t)}}+he^{a_{b,1}^{(t)}}+he^{a_{b,2}^{(t)}}}\pi_{1,2h+1}^{(t)}(\down_\times)\notag\\
    &\geq \left(1+O\left(\frac{1}{k}\right)\right)\frac{a_{p,1}^{(t)}}{3}\Bigg(\frac{he^{a_{b,2}^{(t)}}}{e^{a_{b,0}^{(t)}}+(2h-1)e^{a_{b,2}^{(t)}}}\pi_{1,2h}^{(t)}(\down_\times^1)\notag\\
    &\qquad+\frac{he^{a_{b,1}^{(t)}}}{e^{a_{b,0}^{(t)}}+he^{a_{b,1}^{(t)}}+he^{a_{b,2}^{(t)}}}\pi_{1,2h+1}^{(t)}(\down_\times)-\frac{he^{a_{b,2}^{(t)}}}{e^{a_{b,0}^{(t)}}+(2h-1)e^{a_{b,2}^{(t)}}}\pi_{1,2h}^{(t)}(\down_\times^1)\Bigg),
\end{align}
where the first inequality uses \eqref{eq:a_p_2_bound_phase_1.2}, and the third inequality uses \eqref{eq:ea_b_2_bound_T_14} in Induction Hypothesis II and \eqref{eq:a_b_2_approx_0_t<=T_1_4}.
By \eqref{eq:e_a_b_0_bound_T_1_4}, the first relation in \eqref{eq:pi_2h_down_1_bound_phase1.2} and \eqref{eq:a_p_1_ub_T_1_4} we know that 
\begin{align}\label{eq:pi_1_2h_down_1_ub_phase1.2_simplified}
    \forall h\in[k-1]:\quad\pi_{1,2h}^{(s-1)}(\down_\times^1)&\leq \exp\left(-r_{\up,h,2}^{(s-1)}\left(a_{p,1}^{(s-1)}+a_{p,2}^{(s-1)}\right)\right)\left(1-\pi_{1,2h}^{(s-1)}(\up)\right)\notag\\
    &\leq \exp\left(-\frac{a_{p,1}^{(s-1)}+a_{p,2}^{(s-1)}}{2h+\frac{c}{Nk}\left(a_{p,1}^{(s-1)}\right)^2}\right)\left(1-\pi_{1,2h}^{(s-1)}(\up)\right)\notag\\
    &\leq \exp\left(-\left(1+O\left(\frac{1}{k}\right)\right)\frac{1}{2\frac{k}{a_{p,1}^{(s-1)}}+\frac{c}{Nk}a_{p,1}^{(s-1)}}\right)\left(1-\pi_{1,2h}^{(s-1)}(\up)\right).
\end{align}
By our choice of $T_{1,3}$, \eqref{eq:a_p_1_ub_T_1_4} and the fact that $a_{p,1}^{(t)}$ strictly increases, we have
\begin{align}
    \forall t\in[T_{1,3}, s-1]:\quad 21k\ln k\leq a_{p,1}^{(t)}\leq k^{9/4}\sqrt{N}.
\end{align}
Combining the above two inequalities, we have
(i) if $a_{p,1}^{(s-1)}\leq k^2$, then
\begin{align}
    \pi_{1,2h}^{(s-1)}(\down_\times^1)\leq \frac{1}{k^{10}}\left(1-\pi_{1,2h}^{(s-1)}(\up)\right)\leq \frac{1-\pi_{1,2h}^{(s-1)}(\up)}{k^8 a_{p,1}^{(s-1)}},
\end{align}
and (ii) if $a_{p,1}^{(s-1)}\geq k^2$, then 
\begin{align}
    \pi_{1,2h}^{(s-1)}(\down_\times^1)\leq e^{-k}\left(1-\pi_{1,2h}^{(s-1)}(\up)\right)\overset{\eqref{eq:a_p_1_ub_T_1_4}}\leq \frac{1-\pi_{1,2h}^{(t)}(\up)}{k^8a_{p,1}^{(s-1)}}.
\end{align}
Thereafter, we have
\begin{align}\label{eq:pi_1_2h_down_times_1_small<=T_1_4}
    \forall h\in[k-1],\,\,t\in[T_{1/2}, s-1]:\quad\pi_{1,2h}^{(t)}(\down_\times^1)&\leq\frac{1}{k^{8}a_{p,1}^{(t)}}\left(1-\pi_{1,2h}^{(t)}(\up)\right).
\end{align}
 By \eqref{eq:e_ab1_I_h_1_lb_T_13-T_14}, \eqref{eq:delta_ea_b_i}, \eqref{eq:delta_a_p_1_simplified_balanced_phase_1.2}, \eqref{eq:pi_1_2h_down_times_1_small<=T_1_4} and \eqref{eq:delta_a_q_x_simplified_T_13-T_14} we have for all $t\in[T_{1,3}, s-1]$:
\begin{align}\label{eq:delta_ea_b_1_lb<=T_1_4}
    \delta e^{a_{b,1}^{(t)}}&\geq\frac{1+o(1)}{N}\left(\frac{k a_{p,1}^{(t)}}{3}\delta a_{p,1}^{(t)}-\frac{1}{3k^{8}}\delta a_{q,\times}^{(t)}\right).
\end{align}
This indicates that \eqref{eq:ea_b_1_bound_T_14} holds at step $s$.

On the other hand, similar as how we obtain \eqref{eq:delta_a_b_1_bound_phase1.2}, from \eqref{eq:I_h_1_balanced_phase1.2_simplified} we can obtain that for any $t\in[T_{1,3},s-1]$,
\begin{align}
    \delta e^{a_{b,1}^{(t)}}&\leq (1+o(1)) \frac{e^{a_{b,1}^{(t)}}\left(e^{a_{b,0}^{(t)}}+he^{a_{b,2}^{(t)}}\right)}{e^{a_{b,0}^{(t)}}+he^{a_{b,1}^{(t)}}+he^{a_{b,2}^{(t)}}}\left(\frac{k^2 a_{p,2}^{(t)}}{N}\delta a_{p,1}^{(t)}+\frac{ a_{p,2}^{(t)}}{N\exp\left(0.1a_{p,2}^{(t)}\right)}\delta a_{q,\times}^{(t)}\right)\notag\\
    &\leq (1+o(1))\frac{k}{\ln k}\left(\frac{k^2 a_{p,2}^{(t)}}{N}\delta a_{p,1}^{(t)}+\frac{ a_{p,2}^{(t)}}{Nk^2}\delta a_{q,\times}^{(t)}\right)\notag\\
    &\overset{\eqref{eq:a_p_2_bound_phase_1.2}}=(1+o(1))\frac{k}{\ln k}\left(\frac{k a_{p,1}^{(t)}}{N}\delta a_{p,1}^{(t)}+\frac{ a_{p,1}^{(t)}}{Nk^3}\delta a_{q,\times}^{(t)}\right),
\end{align}
where the second line follows from \eqref{eq:a_b_2_approx_0_t<=T_1_4},  and our choice of $T_{1,4}$ and $T_{1/2}$. Therefore, we have
\begin{align}
    e^{a_{b,1}^{(s)}}-e^{a_{b,1}^{(T_{1,3})}}&\leq (1+o(1))\frac{k}{N\ln k}a_{p,1}^{(s)}\left(k\left(a_{p,1}^{(s)}-a_{p,1}^{(T_{1,3})}\right)+\frac{1}{k^3}\left(a_{q,\times}^{(s)}-a_{q,\times}^{(T_{1,3})}\right)\right)\notag\\
    &\leq (1+o(1))\frac{k^2}{N\ln k}\left(a_{p,1}^{(s)}\right)^2\overset{\eqref{eq:a_p_1_ub_T_1_4}}\lesssim \frac{k^{6.5}}{\ln k},
\end{align}
where the second inequality follows from \eqref{eq:a_q_x_q_0_p_2_asymp_phase_1.2} and \eqref{eq:a_p_2_bound_phase_1.2}. This shows \eqref{eq:ea_b_1_ub_T_14} holds at step $s$.

We now show \eqref{eq:ea_b_2_bound_T_14} holds at step $s$. First, when $s=T_{1,3}$, by \eqref{eq:a_b_0_bound_<T_1_3} and \eqref{eq:a_b_1_bound_<T_1_3} we know that 
\begin{align*}
    e^{a_{b,0}^{(T_{1,3})}}-1<0<\frac{e^{a_{b,1}^{(T_{1,3})}}-1}{k^2},
\end{align*}
and thus \eqref{eq:ea_b_2_bound_T_14} holds at step $s$. If $s>T_{1,3}$, then by setting $t=s$ in \eqref{eq:ea_b_1_bound_T_14}, and setting $t'\in[T_{1,3}, s-1]$ such that
\begin{align*}
    \frac{1}{2}a_{p,1}^{(t)} \leq a_{p,1}^{(t')} \leq \frac{1}{2}a_{p,1}^{(t)}+1,
\end{align*}
we obtain that 
\begin{align}\label{eq:e_ab_1_lb_intermediate}
    e^{a_{b,1}^{(s)}}-1\geq e^{a_{b,1}^{(s)}}-e^{a_{b,1}^{(t')}}&\gtrsim \frac{k}{N}\left(a_{p,1}^{(s)}\right)^2\overset{\eqref{eq:e_a_b_0_bound_T_1_4}}\gtrsim k^2\left(e^{a_{b,0}^{(s)}}-1\right).
\end{align}
This shows \eqref{eq:ea_b_2_bound_T_14} holds at step $s$.

Moreover, by \eqref{eq:pi_1_2h_down_times_1_small<=T_1_4}, \eqref{eq:I_h_0_balanced_phase1.2} and Induction Hypothesis II we have 
\begin{align}\label{eq:I_h_0<T_14}
    I_{h,0}^{(t)}&\coloneqq \left(1+O\left(\frac{1}{k}\right)\right)r_{\up,h,0}^{(t)}r_{\up,h,2}^{(t)}ha_{p,2}^{(t)}\left(1-\pi_{1,2h}^{(t)}(\up)\right)\notag\\
    &\quad\quad-\left(1+O\left(\frac{1}{k}\right)\right)h
    r_{\down,h,0}^{(t)}r_{\down,h,1}^{(t)}\left(a_{p,1}^{(t)}\pi_{1,2h+1}^{(t)}(\down_\times)+a_{p,2}^{(t)}\pi_{1,2h+1}^{(t)}(\up)\right).
\end{align}
 Therefore, by \eqref{eq:delta_a_q_x_simplified_T_13-T_14} and \eqref{eq:I_h_0<T_14} we have for all $t\leq s-1$: 
\begin{align}\label{eq:delta_a_b_0_bound_phase1.2}
    \delta a_{b,0}^{(t)}&\leq \frac{1+o(1)}{4N}a_{p,2}^{(t)}\delta a_{q,\times}^{(t)},
\end{align}
where we use the fact that $hr_{\up,h,2}^{(t)}\leq 1-r_{\up,h,0}^{(t)}$, and $x(1-x)\leq \frac{1}{4}$ for any $x\in\R$.
And by \eqref{eq:delta_a_p_1_simplified_balanced_phase_1.2} 
and \eqref{eq:sum_1-pi_2h+1_up_vs_sum_1-pi_2h_up} we have for all $t\in [T_{1,3},s-1]$:
\begin{align}\label{eq:-delta_a_b_0_bound_phase1.2}
    -\delta a_{b,0}^{(t)}&\leq (1+o(1))r_{\down,h,0}^{(t)}\left(\frac{k^2 a_{p,2}^{(t)}}{N}\delta a_{p,1}^{(t)}+\frac{ a_{p,2}^{(t)}}{N\exp\left(0.1a_{p,2}^{(t)}\right)}\delta a_{q,\times}^{(t)}\right)\notag\\
    &\lesssim \frac{1}{a_{p,2}^{(t)}\ln k}\delta a_{p,1}^{(t)}+\frac{1}{k^2\ln ka_{p,2}^{(t)}\exp\left(0.1a_{p,2}^{(t)}\right)}\delta a_{q,\times}^{(t)},
\end{align}
where the last relation follows from 
\begin{align}
    r_{\down,h,0}^{(t)}\lesssim \frac{N}{\left(a_{p,1}^{(t)}\right)^2\ln k}
\end{align}
guaranteed by our choice of $T_{1,4}$ and \eqref{eq:ea_b_2_bound_T_14}. 

We next show \eqref{eq:e2a_b_0_lb_T_14} holds at step $s$. First we have
\begin{align}\label{eq:delta_e-a_b_0<T_14}
    \delta e^{-a_{b,0}^{(t)}}&\coloneqq \frac{e^{-a_{b,0}^{(t+1)}}-e^{-a_{b,0}^{(t)}}}{\eta}\notag\\
    &=(1+o(1))e^{-a_{b,0}^{(t)}}\left(-\delta a_{b,0}^{(t)}\right)\notag\\
    &\overset{\eqref{eq:I_h_0<T_14}}=
        \frac{1+o(1)}{N}\sum_{h=1}^{k-1}\frac{k-h}{k}\frac{he^{a_{b,1}^{(t)}}}{\left(e^{a_{b,0}^{(t)}}+he^{a_{b,1}^{(t)}}+he^{a_{b,2}^{(t)}}\right)^2}\left(a_{p,1}^{(t)}\pi_{1,2h+1}^{(t)}(\down_\times)+a_{p,2}^{(t)}\pi_{1,2h+1}^{(t)}(\up)\right)\notag\\
    &\qquad -\frac{1+o(1)}{N}\sum_{h=1}^{k-1}\frac{k-h}{k}\frac{he^{a_{b,2}^{(t)}}}{\left(e^{a_{b,0}^{(t)}}+(2h-1)e^{a_{b,2}^{(t)}}\right)^2}a_{p,2}^{(t)}\left(1-\pi_{1,2h}^{(t)}(\up)\right).
\end{align}
Thus our choice of $T_{1,3}, T_{1,4}$, \eqref{eq:a_b_2_approx_0_t<=T_1_4}, \eqref{eq:sum_1-pi_2h+1_up_vs_sum_1-pi_2h_up}, \eqref{eq:delta_a_q_x_simplified_T_13-T_14} and \eqref{eq:delta_a_p_0_simplified_balanced_phase_1.2} we have
\begin{align}\label{eq:delta_e-a_b_0_ub_T_13-T_14}
    \delta e^{-a_{b,0}^{(t)}}&\leq \frac{1+o(1)}{N}\frac{1}{e^{a_{b,0}^{(t)}}+e^{a_{b,1}^{(t)}}+1}\sum_{h=1}^{k-1}\frac{k-h}{k}\frac{he^{a_{b,1}^{(t)}}}{e^{a_{b,0}^{(t)}}+he^{a_{b,1}^{(t)}}+he^{a_{b,2}^{(t)}}}\left(a_{p,1}^{(t)}\pi_{1,2h+1}^{(t)}(\down_\times)+a_{p,2}^{(t)}\pi_{1,2h+1}^{(t)}(\up)\right)\notag\\
    &\qquad -\frac{1+o(1)}{N}\left(\frac{\ln k}{k}\right)^2\sum_{h=1}^{k-1}\frac{k-h}{k}a_{p,2}^{(t)}\left(1-\pi_{1,2h}^{(t)}(\up)\right)\notag\\
    &\leq \frac{1+o(1)}{N}\frac{1}{e^{a_{b,0}^{(t)}}+e^{a_{b,1}^{(t)}}+1}\left(ka_{p,1}^{(t)}\delta a_{p,1}^{(t)}+\frac{a_{p,2}^{(t)}}{\exp\left(0.1a_{p,2}^{(t)}\right)}\delta a_{q,\times}^{(t)}\right)-\frac{1+o(1)}{N}\left(\frac{\ln k}{k}\right)^2a_{p,2}^{(t)}\delta a_{q,\times}^{(t)}\notag\\
    &\overset{\eqref{eq:a_p_2_bound_phase_1.2}}\leq \frac{1+o(1)}{N}\frac{ka_{p,1}^{(t)}}{e^{a_{b,0}^{(t)}}+e^{a_{b,1}^{(t)}}+1}\delta a_{p,1}^{(t)}-\frac{1+o(1)}{N}\frac{\ln^2 k}{k^3}a_{p,1}^{(t)}\delta a_{q,\times}^{(t)}.
\end{align}
Define
\begin{align}\label{eq:t_i_phase1.2}
    t_i\coloneqq \inf\left\{t\in [T_{1,3},s-1]: e^{a_{b,1}^{(t)}}\geq k^{i/2}\right\},\,\,\forall i\in \NN.
\end{align}
then by \eqref{eq:a_b_1_bound_<T_1_3} we know that $t_0=T_{1,3}$.
Comparing \eqref{eq:delta_e-a_b_0_ub_T_13-T_14} and \eqref{eq:delta_ea_b_1_lb<=T_1_4}, we have for any $t\in [T_{1,3},s-1]$, $i\in \NN$:
\begin{align}\label{eq:delta_e-a_b_0_ub_w.r.t.a_b_1_T_13-T_14}
    \text{If } t_i\leq t\leq t_{i+1}:
    \quad \delta e^{-a_{b,0}^{(t)}}\lesssim k^{-i/2}\delta e^{a_{b,1}^{(t)}},
\end{align}
and by \eqref{eq:ea_b_1_ub_T_14} and the fact that $a_{b,1}^{(t)}$ monotonically increases (c.f.~\eqref{eq:delta_ea_b_1_bound_phase_1.2}), there exists positive integer $j\leq 12$ such that $k^{j/2} \leq e^{a_{b,1}^{(s)}}\leq k^{(j+1)/2}$ holds at step $s$. Thus by \eqref{eq:delta_e-a_b_0_ub_w.r.t.a_b_1_T_13-T_14} we have
\begin{align}
    e^{-a_{b,0}^{(s)}}-e^{-a_{b,0}^{(T_{1,3})}}&\lesssim k^{-j/2}\left(e^{a_{b,1}^{(s)}}-e^{a_{b,1}^{(t_j)}}\right)+\sum_{i=0}^{j-1}k^{-i/2}\left(e^{a_{b,1}^{(t_{i+1})}}-e^{a_{b,1}^{(t_i)}}\right)\overset{\eqref{eq:t_i_phase1.2}}\lesssim \sqrt{k},
\end{align}
which combined with \eqref{eq:a_b_0_bound_<T_1_3} suggests \eqref{eq:e2a_b_0_lb_T_14} holds at step $s$.

Now we show \eqref{eq:a_p_1_ub_T_1_4} holds at step $s$ by contradiction. Suppose 
\begin{align}\label{eq:a_p_1_s_ub_contradiction}
    a_{p,1}^{(s)}>k^{9/4}\sqrt{N},
\end{align}
then we can choose $s_0\in[T_{1,3},s-1]$ such that
\begin{align}\label{eq:a_p_1_s0_contradiction}
    \frac{a_{p,1}^{(s)}}{\ln k}\leq a_{p,1}^{(s_0)}\leq \frac{a_{p,1}^{(s)}}{\ln k}+1.
\end{align}
Then by \eqref{eq:ea_b_2_bound_T_14} we have
\begin{align}\label{eq:ea_b_1_s0_lb_contradiction}
    e^{a_{b,1}^{(s_0)}}\gtrsim \frac{k}{N}\left(a_{p,1}^{(s_0)}\right)^2\gtrsim \frac{k^{5.5}}{\ln^2 k}.
\end{align}
Plugging this into \eqref{eq:delta_e-a_b_0_ub_T_13-T_14}, we obtain
\begin{align}
    \forall t\in[s_0,s-1]:\quad \delta e^{-a_{b,0}^{(t)}}&\leq \frac{c}{N}\frac{a_{p,1}^{(s_0)}\ln^2 k}{k^{4.5}}\delta a_{p,1}^{(t)}-\frac{1+o(1)}{N}\frac{\ln^2 k}{k^3}a_{p,1}^{(s)}\delta a_{q,\times}^{(t)}\notag\\
    &\overset{\eqref{eq:a_p_1_s0_contradiction}}\leq (1+o(1))\frac{c}{N}\frac{\ln k}{k^{4.5}}a_{p,1}^{(s)}\delta a_{p,1}^{(t)}-\frac{1+o(1)}{N}\frac{\ln^2 k}{k^3}a_{p,1}^{(s)}\delta a_{q,\times}^{(t)}
\end{align}
with some constant $c>0$. This implies
\begin{align}
    e^{-a_{b,0}^{(s)}}-e^{-a_{b,0}^{(s_0)}}\leq -\frac{1+o(1)}{N}\frac{\ln^2 k}{k^3}a_{p,1}^{(s)}\left(a_{q,\times}^{(s)}-a_{q,\times}^{(s_0)}\right)+(1+o(1))\frac{c}{N}\frac{\ln k}{k^{4.5}}a_{p,1}^{(s)}\left(a_{p,1}^{(s)}-a_{p,1}^{(s_0)}\right).
\end{align}
By \eqref{eq:a_q_x_q_0_p_2_asymp_phase_1.2}, \eqref{eq:a_p_2_bound_phase_1.2}, we have
\begin{align}
    a_{q,\times}^{(s)}\asymp \frac{a_{p,1}^{(s)}}{k}\overset{\eqref{eq:a_p_1_s_ub_contradiction}}\gtrsim k^{5/4}\sqrt{N},\quad a_{q,\times}^{(s_0)}\asymp \frac{a_{p,1}^{(s_0)}}{k}\overset{\eqref{eq:a_p_1_s0_contradiction}}\asymp \frac{a_{p,1}^{(s)}}{k\ln k}\asymp \frac{a_{q,\times}^{(s)}}{\ln k}.
\end{align}
Combining the above two expressions, we have
\begin{align}
    e^{-a_{b,0}^{(s)}}-e^{-a_{b,0}^{(s_0)}}\leq -\frac{1+o(1)}{N}\frac{\ln^2 k}{k^3}a_{p,1}^{(s)}a_{q,\times}^{(s)}\overset{\eqref{eq:a_p_1_s_ub_contradiction}}\lesssim -\frac{1}{N}\frac{\ln^2 k}{k^4}\left(k^{9/4}\sqrt{N}\right)^2=\sqrt{k}\ln^2 k,
\end{align}
which together with \eqref{eq:e2a_b_0_lb_T_14} suggests there exists $c'>0$ such that
\begin{align}
    e^{-a_{b,0}^{(s)}}\leq c'\sqrt{k}-\sqrt{k}\ln^2 k<0.
\end{align}
This is impossible. Therefore, \eqref{eq:a_p_1_ub_T_1_4} holds at step $s$. 

In addition, by \eqref{eq:a_p_1_ub_T_1_4}, \eqref{eq:delta_a_p_1_simplified_balanced_phase_1.2}, \eqref{eq:sum_pi_1_2h+1_down_times_bound_phase1.2}, \eqref{eq:a_p_2_bound_phase_1.2}, \eqref{eq:r_down_times<=1/k-1} we deduce
\begin{align}
    \delta a_{p,1}^{(t)}\gtrsim \frac{\exp\left(-\left(1+\frac{2}{k}\right)\frac{a_{p,1}^{(t)}}{k}\right)}{k^2}\gtrsim \frac{\exp\left(-\left(1+\frac{2}{k}\right)k^{5/4}\sqrt{N}\right)}{k^2}\geq \exp\left(-k^{1.3}\sqrt{N}\right)
\end{align}
for any $t\in[T_{1,3},s-1]$, which implies \eqref{eq:a_p_1_increase_rate_T_13-T_14} holds at step $s$.

On the other hand, if $s=T_{1,4}$, then \eqref{eq:delta_e-a_b_0_ub_T_13-T_14} suggests
\begin{align*}
    \forall t\in[T_{1,3},T_{1,4}-1]:\quad -\delta e^{-a_{b,0}^{(t)}}\leq \frac{1+o(1)}{N}\frac{\ln^2 k}{k^3}a_{p,1}^{(t)}\delta a_{q,\times}^{(t)},
\end{align*}
and thus
\begin{align*}
    \frac{1+o(1)}{N}\frac{\ln^2 k}{k^3}a_{p,1}^{(T_{1,4})}\left( a_{q,\times}^{(T_{1,4})}-a_{q,\times}^{(T_{1,3})}\right)\geq\eta\sum_{t=T_{1,3}}^{T_{1,4}-1}\left(-\delta e^{-a_{b,0}^{(t)}}\right)=e^{-a_{b,0}^{(T_{1,3})}}-e^{-a_{b,0}^{(T_{1,4})}}=1+o(1),
\end{align*}
where the last relation follows from our choice of $T_{1,4}$ and \eqref{eq:a_b_0_bound_<T_1_3}. From the above inequality and \eqref{eq:a_q_x_q_0_p_2_asymp_phase_1.2}, \eqref{eq:a_p_2_bound_phase_1.2} we deduce
\begin{align*}
    \frac{1}{N}\frac{\ln^2 k}{k^4}\left(a_{p,1}^{(T_{1,4})}\right)^2\gtrsim 1,
\end{align*}
from which we can see that 
\begin{align}\label{eq:a_p_1_lb_s=T_14}
    a_{p,1}^{(T_{1,4})}\gtrsim \frac{k^{2}\sqrt{N}}{\ln k}.
\end{align}
By \eqref{eq:a_p_1_lb_s=T_14} and \eqref{eq:ea_b_2_bound_T_14} we have
\begin{align}\label{eq:ea_b_1_lb_s=T_14}
    e^{a_{b,1}^{(T_{1,4})}}\gtrsim \frac{k^5}{\ln^2 k}.
\end{align}


To show \eqref{eq:a_q_x>a_p_2_bound_T_13}, \eqref{eq:a_q_2>a_q_x_bound_T_13}, \eqref{eq:a_p2>a_q_0_bound_T_13}, \eqref{eq:a_q_0>a_q_x_bound_T_13} in Lemma~\ref{lm:phase_1.2} hold at step $s$, below we first bound the gradients of $\mu_2^{(t)}$, $\mu_{2(k-1)}^{(t)}$ and $r_{\down_\times}^{(t)}\left(a_{p,1}^{(t)}+a_{p,2}^{(t)}\right)$ that appear in those relations.


By \eqref{eq:mu_h_phase1.2},
\small
\begin{align}\label{eq:delta_mu_2h_expression_phase1.2}
    \delta \mu_{2h}^{(t)}     &\coloneqq \frac{\mu_{2h}^{(t+1)}-\mu_{2h}^{(t)}}{\eta}\notag\\
    &=\frac{1}{\eta}\left[\left(-r_{\up,h,0}^{(t+1)}\frac{a_{p,0}^{(t+1)}}{k}+r_{\up,h,2}^{(t+1)}\left(ha_{p,2}^{(t+1)}+\frac{h-1}{k}a_{p,\up}^{(t+1)}\right)\right)-\left(-r_{\up,h,0}^{(t)}\frac{a_{p,0}^{(t)}}{k}+r_{\up,h,2}^{(t)}\left(ha_{p,2}^{(t)}+\frac{h-1}{k}a_{p,\up}^{(t)}\right)\right)\right]\notag\\
    &=-r_{\up,h,0}^{(t+1)}\frac{\delta a_{p,0}^{(t)}}{k}-\frac{r_{\up,h,0}^{(t+1)}-r_{\up,h,0}^{(t)}}{\eta}\frac{a_{p,0}^{(t)}}{k}+r_{\up,h,2}^{(t+1)}\left(h\delta a_{p,2}^{(t)}+\frac{h-1}{k}\delta a_{p,\up}^{(t)}\right)+\frac{r_{\up,h,2}^{(t+1)}-r_{\up,h,2}^{(t)}}{\eta}\left(ha_{p,2}^{(t)}+\frac{h-1}{k}a_{p,\up}^{(t)}\right),
\end{align}
\normalsize
and from \eqref{eq:r_up_h_phase1.2} we deduce
\begin{align}
    \frac{r_{\up,h,0}^{(t+1)}-r_{\up,h,0}^{(t)}}{\eta}&=\frac{1}{\eta}\frac{(2h-1)e^{a_{b,2}^{(t)}-a_{b,0}^{(t)}}\left(1-e^{\eta\left(\delta a_{b,2}^{(t)}-\delta a_{b,0}^{(t)}\right)}\right)}{\left(1+(2h-1)e^{a_{b,2}^{(t)}-a_{b,0}^{(t)}}\right)\left(1+(2h-1)e^{a_{b,2}^{(t+1)}-a_{b,0}^{(t+1)}}\right)}\notag\\
    &=r_{\up,h,0}^{(t+1)}\left(1-r_{\up,h,0}^{(t)}\right)\frac{1-e^{\eta\left(\delta a_{b,2}^{(t)}-\delta a_{b,0}^{(t)}\right)}}{\eta}\notag\\
    &=-\left(1+o(1)\right)r_{\up,h,0}^{(t+1)}\left(1-r_{\up,h,0}^{(t)}\right)\frac{1-e^{\eta\left(\delta a_{b,0}^{(t)}-\delta a_{b,2}^{(t)}\right)}}{\eta}
\end{align}
and
\begin{align}
    \frac{r_{\up,h,2}^{(t+1)}-r_{\up,h,2}^{(t)}}{\eta}&=\frac{1}{\eta}\frac{e^{a_{b,0}^{(t)}-a_{b,2}^{(t)}}\left(1-e^{\eta\left(\delta a_{b,0}^{(t)}-\delta a_{b,2}^{(t)}\right)}\right)}{\left(2h-1+e^{a_{b,0}^{(t)}-a_{b,2}^{(t)}}\right)\left(2h-1+e^{a_{b,0}^{(t+1)}-a_{b,2}^{(t+1)}}\right)}\notag\\
    &=\frac{1}{2h-1}\left(1-r_{\up,h,0}^{(t+1)}\right)r_{\up,h,0}^{(t)}\frac{1-e^{\eta\left(\delta a_{b,0}^{(t)}-\delta a_{b,2}^{(t)}\right)}}{\eta}.
\end{align}
Plugging the above two equations into \eqref{eq:delta_mu_2h_expression_phase1.2}, we obtain
\begin{align}\label{eq:delta_mu_2h_expression_phase1.2_simplified}
    \delta \mu_{2h}^{(t)}=\left(1+o(1)\right)\left(-r_{\up,h,0}^{(t+1)}\frac{\delta a_{p,0}^{(t)}}{k}+r_{\up,h,2}^{(t+1)}h\delta a_{p,2}^{(t)}+\frac{h}{2h-1}r_{\up,h,0}^{(t+1)}\left(1-r_{\up,h,0}^{(t)}\right)a_{p,2}^{(t)}\frac{1-e^{\eta\left(\delta a_{b,0}^{(t)}-\delta a_{b,2}^{(t)}\right)}}{\eta}\right).
\end{align}
where
\begin{align}
    -\frac{1-e^{\eta\left(\delta a_{b,0}^{(t)}-\delta a_{b,2}^{(t)}\right)}}{\eta}&=(1+o(1))\left(\delta a_{b,0}^{(t)}-\delta a_{b,2}^{(t)}\right)\overset{\eqref{eq:delta_a_b_2_simplified_balanced}}=(1+o(1))\left(\frac{N}{N-1}\delta a_{b,0}^{(t)}+\frac{1}{N-1}\delta a_{b,1}^{(t)}\right),
\end{align}
by 
\eqref{eq:delta_a_b_0_bound_phase1.2} and \eqref{eq:delta_a_b_1_bound_phase1.2}, there exists a constant $c>0$ such that
\begin{align}\label{eq:delta_a_b_0-delta_a_b_2_ub}
    -\frac{1-e^{\eta\left(\delta a_{b,0}^{(t)}-\delta a_{b,2}^{(t)}\right)}}{\eta}\leq \frac{c}{a_{p,2}^{(t)}N\ln k}\delta a_{p,1}^{(t)}+\frac{1+o(1)}{4N}a_{p,2}^{(t)}\delta a_{q,\times}^{(t)},
\end{align}
and by \eqref{eq:delta_a_b_1_bound_phase1.2}, \eqref{eq:-delta_a_b_0_bound_phase1.2} we have
\begin{align}\label{eq:delta_a_b_0-delta_a_b_2_lb}
    \frac{1-e^{\eta\left(\delta a_{b,0}^{(t)}-\delta a_{b,2}^{(t)}\right)}}{\eta}\lesssim \frac{1}{a_{p,2}^{(t)}\ln k}\delta a_{p,1}^{(t)}+\frac{1}{k^2\ln ka_{p,2}^{(t)}\exp\left(0.1a_{p,2}^{(t)}\right)}\delta a_{q,\times}^{(t)}.
\end{align}

Combining \eqref{eq:delta_mu_2h_expression_phase1.2_simplified} and \eqref{eq:delta_a_b_0-delta_a_b_2_ub}, we have for all $t\in [T_{1,3},s-1]$:
\begin{align}\label{eq:-delta_mu_2h_ub}
    -\delta \mu_{2h}^{(t)}&\leq (1+o(1))\left(r_{\up,h,0}^{(t)}\frac{\delta a_{p,0}^{(t)}}{k}-r_{\up,h,2}^{(t)}h\delta a_{p,2}^{(t)}+\frac{c}{N\ln k}\delta a_{p,1}^{(t)}+\frac{\left(a_{p,2}^{(t)}\right)^2}{16N}\delta a_{q,\times}^{(t)}\right)\notag\\
    &\overset{\eqref{eq:a_p3}}= (1+o(1))\left(r_{\up,h,0}^{(t+1)}\frac{\delta a_{p,0}^{(t)}}{k}+r_{\up,h,2}^{(t)}h\frac{\delta a_{p,3}^{(t)}}{k}-r_{\up,h,2}^{(t)}h\frac{\delta a_{p,1}^{(t)}}{k}+\frac{c}{N\ln k}\delta a_{p,1}^{(t)}+\frac{\left(a_{p,2}^{(t)}\right)^2}{16N}\delta a_{q,\times}^{(t)}\right)\notag\\
    &\leq (1+o(1))\left(\frac{1}{k}+\frac{\left(a_{p,2}^{(t)}\right)^2}{16N}\right)\delta a_{q,\times}^{(t)}-(1+o(1))\frac{\ln k}{k^2}\delta a_{p,1}^{(t)},
\end{align}
where in the last line we use the fact that 
\begin{align}
    \delta a_{p,0}^{(t)}& \leq (1+o(1))\delta a_{q,\times}^{(t)}, \label{eq:delta_a_p_0<=delta_a_q_x_T_13-T_14} \\
    \delta a_{p,3}^{(t)}& \leq \frac{1+o(1)}{k}\delta a_{q,\times}^{(t)} \label{eq:delta_a_p_3<=1/k_delta_a_q_x_T_13-T_14}
\end{align}
indicated by \eqref{eq:delta_a_p_0_simplified_balanced_phase_1.2}, \eqref{eq:delta_a_p_3_simplified_balanced_phase_1.2} and \eqref{eq:delta_a_q_x_simplified_T_13-T_14}, and the fact that 
\begin{align}
    r_{\up,h,2}h\geq (1+o(1)) \frac{\ln k}{k}
\end{align} 
guaranteed by \eqref{eq:T_1_4} and \eqref{eq:a_b_2_approx_0_t<=T_1_4}.
Combining \eqref{eq:delta_mu_2h_expression_phase1.2_simplified} and \eqref{eq:delta_a_b_0-delta_a_b_2_lb}, we have for all $t\in [T_{1,3},s-1]$:
\begin{align}\label{eq:delta_mu_2h_ub}
    \delta \mu_{2h}^{(t)}&\leq \left(1+o(1)\right)\left(-r_{\up,h,0}^{(t+1)}\frac{\delta a_{p,0}^{(t)}}{k}+r_{\up,h,2}^{(t+1)}h\delta a_{p,2}^{(t)}\right)+\frac{c}{\ln k}\delta a_{p,1}^{(t)}+\frac{c}{k^2\ln k\exp\left(0.1a_{p,2}^{(t)}\right)}\delta a_{q,\times}^{(t)}\notag\\
    &\overset{\eqref{eq:a_p3}}= (1+o(1))\left(-r_{\up,h,0}^{(t+1)}\frac{\delta a_{p,0}^{(t)}}{k}-r_{\up,h,2}^{(t)}h\frac{\delta a_{p,3}^{(t)}}{k}+r_{\up,h,2}^{(t)}h\frac{\delta a_{p,1}^{(t)}}{k}\right)+\frac{c}{\ln k}\delta a_{p,1}^{(t)}\notag\\
    &\hspace{6.8cm}+\frac{c}{k^2\ln k\exp\left(0.1a_{p,2}^{(t)}\right)}\delta a_{q,\times}^{(t)}\notag\\
    &=\left(1+o(1)\right)\left(\frac{c}{\ln k}\delta a_{p,1}^{(t)}+\frac{1}{k}\pi_{1,1}^{(t)}(\up)\right)
\end{align}
for some constant $c>0$, where the last line uses \eqref{eq:delta_a_p_3_simplified_balanced_phase_1.2}, \eqref{eq:delta_a_p_0_simplified_balanced_phase_1.2} and \eqref{eq:pi_1_up_dominate_sum_phase1.2}.

Now we bound the gradient of $r_{\down_\times}^{(t)}\left(a_{p,1}^{(t)}+a_{p,2}^{(t)}\right)$:
\begin{align}\label{eq:delta_r_down_x(a_p_1+a_p_2)_expression_phase1.2}
    \delta\left(r_{\down_\times}^{(t)}\left(a_{p,1}^{(t)}+a_{p,2}^{(t)}\right)\right)&\coloneqq \frac{1}{\eta}\left(r_{\down_\times}^{(t+1)}\left(a_{p,1}^{(t+1)}+a_{p,2}^{(t+1)}\right)-r_{\down_\times}^{(t)}\left(a_{p,1}^{(t)}+a_{p,2}^{(t)}\right)\right)\notag\\
    &=r_{\down_\times}^{(t+1)}\left(\delta a_{p,1}^{(t)}+\delta a_{p,2}^{(t)}\right)+\frac{r_{\down_\times}^{(t+1)}-r_{\down_\times}^{(t)}}{\eta}\left(a_{p,1}^{(t)}+a_{p,2}^{(t)}\right),
\end{align}
and by \eqref{eq:r_down_times_phase1.2}, we have
\begin{align}\label{eq:delta_r_down_x(a_p_1+a_p_2)_expression_phase1.2_2}
    \frac{r_{\down_\times}^{(t+1)}-r_{\down_\times}^{(t)}}{\eta}&=r_{\down, k-1, 0}^{(t)}r_{\down_\times}^{(t+1)}\frac{1-e^{\delta a_{b,0}^{(t)}-\delta a_{b,1}^{(t)}}}{\eta}+(k-1)r_{\down, k-1, 2}^{(t)}r_{\down_\times}^{(t+1)}\frac{1-e^{\delta a_{b,2}^{(t)}-\delta a_{b,1}^{(t)}}}{\eta},
\end{align}
where
\begin{align}\label{eq:delta_a_b_0-delta_a_b_1_ub_simplified}
    -\frac{1-e^{\delta a_{b,0}^{(t)}-\delta a_{b,1}^{(t)}}}{\eta}=(1+o(1))\left(\delta a_{b,0}^{(t)}-\delta a_{b,1}^{(t)}\right),
\end{align}
and
\begin{align}
    -\frac{1-e^{\delta a_{b,2}^{(t)}-\delta a_{b,1}^{(t)}}}{\eta}\overset{\eqref{eq:delta_a_b_2_simplified_balanced}}=-(1+o(1))\left(\frac{N}{N-1}\delta a_{b,1}^{(t)}+\frac{1}{N-1}\delta a_{b,0}^{(t)}\right).
\end{align}
By \eqref{eq:delta_a_b_0_bound_phase1.2}, \eqref{eq:delta_a_b_1_bound_phase1.2} and the above two expressions, we have
\begin{subequations}
\begin{align}
    -\frac{1-e^{\delta a_{b,0}^{(t)}-\delta a_{b,1}^{(t)}}}{\eta} & \leq \frac{1+o(1)}{4N}a_{p,2}^{(t)}\delta a_{q,\times}^{(t)}, \label{eq:delta_a_b_0-delta_a_b_1_ub} \\ 
    \frac{1-e^{\delta a_{b,0}^{(t)}-\delta a_{b,1}^{(t)}}}{\eta}& \lesssim \frac{1}{a_{p,2}^{(t)}\ln k}\delta a_{p,1}^{(t)}+\frac{1}{k^2\ln ka_{p,2}^{(t)}\exp\left(0.1a_{p,2}^{(t)}\right)}\delta a_{q,\times}^{(t)}, \label{eq:delta_a_b_0-delta_a_b_1_lb} \\
    -\frac{1-e^{\delta a_{b,2}^{(t)}-\delta a_{b,1}^{(t)}}}{\eta}& \lesssim \frac{1}{a_{p,2}^{(t)}N\ln k}\delta a_{p,1}^{(t)}+\frac{1}{Nk^2\ln ka_{p,2}^{(t)}\exp\left(0.1a_{p,2}^{(t)}\right)}\delta a_{q,\times}^{(t)}, \label{eq:delta_a_b_2-delta_a_b_1_lb}
\end{align}
\end{subequations}
and there exists a constant $c'>0$ such that
\begin{align}\label{eq:delta_a_b_2-delta_a_b_1_ub}
    \frac{1-e^{\delta a_{b,2}^{(t)}-\delta a_{b,1}^{(t)}}}{\eta}\leq \frac{c'}{a_{p,2}^{(t)}\ln k}\delta a_{p,1}^{(t)}+
\left(\frac{c'}{k^2\ln ka_{p,2}^{(t)}\exp\left(0.1a_{p,2}^{(t)}\right)}+\frac{1+o(1)}{4N^2}a_{p,2}^{(t)}\right)\delta a_{q,\times}^{(t)}.
\end{align}

By \eqref{eq:delta_a_b_0-delta_a_b_1_ub} and \eqref{eq:delta_a_b_2-delta_a_b_1_lb}, \eqref{eq:delta_r_down_x(a_p_1+a_p_2)_expression_phase1.2_2} and \eqref{eq:a_p_2_bound_phase_1.2} we have
\begin{align}\label{eq:delta_r_down_x_lb}
    \frac{r_{\down_\times}^{(t+1)}-r_{\down_\times}^{(t)}}{\eta}&\geq -\frac{1}{k-1}\left(1-(k-1)r_{\down, k-1, 1}^{(t)}\right)\left(\frac{1+o(1)}{4N}a_{p,2}^{(t)}\delta a_{q,\times}^{(t)}+\frac{c_1}{a_{p,2}^{(t)}N\ln k}\delta a_{p,1}^{(t)}\right)\notag\\
    &\overset{\eqref{eq:1-hr_down_h_1_bound_phase1.2}}\gtrsim -\frac{1}{k-1}\frac{N}{\left(a_{p,1}^{(t)}\right)^2\ln k}\left(\frac{1}{N}a_{p,2}^{(t)}\delta a_{q,\times}^{(t)}+\frac{1}{a_{p,2}^{(t)}N\ln k}\delta a_{p,1}^{(t)}\right)\notag\\
    &\asymp -\left(\frac{1}{k^3\ln k a_{p,2}^{(t)}}\delta a_{q,\times}^{(t)}+\frac{1}{k^3\ln^2 k \left(a_{p,2}^{(t)}\right)^3}\delta a_{p,1}^{(t)}\right)
\end{align}
for some constant $c_1>0$, where in the first relation we also use the fact that 
\begin{align}\label{eq:r_down_times<=1/k-1}
    r_{\down_\times}^{(t+1)}\geq \frac{1}{k-1}.
\end{align}
Plugging \eqref{eq:delta_r_down_x_lb} into \eqref{eq:delta_r_down_x(a_p_1+a_p_2)_expression_phase1.2}, and use the fact that
\begin{align}
    r_{\down_\times}^{(t+1)}\geq \frac{1}{2k+1}
\end{align} 
guaranteed by \eqref{eq:ea_b_2_bound_T_14}, \eqref{eq:a_b_2_approx_0_t<=T_1_4} and \eqref{eq:a_b_1_bound_<T_1_3},
 we have
\begin{align}\label{eq:delta_r_down_x_lb_intermediate}
    \delta\left(r_{\down_\times}^{(t)}\left(a_{p,1}^{(t)}+a_{p,2}^{(t)}\right)\right)\geq \frac{1}{2k+1}\left(\delta a_{p,1}^{(t)}+\delta a_{p,2}^{(t)}\right)-\frac{c_2}{k^2\ln k}\delta a_{q,\times}^{(t)},
\end{align}
for some constant $c_2>0$. 
By \eqref{eq:a_p3} and the above inequality we have
\begin{align}\label{eq:delta_r_down_x_lb_simplified}
    \delta \left(r_{\down_\times}^{(t)}\left(a_{p,1}^{(t)}+a_{p,2}^{(t)}\right)\right)&\geq \left(1+o(1)\right)\left(\frac{1}{2k}\delta a_{p,1}^{(t)}-\frac{1}{2k^2}\delta a_{p,3}^{(t)}-\frac{c_2}{k^2\ln k}\delta a_{q,\times}^{(t)}\right)\notag\\
    &\overset{\eqref{eq:delta_a_p_3<=1/k_delta_a_q_x_T_13-T_14}}\geq \left(1+o(1)\right)\left(\frac{1}{2k}\delta a_{p,1}^{(t)}-\frac{c_2}{k^2\ln k}\delta a_{q,\times}^{(t)}\right).
\end{align}

Similar as how we compute \eqref{eq:delta_r_down_x_lb}, plugging \eqref{eq:delta_a_b_0-delta_a_b_1_lb} and \eqref{eq:delta_a_b_2-delta_a_b_1_ub} into \eqref{eq:delta_r_down_x(a_p_1+a_p_2)_expression_phase1.2_2}, and using \eqref{eq:1-hr_down_h_1_bound_phase1.2}, \eqref{eq:r_down_times<=1/k-1} we deduce
\small
\begin{align}
   \frac{r_{\down_\times}^{(t+1)}-r_{\down_\times}^{(t)}}{\eta} 
    &\lesssim \frac{1}{k-1}\min\left\{\frac{N}{\left(a_{p,1}^{(t)}\right)^2\ln k},1\right\}\Bigg(\frac{1}{a_{p,2}^{(t)}\ln k}\delta a_{p,1}^{(t)}+
    \Bigg(\frac{1}{k^2\ln k a_{p,2}^{(t)}\exp\left(0.1a_{p,2}^{(t)}\right)}+\frac{1}{N^2}a_{p,2}^{(t)}\Bigg)\delta a_{q,\times}^{(t)}\Bigg)\notag\\
    &\lesssim  \frac{1}{a_{p,2}^{(t)}k\ln k}\delta a_{p,1}^{(t)}+\left(\frac{1}{k^3\ln k a_{p,2}^{(t)}\exp\left(0.1a_{p,2}^{(t)}\right)}+\frac{1}{Nk^3\ln k a_{p,2}^{(t)}}\right)\delta a_{q,\times}^{(t)}.
\end{align}
\normalsize
Plugging this back into \eqref{eq:delta_r_down_x(a_p_1+a_p_2)_expression_phase1.2} and using \eqref{eq:r_down_times<=1/k-1}, \eqref{eq:a_p_2_bound_phase_1.2}, we have
\begin{align}\label{eq:delta_r_down_x(a_p1+a_p2)_ub}
    \delta\left(r_{\down_\times}^{(t)}\left(a_{p,1}^{(t)}+a_{p,2}^{(t)}\right)\right)
    &\leq \frac{\delta a_{p,1}^{(t)}+\delta a_{p,2}^{(t)}}{k-1}+\frac{1}{\ln k}\delta a_{p,1}^{(t)}+c\left(\frac{1}{k^2\ln k \exp\left(0.1a_{p,2}^{(t)}\right)}+\frac{1}{Nk^2\ln k }\right)\delta a_{q,\times}^{(t)}\notag\\
    &\overset{\eqref{eq:a_p3}}=\left(1+o(1)\right)\left(\frac{1}{\ln k}\delta a_{p,1}^{(t)}-\frac{a_{p,3}^{(t)}}{k^2}\right)+c\left(\frac{1}{k^2\ln k \exp\left(0.1a_{p,2}^{(t)}\right)}+\frac{1}{Nk^2\ln k }\right)\delta a_{q,\times}^{(t)}\notag\\
    &\leq \frac{1+o(1)}{\ln k}\delta a_{p,1}^{(t)}+c\left(\frac{1}{k^2\ln k \exp\left(0.1a_{p,2}^{(t)}\right)}+\frac{1}{Nk^2\ln k }\right)\delta a_{q,\times}^{(t)},
\end{align}
for some constant $c>0$, where the last relation follows from the fact that 
\begin{align}
    \forall t\in[T_{1,3},s-1]:\quad a_{p,3}^{(t)}>0
\end{align}
 guaranteed by \eqref{eq:delta_a_p_3_simplified_balanced_phase_1.2} and \eqref{eq:sum_1-pi_2h_up>>sum_1-pi_2h+1_up}. 

With the above gradient bounds, we next show \eqref{eq:a_q_x>a_p_2_bound_T_13}, \eqref{eq:a_q_2>a_q_x_bound_T_13}, \eqref{eq:a_p2>a_q_0_bound_T_13}, \eqref{eq:a_q_0>a_q_x_bound_T_13} hold at step $s$ when $s-1\geq T_{1,3}$. 
We first show that \eqref{eq:a_q_2>a_q_x_bound_T_13} holds at step $s$. Suppose otherwise, i.e., we have
\begin{align}\label{eq:a_q_2<a_q_x_t_1_contradiction}
    r_{\down_\times}^{(s)}\left(a_{p,1}^{(s)}+a_{p,2}^{(s)}\right)+0.1\sqrt{a_{p,1}^{(s)}/k}+5\ln k &<\min\left\{a_{q,\times}^{(s)}-\mu_2^{(s)},a_{q,\times}^{(s)}-\mu_{2(k-1)}^{(s)}\right\}.
\end{align}
By Induction Hypothesis II we have
\begin{align}\label{eq:a_q_2<a_q_x_t_1-1_contradiction}
   r_{\down_\times}^{(s-1)}\left(a_{p,1}^{(s-1)}+a_{p,2}^{(s-1)}\right)+0.1\sqrt{a_{p,1}^{(s-1)}/k}+5\ln k
    &\geq \min\left\{a_{q,\times}^{(s-1)}-\mu_2^{(s-1)},a_{q,\times}^{(s-1)}-\mu_{2(k-1)}^{(s-1)}\right\}.
\end{align}
Meanwhile, by our gradient update rule and \eqref{eq:a_q_2<a_q_x_t_1_contradiction}, we have
\begin{align}
    r_{\down_\times}^{(s-1)}\left(a_{p,1}^{(s-1)}+a_{p,2}^{(s-1)}\right)+0.1\sqrt{a_{p,1}^{(s-1)}/k}+5\ln k
    &\leq \min\left\{a_{q,\times}^{(s-1)}-\mu_2^{(s-1)},a_{q,\times}^{(s-1)}-\mu_{2(k-1)}^{(s-1)}\right\}+1,
\end{align}
and thus similar as \eqref{eq:delta_a_p_1_contradiction_T_13}, we can compute that
\begin{align}\label{eq:delta_a_p_1_t_1-1_contradiction}
    \delta a_{p,1}^{(s-1)}\gtrsim k\exp\left(0.1\sqrt{a_{p,1}^{(s-1)}/k}\right)\delta a_{q,\times}^{(s-1)}.
\end{align}
Then we have 
\begin{align}
    &r_{\down_\times}^{(s)}\left(a_{p,1}^{(s)}+a_{p,2}^{(s)}\right)+0.1\sqrt{a_{p,1}^{(s)}/k}+5\ln k - \min\left\{a_{q,\times}^{(s)}-\mu_2^{(s)},a_{q,\times}^{(s)}-\mu_{2(k-1)}^{(s)}\right\}\notag\\
    &\overset{\eqref{eq:a_q_2<a_q_x_t_1-1_contradiction}}\geq \delta\left(r_{\down_\times}^{(s-1)}\left(a_{p,1}^{(s-1)}+a_{p,2}^{(s-1)}\right)\right)-\delta a_{q,\times}^{(s-1)}+\min\{\delta \mu_2^{(s)},\delta \mu_{2(k-1)}^{(s)}\}\notag\\
    &\geq \left(1+o(1)\right)\left(\frac{1}{2k}\delta a_{p,1}^{(s-1)}-\frac{c_2}{k^2\ln k}\delta a_{q,\times}^{(s-1)}\right)\notag\\
    &\quad-\delta a_{q,\times}^{(s-1)}-(1+o(1))\left(\frac{1}{k}+\frac{\left(a_{p,2}^{(s-1)}\right)^2}{16N}\right)\delta a_{q,\times}^{(s-1)}+(1+o(1))\frac{\ln k}{k^2}\delta a_{p,1}^{(s-1)}\notag\\
    &\geq (1+o(1))\frac{1}{2k}\delta a_{p,1}^{(s-1)}-(1+o(1))\left(1+\frac{\left(a_{p,2}^{(s-1)}\right)^2}{16N}\right)\delta a_{q,\times}^{(s-1)}\notag\\
    &\overset{\eqref{eq:delta_a_p_1_t_1-1_contradiction}}\gtrsim \exp\left(0.1\sqrt{a_{p,1}^{(s-1)}/k}\right)\delta a_{q,\times}^{(s-1)}>0,
\end{align}
where the third line follows from \eqref{eq:-delta_mu_2h_ub} and \eqref{eq:delta_r_down_x_lb_simplified}. This contradicts \eqref{eq:a_q_2<a_q_x_t_1_contradiction}. 
Therefore, we have shown that \eqref{eq:a_q_2>a_q_x_bound_T_13} holds at step $s$.

We now show \eqref{eq:a_q_0>a_q_x_bound_T_13} holds at step $s$. Suppose otherwise, i.e.,
\begin{align}\label{eq:a_q_0<a_q_x_bound_T_13_contradiction_2}
    \left(1+\frac{1}{k}\right)a_{q,0}^{(s)}-\frac{a_{p,0}^{(s)}}{k}+0.1\sqrt{a_{p,1}^{(s)}/k}+3\ln k<  \min\left\{a_{q,\times}^{(s)}-\mu_2^{(s)},a_{q,\times}^{(s)}-\mu_{2(k-1)}^{(s)}\right\}.
\end{align}
By Induction Hypothesis II we have
\begin{align}\label{eq:a_q_0<a_q_x_bound_T_13_contradiction_2_2}
    \left(1+\frac{1}{k}\right)a_{q,0}^{(s-1)}-\frac{a_{p,0}^{(s-1)}}{k}+0.1\sqrt{a_{p,1}^{(s-1)}/k}+3\ln k
    &\geq \min\left\{a_{q,\times}^{(s-1)}-\mu_2^{(s-1)},a_{q,\times}^{(s-1)}-\mu_{2(k-1)}^{(s-1)}\right\}.
\end{align}
Meanwhile, by our gradient update rule and \eqref{eq:a_q_0<a_q_x_bound_T_13_contradiction_2}, we have
\begin{align}
    \left(1+\frac{1}{k}\right)a_{q,0}^{(s)}-\frac{a_{p,0}^{(s)}}{k}+0.1\sqrt{a_{p,1}^{(s)}/k}+3\ln k
    \leq \min\left\{a_{q,\times}^{(s)}-\mu_2^{(s)},a_{q,\times}^{(s)}-\mu_{2(k-1)}^{(s)}\right\}+1,
\end{align}
Then similar as how we obtain \eqref{eq:delta_a_q_0_T_13_contradiction}, from the above relation and \eqref{eq:delta_a_q_0_simplified_balanced_phase_1.2}, we can compute that
\begin{align}
    \delta a_{q,0}^{(s-1)}& \gtrsim \exp\left(0.1\sqrt{a_{p,1}^{(s-1)}/k}\right)\delta a_{q,\times}^{(s-1)}. \label{eq:delta_a_q_0_s-1_contradiction} \\
    -\delta a_{p,0}^{(s-1)} & =(1+o(1))\delta a_{q,0}^{(s-1)}. \label{eq:-delta_a_p_0_s-1_contradiction}
\end{align}
On the other hand, by \eqref{eq:-delta_mu_2h_ub} we know that
\begin{align}\label{eq:-delta_mu_2h_s-1_contradiction}
    -\delta \mu_{2h}^{(s-1)}\leq (1+o(1))\left(\frac{1}{k}+\frac{\left(a_{p,2}^{(s)}\right)^2}{16N}\right)\delta a_{q,\times}^{(s)},
\end{align}
and thus we have
\begin{align}
    &\left(1+\frac{1}{k}\right)a_{q,0}^{(s)}-\frac{a_{p,0}^{(s)}}{k}+0.1\sqrt{a_{p,1}^{(s)}/k}+3\ln k-\min\left\{a_{q,\times}^{(s)}-\mu_2^{(s)},a_{q,\times}^{(s)}-\mu_{2(k-1)}^{(s)}\right\}\notag\\
    &\overset{\eqref{eq:a_q_0<a_q_x_bound_T_13_contradiction_2_2}}\geq \left(1+\frac{1}{k}\right)\delta a_{q,0}^{(s-1)}-\frac{\delta a_{p,0}^{(s-1)}}{k}-\delta a_{q,\times}^{(s-1)}+\min\{\delta \mu_2^{(s-1)},\delta \mu_{2(k-1)}^{(s-1)}\}\notag\\
    &\gtrsim \exp\left(0.1\sqrt{a_{p,1}^{(s-1)}/k}\right)\delta a_{q,\times}^{(s-1)}>0,
\end{align}
where the last relation follows from \eqref{eq:delta_a_q_0_s-1_contradiction}, \eqref{eq:-delta_a_p_0_s-1_contradiction} and \eqref{eq:-delta_mu_2h_s-1_contradiction}. This contradicts \eqref{eq:a_q_0<a_q_x_bound_T_13_contradiction_2}. Therefore, \eqref{eq:a_q_0>a_q_x_bound_T_13} holds at step $s$.

We next show \eqref{eq:a_p2>a_q_0_bound_T_13} holds at step $s$. We first give a new lower bound of $\delta \left(r_{\down_\times}^{(t)}\left(a_{p,1}^{(t)}+a_{p,2}^{(t)}\right)\right)$ for all $t\in [T_{1,3},s-1]$. Similar as how we obtain \eqref{eq:delta_a_b_0_bound_phase1.2}, by \eqref{eq:I_h_0<T_14} we can get a different upper bound for $\delta a_{b,0}^{(t)}$ for all $t\in [T_{1/2},s-1]$:
\begin{align}\label{eq:delta_a_b_0_bound_phase1.2_new}
    \delta a_{b,0}^{(t)}&\leq \frac{1+o(1)}{N}a_{p,2}^{(t)}\sum_{h=1}^{k-1}r_{\up,h,0}^{(t)}\frac{k-h}{k}\left(1-\pi_{1,2h}^{(t)}(\up)\right),
\end{align}
and by \eqref{eq:delta_a_p_3_simplified_balanced_phase_1.2} we have 
\begin{align}\label{eq:delta_a_p_3_bound_phase1.2_new}
    \delta a_{p,3}^{(t)}&\leq \frac{1+o(1)}{k}\sum_{h=1}^{k-1}\frac{k-h}{k}\left(1-\pi_{1,2h}^{(t)}(\up)\right)\lesssim k^{1/2}  \sum_{h=1}^{k-1}\frac{k-h}{k}r_{\up,h,0}^{(t)}\left(1-\pi_{1,2h}^{(t)}(\up)\right),
\end{align}
where we use the fact that 
$r_{\up,h,0}^{(t)}\lesssim k^{3/2}$
guaranteed by \eqref{eq:e2a_b_0_lb_T_14} and \eqref{eq:a_b_2_approx_0_t<=T_1_4}.
Then similar as how we obtain \eqref{eq:delta_r_down_x_lb_simplified}, by the new bounds \eqref{eq:delta_a_b_0_bound_phase1.2_new} and \eqref{eq:delta_a_p_3_bound_phase1.2_new}, we can compute that for all $t\in [T_{1,3},s-1]$:
\begin{align}\label{eq:delta_r_down_x_lb_phase1.2_new}
    \delta \left(r_{\down_\times}^{(t)}\left(a_{p,1}^{(t)}+a_{p,2}^{(t)}\right)\right)\geq \left(1+o(1)\right)\left(\frac{1}{2k}\delta a_{p,1}^{(t)}-\frac{c_3}{k^{3/2}}\sum_{h=1}^{k-1}\frac{k-h}{k}r_{\up,h,0}^{(t)}\left(1-\pi_{1,2h}^{(t)}(\up)\right)\right)
\end{align}
for some constant $c_3>0$.  Now we show that \eqref{eq:a_p2>a_q_0_bound_T_13} holds at step $s$ by contradiction. Suppose otherwise, i.e.,
\begin{align}\label{eq:a_p2>a_q_0_bound_T_13_contradiction_1_2}
    r_{\down_\times}^{(s)}\left(a_{p,1}^{(s)}+a_{p,2}^{(s)}\right)+2.1\ln k < \left(1+\frac{1}{k}\right)a_{q,0}^{(s)}-\frac{a_{p,0}^{(s)}}{k}.
\end{align}
By Induction Hypothesis II we have
\begin{align}\label{eq:a_p2>a_q_0_bound_T_13_contradiction_1_3}
    r_{\down_\times}^{(s-1)}\left(a_{p,1}^{(s-1)}+a_{p,2}^{(s-1)}\right)+2.1\ln k \geq \left(1+\frac{1}{k}\right)a_{q,0}^{(s-1)}-\frac{a_{p,0}^{(s-1)}}{k}.
\end{align}
Meanwhile, by our update rule and \eqref{eq:a_p2>a_q_0_bound_T_13_contradiction_1_2}, we have
\begin{align*}
    r_{\down_\times}^{(s-1)}\left(a_{p,1}^{(s-1)}+a_{p,2}^{(s-1)}\right)+2.1\ln k \leq \left(1+\frac{1}{k}\right)a_{q,0}^{(s-1)}-\frac{a_{p,0}^{(s-1)}}{k}+1,
\end{align*}
and similar as how we obtain \eqref{eq:delta_a_p_1_bound_T_13_contradiction_1}, we have
\begin{align}\label{eq:delta_a_p_1_bound_T_13_contradiction_1_2}
    \delta a_{p,1}^{(s-1)}\gtrsim k^{1.1}\pi_{1,1}^{(s-1)}(\up) \gtrsim k^{1.1}\delta a_{q,0}^{(s-1)}.
\end{align}
Therefore, we have
\begin{align}
    &r_{\down_\times}^{(s)}\left(a_{p,1}^{(s)}+a_{p,2}^{(s)}\right)+2.1\ln k - \left(\left(1+\frac{1}{k}\right)a_{q,0}^{(s)}-\frac{a_{p,0}^{(s)}}{k}\right)\notag\\
    &\overset{\eqref{eq:a_p2>a_q_0_bound_T_13_contradiction_1_3}}\geq \delta\left(r_{\down_\times}^{(s-1)}\left(a_{p,1}^{(s-1)}+a_{p,2}^{(s-1)}\right)\right)-\left(1+\frac{1}{k}\right)\delta a_{q,0}^{(s-1)}+\frac{\delta a_{p,0}^{(s-1)}}{k}\notag\\
    &\geq \left(1+o(1)\right)\left(\frac{1}{2k}\delta a_{p,1}^{(s-1)}-\frac{c_3}{k^{3/2}}\sum_{h=1}^{k-1}\frac{k-h}{k}r_{\up,h,0}^{(s-1)}\left(1-\pi_{1,2h}^{(s-1)}(\up)\right)\right)-\left(1+\frac{1}{k}\right)\delta a_{q,0}^{(s-1)}\notag\\
    &\quad+\frac{1+o(1)}{k}\left(-\pi_{1,1}^{(s-1)}(\up)+\sum_{h=1}^{k-1}\frac{k-h}{k}r_{\up,h,0}^{(s-1)}\left(1-\pi_{1,2h}^{(s-1)}(\up)\right)\right)\notag\\
    &\overset{\eqref{eq:delta_a_p_1_bound_T_13_contradiction_1_2}}\gtrsim k^{0.1}\delta a_{q,0}^{(s-1)}>0,
\end{align}
where the third line follows from \eqref{eq:delta_r_down_x_lb_phase1.2_new}, \eqref{eq:delta_a_p_0_simplified_balanced_phase_1.2} and \eqref{eq:pi_1_up_dominate_sum_phase1.2}. This contradicts \eqref{eq:a_p2>a_q_0_bound_T_13_contradiction_1_2}, and thus \eqref{eq:a_p2>a_q_0_bound_T_13} holds at step $s$.

We now show \eqref{eq:a_q_x>a_p_2_bound_T_13} holds at step $s$ by contradiction. Suppose otherwise, i.e., 
\begin{align}\label{eq:a_q_x>a_p_2_bound_contradiction}
    \min\left\{a_{q,\times}^{(s)}-\mu_2^{(s)}-\ln k,a_{q,\times}^{(s)}-\mu_{2(k-1)}^{(s)}+\ln k\right\}+\frac{2}{k}a_{q,0}^{(s)}< r_{\down_\times}^{(s)}\left(a_{p,1}^{(s)}+a_{p,2}^{(s)}\right)+\ln k,
\end{align}
By Induction Hypothesis II we have
\begin{align}\label{eq:a_q_x>a_p_2_bound_contradiction_2}
    \min\left\{a_{q,\times}^{(s-1)}-\mu_2^{(s-1)}-\ln k,a_{q,\times}^{(s-1)}-\mu_{2(k-1)}^{(s-1)}+\ln k\right\}+\frac{2}{k}a_{q,0}^{(s-1)}\geq r_{\down_\times}^{(s-1)}\left(a_{p,1}^{(s-1)}+a_{p,2}^{(s-1)}\right)+\ln k,
\end{align}
Meanwhile, by our gradient update rule and \eqref{eq:a_q_x>a_p_2_bound_contradiction}, we have
\begin{align}
    \min\left\{a_{q,\times}^{(s)}-\mu_2^{(s)}-\ln k,a_{q,\times}^{(s)}-\mu_{2(k-1)}^{(s)}+\ln k\right\}+\frac{2}{k}a_{q,0}^{(s)}\leq r_{\down_\times}^{(s)}\left(a_{p,1}^{(s)}+a_{p,2}^{(s)}\right)+\ln k+1.
\end{align}
Then similar as how we prove \eqref{eq:delta_a_p_1_bound_T_13_contradiction_2}, we can compute that 
\begin{align}\label{eq:delta_a_p_1_bound_contradiction_2}
    \delta a_{p,1}^{(s-1)}\lesssim \delta a_{q,\times}^{(s-1)}.
\end{align}
Then we have
\begin{align}
    &\min\left\{a_{q,\times}^{(s)}-\mu_2^{(s)}-\ln k,a_{q,\times}^{(s)}-\mu_{2(k-1)}^{(s)}+\ln k\right\}+\frac{2}{k}a_{q,0}^{(s)}-\left( r_{\down_\times}^{(s)}\left(a_{p,1}^{(s)}+a_{p,2}^{(s)}\right)+\ln k\right)\notag\\
    &\overset{\eqref{eq:a_q_x>a_p_2_bound_contradiction_2}}\geq \delta a_{q,\times}^{(s-1)}-\max\{\mu_2^{(s-1)},\mu_{2(k-1)}^{(s-1)}\}+\frac{2}{k}\delta a_{q,0}^{(s-1)}-\delta\left(r_{\down_\times}^{(s-1)}\left(a_{p,1}^{(s-1)}+a_{p,2}^{(s-1)}\right)\right)\notag\\
    &\geq \left(1+o(1)\right)\delta a_{q,\times}^{(s-1)}-\left(1+o(1)\right)\left(\frac{c}{\ln k}\delta a_{p,1}^{(s-1)}+\frac{1}{k}\pi_{1,1}^{(s-1)}(\up)\right)+\frac{2+o(1)}{k}\pi_{1,1}^{(s-1)}(\up)-\frac{1+o(1)}{\ln k}\delta a_{p,1}^{(s-1)}\notag\\
    &\overset{\eqref{eq:delta_a_p_1_bound_contradiction_2}}\gtrsim \delta a_{q,\times}^{(s-1)}>0,
\end{align}
where the third line follows from \eqref{eq:delta_a_q_0_simplified_balanced_phase_1.2}, \eqref{eq:delta_r_down_x(a_p1+a_p2)_ub} and \eqref{eq:delta_mu_2h_ub}. This contradicts \eqref{eq:a_q_x>a_p_2_bound_contradiction}. Therefore, \eqref{eq:a_q_x>a_p_2_bound_T_13} must hold at step $s$.

Now we have shown that  \eqref{eq:a_q_x>a_p_2_bound_T_13}, \eqref{eq:a_q_2>a_q_x_bound_T_13}, \eqref{eq:a_p2>a_q_0_bound_T_13}, \eqref{eq:a_q_0>a_q_x_bound_T_13} and \eqref{eq:a_p_0_bound_phase_1.2} hold at step $s$. They together guarantee that \eqref{eq:a_q_x_q_0_p_2_asymp_phase_1.2} holds at step $s$. 

By \eqref{eq:a_p_1_increase_rate_T_13-T_14}, \eqref{eq:a_p_1_ub_T_1_4}, \eqref{eq:a_p_1_bound_<T_1_3} and \eqref{eq:T_1/2_bound} we have 
\begin{align}\label{eq:T_1_4_bound}
    T_{1,4}\lesssim \frac{k^{9/4}N\exp\left( k^{1.3}\sqrt{N}\right)}{\eta}.
\end{align}

\paragraph{When $T_{1,4}+1\leq s\leq T_{1}$.} 
By Induction Hypothesis II we have \eqref{eq:ea_b_1>=ea_b_0_T_15}, \eqref{eq:e_ab_2_lb>T14}, \eqref{eq:ea_b_1_bound_T_15} hold for all $t\in[T_{1,4},s-1]$. Combining them with \eqref{eq:r_down_times_phase1.2}, we have
    \begin{align}\label{eq:r_down_times_bound_phase1.2_pre}
        \forall t\in[T_{1,4},s-1]:\quad \frac{1}{k-1}-\frac{c_1}{k^4}\leq r_{\down_\times}^{(t)}\leq \frac{1}{k-1}
    \end{align}
for some constant $c_1>0$, and
\begin{align}
    \max_{h\in[k-1]}\{\mu_{2h}^{(t)}\}\leq a_{p,2}^{(t)}.
\end{align}
Then use a similar argument as how we show \eqref{eq:a_q_x>a_p_2_bound_T_13}, 
\eqref{eq:a_q_2>a_q_x_bound_T_13} and \eqref{eq:a_q_0>a_q_x_bound_T_13}, we can show that \eqref{eq:a_q_x>a_p_2_bound_T_15}, 
\eqref{eq:a_p2>a_q_0_bound_T_15} and \eqref{eq:a_q_0>a_q_x_bound_T_15} hold at step $s$. And from these three expressions we know that \eqref{eq:a_q_x_q_0_p_2_asymp_phase_1.2} holds at step $s$.
From \eqref{eq:a_p_1_lb_s=T_14}, \eqref{eq:ea_b_1_lb_s=T_14} and the fact that $a_{p,1}^{(t)}, a_{b,1}^{(t)}$ monotonically increase we immediately know that \eqref{eq:a_p_1_lb_T_1_5} and \eqref{eq:ea_b_1_bound_T_15} hold at step $s$.
Then by the same argument as in \eqref{eq:sum_1-pi_2h_up>>sum_1-pi_2h+1_up}, from  \eqref{eq:a_q_0>a_q_x_bound_T_15} we deduce \eqref{eq:delta_a_q_x_simplified_T_13-T_14} still holds here, i.e.,
\begin{align}\label{eq:delta_a_q_x_simplified_T_14-T_15}
    \delta a_{q,\times}^{(s-1)}=(1+o(1))\sum_{h=1}^{k-1}\frac{k-h}{k}\left(1-\pi_{1,2h}^{(s-1)}(\up)\right),
\end{align}
And \eqref{eq:sum_1-pi_2h+1_up_vs_sum_1-pi_2h_up}, \eqref{eq:delta_e2a_c_0_simplified_balanced_ub_T_13-T_14} also still holds here, i.e., for all $t\in [T_{1,4},s-1]$:
\begin{align}\label{eq:delta_e2a_c_0_simplified_balanced_ub_T_14-T_15}
    \delta e^{2a_{c,0}^{(t)}}\lesssim \frac{a_{p,2}^{(t)}}{\exp\left(0.1a_{p,2}^{(t)}\right)}\delta a_{q,\times}^{(t)}.
\end{align}
From \eqref{eq:delta_e2a_c_0_simplified_balanced_ub_T_14-T_15}, \eqref{eq:a_p_1_lb_T_1_5}, \eqref{eq:a_p_2_bound_phase_1.2} and \eqref{eq:a_q_x_q_0_p_2_asymp_phase_1.2} we know that
\begin{align}\label{eq:e2ac_0<=ln^2k_T_14-T_15}
    e^{2a_{c,0}^{(s)}}-e^{2a_{c,0}^{(T_{1,4})}}\lesssim \sum_{i=0}^\infty \frac{a_{p,1}^{(T_{1,4})}k^{i-1}}{\exp\left(0.1a_{p,1}^{(T_{1,4})}k^{i-1}\right)}a_{p,1}^{(T_{1,4})}k^i\lesssim \frac{\left(a_{p,1}^{(T_{1,4})}\right)^2/k}{\exp\left(0.1a_{p,1}^{(T_{1,4})}/k\right)}=o(1).
\end{align}
Therefore, from \eqref{eq:e2ac_0<=ln^2k_T_14-T_15}, \eqref{eq:a_c_0_increase_phase1.2} \eqref{eq:e2ac_0<=ln^2k_T_13-T_14} we know that
\eqref{eq:e2a_c_0_balanced_phase_1.2} holds at step $s$.

And similar as how we obtain \eqref{eq:I_h_0<T_14}, from \eqref{eq:I_h_0_balanced_phase1.2} and Induction Hypothesis II we deduce for all $t\in [T_{1,4},s-1]$:
\begin{align}\label{eq:I_h_0>T_14}
    I_{h,0}^{(t)}&\coloneqq \left(1+O\left(\frac{1}{k}\right)\right)r_{\up,h,0}^{(t)}r_{\up,h,2}^{(t)}ha_{p,2}^{(t)}\left(1-\pi_{1,2h}^{(t)}(\up)\right)\notag\\
    &\quad \quad-\left(1+O\left(\frac{1}{k}\right)\right)r_{\up,h,0}^{(t)}r_{\up,h,2}^{(t)}ha_{p,1}^{(t)}\pi_{1,2h}^{(t)}(\down_\times^1)\notag\\
    &\quad \quad-\left(1+O\left(\frac{1}{k}\right)\right)h
    r_{\down,h,0}^{(t)}r_{\down,h,1}^{(t)}\left(a_{p,1}^{(t)}\pi_{1,2h+1}^{(t)}(\down_\times)+a_{p,2}^{(t)}\pi_{1,2h+1}^{(t)}(\up)\right).
\end{align}
Also note that \eqref{eq:I_h_1_balanced_phase1.2_simplified} still holds here, i.e., we have for all $t\in [T_{1,4},s-1]$:
\begin{align}\label{eq:I_h_1_balanced>T_14}
    I_{h,1}^{(t)}&= \left(1+O\left(\frac{1}{k}\right)\right)\left(1-hr_{\down,h,1}^{(t)}\right)\notag\\
    &\quad\quad \cdot\left(ka_{p,2}^{(t)}\frac{he^{a_{b,1}^{(t)}}}{e^{a_{b,0}^{(t)}}+he^{a_{b,1}^{(t)}}+he^{a_{b,2}^{(t)}}}\pi_{1,2h+1}^{(t)}(\down_\times)+hr_{\down,h,1}^{(t)}a_{p,2}^{(t)}\pi_{1,2h+1}^{(t)}(\up)\right).
\end{align}

Consider
\begin{align}\label{eq:delta_e_1+1/N_a_b_i}
    \delta e^{\frac{N}{N-1}a_{b,i}^{(t)}}\coloneqq \frac{e^{\frac{N}{N-1}a_{b,i}^{(t+1)}}-e^{\frac{N}{N-1}a_{b,i}^{(t)}}}{\eta}=(1+o(1))e^{\frac{N}{N-1}a_{b,i}^{(t)}}\delta a_{b,i}^{(t)}.
\end{align}
Then by \eqref{eq:e_ab_2_lb>T14}, \eqref{eq:I_h_0>T_14}, \eqref{eq:delta_a_q_x_simplified_T_14-T_15} and \eqref{eq:A_M_sum0} we have for all $t\in [T_{1,4},s-1]$:
\begin{align}
    \delta e^{\frac{N}{N-1}a_{b,0}^{(t)}}\leq (1+o(1))\frac{a_{p,1}^{(t)}}{N}\delta a_{q,\times}^{(t)}.
\end{align}
And same as how we obtain \eqref{eq:e_a_b_0_bound_T_1_4}, the above expression suggests
\begin{align}\label{eq:e_a_b_0_bound_T_15}
    e^{\frac{N}{N-1}a_{b,0}^{(t)}}\lesssim \frac{1}{Nk}\left(a_{p,1}^{(t)}\right)^2.
\end{align}
Moreover, similar as how we obtain \eqref{eq:delta_e-a_b_0<T_14}, by \eqref{eq:I_h_0>T_14} we have for all $t\in [T_{1,4},s-1]$:
\small
\begin{align}\label{eq:delta_e-a_b_0>T_14}
    \delta e^{-a_{b,0}^{(t)}}&=
        \frac{1+o(1)}{N}\sum_{h=1}^{k-1}\frac{k-h}{k}\Bigg[\frac{he^{a_{b,1}^{(t)}}}{\left(e^{a_{b,0}^{(t)}}+he^{a_{b,1}^{(t)}}+he^{a_{b,2}^{(t)}}\right)^2}\left(a_{p,1}^{(t)}\pi_{1,2h+1}^{(t)}(\down_\times)+a_{p,2}^{(t)}\pi_{1,2h+1}^{(t)}(\up)\right)\notag\\
    &\hspace{5.0cm}+\frac{he^{a_{b,2}^{(t)}}}{\left(e^{a_{b,0}^{(t)}}+(2h-1)e^{a_{b,2}^{(t)}}\right)^2}a_{p,1}^{(t)}\pi_{1,2h}^{(t)}(\down_\times^1)\Bigg]\notag\\
    &\qquad -\frac{1+o(1)}{N}\sum_{h=1}^{k-1}\frac{k-h}{k}\frac{he^{a_{b,2}^{(t)}}}{\left(e^{a_{b,0}^{(t)}}+(2h-1)e^{a_{b,2}^{(t)}}\right)^2}a_{p,2}^{(t)}\left(1-\pi_{1,2h}^{(t)}(\up)\right).
\end{align}
\normalsize

Similar as how we show \eqref{eq:e_ab1_I_h_1_lb_T_13-T_14}, from \eqref{eq:I_h_1_balanced>T_14} we can compute that for all $t\in [T_{1,4},s-1]$ and $h\in[k-1]$:
\begin{align}\label{eq:e_ab1_I_h_1_lb>T_14}
    e^{\frac{N}{N-1}a_{b,1}^{(t)}}I_{h,1}^{(t)} 
    &\overset{\eqref{eq:A_M_sum0}}\geq \left(1+O\left(\frac{1}{k}\right)\right)\frac{e^{a_{b,1}^{(t)}}\left(e^{a_{b,0}^{(t)}+\frac{1}{N-1}a_{b,1}^{(t)}}+he^{\frac{-a_{b,0}^{(t)}}{N-1}}\right)}{e^{a_{b,0}^{(t)}}+he^{a_{b,1}^{(t)}}+he^{a_{b,2}^{(t)}}}a_{p,1}^{(t)}\cdot\frac{he^{a_{b,1}^{(t)}}}{e^{a_{b,0}^{(t)}}+he^{a_{b,1}^{(t)}}+he^{a_{b,2}^{(t)}}}\pi_{1,2h+1}^{(t)}(\down_\times)\notag\\
    &\gtrsim\frac{he^{a_{b,1}^{(t)}}}{e^{a_{b,0}^{(t)}}+he^{a_{b,1}^{(t)}}+he^{a_{b,2}^{(t)}}}a_{p,1}^{(t)}\cdot\frac{he^{a_{b,1}^{(t)}}}{e^{a_{b,0}^{(t)}}+he^{a_{b,1}^{(t)}}+he^{a_{b,2}^{(t)}}}\pi_{1,2h+1}^{(t)}(\down_\times)\notag\\
    &\gtrsim a_{p,1}^{(t)}\Bigg(\frac{he^{a_{b,2}^{(t)}}}{e^{a_{b,0}^{(t)}}+(2h-1)e^{a_{b,2}^{(t)}}}\pi_{1,2h}^{(t)}(\down_\times^1)+\frac{he^{a_{b,1}^{(t)}}}{e^{a_{b,0}^{(t)}}+he^{a_{b,1}^{(t)}}+he^{a_{b,2}^{(t)}}}\pi_{1,2h+1}^{(t)}(\down_\times)\notag\\
    &\hspace{5.8cm}-\frac{he^{a_{b,2}^{(t)}}}{e^{a_{b,0}^{(t)}}+(2h-1)e^{a_{b,2}^{(t)}}}\pi_{1,2h}^{(t)}(\down_\times^1)\Bigg),
\end{align}
where the second relation uses the following fact
\begin{align}
    e^{a_{b,0}^{(t)}+\frac{1}{N-1}a_{b,1}^{(t)}}+he^{\frac{-a_{b,0}^{(t)}}{N-1}}\gtrsim h.
\end{align}
Combining \eqref{eq:delta_e_1+1/N_a_b_i}, \eqref{eq:e_ab1_I_h_1_lb>T_14} and \eqref{eq:pi_2h_down_1_bound_phase1.2}, we obtain for all $t\in [T_{1,4},s-1]$
\begin{align}\label{eq:delta_ea_b_1_lb<=T_1_5}
    \delta e^{\frac{N}{N-1}a_{b,1}^{(t)}}\overset{\eqref{eq:delta_a_q_x_simplified_T_14-T_15}}\gtrsim \frac{a_{p,1}^{(t)}}{N}\left(k\delta a_{p,1}^{(t)}-\frac{1}{2k}\delta a_{q,\times}^{(t)}\right).
\end{align}
And similar as how we obtain \eqref{eq:e_ab_1_lb_intermediate}, from the above expression we deduce
\begin{align}\label{eq:e_ab_1_lb_intermediate>T_14}
    e^{\frac{N}{N-1}a_{b,1}^{(s)}}&\gtrsim \frac{k}{N}\left(a_{p,1}^{(s)}\right)^2\overset{\eqref{eq:e_a_b_0_bound_T_15}}\gtrsim k^2e^{\frac{N}{N-1}a_{b,0}^{(s)}},
\end{align}
from which we know that \eqref{eq:ea_b_1>=ea_b_0_T_15} holds at step $s$.

Meanwhile, from \eqref{eq:I_h_1_balanced>T_14}, \eqref{eq:ea_b_1>=ea_b_0_T_15} and \eqref{eq:e_ab_2_lb>T14} we deduce for all $t\in[T_{1,4},s-1]$:
\small
\begin{align}
    \delta e^{a_{b,1}^{(t)}}&\leq \frac{1+o(1)}{N}\sum_{h=1}^{k-1}\frac{k-h}{k}\left(e^{a_{b,0}^{(t)}}+he^{a_{b,2}^{(t)}}\right)\Bigg(ka_{p,2}^{(t)}\frac{he^{a_{b,1}^{(t)}}}{e^{a_{b,0}^{(t)}}+he^{a_{b,1}^{(t)}}+he^{a_{b,2}^{(t)}}}\pi_{1,2h+1}^{(t)}(\down_\times)+hr_{\down,h,1}^{(t)}a_{p,2}^{(t)}\pi_{1,2h+1}^{(t)}(\up)\Bigg)\notag\\
    &\lesssim\frac{1}{N} \frac{\left(a_{p,1}^{(t)}\right)^{\frac{2(N-1)}{N}}}{Nk}\left(ka_{p,1}^{(t)}\delta a_{p,1}^{(t)}+\frac{ka_{p,2}^{(t)}}{\exp\left(0.1a_{p,2}^{(t)}\right)}\delta a_{q,\times}^{(t)}\right)
\end{align}
\normalsize
where the second line also use \eqref{eq:delta_a_p_1_simplified_balanced_phase_1.2} and \eqref{eq:sum_1-pi_2h+1_up_vs_sum_1-pi_2h_up} (which also holds here). This together with \eqref{eq:ea_b_1_ub_T_14}, \eqref{eq:a_p_1_lb_T_1_5} suggests for all $t\in[T_{1,4},s]$: 
\begin{align}\label{eq:e_ab_1_ub_T_1_5}
    e^{a_{b,1}^{(t)}}\lesssim \frac{\left(a_{p,1}^{(t)}\right)^{4-\frac{2}{N}}}{N^2}.
\end{align}
Then by \eqref{eq:e_ab_1_ub_T_1_5}, \eqref{eq:ea_b_1>=ea_b_0_T_15} and \eqref{eq:A_M_sum0} we have 
\begin{align}\label{eq:e_ab_2_lb_T_1_5}
    \forall t\in[T_{1,4},s]:\quad e^{a_{b,2}^{(t)}}&\geq (1+o(1))\left(a_{p,1}^{(t)}\right)^{-\frac{6}{N-1}}.
\end{align}
This indicates the first relation of \eqref{eq:e_ab_2_lb>T14} holds at step $s$.

Define 
\begin{align}\label{eq:s_0}
    s_0\coloneqq \sup\left\{t\in[T_{1,4},T_1]: a_{p,1}^{(t)}\leq Nk^3\right\}.
\end{align}
Below we prove \eqref{eq:ea_b_0>=k/lnk_T_15} holds for $s\in [T_{1,4}+1,s_0]$, and \eqref{eq:a_b_0>=k^2_T_15} holds for $s\in [s_0,T_1]$. 
We first assume $s\in [T_{1,4}+1,s_0]$, and prove \eqref{eq:ea_b_0>=k/lnk_T_15} holds by showing 
\begin{align}\label{eq:ea_b_0>=Ck/lnk}
    \forall t\in[T_{1,4},s]: \quad e^{a_{b,0}^{(t)}}\geq \frac{Ck}{\ln k}
\end{align} 
for some small enough constant $C\in(0,1/2)$.
Suppose otherwise, then from our choice of $T_{1,4}$ we know that there exists the largest $t_0',t_0\in[T_{1,4}+1,s]$, $t_0'<t_0$ such that
\begin{align}\label{eq:e2a_b_0_t0_contradiction}
    e^{a_{b,0}^{(t_0')}}\geq 2C\frac{k}{\ln k},\quad e^{a_{b,0}^{(t_0)}}\leq C\frac{k}{\ln k},
\end{align}
and 
\begin{align}\label{eq:e2a_b_0_t0_contradiction_2}
    \forall t\in[t_0'+1,t_0-1]: \quad C\frac{k}{\ln k}<e^{a_{b,0}^{(t)}}<2C\frac{k}{\ln k}.
\end{align}
From the first line of \eqref{eq:pi_2h_down_1_bound_phase1.2}, \eqref{eq:e_ab_2_lb_T_1_5} \eqref{eq:e2a_b_0_t0_contradiction} and \eqref{eq:a_p_2_bound_phase_1.2} and \eqref{eq:a_p_1_lb_T_1_5} we know that for all $t\in[t_0',t_0]$ and $h\in[k-1]$:
\begin{align}\label{eq:pi_2h_down_1_bound_phase1.2_cond}
    \pi_{1,2h}^{(t)}(\down_\times^1)&\leq\left(1-\pi_{1,2h}^{(t)}(\up)\right)\exp\left(-r_{\up,h,2}^{(t)}\left(a_{p,1}^{(t)}+a_{p,2}^{(t)}\right)\right)= o\left(\frac{1}{k}\right)\left(1-\pi_{1,2h}^{(t)}(\up)\right).
\end{align}
Note that from \eqref{eq:e_ab_2_lb_T_1_5} and the fact that $s\leq s_0$ we know that 
\begin{align}\label{eq:e2a_b_2>=c}
    e^{a_{b,2}^{(t_0)}}\geq 1+o(1).
\end{align}
Then from \eqref{eq:I_h_0>T_14}, \eqref{eq:e2a_b_2>=c} and our choice of $t_0$ we deduce for all $t\in[t_0',t_0]$:
\begin{align}
    \delta e^{-a_{b,0}^{(t)}}
    &=-\frac{1+o(1)}{N}\sum_{h=1}^{k-1}\frac{k-h}{k} \frac{he^{a_{b,2}^{(t)}}}{\left(e^{a_{b,0}^{(t)}}+(2h-1)e^{a_{b,2}^{(t)}}\right)^2}a_{p,2}^{(t)}\left(1-\pi_{1,2h}^{(t)}(\up)\right)\notag\\
    &\quad +\frac{1+o(1)}{N}\sum_{h=1}^{k-1}\frac{k-h}{k}\frac{he^{a_{b,1}^{(t)}}}{\left(e^{a_{b,0}^{(t)}}+he^{a_{b,1}^{(t)}}+he^{a_{b,2}^{(t)}}\right)^2}\left(a_{p,1}^{(t)}\pi_{1,2h+1}^{(t)}(\down_\times)+a_{p,2}^{(t)}\pi_{1,2h+1}^{(t)}(\up)\right)\notag\\
    &\leq -\frac{1+o(1)}{(2C)^2}\frac{\ln^2 k}{Nk^3}a_{p,1}^{(t)}\delta a_{q,\times}^{(t)}+\frac{1+o(1)}{N}\frac{1}{e^{a_{b,0}^{(t)}}+e^{a_{b,1}^{(t)}}+e^{a_{b,2}^{(t)}}}\left(ka_{p,1}^{(t)}\delta a_{p,1}^{(t)}+\frac{a_{p,2}^{(t)}}{\exp\left(0.1a_{p,2}^{(t)}\right)}\delta a_{q,\times}^{(t)}\right)\notag\\
    &\leq -\frac{1+o(1)}{(2C)^2}\frac{\ln^2 k}{Nk^3}a_{p,1}^{(t)}\delta a_{q,\times}^{(t)}+\frac{c_1+o(1)}{a_{p,1}^{(t)}}\delta a_{p,1}^{(t)}
\end{align}
for some constant $c_1>0$, where we use \eqref{eq:e2a_b_0_t0_contradiction_2}, \eqref{eq:delta_a_q_x_simplified_T_14-T_15}, \eqref{eq:delta_a_p_1_simplified_balanced_phase_1.2} and \eqref{eq:sum_1-pi_2h+1_up_vs_sum_1-pi_2h_up} (which also holds here) to obtain the second relation, and use \eqref{eq:ea_b_1>=ea_b_0_T_15} to obtain the third relation.
The above expression implies
\begin{align}\label{eq:e-_a_b_0_diff_T_15}
    e^{-a_{b,0}^{(t_0)}}-e^{-a_{b,0}^{(t_0')}}\leq-\frac{1+o(1)}{(2C)^2}\frac{\ln^2 k}{Nk^3}a_{p,1}^{(t_0')}\left(a_{q,\times}^{(t_0)}-a_{q,\times}^{(t_0')}\right)+\frac{c_1+o(1)}{a_{p,1}^{(t_0')}}\left(a_{p,1}^{(t_0)}-a_{p,1}^{(t_0')}\right).
\end{align}
From \eqref{eq:e2a_b_0_t0_contradiction_2}, \eqref{eq:mu_h_phase1.2}, \eqref{eq:a_q_x_q_0_p_2_asymp_phase_1.2}, \eqref{eq:a_p_0_bound_phase_1.2}, \eqref{eq:a_p_up_bound_phase_1.2} and \eqref{eq:a_p_2_bound_phase_1.2} we know that for all $t\in[t_0',t_0]$:
\begin{align}
    \max_{h\in[k-1]}\left\{\mu_{2h}^{(t)}\right\} & =\mu_{2(k-1)}^{(t_0)}, \label{eq:mu_2k-1=max_h_mu_2h} \\
    \mu_{2(k-1)}^{(t_0')}& =\left(1-\frac{C+o(1)}{\ln k}\right)\frac{a_{p,1}^{(t_0')}}{2k}, \label{eq:mu_2k-1_t0'}  \\ 
    \mu_{2(k-1)}^{(t_0)} & =\left(1-\frac{C+o(1)}{2\ln k}\right)\frac{a_{p,1}^{(t_0)}}{2k}. \label{eq:mu_2k-1_t0} 
\end{align}
Also note that by \eqref{eq:mu_2k-1=max_h_mu_2h}, \eqref{eq:a_q_x>a_p_2_bound_T_15}, \eqref{eq:a_p2>a_q_0_bound_T_15}, \eqref{eq:a_q_0>a_q_x_bound_T_15}, \eqref{eq:a_p_1_lb_T_1_5}, \eqref{eq:r_down_times_bound_phase1.2_pre}, \eqref{eq:a_q_x_q_0_p_2_asymp_phase_1.2}, \eqref{eq:a_p_0_bound_phase_1.2}, and \eqref{eq:a_p_2_bound_phase_1.2} we have for all $t\in[t_0',t_0]$:
\begin{align}\label{eq:a_q_x=mu_2k-1+1/k_a_p_1}
    a_{q,\times}^{(t)}=\mu_{2(k-1)}^{(t)}+\frac{1}{k}a_{p,1}^{(t)}+O\left(\frac{1}{k^2}\right)a_{p,1}^{(t)}.
\end{align}
Combining \eqref{eq:mu_2k-1_t0'}, \eqref{eq:mu_2k-1_t0}, \eqref{eq:a_q_x=mu_2k-1+1/k_a_p_1} we have
\begin{align}
    a_{q,\times}^{(t_0)}-a_{q,\times}^{(t_0')}&=\frac{3}{2k}\left(a_{p,1}^{(t_0)}-a_{p,1}^{(t_0')}\right)-\frac{C+o(1)}{4k\ln k }a_{p,1}^{(t_0)}+\frac{C+o(1)}{2k\ln k} a_{p,1}^{(t_0')}\notag\\
    &\geq \frac{3}{2k}\left(a_{p,1}^{(t_0)}-a_{p,1}^{(t_0')}\right)-\frac{C+o(1)}{4k\ln k }\left(a_{p,1}^{(t_0)}-a_{p,1}^{(t_0')}\right)=\frac{3+o(1)}{2k}\left(a_{p,1}^{(t_0)}-a_{p,1}^{(t_0')}\right).
\end{align}
Plugging this back into \eqref{eq:e-_a_b_0_diff_T_15}, we have 
\begin{align}\label{eq:e-_a_b_0_diff_T_15_bound}
    e^{-a_{b,0}^{(t_0)}}-e^{-a_{b,0}^{(t_0')}}&\leq -\frac{3+o(1)}{8C^2}\frac{\ln^2 k}{Nk^4}a_{p,1}^{(t_0')}\left(a_{p,1}^{(t_0)}-a_{p,1}^{(t_0')}\right)+\frac{c_1+o(1)}{a_{p,1}^{(t_0')}}\left(a_{p,1}^{(t_0)}-a_{p,1}^{(t_0')}\right)\notag\\
    &\overset{\eqref{eq:a_p_1_lb_T_1_5}}\leq \left(-\frac{3c_2+o(1)}{8C^2}+\frac{c_1+o(1)}{c_2}\right)\frac{\ln k}{\sqrt{N}k^2}\left(a_{p,1}^{(t_0)}-a_{p,1}^{(t_0')}\right)
\end{align}
for some constant $c_2>0$. From the above inequality we can see that we can set $C$ small enough such that 
\begin{align}
    e^{-a_{b,0}^{(t_0)}}-e^{-a_{b,0}^{(t_0')}}< 0.
\end{align}
However, from \eqref{eq:e2a_b_0_t0_contradiction} we have
\begin{align}
    e^{-a_{b,0}^{(t_0)}}-e^{-a_{b,0}^{(t_0')}}\geq \frac{\ln k}{2Ck}.
\end{align}
The above two inequalities are contradictory. Therefore,  \eqref{eq:ea_b_0>=Ck/lnk} must hold. This way we have shown \eqref{eq:ea_b_0>=k/lnk_T_15}.

Below we prove \eqref{eq:a_b_0>=k^2_T_15} assuming $s\geq s_0$. Define 
\begin{align}\label{eq:s_0'}
    s'=\sup\left\{t\in[T_{1,4},s-1]: a_{p,1}^{(t)}\leq C_1a_{p,1}^{(s)}\right\}
\end{align}
for some small enough constant $C_1>0$.
We first show there exists $t_1\in[ s',s]$ such that 
\begin{align}\label{eq:e2a_b_0_t1}
    e^{a_{b,0}^{(t_1)}}&\geq \frac{C\left(a_{p,1}^{(t_1)}\right)^{1-\frac{6}{N-1}}}{N k}
\end{align}
for some small enough constant $C>0$.
Suppose this does not hold, i.e., for all $t\in[s',s]$, we have
\begin{align}\label{eq:e2a_b_0_t1_contradiction}
    e^{a_{b,0}^{(t)}}< \frac{C\left(a_{p,1}^{(t)}\right)^{1-\frac{6}{N-1}}}{N k}.
\end{align}
Then same as for \eqref{eq:pi_2h_down_1_bound_phase1.2_cond}, we have for all $t\in[s',s]$ and $h\in[k-1]$:
\begin{align}
    \pi_{1,2h}^{(t)}(\down_\times^1)&\leq\left(1-\pi_{1,2h}^{(t)}(\up)\right)\exp\left(-r_{\up,h,2}^{(t)}\left(a_{p,1}^{(t)}+a_{p,2}^{(t)}\right)\right)= o\left(\frac{1}{k}\right)\left(1-\pi_{1,2h}^{(t)}(\up)\right).
\end{align}
Then by \eqref{eq:I_h_0>T_14} we have for all $t\in[s_0',s_0]$:
\begin{align}\label{eq:delta_e_a_b_0_s_0_contradiction}
    \delta e^{a_{b,0}^{(t)}}&=\frac{1+o(1)}{N}\sum_{h=1}^{k-1}\frac{k-h}{k}\left(r_{\up,h,0}^{(t)}\right)^2 he^{a_{b,2}^{(t)}}a_{p,2}^{(t)}\left(1-\pi_{1,2h}^{(t)}(\up)\right)\notag\\
    &\quad -\frac{1+o(1)}{N}\sum_{h=1}^{k-1}\frac{k-h}{k}\frac{he^{a_{b,1}^{(t)}}e^{2a_{b,0}^{(t)}}}{\left(e^{a_{b,0}^{(t)}}+he^{a_{b,1}^{(t)}}+he^{a_{b,2}^{(t)}}\right)^2}\left(a_{p,1}^{(t)}\pi_{1,2h+1}^{(t)}(\down_\times)+a_{p,2}^{(t)}\pi_{1,2h+1}^{(t)}(\up)\right)\notag\\
    &\geq \frac{1+o(1)}{Nk}\left(a_{p,1}^{(t)}\right)^{1-\frac{6}{N-1}}\delta a_{q,\times}^{(t)}\notag\\
    &\quad -\frac{1+o(1)}{N}\frac{e^{2a_{b,0}^{(t)}}}{e^{a_{b,0}^{(t)}}+e^{a_{b,1}^{(t)}}+e^{a_{b,2}^{(t)}}}\left(ka_{p,1}^{(t)}\delta a_{p,1}^{(t)}+\frac{a_{p,2}^{(t)}}{\exp\left(0.1a_{p,2}^{(t)}\right)}\delta a_{q,\times}^{(t)}\right)\notag\\
    &\geq \frac{1+o(1)}{Nk}\left(a_{p,1}^{(t)}\right)^{1-\frac{6}{N-1}}\delta a_{q,\times}^{(t)}-\frac{C^2+o(1)}{Nk^2}\left(a_{p,1}^{(t)}\right)^{1-\frac{10}{N-1}}\delta a_{p,1}^{(t)},
\end{align}
where the second relation follows from \eqref{eq:ea_b_0>=k/lnk_T_15}, \eqref{eq:e_ab_2_lb_T_1_5}, \eqref{eq:delta_a_q_x_simplified_T_14-T_15}, \eqref{eq:delta_a_p_1_simplified_balanced_phase_1.2} and \eqref{eq:sum_1-pi_2h+1_up_vs_sum_1-pi_2h_up} (which also holds here), and the third relation follows from \eqref{eq:ea_b_1>=ea_b_0_T_15}, \eqref{eq:e_ab_2_lb>T14} and \eqref{eq:e2a_b_0_t1_contradiction}. 
From \eqref{eq:delta_e_a_b_0_s_0_contradiction} we deduce
\begin{align}\label{eq:e2a_b_0_s_0_contradiction}
    e^{a_{b,0}^{(s)}}-e^{a_{b,0}^{(s')}}\geq \frac{1+o(1)}{Nk}\left(a_{p,1}^{(s')}\right)^{1-\frac{6}{N-1}}\left(a_{q,\times}^{(s)}-a_{q,\times}^{(s')}\right)-\frac{C^2+o(1)}{Nk^2}\left(a_{p,1}^{(s)}\right)^{1-\frac{10}{N-1}}\left(a_{p,1}^{(s)}-a_{p,1}^{(s')}\right).
\end{align}
From \eqref{eq:a_q_x_q_0_p_2_asymp_phase_1.2}, \eqref{eq:a_p_2_bound_phase_1.2} we know that there exist constants $c_1,c_2>0$, $c_2>c_1$ such that for all $t\in[s',s]$: 
\begin{align}
    c_1\frac{a_{p,1}^{(t_1)}}{k}\leq a_{q,\times}^{(t_1)}\leq c_2\frac{a_{p,1}^{(t_1)}}{k}.
\end{align}
We can choose $C_1\leq\frac{c_1}{2c_2}$ in \eqref{eq:s_0'}, which guarantees 
\begin{align}
    a_{q,\times}^{(s_0)}-a_{q,\times}^{(s_0')}\geq c_1\frac{a_{p,1}^{(s_0)}}{k}-c_2\frac{a_{p,1}^{(s_0')}}{k}\geq c_1\frac{a_{p,1}^{(s_0)}}{2k}.
\end{align}
Plugging this into \eqref{eq:e2a_b_0_s_0_contradiction}, we have
\begin{align}
    e^{a_{b,0}^{(s)}}-e^{a_{b,0}^{(s')}}&\geq \frac{1+o(1)}{Nk^2}\frac{c_1^2}{4c_2}\left(a_{p,1}^{(s)}\right)^{2-\frac{6}{N-1}}-\frac{C^2+o(1)}{Nk^2}\left(a_{p,1}^{(s)}\right)^{2-\frac{10}{N-1}}\notag\\
    &\gtrsim \frac{1}{Nk^2}\left(a_{p,1}^{(s)}\right)^{2-\frac{6}{N-1}}\overset{\eqref{eq:s_0}}\gg \frac{\left(a_{p,1}^{(s)}\right)^{1-\frac{6}{N-1}}}{Nk}
\end{align}
if we set $C$ small enough. This contradicts \eqref{eq:e2a_b_0_t1_contradiction}. Therefore, there exists $t_1\in[ s',s]$ such that \eqref{eq:e2a_b_0_t1} holds. Then by a similar argument as how we show \eqref{eq:ea_b_0>=k/lnk_T_15}, we can show \eqref{eq:a_b_0>=k^2_T_15} holds for all $t\in[s',s]$ again by contradiction.

By \eqref{eq:ea_b_1_bound_T_15}, \eqref{eq:ea_b_0>=k/lnk_T_15} and \eqref{eq:A_M_sum0} we know that
\begin{align}\label{eq:a_b_2<0}
    \forall t\in [T_{1,4},s]:\quad a_{b,2}^{(t)}<0.
\end{align}
This shows the second relation of \eqref{eq:e_ab_2_lb>T14} holds at step $s$.
Therefore, at time $s$ we have established 
\eqref{eq:ea_b_1_bound_T_15}, \eqref{eq:ea_b_1>=ea_b_0_T_15}, and both relations in
\eqref{eq:e_ab_2_lb>T14}. Applying the same algebraic estimate from
\eqref{eq:r_down_times_phase1.2} at time $s$, we obtain
\[
    \frac{1}{k-1}-\frac{c_1}{k^4}
    \leq r_{\down_\times}^{(s)}
    \leq \frac{1}{k-1},
\]
possibly after enlarging the constant $c_1$. Combining this with
\eqref{eq:r_down_times_bound_phase1.2_pre}, we conclude that
\begin{align}\label{eq:r_down_times_bound_phase1.2}
    \forall t\in[T_{1,4},s]:\quad
    \frac{1}{k-1}-\frac{c_1}{k^4}
    \leq r_{\down_\times}^{(t)}
    \leq \frac{1}{k-1}.
\end{align}

\paragraph{At $s=T_1$:} 
We first show \eqref{eq:T_1_bound}.
Same as in \eqref{eq:R_pi_s_1_expression_phase_1}, we have for any $t\in\NN$:
\begin{align}\label{eq:R_pi_s_1_expression_phase_1.2}
    R(\pi^{(t)};\gP_1) 
    &=\frac{1}{k}\left(1-\pi_{1,1}^{(t)}(\up)\right)+\sum_{h=1}^{k-1}\frac{1}{k}\bigg(\prod_{i=1}^{h-1}\pi_{1,2i}^{(t)}(\up)\prod_{i=1}^h\left(1-(i-1)\pi_{1,2i-1}^{(t)}(\down_\times)-\pi_{1,2i-1}^{(t)}(\up)\right)\notag\\
    &\hspace{5.0cm}\cdot\pi_{1,2h}^{(t)}(\up)\left(1-\pi_{1,2h+1}^{(t)}(\up)-h\pi_{1,2h+1}^{(t)}(\down_\times)\right)\bigg).
\end{align}
From \eqref{eq:1-R_pi_T_1_2_phase_1_end} we know that a sufficient condition for $R(\pi^{(t)};\gP_1)\geq 1-\epsilon^{1.05}$ is
\begin{align}\label{eq:sufficient_condition_for_end_phase1.2}
    \pi_{1,1}^{(t)}(\up)\leq \frac{\epsilon^{1.05}}{4k},\quad \max_{h\in[k-1]}\{1-\pi_{1,2h}^{(t)}(\up)\}\leq \frac{\epsilon^{1.05}}{2k},\quad \max_{h\in[k-1]}\{h\pi_{1,2h+1}^{(t)}(\down_\times)\}\leq \frac{\epsilon^{1.05}}{4k},
\end{align}
where we use the fact that
\begin{align}
    \max\{\pi_{1,2h-1}^{(t)}(\up)\}_{h\in[k]}=\pi_{1,1}^{(t)}(\up)
\end{align}
guaranteed by \eqref{eq:tiny_pi_1_2h+1_up_phase1.2}.
Therefore, by our curriculum design, we have
\begin{align}\label{eq:max_pi_1_2h_up_down_times_T_1}
    \max\bigg\{\pi_{1,1}^{(T_1-1)}(\up),\max_{h\in[k-1]}\{1-\pi_{1,2h}^{(T_1-1)}(\up)\},\max_{h\in[k-1]}\{h\pi_{1,2h+1}^{(T_1-1)}(\down_\times)\}\bigg\}\geq \frac{\epsilon^{1.05}}{4k}.
\end{align}
By \eqref{eq:a_q_x>a_p_2_bound_T_15}, \eqref{eq:a_p2>a_q_0_bound_T_15}, \eqref{eq:a_q_0>a_q_x_bound_T_15}, \eqref{eq:pi_1_2h+1_down_times_simplified_phase1.2}, \eqref{eq:1-pi_2h_bound_phase1.2}, \eqref{eq:pi_1_1_up_phase1.2}, \eqref{eq:a_q_x_q_0_p_2_asymp_phase_1.2} and \eqref{eq:a_p_1_lb_T_1_5} we know that \eqref{eq:max_pi_1_2h_up_down_times_T_1} indicates
\begin{align}\label{eq:a_p_1_bound_phase1.2_T_1_intermediate}
   r_{\down_\times}^{(T_1-1)}\left(a_{p,1}^{(T_1-1)}+a_{p,2}^{(T_1-1)}\right)\leq \left(1+\frac{c_1}{k}\right)\ln (k^{2}/\epsilon^{1.05}),
\end{align}
for some constant $c_1>0$. This combined with \eqref{eq:r_down_times_bound_phase1.2}, \eqref{eq:a_p_2_bound_phase_1.2} further gives
\begin{align}\label{eq:a_p_1_bound_phase1.2_T_1}
    a_{p,1}^{(T_1-1)}\lesssim k\ln(1/\epsilon).
\end{align}
By \eqref{eq:a_p_1_bound_phase1.2_T_1_intermediate}, \eqref{eq:r_down_times_bound_phase1.2}, \eqref{eq:a_p_2_bound_phase_1.2} and the fact that $a_{p,1}^{(t)}$ is increasing in $t$ (c.f.~\eqref{eq:a_p_1_increase_phase_1.2}), we have for all $t\in[T_{1,4},T_1-1]$:
    \begin{align}
        \sum_{h=1}^{k-1}\frac{k-h}{k}\pi_{1,2h+1}^{(t)}(\down_\times)
        &\overset{\eqref{eq:sum_pi_1_2h+1_down_times_bound_phase1.2}}\geq \frac{1+o(1)}{k}\exp\left(-r_{\down_\times}^{(t)}\left(a_{p,1}^{(t)}+a_{p,2}^{(t)}\right)\right)\geq (1+o(1))\frac{\epsilon^{1.05+\frac{c_2}{k}}}{k^{3}}
    \end{align}
    for some constant $c_2>0$.
By this and \eqref{eq:delta_a_p_1_simplified_balanced_phase_1.2} we know that for all $t\in[T_{1,4},T_1-1]$:
\begin{align}\label{eq:delta_a_p_1_bound_phase1.2_T_1}
    \delta a_{p,1}^{(t)}\gtrsim \frac{\epsilon^{1.05+\frac{c_2}{k}}}{k^{4}}
\end{align}
    \eqref{eq:delta_a_p_1_bound_phase1.2_T_1} together with \eqref{eq:a_p_1_bound_phase1.2_T_1} suggests
    \begin{align}
        T_1-T_{1,4}\lesssim \frac{k^{5}\ln (1/\epsilon)}{\eta\epsilon^{1.05+\frac{c_2}{k}}}
    \end{align}
    By \eqref{eq:T_1_4_bound} and our assumption that $\ln(1/\epsilon)\geq e^k$, we have
    \begin{align}\label{eq:T_1_bound_a_p_1}
        T_1\lesssim \frac{k^{5}\ln (1/\epsilon)}{\eta\epsilon^{1.05+\frac{c_2}{k}}}\lesssim \frac{1}{\eta\epsilon^{1.1}}.
    \end{align}
This gives \eqref{eq:T_1_bound_a_p_1}.

On the other hand, by \eqref{eq:R_pi_s_1_expression_phase_1.2} and the fact that $R\left(\pi^{(T_1)};\gP_1\right)\geq 1-\frac{\epsilon}{k^L}$ we know that 
\begin{align}\label{eq:pi_ub_T_1}
    \pi_{1,1}^{(T_1)}(\up)\leq\epsilon^{1.05},\quad 1-\pi_{1,2}^{(T_1)}(\up)\leq \frac{k}{k-1}\epsilon^{1.05},\quad \pi_{1,2k-1}^{(T_1)}(\down_\times)\leq k\epsilon^{1.05}.
\end{align}
Thus by the first relation and \eqref{eq:pi_1_1_up_phase1.2} we have
\begin{align}
    \left(1+\frac{1}{k}\right)a_{q,0}^{(T_1)}-\frac{a_{p,0}^{(T_1)}}{k}\geq \ln\left(\frac{1}{k\epsilon^{1.05}}-\frac{1}{k}\right),
\end{align}
which gives \eqref{eq:a_q_0_bound_T_1}. \eqref{eq:a_q_0_bound_T_1} combined with \eqref{eq:e^2ac1=(a_q_x+a_q_0)^2} gives \eqref{eq:a_c_1_bound_T_1}. 
\eqref{eq:a_q_0_bound_T_1} together with Induction Hypothesis II and \eqref{eq:a_p_1_bound_phase1.2_T_1} also yields \eqref{eq:a_p_1_bound_T_1}. From \eqref{eq:a_p_1_bound_T_1} and \eqref{eq:a_b_0>=k^2_T_15} we know that \eqref{eq:e_ab_0_lb_T_1} holds.

When $t\geq s_0$ (where $s_0$ is defined in \eqref{eq:s_0}), similar to \eqref{eq:e_ab1_I_h_1_lb>T_14}, by \eqref{eq:a_b_0>=k^2_T_15} and \eqref{eq:I_h_1_balanced>T_14} we have
\begin{align}
    e^{a_{b,1}^{(t)}}I_{h,1}^{(t)}
    &\geq \left(1+O\left(\frac{1}{k}\right)\right)\frac{e^{a_{b,1}^{(t)}}e^{a_{b,0}^{(t)}}}{e^{a_{b,0}^{(t)}}+he^{a_{b,1}^{(t)}}+he^{a_{b,2}^{(t)}}}a_{p,1}^{(t)}\cdot\frac{he^{a_{b,1}^{(t)}}}{e^{a_{b,0}^{(t)}}+he^{a_{b,1}^{(t)}}+he^{a_{b,2}^{(t)}}}\pi_{1,2h+1}^{(t)}(\down_\times)\notag\\
    &\geq\left(1+O\left(\frac{1}{k}\right)\right)\frac{\left(a_{p,1}^{(t)}\right)^{2-\frac{6}{N-1}}}{Nk^2}\frac{he^{a_{b,1}^{(t)}}}{e^{a_{b,0}^{(t)}}+he^{a_{b,1}^{(t)}}+he^{a_{b,2}^{(t)}}}\pi_{1,2h+1}^{(t)}(\down_\times)\notag\\
    &\geq\left(1+O\left(\frac{1}{k}\right)\right) \frac{\left(a_{p,1}^{(t)}\right)^{2-\frac{6}{N-1}}}{Nk^2}\Bigg(\frac{he^{a_{b,2}^{(t)}}}{e^{a_{b,0}^{(t)}}+(2h-1)e^{a_{b,2}^{(t)}}}\pi_{1,2h}^{(t)}(\down_\times^1)\notag\\
    &\qquad+\frac{he^{a_{b,1}^{(t)}}}{e^{a_{b,0}^{(t)}}+he^{a_{b,1}^{(t)}}+he^{a_{b,2}^{(t)}}}\pi_{1,2h+1}^{(t)}(\down_\times)-\frac{he^{a_{b,2}^{(t)}}}{e^{a_{b,0}^{(t)}}+(2h-1)e^{a_{b,2}^{(t)}}}\pi_{1,2h}^{(t)}(\down_\times^1)\Bigg),
\end{align}
Then similar to \eqref{eq:delta_ea_b_1_lb<=T_1_5}, we have 
\begin{align}
    \delta e^{a_{b,1}^{(t)}}&\geq \frac{1+o(1)}{N^2k^2}\left(a_{p,1}^{(t)}\right)^{2-\frac{6}{N-1}}\left(k\delta a_{p,1}^{(t)}-\frac{1}{2k}\delta a_{q,\times}^{(t)}\right),
\end{align}
and same as for \eqref{eq:e_ab_1_lb_intermediate>T_14}, this yields
\begin{align}
    e^{a_{b,1}^{(T_1)}}\gtrsim \frac{1}{N^2k}\left(a_{p,1}^{(T_1)}\right)^{3-\frac{6}{N-1}}\overset{\eqref{eq:ea_b_1>=ea_b_0_T_15}}\gtrsim \frac{1}{N}\left(a_{p,1}^{(T_1)}\right)^{1-\frac{5}{N}}e^{a_{b,0}^{(T_1)}}.
\end{align}
This combined with \eqref{eq:a_p_1_bound_T_1} gives \eqref{eq:ea_b_1_lb_T_1}. And by \eqref{eq:e_ab_0_lb_T_1}, \eqref{eq:ea_b_1_lb_T_1}, \eqref{eq:e_ab_2_lb>T14} \eqref{eq:A_M_sum0} we obtain \eqref{eq:e_ab_2_lb_T_1}.

Further, by \eqref{eq:e_ab_2_lb>T14}, \eqref{eq:a_b_0>=k^2_T_15}, \eqref{eq:a_p_2_bound_phase_1.2}, \eqref{eq:a_p_0_bound_phase_1.2} and \eqref{eq:mu_h_phase1.2} we have
\begin{align}\label{eq:mu_2_ub_T_1}
    -\frac{a_{p,0}^{(T_1)}}{k}\leq \mu_2^{(T_1)}&\leq-\frac{a_{p,0}^{(T_1)}}{k}+\left(\frac{a_{p,0}^{(T_1)}}{k}+a_{p,2}^{(T_1)}\right)\frac{Nk}{\left(a_{p,1}^{(T_1)}\right)^{1-\frac{6}{N-1}}}\notag\\
    &=-\frac{a_{p,0}^{(T_1)}}{k}+\left(1+O\left(\frac{1}{k}\right)\right)N\left(a_{p,1}^{(T_1)}\right)^{\frac{6}{N-1}}.
\end{align}
By Lemma~\ref{lm:gradients_simplified_phase1.2_balanced}, the second relation in \eqref{eq:pi_ub_T_1}, \eqref{eq:a_q_0_bound_T_1}, \eqref{eq:xi_2h=a_q_x_phase1.2} and \eqref{eq:mu_2_ub_T_1} we have \eqref{eq:a_q_x_bound_T_1} holds.
From the third relation in \eqref{eq:pi_ub_T_1}, \eqref{eq:a_q_0_bound_T_1}, Lemma~\ref{lm:gradients_simplified_phase1.2_balanced} and Induction Hypothesis II, we have
\begin{align}
    k\epsilon^{1.05}\geq \pi_{1,2k-1}^{(T_1)}(\down_\times)=\frac{\exp\left(-r_{\down_\times}^{(T_1)}\left(a_{p,1}^{(T_1)}+a_{p,2}^{(T_1)}\right)\right)}{(k-1)\exp\left(-r_{\down_\times}^{(T_1)}\left(a_{p,1}^{(T_1)}+a_{p,2}^{(T_1)}\right)\right)+1+o(\epsilon)}.
\end{align}
By \eqref{eq:ea_b_1_lb_T_1}, \eqref{eq:e_ab_2_lb>T14} and \eqref{eq:r_down_times_phase1.2} we have
\begin{align}\label{eq:r_down_times_T_1}
    r_{\down,\times}^{(T_1)}=\frac{1}{k-1}+O\left(\frac{1}{k^2e^{a_{b,1}^{(T_1)}}}\right)\overset{\eqref{eq:ea_b_1_lb_T_1}}=\frac{1}{k-1}+o\left(\frac{1}{k\ln(1/\epsilon)}\right).
\end{align}
By the above two expressions and \eqref{eq:a_p_1_bound_phase1.2_T_1_intermediate}, we obtain \eqref{eq:a_p_1_2_bound_T_1}. 

Finally, we show \eqref{eq:win_rate_l>1}. Specifically, we'll show that given any tree with depth $l\in\{2,\cdots, L\}$ drawn from the balanced training distribution $\gP_l$, under the event that the goal is not at the first $k$ leaves $\pi^{(T_1)}$ visits (which happens with probability $1-1/k^{l-1}$), with probabillity $1-o(1)$, the agent will go all the way down to a leaf, and visit all its siblings and go back to its parent node. At this node, with probability $1-o(1)$, the agent will choose to go down again and visit a leaf that's already been visited, and the task fails.

Below we give the formal proof. All probabilities here are under policy $\pi^{(T_1)}$ and the balanced training distribution $\gP_l$ that satisfies \eqref{eq:balanced_goal_distribution} for a fixed $l\in \{2,3,\cdots,L\}$. We define events 
\begin{align}
    \gC_1&\coloneqq \left\{\text{goal is not at the first $k$ leaves $\pi^{(T_1)}$ visits, and } a_h\in[k],\forall h\in[l]\right\},\label{eq:event_C_1}\\
    \gC_2&\coloneqq \left\{a_l=\up\right\},\label{eq:event_C_2}
\end{align}
and for all $i\in\{2,3,\cdots, k\}$:
\begin{align}
    \gC_{2i-1}&\coloneqq \left\{a_{l+2(i-1)}\in[k]\setminus\gA_{l+2(i-1),\times}\right\},\label{eq:event_C_2i-1}\\
    \gC_{2i}&\coloneqq \left\{a_{l+2i-1}=\up\right\}.\label{eq:event_C_2i}
\end{align}
Then
\small
\begin{align}\label{eq:R_pi_T_1_l_phase_1_end_proof_1}
    1-R\left(\pi^{(T_1)};\gP_l\right)&\geq \PP(\gC_1)\Big(\underbrace{\PP(a_{l+1}\neq \up|\gC_1)}_{=1-\PP(\gC_2|\gC_1)}+\PP(\gC_2|\gC_1)\PP(\{\text{$\pi^{(T_1)}$ fails at some step $h>l+1$}\}|\gC_1\gC_2)\Big)\notag\\
    &\geq \PP(\gC_1)\PP(\{\text{$\pi^{(T_1)}$ fails at some step $h>l+1$}\}|\gC_1\gC_2)\notag\\
    &\geq \PP(\gC_1)\left(\PP(a_{l+2}\in \gA_{l+2,\times}|\gC_1\gC_2)+\PP(\gC_3|\gC_1\gC_2)\PP\left(\{\text{$\pi^{(T_1)}$ fails at some step $h>l+2$}\}\Big|\bigcap_{j=1}^3\gC_j\right)\right)\notag\\
    &\geq \PP(\gC_1)\PP(a_{l+2}\neq\up|\gC_1\gC_2)\PP\left(\{\text{$\pi^{(T_1)}$ fails at some step $h>l+2$}\}\Big|\bigcap_{j=1}^3\gC_i\right)\notag\\
    &\geq \cdots \geq \PP(\gC_1)\prod_{i=1}^{k}\PP\left(a_{l+2i}\neq\up\Big|\bigcap_{j=1}^{2i}\gC_{2j}\right).
\end{align}
\normalsize
Now we bound $\PP(\gC_1)$ and $\PP(a_{l+2i}\neq\up|\gC_{2i})$ ($i\in[k]$) that appear in the last line of the above inequality. First, we have
\begin{align}
    \PP\left(\text{goal is not at the first $k$ leaves $\pi^{(T_1)}$ visits}\right)=1-\frac{1}{k^{l-1}}.
\end{align}
And by Lemma~\ref{lm:gradients}, Induction Hypothesis II, what we have proven to hold at time $T_1$ in Lemma~\ref{lm:phase_1.2}, and our assumption that $\ln(1/\epsilon)\geq e^k$, we have for all $h\in[l]$, conditioned on $\gC_{1}$:
\begin{align}
    \pi_{h}(\up)&=\frac{\exp\left(-\left(1+\frac{1}{k}\right)a_{q,0}^{(T_1)}\right)}{\exp\left(-\left(1+\frac{1}{k}\right)a_{q,0}^{(T_1)}\right)+\sum_{i=1}^{k}\exp\left(\frac{-\frac{a_{p,0}^{(T_1)}}{k}e^{a_{b,0}^{(T_1)}}+\left(-N_h(i)(a_{p,1}^{(T_1)}+a_{p,2}^{(T_1)})+(h-1)a_{p,2}^{(T_1)}\right)e^{a_{b,2}^{(T_1)}}}{e^{a_{b,0}^{(T_1)}}+(h-1)e^{a_{b,2}^{(T_1)}}}\right)}\notag\\
    &\leq \frac{\exp\left(-\left(1+\frac{1}{k}\right)a_{q,0}^{(T_1)}+\frac{a_{p,0}^{(T_1)}}{k}\right)}{\exp\left(-\left(1+\frac{1}{k}\right)a_{q,0}^{(T_1)}+\frac{a_{p,0}^{(T_1)}}{k}\right)+\sum_{i=1}^{k}\exp\left(\frac{\Big(-a_{p,1}^{(T_1)}+\frac{a_{p,0}^{(T_1)}}{k}\Big)(h-1)e^{a_{b,2}^{(T_1)}}}{e^{a_{b,0}^{(T_1)}}+(h-1)e^{a_{b,2}^{(T_1)}}}\right)}=\epsilon^{1.05+o(1)}.
\end{align}
Combining the above two facts, we have 
\begin{align}\label{eq:PP_gC_1_bound_T_1}
    \PP(\gC_1)\geq \left(1-\frac{1}{k^{l-1}}\right)\left(1-\left(\frac{\epsilon}{k^L}\right)^{1+o(1)}\right)^l=1-\frac{1+o(1)}{k^{l-1}}.
\end{align}
Similarly, by Lemma~\ref{lm:gradients} and Induction Hypothesis II, we have for all $i\in[k]$:
\small
\begin{align}\label{eq:PP_a_l+2i_up_bound_T_1}
    &\PP\Big(a_{l+2i}=\up\Big|\bigcap_{j=1}^{2i}\gC_{2j}\Big)\notag\\
    &\leq\frac{\exp\left(-\left(1+O\left(\frac{1}{\ln^2 k}\right)\right)a_{q,0}^{(T_1)}\right)}{\exp\left(-\left(1+O\left(\frac{1}{\ln^2k}\right)\right)a_{q,0}^{(T_1)}\right)+(k-i)\exp\left((1+o(1))a_{p,2}^{(T_1)}\right)+i\exp\left(\frac{1}{i}\left((i-1)a_{p,2}^{(T_1)}-a_{p,1}^{(T_1)}+o\left(\frac{1}{k^2}\right)a_{p,1}^{(T_1)}\right)\right)}\notag\\
    &\overset{\eqref{eq:a_p_2_bound_phase_1.2}}\leq \frac{\exp\left(-\left(1+O\left(\frac{1}{\ln^2 k}\right)\right)a_{q,0}^{(T_1)}\right)}{k\exp\left(O\left(\frac{1}{k^2}\right)a_{p,1}^{(T_1)}\right)}\overset{\eqref{eq:a_q_0_bound_T_1}}\leq \epsilon^{1.05+o(1)}.
\end{align}
\normalsize
Combining \eqref{eq:PP_gC_1_bound_T_1}, \eqref{eq:PP_a_l+2i_up_bound_T_1} and \eqref{eq:R_pi_T_1_l_phase_1_end_proof_1}, we have
\begin{align}
    1-R\left(\pi^{(T_1)};\gP_l\right)\geq 1-\frac{1+o(1)}{k^{l-1}}.
\end{align}
This gives \eqref{eq:win_rate_l>1}.

\subsubsection{Proof of Lemma~\ref{lm:gradients_simplified_phase1.2_balanced}}\label{sec:proof_gradients_simplified_phase1.2_balanced}
From the proof of Lemma~\ref{lm:gradients_simplified_phase1.1_balanced} in Appendix~\ref{sec:proof_gradients_simplified_phase1_balanced} we know that \eqref{eq:delta_a_q_0_simplified_balanced_phase_1}, \eqref{eq:delta_a_q_x_simplified_balanced_phase_1}, \eqref{eq:delta_a_c_0_simplified_balanced_phase_1} and \eqref{eq:delta_a_c_1_simplified_balanced_phase_1} still hold at Phase 1.2.
Note that when $l=1$, for all $h\in[k-1]$, we have
\begin{align}
    \PP^{(t)}\left(\gE_{1,2h}\right)=\underbrace{\frac{k-h}{k}}_{\text{(a)}}\prod_{i=1}^{h}\left(1-(i-1)\pi_{1,2i-1}^{(t)}(\down_\times)-\pi_{1,2i-1}^{(t)}(\up)\right)\prod_{i=1}^{h-1}\pi_{1,2i}^{(t)}(\up),
\end{align}
where (a) represents the probability of the event that the goal is not at the first $h$ leaves visited.
By Induction Hypothesis II and Lemma~\ref{lm:gradients} we can compute that for all $h\in[k-1]$,
\begin{align}\label{eq:PP_gE_1_2h_balanced_phase1.2}
    \PP^{(t)}\left(\gE_{1,2h}\right)=\left(1+o\left(\frac{1}{k}\right)\right)\frac{k-h}{k},
\end{align}
and
\begin{align}\label{eq:V_approx_1}
    1-V_{1,2h+1}^{(t)}=o(1).
\end{align}
Combining \eqref{eq:delta_a_q_0_simplified_balanced_phase_1}, \eqref{eq:delta_a_q_x_simplified_balanced_phase_1}, \eqref{eq:delta_a_c_0_simplified_balanced_phase_1} and \eqref{eq:delta_a_c_1_simplified_balanced_phase_1} with \eqref{eq:PP_gE_1_2h_balanced_phase1.2} and \eqref{eq:V_approx_1}, we obtain \eqref{eq:delta_a_q_0_simplified_balanced_phase_1.2}, \eqref{eq:delta_a_q_x_simplified_balanced_phase_1.2}, \eqref{eq:delta_a_c_0_simplified_balanced_phase_1.2} and \eqref{eq:delta_a_c_1_simplified_balanced_phase_1.2}.

Note that by our notation, we have for all $h\in[k]$:
\begin{align}\label{eq:A_computation_phase1}
    |\gA_{2h,\times}|\mid \gE_{2h} = 0,\quad |\gA_{2h-1,\times}|\mid \gE_{2h-1} = h-1,\quad N_{2h-1}(\up)=N_{2h}(\up)=h-1.
\end{align}
Thus we have
\begin{subequations}
\begin{align}\label{eq:delta_a_p_0_simplified_balanced_phase1.2_computation}
    \delta a_{p,0}^{(t)}&\overset{\eqref{eq:delta_a_p_0_simplified_balanced}}=\EE\left[\sum_{h=1}^{k-1} \frac{e^{a_{b,0}^{(t)}}}{e^{a_{b,0}^{(t)}}+(2h-1)e^{a_{b,2}^{(t)}}}\pi_{1,2h}^{(t)}(\up)\left(1-\pi_{1,2h}^{(t)}(\up)\right)Q_{2h}^{(t)}(\up)\bigg|l=1\right]\notag\\
    &\qquad-\EE\left[\pi_{1,1}^{(t)}(\up)G_1^{(t)}|l=1\right]-\EE\left[\sum_{h=1}^{k-1}\frac{e^{a_{b,0}^{(t)}}}{e^{a_{b,0}^{(t)}}+he^{a_{b,1}^{(t)}}+he^{a_{b,2}^{(t)}}}\pi_{1,2h+1}^{(t)}(\up)G_{2h+1}^{(t)}\bigg|l=1\right],\notag\\
    &\overset{\eqref{eq:PP_gE_1_2h_gE_1_2h+1_balanced}}=-\pi_{1,1}^{(t)}(\up)V_{1,1}^{(t)}+\sum_{h=1}^{k-1} \PP^{(t)}\left(\gE_{1,2h}\right)\pi_{1,2h}^{(t)}(\up)\Bigg(\frac{e^{a_{b,0}^{(t)}}}{e^{a_{b,0}^{(t)}}+(2h-1)e^{a_{b,2}^{(t)}}}\left(1-\pi_{1,2h}^{(t)}(\up)\right)\notag\\
    &\hspace{6cm}-\frac{e^{a_{b,0}^{(t)}}}{e^{a_{b,0}^{(t)}}+he^{a_{b,1}^{(t)}}+he^{a_{b,2}^{(t)}}}\pi_{1,2h+1}^{(t)}(\up)\Bigg)V_{1,2h+1}^{(t)}, \\
\delta a_{p,1}^{(t)} &\overset{\eqref{eq:delta_a_p_1_simplified_balanced}}=\EE\left[\sum_{h=1}^{k-1} \frac{h}{k}\frac{e^{a_{b,1}^{(t)}}}{e^{a_{b,0}^{(t)}}+he^{a_{b,1}^{(t)}}+he^{a_{b,2}^{(t)}}}\pi_{1,2h+1}^{(t)}(j)G_{2h+1}^{(t)}\bigg|j\in\gA_{2h+1,\times},l=1\right]\notag\\
    &\qquad +\EE\left[\sum_{h=1}^{k-1} \frac{N_{2h}(j)e^{a_{b,2}^{(t)}}}{e^{a_{b,0}^{(t)}}+(2h-1)e^{a_{b,2}^{(t)}}}\pi_{1,2h}^{(t)}(j)\pi_{1,2h}^{(t)}(\up)Q_{2h}^{(t)}(\up)\bigg|j\in[k],l=1\right]\notag\\
    &=\EE\left[\sum_{h=1}^{k-1} \frac{h}{k}\frac{e^{a_{b,1}^{(t)}}}{e^{a_{b,0}^{(t)}}+he^{a_{b,1}^{(t)}}+he^{a_{b,2}^{(t)}}}\pi_{1,2h+1}^{(t)}(j)G_{2h+1}^{(t)}\bigg|j\in\gA_{2h+1,\times},l=1\right]\notag\\
    &\qquad +\EE\left[\sum_{h=1}^{k-1} \frac{h}{k}\frac{e^{a_{b,2}^{(t)}}}{e^{a_{b,0}^{(t)}}+(2h-1)e^{a_{b,2}^{(t)}}}\pi_{1,2h}^{(t)}(j)\pi_{1,2h}^{(t)}(\up)Q_{2h}^{(t)}(\up)\bigg|j\in\gA_{2h-1,\times}\cup\{a_{2h-1}\},l=1\right]\notag\\
    &\overset{\eqref{eq:PP_gE_1_2h_gE_1_2h+1_balanced}}=\frac{1}{k}\sum_{h=1}^{k-1} \PP^{(t)}\left(\gE_{1,2h}\right)\pi_{1,2h}^{(t)}(\up)\Bigg(\frac{he^{a_{b,2}^{(t)}}}{e^{a_{b,0}^{(t)}}+(2h-1)e^{a_{b,2}^{(t)}}}\pi_{1,2h}^{(t)}(\down_\times^1)\notag\\
    &\hspace{4cm}+\frac{he^{a_{b,1}^{(t)}}}{e^{a_{b,0}^{(t)}}+he^{a_{b,1}^{(t)}}+he^{a_{b,2}^{(t)}}}\pi_{1,2h+1}^{(t)}(\down_\times)\Bigg)V_{1,2h+1}^{(t)}, \label{eq:delta_a_p_1_simplified_balanced_phase1.2_computation} \\
    \delta a_{p,3}^{(t)} &\overset{\eqref{eq:delta_a_p_3_simplified_balanced}}=-\EE\left[\sum_{h=1}^{k-1} \frac{h}{k}\frac{e^{a_{b,1}^{(t)}}}{e^{a_{b,0}^{(t)}}+he^{a_{b,1}^{(t)}}+he^{a_{b,2}^{(t)}}}\pi_{1,2h+1}^{(t)}(\up)G_{2h+1}^{(t)}\bigg|j\in\gA_{2h+1,\times},l=1\right]\notag\\
    &\quad+\EE\left[\sum_{h=1}^{k-1} \frac{N_{2h}(j)e^{a_{b,2}^{(t)}}}{e^{a_{b,0}^{(t)}}+(2h-1)e^{a_{b,2}^{(t)}}}\pi_{1,2h}^{(t)}(\up)\left(1-\pi_{1,2h}^{(t)}(\up)\right)Q_{2h}^{(t)}(\up)\bigg|j\in[k],l=1\right]\notag\\
    &=-\EE\left[\sum_{h=1}^{k-1} \frac{h}{k}\frac{e^{a_{b,1}^{(t)}}}{e^{a_{b,0}^{(t)}}+he^{a_{b,1}^{(t)}}+he^{a_{b,2}^{(t)}}}\pi_{1,2h+1}^{(t)}(\up)G_{2h+1}^{(t)}\bigg|j\in\gA_{2h+1,\times},l=1\right]\notag\\
    &\quad+\EE\left[\sum_{h=1}^{k-1} \frac{h}{k}\frac{e^{a_{b,2}^{(t)}}}{e^{a_{b,0}^{(t)}}+(2h-1)e^{a_{b,2}^{(t)}}}\pi_{1,2h}^{(t)}(\up)\left(1-\pi_{1,2h}^{(t)}(\up)\right)Q_{2h}^{(t)}(\up)\bigg|j\in\gA_{2h-1,\times}\cup\{a_{2h-1}\},l=1\right]\notag\\
    &\overset{\eqref{eq:PP_gE_1_2h_gE_1_2h+1_balanced}}=\frac{1}{k}\sum_{h=1}^{k-1} \PP^{(t)}\left(\gE_{1,2h}\right)\pi_{1,2h}^{(t)}(\up)\Bigg(\frac{he^{a_{b,2}^{(t)}}}{e^{a_{b,0}^{(t)}}+(2h-1)e^{a_{b,2}^{(t)}}}\left(1-\pi_{1,2h}^{(t)}(\up)\right)\notag\\
    &\hspace{4cm}-\frac{he^{a_{b,1}^{(t)}}}{e^{a_{b,0}^{(t)}}+he^{a_{b,1}^{(t)}}+he^{a_{b,2}^{(t)}}}\pi_{1,2h+1}^{(t)}(\up)\Bigg)V_{1,2h+1}^{(t)}, \label{eq:delta_a_p_3_simplified_balanced_phase1.2_computation} \\
    \delta a_{p,\up}^{(t)} &\overset{\eqref{eq:delta_a_p_up_simplified_balanced}}=\EE\left[\sum_{h=1}^{k-1} \frac{he^{a_{b,2}^{(t)}}}{e^{a_{b,0}^{(t)}}+he^{a_{b,1}^{(t)}}+he^{a_{b,2}^{(t)}}}\pi_{1,2h+1}^{(t)}(\up)G_{2h+1}^{(t)}\bigg|l=1\right]\notag\\
    &\qquad-\EE\left[\sum_{h=1}^{k-1} \frac{(h-1)e^{a_{b,2}^{(t)}}}{e^{a_{b,0}^{(t)}}+(2h-1)e^{a_{b,2}^{(t)}}}\pi_{1,2h}^{(t)}(\up)\left(1-\pi_{1,2h}^{(t)}(\up)\right)Q_{2h}^{(t)}(\up)\bigg|l=1\right]\notag\\
    &=\sum_{h=1}^{k-1} \PP^{(t)}\left(\gE_{1,2h}\right)\pi_{1,2h}^{(t)}(\up)\Bigg(\frac{he^{a_{b,2}^{(t)}}}{e^{a_{b,0}^{(t)}}+he^{a_{b,1}^{(t)}}+he^{a_{b,2}^{(t)}}}\pi_{1,2h+1}^{(t)}(\up)\notag\\
    &\hspace{4cm}-\frac{(h-1)e^{a_{b,2}^{(t)}}}{e^{a_{b,0}^{(t)}}+(2h-1)e^{a_{b,2}^{(t)}}}\left(1-\pi_{1,2h}^{(t)}(\up)\right)\Bigg)V_{1,2h+1}^{(t)}.\label{eq:delta_a_p_up_simplified_balanced_phase1.2_computation}
\end{align}
\end{subequations}
Combining the above expressions with \eqref{eq:PP_gE_1_2h_balanced_phase1.2}, \eqref{eq:V_approx_1}, we obtain \eqref{eq:delta_a_p_0_simplified_balanced_phase_1.2}, \eqref{eq:delta_a_p_1_simplified_balanced_phase_1.2}, \eqref{eq:delta_a_p_3_simplified_balanced_phase_1.2} and \eqref{eq:delta_a_p_up_simplified_balanced_phase_1.2}.

Note that \eqref{eq:beta_2h-1_0_l=1_balanced}-\eqref{eq:beta_2h_2_l=1_balanced} computed in the proof of Lemma~\ref{lm:gradients_simplified_phase1.1_balanced} in Appendix~\ref{sec:proof_gradients_simplified_phase1_balanced} still hold at Phase 1.2, and 
\small
\begin{align*}
    \delta a_{b,0}^{(t)}&\overset{\eqref{eq:delta_a_b_0_simplified_balanced}}=\frac{1}{N}\Bigg(\EE\Bigg[\sum_{h=1}^{k-1} \frac{e^{a_{b,0}^{(t)}}}{(e^{a_{b,0}^{(t)}}+(2h-1)e^{a_{b,2}^{(t)}})^2}\left((2h-1)e^{a_{b,2}^{(t)}}\beta_{2h,0}^{(t)}-e^{a_{b,1}^{(t)}}\beta_{2h,1}^{(t)}-e^{a_{b,2}^{(t)}}\beta_{2h,2}^{(t)}\right)\Bigg|l=1\Bigg]\notag\\
    &\qquad+\EE\Bigg[\sum_{h=1}^{k-1} \frac{e^{a_{b,0}^{(t)}}}{(e^{a_{b,0}^{(t)}}+he^{a_{b,1}^{(t)}}+he^{a_{b,2}^{(t)}})^2}\left(\left(he^{a_{b,1}^{(t)}}+he^{a_{b,2}^{(t)}}\right)\beta_{2h+1,0}^{(t)}-e^{a_{b,1}^{(t)}}\beta_{2h+1,1}^{(t)}-e^{a_{b,2}^{(t)}}\beta_{2h+1,2}^{(t)}\right)\Bigg|l=1\Bigg]\Bigg), \\
    \delta a_{b,1}^{(t)}&\overset{\eqref{eq:delta_a_b_1_simplified_balanced}}=\frac{1}{N}\Bigg(\EE\Bigg[\sum_{h=1}^{k-1} \frac{e^{a_{b,1}^{(t)}}}{(e^{a_{b,0}^{(t)}}+(2h-1)e^{a_{b,2}^{(t)}})^2}\left(e^{a_{b,0}^{(t)}}+(2h-1)e^{a_{b,2}^{(t)}}\right)\beta_{2h,1}^{(t)}\Bigg|l=1\Bigg]\notag\\
    &\qquad+\EE\Bigg[\sum_{h=1}^{k-1} \frac{e^{a_{b,1}^{(t)}}}{(e^{a_{b,0}^{(t)}}+he^{a_{b,1}^{(t)}}+he^{a_{b,2}^{(t)}})^2}\left(\left(e^{a_{b,0}^{(t)}}+he^{a_{b,2}^{(t)}}\right)\beta_{2h+1,1}^{(t)}-he^{a_{b,0}^{(t)}}\beta_{2h+1,0}^{(t)}-he^{a_{b,2}^{(t)}}\beta_{2h+1,2}^{(t)}\right)\Bigg|l=1\Bigg]\Bigg).
\end{align*}
\normalsize
Plugging \eqref{eq:beta_2h-1_0_l=1_balanced}-\eqref{eq:beta_2h_2_l=1_balanced} into the above two expressions, and utilizing \eqref{eq:PP_gE_1_2h_balanced_phase1.2}, \eqref{eq:V_approx_1}, we obtain \eqref{eq:delta_a_b_0_simplified_balanced_phase_1.2} and \eqref{eq:delta_a_b_1_simplified_balanced_phase_1.2}.

Finally, plugging \eqref{eq:A_computation_phase1} into \eqref{eq:pi_h_simplified_balanced}-\eqref{eq:Z_h_balanced} in Lemma~\ref{lm:gradients}, we obtain the policy expressions in Lemma~\ref{lm:gradients_simplified_phase1.2_balanced}.

\subsubsection{Proof of Lemma~\ref{lm:phase_2}}\label{sec_app:phase_2}
During depth-2 curriculum training, we only train matrix $P$ starting from $P^{(T_1)}$. We fix $B=B^{(T_1)},C=C^{(T_1)},Q=Q^{(T_1)}$ and drop the superscript $(T_1)$ corresponding to these variables. All the expectations/probabilities are taken over training distribution $P_2$. Below we show Lemma~\ref{lm:phase_2} by induction.

\paragraph{Base case. } At step $T_1$, \eqref{eq:a_p_0_bound_phase2}, \eqref{eq:a_p_up_bound_phase2}, \eqref{eq:a_p_1_increase_rate_phase2}, \eqref{eq:a_p_3_increase_rate<T_2_1} hold trivially at step $T_1$. By \eqref{eq:delta_a_p_3_bound_phase1.2} and \eqref{eq:a_q_x_q_0_p_2_asymp_phase_1.2} we have
\begin{align}
    |a_{p,3}^{(T_1)}|\lesssim \frac{a_{q,0}}{k}.
\end{align}
Thus \eqref{eq:a_p_3_vs_a_q_0_phase2} holds at step $T_1$. By \eqref{eq:a_p_1_bound_T_1}, \eqref{eq:a_q_x_q_0_p_2_asymp_phase_1.2} and \eqref{eq:a_p_2_bound_phase_1.2} we know that \eqref{eq:a_p_1_vs_a_q_0_phase2} holds at step $T_1$.

\paragraph{Induction. } We make the following induction hypothesis:
\begin{center}
\textbf{Induction Hypothesis III:} For all $t\leq s-1$ ($s\in [T_1+1,T_2]$), the relations in Lemma~\ref{lm:phase_2} hold. 
\end{center}

Below we show Lemma~\ref{lm:phase_2} holds for $t=s$ under Induction Hypothesis III. Define 
\begin{align}\label{eq:overline_pi_2}
    \overline{\pi}_{2}^{(t)}(\up)\coloneqq \frac{\exp\left(\left(1+\frac{1}{k}\right)\frac{ke^{-2a_{c,0}}a_{q,\times}-(k+2)a_{q,0}}{ke^{-2a_{c,0}}+k+2}\right)}{k\exp\left(-\frac{a_{p,3}^{(t)}}{k}+\frac{1}{e^{a_{b,1}-a_{b,0}}}\left(-\frac{a_{p,0}^{(t)}}{k^2}+\frac{a_{p,3}^{(t)}}{k}\right)\right)+\exp\left(\left(1+\frac{1}{k}\right)\frac{ke^{-2a_{c,0}}a_{q,\times}-(k+2)a_{q,0}}{ke^{-2a_{c,0}}+k+2}\right)}.
\end{align}
Let $h_i$ denote the first timestep that the $i$-th child arranged in order of occurrence is visited, i.e.,
\begin{align}\label{eq:h_i}
   \forall i\in[k]:\quad h_i\coloneqq \inf_{h\in\NN_+}\{\text{at step $h$, $i$ chilrden of the root is visited}\}.
\end{align}
Then $h_i=+\infty$ if the event that $i$ children of the root are visited never happens. 
Define
\begin{align}\label{eq:overline_gE_2_i}
    \overline{\gE}_{2,i}\coloneqq \{\text{the agent traverses all children of $n_{h_j}$ for all $j\in[i]$ at the first $h_i+2k-1$ steps}, \notag\\
    \text{and arrives at $n_{h_i}$ at step $h_i+2k$, $l=2$}\}
\end{align}
($\overline{\gE}_{2,k}$ is an impossible event). And define 
\begin{align}
    Q_{2,i}^{(t)}(\up)\coloneqq Q_{h_{i}+2k}^{(t)}(\up)|\overline{\gE}_{2,i},\,\,\forall i\in[k-1].
\end{align}

We first simplify the gradient expressions related to $P$ in Lemma~\ref{lm:gradients} when $l=2$.
\begin{lm}\label{lm:gradients_simplified_phase2_balanced}
    Under Induction Hypothesis III, we have for any $t\in[T_1,s-1]$:
    \begin{subequations}
    \begin{align}
        \delta a_{p,1}^{(t)}&=\frac{1+o(1)}{k^2}\sum_{i=1}^{k-1}\frac{k-i}{k}\left(\overline\pi_{2}^{(t)}(\up)\right)^{i}\left(1-\overline\pi_{2}^{(t)}(\up)\right)Q_{2,i}^{(t)}(\up),\label{eq:delta_a_p_1_simplified_balanced_phase2}\\
        \delta a_{p,3}^{(t)}&=\frac{1}{k}\sum_{i=1}^{k-1}\frac{k-i}{k}\left(\overline{\pi}_{2}^{(t)}(\up)\right)^{i}\Bigg((1+o(1))\left(1-\overline{\pi}_{2}^{(t)}(\up)\right)\notag\\
        &\qquad\qquad-\frac{1}{(k-i)\exp\left(\frac{a_{p,1}^{(t)}-a_{p,3}^{(t)}}{k-1}+\left(1+O\left(\frac{1}{\ln^2 k}\right)\right)a_{q,0}\right)}\Bigg)Q_{2,i}^{(t)}(\up),\label{eq:delta_a_p_3_simplified_balanced_phase2}\\
        |\delta a_{p,0}^{(t)}|&\lesssim \frac{1}{\left(\ln (1/\epsilon)\right)^{1+o(1)}}\left(\delta a_{p,1}^{(t)}-\frac{1}{k^2}\delta a_{p,3}^{(t)}\right),\label{eq:delta_a_p_0_simplified_balanced_phase2}\\
        |\delta a_{p,\up}^{(t)}|&\lesssim \frac{1}{\left(\ln (1/\epsilon)\right)^{3+o(1)}}\left(\delta a_{p,1}^{(t)}-\frac{1}{k^2}\delta a_{p,3}^{(t)}\right).\label{eq:delta_a_p_up_simplified_balanced_phase2}
    \end{align}
    \end{subequations}
\end{lm}
The proof of Lemma~\ref{lm:gradients_simplified_phase2_balanced} is postponed to Appendix~\ref{sec_app:gradients_simplified_phase2_balanced}.

First, by \eqref{eq:delta_a_p_3_bound_phase1.2} and \eqref{eq:a_q_x_q_0_p_2_asymp_phase_1.2} we have
\begin{align}\label{eq:a_p_3_T_1_bound_wrt_a_q_0}
    |a_{p,3}^{(T_1)}|\lesssim \frac{a_{q,0}}{k}.
\end{align}
And \eqref{eq:a_q_x_q_0_p_2_asymp_phase_1.2}, \eqref{eq:a_p_2_bound_phase_1.2} guarantee 
\begin{align}\label{eq:a_p_1_T_1_bound_wrt_a_q_0}
    a_{p,1}^{(T_1)}\asymp k a_{q,0}^{(T_1)}.
\end{align}
Thus by the above two facts, \eqref{eq:delta_a_p_0_simplified_balanced_phase2}, \eqref{eq:delta_a_p_up_simplified_balanced_phase2}, our assumption that $\ln(1/\epsilon)\geq e^k$, which guarantees $k=\ln(1/\epsilon)^{o(1)}$, and that \eqref{eq:a_p_3_vs_a_q_0_phase2}, \eqref{eq:a_p_1_vs_a_q_0_phase2} holds at step $s-1$ by Induction Hypothesis III, we have \eqref{eq:a_p_0_bound_phase2}, \eqref{eq:a_p_up_bound_phase2} hold at step $s$. In addition, from \eqref{eq:a_p_1_increase_rate_phase2}, \eqref{eq:a_p_3_increase_rate<T_2_1}, \eqref{eq:a_p_3_bound_phase2}, \eqref{eq:a_p_1_vs_a_q_0_phase2} and \eqref{eq:a_p_3_T_1_bound_wrt_a_q_0} we know that the first inequality in \eqref{eq:a_p_3_vs_a_q_0_phase2} holds at step $s$. By \eqref{eq:a_p_1_vs_a_q_0_phase2}, \eqref{eq:a_p_3_increase_rate<T_2_1}, \eqref{eq:a_p_3_bound_phase2} we deduce that the second inequality in \eqref{eq:a_p_3_vs_a_q_0_phase2} also holds at step $s$ for any $s\in[T_1+1,T_2]$.

By the expression of $\overline\pi_{2}^{(t)}(\up)$ in \eqref{eq:overline_pi_2} we know that there exists a constant $C\in\RR$ such that for $T_{2,1}$ defined in \eqref{eq:T_2_1_definition}, if $s\leq T_{2,1}$, we have 
\begin{align}\label{eq:overline_pi_2_ub_wrt_a_p_1_a_p_3}
    \forall t\leq s-1:\quad \frac{1}{\exp\left(\frac{a_{p,1}^{(t)}-a_{p,3}^{(t)}}{k-1}+\left(1+O\left(\frac{1}{\ln^2 k}\right)\right)a_{q,0}\right)}&=\exp\left(-c\frac{a_{q,0}}{\ln^2 k}\right)\left(1-\overline\pi_{2}^{(t)}(\up)\right)\notag\\
    &=o\left(1-\overline\pi_{2}^{(t)}(\up)\right)
\end{align}
for some constant $c>0$.
By \eqref{eq:overline_pi_2_ub_wrt_a_p_1_a_p_3} and \eqref{eq:delta_a_p_3_simplified_balanced_phase2} we have
\begin{align}\label{eq:delta_a_p_3_vs_a_p_1<T_2_1}
    \forall t\leq s-1:\quad \delta a_{p,3}^{(t)}&=\frac{1+o(1)}{k}\sum_{i=1}^{k-1}\frac{k-i}{k}\left(\overline\pi_{2}^{(t)}(\up)\right)^{i}\left(1-\overline\pi_{2}^{(t)}(\up)\right)Q_{2,i}^{(t)}(\up)\overset{\eqref{eq:delta_a_p_1_simplified_balanced_phase2}}\notag\\
    &=(1+o(1))k\delta a_{p,1}^{(t)}.
\end{align}
This proves \eqref{eq:a_p_3_increase_rate<T_2_1} holds for $t=s$ when $s\leq T_{2,1}$.

When $s>T_{2,1}$, we show
\begin{align}\label{eq:a_p_3_lb_wrt_a_p_1<T_2_1}
    \forall t\in[T_{2,1},s]:\quad a_{p,3}^{(t)}\geq \frac{a_{p,1}^{(t)}}{2}+k\left(1+\frac{C}{\ln^2 k}\right)a_{q,0}
\end{align}
by contradiction. Suppose this does not hold, then we could choose the smallest $t_0\in[T_{2,1}+1,s]$ such that the above is violated, i.e.,
\begin{align}\label{eq:a_p_3_lb_wrt_a_p_1<T_2_1_contradiction}
    a_{p,3}^{(t_0)}<\frac{a_{p,1}^{(t_0)}}{2}+k\left(1+\frac{C}{\ln^2 k}\right)a_{q,0},
\end{align}
but 
\begin{align}\label{eq:a_p_3_ub_wrt_a_p_1<T_2_1}
    \forall t\in[T_{2,1},t_0-1]:\quad a_{p,3}^{(t)}\geq \frac{a_{p,1}^{(t)}}{2}+k\left(1+\frac{C}{\ln^2 k}\right)a_{q,0}.
\end{align}
Then by our update rule, we have 
\begin{align}
    a_{p,3}^{(t_0-1)}= \frac{a_{p,1}^{(t_0-1)}}{2}+k\left(1+\frac{C+o(1)}{\ln^2 k}\right)a_{q,0},
\end{align}
which, together with \eqref{eq:overline_pi_2_ub_wrt_a_p_1_a_p_3} suggests 
\begin{align}
    \frac{1}{\exp\left(\frac{a_{p,1}^{(t_0-1)}-a_{p,3}^{(t_0-1)}}{k-1}+\left(1+O\left(\frac{1}{\ln^2 k}\right)\right)a_{q,0}\right)}&=\exp\left(-(c+o(1))\frac{a_{q,0}}{\ln^2 k}\right)\left(1-\overline\pi_{2}^{(t_0-1)}(\up)\right)\notag\\
    &=o\left(1-\overline\pi_{2}^{(t_0-1)}(\up)\right),
\end{align}
and thus \eqref{eq:delta_a_p_3_vs_a_p_1<T_2_1} also holds for $t=t_0-1$, i.e., 
\begin{align}
    \delta a_{p,3}^{(t_0-1)}=k(1+o(1))\delta a_{p,1}^{(t_0-1)}.
\end{align}
This suggests 
\begin{align}
    a_{p,3}^{(t_0)}-\frac{a_{p,1}^{(t_0)}}{2}>a_{p,3}^{(t_0-1)}-\frac{a_{p,1}^{(t_0-1)}}{2}\overset{\eqref{eq:a_p_3_ub_wrt_a_p_1<T_2_1}}\geq k\left(1+\frac{C}{\ln^2 k}\right)a_{q,0}.
\end{align}
This contradicts \eqref{eq:a_p_3_lb_wrt_a_p_1<T_2_1_contradiction}. Thus \eqref{eq:a_p_3_lb_wrt_a_p_1<T_2_1} must hold.
On thet other hand, there exists constant $C_1>C$ such that for all $t\in[T_{2,1},s]$, if 
\begin{align}\label{eq:a_p_3_lb_wrt_a_p_1<T_2_1_condition}
    a_{p,3}^{(t)}\geq \frac{a_{p,1}^{(t)}}{2}+k\left(1+\frac{C_1}{\ln^2 k}\right)a_{q,0},
\end{align}
then we have 
\begin{align}
    \frac{1}{\exp\left(\frac{a_{p,1}^{(t)}-a_{p,3}^{(t)}}{k-1}+\left(1+O\left(\frac{1}{\ln^2 k}\right)\right)a_{q,0}\right)}=\exp\left(c\frac{a_{q,0}}{\ln^2 k}\right)\left(1-\overline\pi_{2}^{(t)}(\up)\right).
\end{align}
Then we can similarly show by contradiction that 
\begin{align}\label{eq:a_p_3_ub_wrt_a_p_1>T_2_1}
    \forall t\in[T_{2,1},s]:\quad a_{p,3}^{(t)}\leq \frac{a_{p,1}^{(t)}}{2}+k\left(1+\frac{C_1}{\ln^2 k}\right)a_{q,0}.
\end{align}
By \eqref{eq:a_p_3_lb_wrt_a_p_1<T_2_1} and \eqref{eq:a_p_3_ub_wrt_a_p_1>T_2_1} we know that \eqref{eq:a_p_3_bound_phase2} holds at step $s$.

\paragraph{When $s=T_2$.} 
Note that 
\begin{align}\label{eq:1-R_pi_T_2_2}
    1-R(\pi^{(T_2)};\gP_2)&\geq \PP^{\pi^{(T_2)}}\left(\overline{\gE}_{2,1}\right)\left(1-\pi_{2k+2}^{(T_2)}(\up)|\overline{\gE}_{2,1}\right),
\end{align}
where $\overline{\gE}_{2,1}$ is defined in \eqref{eq:overline_gE_2_i}.
Then from the policy computation in the proof of Lemma~\ref{lm:gradients_simplified_phase2_balanced} (c.f.~\eqref{eq:pi_h_up_Ax=0_c=0_phase2}, \eqref{eq:1-pi_h_up_c=x_phase2}, \eqref{eq:pi_h_up_1<=Ax<=k-1_phase2}, \eqref{eq:pi_h_up_Ax=k_phase2}, \eqref{eq:pi_h_down_x_phase2}) we know that 
\begin{align}\label{eq:PP_overline_gE_2_1}
    \PP^{\pi^{(T_2)}}\left(\overline{\gE}_{2,1}\right)\geq (1+o\left(\epsilon^{0.5}\right))\frac{k-1}{k},
\end{align}
and from \eqref{eq:pi_2_i_up_equal} we know that 
\begin{align}\label{eq:pi_2k+2_up_overline_gE_2_1}
    \pi_{2k+2}^{(T_2)}(\up)|\overline{\gE}_{2,1}=\left(1+O\left(\frac{1}{\left(\ln (1/\epsilon)\right)^{1+o(1)}}\right)\right)\overline{\pi}_{2}^{(T_2)}(\up).
\end{align}
Plugging \eqref{eq:PP_overline_gE_2_1}, \eqref{eq:pi_2k+2_up_overline_gE_2_1} into \eqref{eq:1-R_pi_T_2_2}, and recall we set $\delta_2=\frac{\epsilon^{1.02}}{k^L}$, we have
\begin{align}\label{eq:1-overline_pi_2_T_2_ub}
    \left(1+O\left(\frac{1}{\left(\ln (1/\epsilon)\right)^{1+o(1)}}\right)\right)\frac{k-1}{k}\left(1-\overline{\pi}_{2}^{(T_2)}(\up)\right)\leq \frac{\epsilon^{1.02}}{k^L}.
\end{align}
This combined with \eqref{eq:overline_pi_2} suggests 
\begin{align}
    \frac{a_{p,3}^{(T_2)}}{k}-\left(1-\frac{c_1}{\ln^2 k}\right)a_{q,0}\geq \ln\left(\frac{k^L}{\epsilon^{1.02}}\right)
\end{align}
for some constant $c_1>0$. This shows \eqref{eq:a_p_3_bound_phase2_T1} holds. 
In addition, by \eqref{eq:a_p_3_bound_phase2} and our choice of $T_{2,1}$ (c.f.~\eqref{eq:T_2_1_definition}) we know that when $s=T_{2}$, for any $t\in[T_{2,1},T_2]$:
\begin{align}\label{eq:a_p_3<=}
    a_{p,3}^{(t)}\leq \frac{a_{p,1}^{(t)}}{2}+k\left(1+\frac{C_1}{\ln^2 k}\right)a_{q,0}
\end{align}
for constant $C_1$ defined in \eqref{eq:a_p_3_ub_wrt_a_p_1>T_2_1}.
By \eqref{eq:a_p_3_bound_phase2_T1} and \eqref{eq:a_p_3<=} we have \eqref{eq:a_p_1_bound_phase2_T1} holds.

We next show \eqref{eq:a_p_3_ub_T2} and \eqref{eq:a_p_1_ub_T2}. 
For all $i,j\in[k]$, define events
\begin{align}\label{eq:gC_2_i_j}
    \gC_{2,i}^j&\coloneqq \{\text{the agent traverses all children of $n_{h_m}$ for all $m\in[i-1]$ (if $i>1$),} \notag\\
    &\qquad\text{then visits $j$ children of $n_{h_i}$ and reach the goal after taking $a_{h_i+2(j-1)}$, $\gT\sim P_2$}\},
\end{align}
where $h_i$ is defined in \eqref{eq:h_i}.
And let $[\gC_{2,i}^j]_{\leq h}$ be the "mask of $\gC_{2,i}^j$ up to step $h$", i.e.,
\begin{align}
    [\gC_{2,i}^j]_{\leq h}\coloneqq \left\{\omega\in\Omega:\exists \omega'\in\gC_{2,i}^j,\text{ such that } s_h(\omega')\leq s_h(\omega)\right\},
\end{align}
where $\Omega$ is the sample space of one full episode. And let 
\begin{align}
    \widehat\pi_h\coloneqq \pi_h|[\gC_{2,i}^j]_{\leq h}.
\end{align}
Then we have 
\small
\begin{align}\label{eq:R_pi_T_2_lb}
    &R(\pi^{(t)};\gP_2)\notag\\
    &\geq \sum_{i=1}^{k}\sum_{j=1}^{k} \PP^{\pi^{(t)}}\left(\gC_{2,i}^j\right)\notag\\
    &=\frac{1}{k^2}\sum_{i=1}^{k}\sum_{j=1}^{k}(1-\widehat\pi_1^{(t)}(\up))\Bigg[\prod_{i'=1}^{i-1}\bigg(\prod_{j'=1}^{k}\Big(1-\widehat\pi_{h_{i'}+2(j'-1)}^{(t)}(\up)-\sum_{m\in\gA_{h_{i'}+2(j'-1),\times}}\widehat\pi_{h_{i'}+2(j'-1)}^{(t)}(m)\Big)\widehat\pi_{h_i+2j'-1}^{(t)}(\up)\bigg)\notag\\
    &\qquad\qquad\cdot\widehat\pi_{h_{i'}+2k}^{(t)}(\up)\Big(1-\widehat\pi_{h_{i'}+2k+1}^{(t)}(\up)-\sum_{m\in\gA_{h_{i'}+2k+1,\times}}\widehat\pi_{h_{i'}+2k+1}^{(t)}(m)\Big)\Bigg]\notag\\
    &\qquad\qquad\cdot \prod_{j'=1}^{j}\left(1-\widehat\pi_{h_i+2(j'-1)}^{(t)}(\up)-\sum_{m\in\gA_{h_i+2(j'-1),\times}}\widehat\pi_{h_i+2(j'-1)}^{(t)}(m)\right)\prod_{j'=1}^{j-1}\pi_{h_i+2j'-1}^{(t)}(\up)\notag\\
    &\geq \frac{1-\epsilon^{1.05+o(1)}}{k^2}\sum_{i=1}^{k}\sum_{j=1}^{k}\Bigg[\prod_{i'=1}^{i-1}\bigg(\prod_{j'=1}^{k-1}\Big(1-\epsilon^{1.05+o(1)}-\widehat\pi_{h_{i'}+2j'}^{(t)}(\up)\Big)\bigg)\cdot\widehat\pi_{h_{i'}+2k}^{(t)}(\up)\Big(1-\epsilon^{1.05+o(1)}-\widehat\pi_{h_{i'}+2k+1}^{(t)}(\up)\Big)\Bigg]\notag\\
    &\hspace{8cm}\cdot \prod_{j'=1}^{j-1}\left(1-\epsilon^{1.05+o(1)}-\widehat\pi_{h_i+2j'}^{(t)}(\up)\right).
\end{align}
\normalsize
where the last line follows from  \eqref{eq:pi_h_up_Ax=0_c=0_phase2}, \eqref{eq:1-pi_h_up_c=x_phase2}, \eqref{eq:pi_h_down_x_phase2} in the proof of Lemma~\ref{lm:gradients_simplified_phase2_balanced} and our assumption that $\epsilon$ is small enough. And from \eqref{eq:pi_h_up_1<=Ax<=k-1_phase2}, \eqref{eq:pi_h_up_Ax=k_phase2} we know that there exists some constant $c>0$ such that for all $i'\in[i-1]$, $j'\in[k-1]$:
\begin{align}
    \widehat\pi_{h_{i'}+2j'}^{(t)}(\up)\leq\frac{\exp\left(-\left(1-\frac{c}{\ln^2 k}\right)a_{q,0}\right)}{(k-j')\exp\left(\frac{a_{p,1}^{(t)}-a_{p,3}^{(t)}}{k-1}\right)+\exp\left(-\left(1-\frac{c}{\ln^2 k}\right)a_{q,0}\right)}, \label{eq:pi_h_i'_2j'_up_T2} \\
    \widehat\pi_{h_{i'}+2k+1}^{(t)}(\up)\leq\frac{\exp\left(-\left(1-\frac{c}{\ln^2 k}\right)a_{q,0}\right)}{(k-i')\exp\left(\frac{a_{p,1}^{(t)}-a_{p,3}^{(t)}}{k-1}\right)+\exp\left(-\left(1-\frac{c}{\ln^2 k}\right)a_{q,0}\right)}, \label{eq:pi_h_i'_2k+1_up_T2}
\end{align}
for all $j'\in[j-1]$ (if $j>1$):
\begin{align}\label{eq:pi_h_i_2j'_up_T2}
    \widehat\pi_{h_i+2j'}^{(t)}(\up)\leq\frac{\exp\left(-\left(1-\frac{c}{\ln^2 k}\right)a_{q,0}\right)}{(k-j')\exp\left(\frac{a_{p,1}^{(t)}-a_{p,3}^{(t)}}{k-1}\right)+\exp\left(-\left(1-\frac{c}{\ln^2 k}\right)a_{q,0}\right)},
\end{align}
for all $i'\in[i-1]$:
\begin{align}\label{eq:pi_h_i'_2k_up_T2}
    1-\widehat\pi_{h_{i'}+2k}^{(t)}(\up)\leq\frac{k}{k+\exp\left(-\left(1-\frac{c}{\ln^2 k}\right)a_{q,0}+\frac{a_{p,3}^{(t)}}{k}\right)}.
\end{align}
From \eqref{eq:R_pi_T_2_lb}, \eqref{eq:pi_h_i'_2j'_up_T2}, \eqref{eq:pi_h_i'_2k+1_up_T2}, \eqref{eq:pi_h_i_2j'_up_T2}, \eqref{eq:pi_h_i'_2k_up_T2} and our choice of $\delta_2$ we know that for all $t\in[T_{2,1},T_2-1]$, the following must hold:
\begin{align}\label{eq:pi_h_up_T2_lb}
    &\max\left\{\frac{\exp\left(-\left(1-\frac{c}{\ln^2 k}\right)a_{q,0}\right)}{\exp\left(\frac{a_{p,1}^{(t)}-a_{p,3}^{(t)}}{k-1}\right)+\exp\left(-\left(1-\frac{c}{\ln^2 k}\right)a_{q,0}\right)},\frac{k}{k+\exp\left(-\left(1-\frac{c}{\ln^2 k}\right)a_{q,0}+\frac{a_{p,3}^{(t)}}{k}\right)}\right\}\notag\\
    &\geq \frac{\epsilon^{1.02+o(1)}}{k^L}
\end{align}
or otherwise $R(\pi^{(t)};\gP_2)>1-\delta_2$ for some $t\in[T_{2,1},T_2-1]$, which contradicts our training setting. From \eqref{eq:a_p_3_bound_phase2} and \eqref{eq:pi_h_up_T2_lb} we deduce that there eixsts constants $c_3,c_4>0$ such that 
\begin{align}
    \frac{a_{p,3}^{(T_2)}}{k}-\left(1-\frac{c_3}{\ln^2 k}\right)a_{q,0}\leq \ln \left(\frac{k^L}{\epsilon^{1.02+o(1)}}\right),\quad \frac{a_{p,1}^{(T_2)}}{k}\leq 2\ln \left(\frac{k^L}{\epsilon^{1.02+o(1)}}\right)+\frac{c_4}{\ln^2 k}a_{q,0}.
\end{align}
This gives \eqref{eq:a_p_3_ub_T2} and \eqref{eq:a_p_1_ub_T2}. From \eqref{eq:a_p_1_ub_T2}, \eqref{eq:a_p_1_increase_rate_phase2} and Lemma~\ref{lm:phase_1.2} we immediately deduce that \eqref{eq:a_p_1_vs_a_q_0_phase2} holds at step $s$ for any $s\in[T_1+1,T_2]$.

Moreover, from \eqref{eq:pi_h_up_T2_lb}, \eqref{eq:a_p_3<=} and \eqref{eq:overline_pi_2} we also know that for all $t\in[T_1,T_2-1]$:
\begin{align}\label{eq:1-overline_pi_2_up_lb_T2}
    1-\overline\pi_{2}^{(t)}(\up)\geq \frac{\epsilon^{1.02+o(1)}}{k^L}.
\end{align}
And from Lemma~\ref{lm:phase_1.2} and Induction Hypothesis III we also know that 
\begin{align}\label{eq:overline_pi_2_lb_T2}
    \overline{\pi}_{2}^{(t)}(\up)\geq \epsilon^{1.05+o(1)}.
\end{align}
By \eqref{eq:overline_pi_2_lb_T2}, \eqref{eq:1-overline_pi_2_up_lb_T2}, \eqref{eq:delta_a_p_1_simplified_balanced_phase2} and the fact that (c.f.~\eqref{eq:Q_2_i>=1/k})
$$Q_{2,i}^{(t)}(\up)\geq \frac{1+o(1)}{k},\,\,\forall i\in[k-1]$$ 
guaranteed by our policy computation in the proof of Lemma~\ref{lm:gradients_simplified_phase2_balanced},
we have 
\begin{align}\label{eq:delta_a_p_1_lb_T2}
    \forall t\in[T_1,T_2-1]:\quad \delta a_{p,1}^{(t)}\geq \min\left\{\epsilon^{1.05+o(1)}, \frac{\epsilon^{1.02+o(1)}}{k^L}\right\}\geq \frac{\epsilon^{1.1}}{k^L}.
\end{align}
\eqref{eq:delta_a_p_1_lb_T2} implies \eqref{eq:a_p_1_increase_rate_phase2} holds.
\eqref{eq:delta_a_p_1_lb_T2} combined with \eqref{eq:a_p_1_ub_T2} and Lemma~\ref{lm:phase_1.2} also yields \eqref{eq:T_2_bound}.


Finally, we show \eqref{eq:success_rate_generalization_phase2} holds for any full $k$-ary tree $\gT$ with depth $l\in[L]$ (that may not be a perfect $k$-ary tree). First note that by a similar policy computation as in the proof of Lemma~\ref{lm:gradients_simplified_phase2_balanced} we know that given any $k$-ary tree $\gT$ with depth $l\in[L]$, \eqref{eq:pi_h_up_Ax=0_c=0_phase2}, \eqref{eq:1-pi_h_up_c=x_phase2}, \eqref{eq:pi_h_down_x_phase2} still hold, i.e., 
\begin{align}\label{eq:small_prob_l}
    \text{When }c_h=0 \text{ and }|\gA_{h,\times}|=0:\quad \pi_h^{(T_2)}(\up)&\leq \epsilon^{1.05+o(1)},\\
    \text{when }c_h=\times:\quad 1-\pi_{h}^{(T_2)}(\up)&\leq \epsilon^{1.05+o(1)},\\
    \text{when }|\gA_{h,\times}|\in[k-1]:\quad \pi_{h}^{(T_2)}(j)&\leq \epsilon^{1.05+o(1)},\,\,\forall j\in\gA_{h,\times}.
\end{align}
And similar as how we compute \eqref{eq:pi_h_up_1<=Ax<=k-1_phase2}, \eqref{eq:pi_h_up_Ax=k_phase2} in the proof of Lemma~\ref{lm:gradients_simplified_phase2_balanced}, by \eqref{eq:a_p_3_bound_phase2_T1}, \eqref{eq:a_p_1_bound_phase2_T1} and our choice of $\delta_2$ we can compute that, for any $k$-ary tree $\gT$ with depth $l\in[L]$, we have
\begin{align}\label{eq:small_prob_l_2}
    \text{When } |\gA_{h,\times}|\in[k-1]:\quad \pi_h^{(T_2)}(\up)\leq \frac{\epsilon^{1.02+o(1)}}{k^L},\notag\\
    \text{When } |\gA_{h,\times}|=k:\quad 1-\pi_h^{(T_2)}(\up)\leq \frac{\epsilon^{1.02+o(1)}}{k^L}.
\end{align}
Similar as how we define $\gC_{2,i}^j$ in \eqref{eq:gC_2_i_j} and use it to lower bound $R(\pi^{(t)};\gP_2)$ in \eqref{eq:R_pi_T_2_lb}, here we define the event 
\begin{align}
    \gE(\gT;g)\coloneqq \{\text{the agent follows the DFS rule on $\gT$ until it first reaches } g\},
\end{align}
and let $\gV_{\text{leaf}}(\gT)$ denote the set of leaves of $\gT$. 
%
For any fixed goal node $g_\star\in \gV_{\mathrm{leaf}}(\gT)$, 
on the event $\gE(\gT;g_\star)$, the agent reaches the target goal node $g_\star$.
Therefore,
\begin{align}\label{eq:R_pi_T_2_lb_fixed_goal_phase2}
    R(\pi^{(T_2)};(\gT,g_\star))
    \geq
    \PP^{\pi^{(T_2)}}\left(\gE(\gT;g_\star)\right).
\end{align}
For each step $h$, let $\gF_h(g_\star)$ denote the event that the trajectory has
followed the DFS rule toward $g_\star$ up to step $h$. Equivalently,
$\gF_h(g_\star)$ is the prefix event induced by $\gE(\gT;g_\star)$ up to step $h$.
Let
\begin{align}
    \widehat\pi_h^{(T_2)}
    \coloneqq
    \pi_h^{(T_2)}\mid \gF_h(g_\star).
\end{align}
Conditioned on $\gF_h(g_\star)$, the one-step probability of deviating from the
DFS rule is
\begin{align}
    \eta_h(\gT;g_\star)
    \coloneqq
    \begin{cases}
        \widehat\pi_h^{(T_2)}(\up)
        +\sum_{i\in\gA_{h,\times}}\widehat\pi_h^{(T_2)}(i),
        & \text{if } c_h=0,\ |\gA_{h,\times}|\leq k-1,\\
        1-\widehat\pi_h^{(T_2)}(\up),
        & \text{if } c_h=\times
        \text{ or } (c_h=0 \text{ and } |\gA_{h,\times}|=k).
    \end{cases}
\end{align}
%

By \eqref{eq:small_prob_l} and \eqref{eq:small_prob_l_2}, for every $h$,
\begin{align}
    \EE[\eta_h(\gT;g_\star)]
    \leq
    \frac{\epsilon^{1.02+o(1)}}{k^L}.
\end{align}
Since a deviation from $\gE(\gT;g_\star)$ must occur at some step $h\leq H$,
we have
\begin{align}
    \PP^{\pi^{(T_2)}}\left(\gE(\gT;g_\star)^c\right)
    &\leq
    \sum_{h=1}^{H}
    \EE\left[\mathbbm{1}\{\gF_h(g_\star)\}\eta_h(\gT;g_\star)\right] \notag\\
    &\leq
    \sum_{h=1}^{H}
    \EE[\eta_h(\gT;g_\star)]  \leq
    H\cdot \frac{\epsilon^{1.02+o(1)}}{k^L}
    \leq
    \epsilon^{1.02+o(1)}
    \leq
    \epsilon,
\end{align}
where we used $H\lesssim k^L$. Therefore,
\begin{align}
    R(\pi^{(T_2)};(\gT,g_\star))
    \overset{\eqref{eq:R_pi_T_2_lb_fixed_goal_phase2}}\geq
    \PP^{\pi^{(T_2)}}\left(\gE(\gT;g_\star)\right)
    \geq
    1-\epsilon.
\end{align}
Since $g_\star\in\gV_{\mathrm{leaf}}(\gT)$ was arbitrary,
\eqref{eq:success_rate_generalization_phase2} follows.


\subsubsection{Proof of Lemma~\ref{lm:gradients_simplified_phase2_balanced}}\label{sec_app:gradients_simplified_phase2_balanced}
\paragraph{Policy computation.} To compute the gradients, we first compute the policies on trees with depth 2.
Note that under our setting,
on any perfect $k$-ary tree of depth $l$, $c_h=0$ if $h\leq l$, and if $c_h=\times$ at some step $h$, then $c_{h-1}=0$, and $c_{h+1}=0$ if the environment doesn't terminate at step $h$. 
Therefore, at any step $h\in[H]$, we have
\begin{align}\label{eq:N_h_times_ub}
    N_{h,\times}\leq\begin{cases}
        \max\{0,\frac{h-l}{2}\}, & \text{if } c_h=0,\\
        \max\{0,\frac{h-l+1}{2}\}, & \text{if } c_h=\times.
    \end{cases}
\end{align}
In addition, the level the agent is at a tree of depth $l$ (root is level 0, and leaves are level $l$) at step $h$ equals $\sum_{i=1}^{k} N_h(i)-N_h(\up)$. As a result, we have
\begin{align}
    0\leq \sum_{i=1}^{k} N_h(i)-N_h(\up)\leq l,\quad \sum_{i=1}^{k} N_h(i)+N_h(\up)= h-1.
\end{align}
The above two relations yield
\begin{align}\label{eq:N_h_up_sum_i_bound}
    \frac{h-1}{2}\leq\sum_{i=1}^{k} N_h(i)\leq \frac{h-1+l}{2},\quad \max\left\{0,\frac{h-1-l}{2}\right\}\leq N_h(\up)\leq \frac{h-1}{2}.
\end{align}
By \eqref{eq:N_h_times_ub}, \eqref{eq:xi_h_balanced} and Lemma~\ref{lm:phase_1.2}, we have
\begin{align}\label{eq:xi_h_balanced_phase2}
    \xi_h=\begin{cases}
        -\left(1+O\left(\frac{1}{\ln^2k}\right)\right)a_{q,0}, & \text{if } c_h=0,\\
        a_{q,\times}+O\left(\frac{k^2}{\ln(1/\epsilon)}\right), & \text{if } c_h=\times.
    \end{cases}
\end{align}
Let $\widetilde{\phi}_h\coloneqq (\widetilde{\phi}_h(1),\ldots,\widetilde{\phi}_h(k),\widetilde{\phi}_h(\up))^\top$ with
\begin{align}\label{eq:tilde_phi_h_balanced}
    \widetilde{\phi}_h(i)\coloneqq \frac{\varphi_h(i)}{e^{a_{b,0}}+|\gA_{h,\times}|e^{a_{b,1}}+(h-1-|\gA_{h,\times}|)e^{a_{b,2}}},\,\,\forall i\in[k],\quad \widetilde{\phi}_h(\up)\coloneqq \left(1+\frac{1}{k}\right)\xi_h.
\end{align}
Then by Lemma~\ref{lm:gradients} we have 
\begin{align}\label{eq:pi_h_simplified_balanced_phase2}
    \pi_h=\sm\left(\widetilde{\phi}_h\right).
\end{align}
In addition, when $|\gA_{h,\times}|=0$, for all $i\in[k]$, by Lemma~\ref{lm:gradients}, Lemma~\ref{lm:phase_1.2}, Induction Hypothesis III and \eqref{eq:N_h_up_sum_i_bound} we have
\begin{align}\label{eq:tilde_phi_h_i_Ax=0}
    \widetilde{\phi}_h(i)&=-\frac{a_{p,0}^{(t)}}{k}+\frac{1}{e^{a_{b,0}-a_{b,2}}}\left(N_h(\up)\frac{a_{p,\up}^{(t)}}{k}+\frac{a_{p,0}^{(t)}}{k}+(h-1-N_{h}(\up))a_{p,2}^{(t)}-N_h(i)\left(a_{p,1}^{(t)}+a_{p,2}^{(t)}\right)\right)\notag\\
    &\qquad +O\left(\frac{1}{\left(\ln (1/\epsilon)\right)^{1+o(1)}}\right)\notag\\
    &=-\frac{a_{p,0}^{(t)}}{k}+O\left(\frac{k}{\left(\ln (1/\epsilon)\right)^{1+O(1/N)}}\right)a_{p,1}^{(t)}.
\end{align}
By \eqref{eq:pi_h_simplified_balanced_phase2}, \eqref{eq:tilde_phi_h_i_Ax=0}, \eqref{eq:xi_h_balanced_phase2}, \eqref{eq:a_q_0_bound_T_1} we have 
\begin{align}\label{eq:pi_h_up_Ax=0_c=0_phase2}
    \text{When }c_h=0 \text{ and }|\gA_{h,\times}|=0:\quad \pi_h(\up)&= 
    \frac{1}{k}\exp\left(-\left(1+O\left(\frac{1}{\ln^2 k}\right)\right)a_{q,0}+\frac{a_{p,0}^{(t)}}{k}\right)\notag\\
    &\leq\epsilon^{1.05+O\left(1/\ln^2 k\right)}.
\end{align}
Note $|\gA_{h,\times}|=0$, as in \eqref{eq:gA_h_times} we define $\gA_{h,\times}=\emptyset$ if $c_h=\times$. Then by \eqref{eq:pi_h_simplified_balanced_phase2}, \eqref{eq:tilde_phi_h_i_Ax=0}, \eqref{eq:xi_h_balanced_phase2} and \eqref{eq:a_q_x_bound_T_1} we have 
\begin{align}\label{eq:1-pi_h_up_c=x_phase2}
    \text{When }c_h=\times:\,\, 1-\pi_{h}(\up)&=
    k\exp\left(-\left(1+\frac{1}{k}+O\left(\frac{1}{\left(\ln(1/\epsilon)\right)^{1+o(1)}}\right)\right)a_{q,\times}-\frac{a_{p,0}^{(t)}}{k}\right)\notag\\
    &\leq \epsilon^{1.05+O\left(1/k\right)}.
\end{align}
Similarly, when $|\gA_{h,\times}|\geq 1$, by Lemma~\ref{lm:gradients}, Lemma~\ref{lm:phase_1.2} and Induction Hypothesis III, we have
\begin{align}\label{eq:tilde_phi_h_i_Ax>=1}
    \widetilde{\phi}_h(i)&=a_{p,2}^{(t)}-\frac{1}{|\gA_{h,\times}|}\mathbbm{1}\{i\in\gA_{h,\times}\}(a_{p,1}^{(t)}+a_{p,2}^{(t)})\notag\\
    &\quad +\frac{1}{e^{a_{b,1}^{(t)}-a_{b,0}^{(t)}}}\left(-\frac{a_{p,0}^{(t)}}{k|\gA_{h,\times}|}-a_{p,2}^{(t)}+\frac{1}{|\gA_{h,\times}|}\mathbbm{1}\{i\in\gA_{h,\times}\}(a_{p,1}^{(t)}+a_{p,2}^{(t)})\right)+O\left(\frac{1}{\left(\ln (1/\epsilon)\right)^{1+o(1)}}\right)\notag\\
    &=a_{p,2}^{(t)}-\frac{1}{|\gA_{h,\times}|}\mathbbm{1}\{i\in\gA_{h,\times}\}(a_{p,1}^{(t)}+a_{p,2}^{(t)})+O\left(\frac{1}{\left(\ln (1/\epsilon)\right)^{1+O(1/N)}}\right)a_{p,1}^{(t)}.
\end{align}
Especially, when $|\gA_{h,\times}|=k$, by \eqref{eq:tilde_phi_h_i_Ax>=1} and \eqref{eq:a_p3} we have 
\begin{align}\label{eq:tilde_phi_h_i_Ax=k}
    \forall i\in[k]:\quad \widetilde{\phi}_h(i)&=-\frac{a_{p,3}^{(t)}}{k}+\frac{1}{e^{a_{b,1}^{(t)}-a_{b,0}^{(t)}}}\left(-\frac{a_{p,0}^{(t)}}{k^2}+\frac{a_{p,3}^{(t)}}{k}\right)+O\left(\frac{1}{\left(\ln (1/\epsilon)\right)^{1+o(1)}}\right)\notag\\
    &=-\frac{a_{p,3}^{(t)}}{k}+O\left(\frac{1}{k\left(\ln (1/\epsilon)\right)^{1+O(1/N)}}\right)a_{p,1}^{(t)}.
\end{align}
Then by \eqref{eq:pi_h_simplified_balanced_phase2}, \eqref{eq:xi_h_balanced_phase2}, \eqref{eq:a_p3} and the above two expressions we have when $|\gA_{h,\times}|\in[k-1]$: 
\begin{align}\label{eq:pi_h_up_1<=Ax<=k-1_phase2}
    \pi_{h}(\up)&=\frac{\exp\left(-\left(1+O\left(\frac{1}{\ln^2 k}\right)\right)a_{q,0}\right)}{(k-|\gA_{h,\times}|)\exp\left(\frac{a_{p,1}^{(t)}-a_{p,3}^{(t)}}{k-1}\right)+\exp\left(-\left(1+O\left(\frac{1}{\ln^2 k}\right)\right)a_{q,0}\right)}\notag\\
    &\overset{\eqref{eq:a_p_3_bound_phase2}}\leq \frac{\exp\left(-\left(1+O\left(\frac{1}{\ln^2 k}\right)\right)a_{q,0}\right)}{(k-|\gA_{h,\times}|)\exp\left(\frac{a_{p,1}^{(t)}}{2(k-1)}-\left(1+O\left(\frac{1}{\ln^2 k}\right)\right)a_{q,0}\right)+\exp\left(-\left(1+O\left(\frac{1}{\ln^2 k}\right)\right)a_{q,0}\right)}\overset{\eqref{eq:a_p_1_2_bound_T_1}}\notag\\
    &\leq \epsilon^{\frac{1.05}{2}+o(1)},
\end{align}
and 
\begin{align}\label{eq:pi_h_up_Ax=k_phase2}
    \text{When }|\gA_{h,\times}|=k:\quad \pi_{h}(\up)&=\frac{\exp\left(-\left(1+O\left(\frac{1}{\ln^2 k}\right)\right)a_{q,0}+\frac{a_{p,3}^{(t)}}{k}\right)}{k+\exp\left(-\left(1+O\left(\frac{1}{\ln^2 k}\right)\right)a_{q,0}+\frac{a_{p,3}^{(t)}}{k}\right)}.
\end{align}
And when $|\gA_{h,\times}|\in[k-1]$:
\begin{align}
    \forall j\in\gA_{h,\times}:\quad \pi_{h}(j)&=\frac{1}{k-|\gA_{h,\times}|}\exp\left(-\left(1+O\left(\frac{1}{\left(\ln (1/\epsilon)\right)^{1+O(1/N)}}\right)\right)\frac{a_{p,1}^{(t)}+a_{p,2}^{(t)}}{|\gA_{h,\times}|}\right),
\end{align}
which, combined with \eqref{eq:a_p_1_2_bound_T_1} and Induction Hypothesis III, further indicates  for all $j\in\gA_{h,\times}$: 
\begin{align}\label{eq:pi_h_down_x_phase2}
    \text{When }|\gA_{h,\times}|=k-1:\quad \pi_{h}(j)&=\exp\left(-\left(1+O\left(\frac{1}{\left(\ln (1/\epsilon)\right)^{1+O(1/N)}}\right)\right)\frac{a_{p,1}^{(t)}+a_{p,2}^{(t)}}{k-1}\right)\notag\\
    &\leq\epsilon^{1.05+O\left(1/k\right)},\notag\\
    \text{when }|\gA_{h,\times}|\in[k-2]:\quad \pi_{h}(j)&\leq\epsilon^{\frac{C}{k}}\exp\left(-\left(1+O\left(\frac{1}{\left(\ln (1/\epsilon)\right)^{1+O(1/N)}}\right)\right)\frac{a_{p,1}^{(t)}+a_{p,2}^{(t)}}{k-1}\right)
\end{align}
for some constant $C>0$.
When $|\gA_{h,\times}|=k$, from \eqref{eq:tilde_phi_h_i_Ax>=1} we can see that 
\begin{align}\label{eq:pi_h_j_in_Ax=k_phase2}
    \forall j\in[k]:\quad \pi_h(j)&=(1+o(1/k^2))\frac{1-\pi_h(\up)}{k}\overset{\eqref{eq:pi_h_up_Ax=k_phase2}}=\frac{1}{k+\exp\left(-\left(1+O\left(\frac{1}{\ln^2 k}\right)\right)a_{q,0}+\frac{a_{p,3}^{(t)}}{k}\right)}.
\end{align}

\paragraph{Gradient computation.} 
For all $i\in[k]$, we define events
\begin{align}
    \gE_{2,i}\coloneqq \{\text{the environment has not terminated at step } h_i, l=2\},
\end{align}
where $h_i$ is defined in \eqref{eq:h_i}. We also define $h_{k+1}=+\infty$.
Let
\begin{align}\label{eq:pi_2_i_up_and_Q_2_i_up}
    \forall i\in[k-1]:\quad\pi_{2,i}(\up)\coloneqq \pi_{h_{i}+2k}(\up)|\overline{\gE}_{2,i}, \quad Q_{2,i}^{(t)}(\up)\coloneqq Q_{h_{i}+2k}^{(t)}(\up)|\overline{\gE}_{2,i},\notag\\
    \pi_{2,k}(\up)\coloneqq 0, \quad Q_{2,k}^{(t)}(\up)\coloneqq 0.
\end{align}
Note that $\pi_{2,i}(\up)$ and $Q_{2,i}^{(t)}(\up)$ are well-defined because they are invariant to the order of children of $n_{h_j}$ ($j\in[i]$) and the root visited under the balanced distribution.
We also define the following events for all $i',i\in[k-1]$ and $j\in\{0,1,\ldots,k-1\}$:
\begin{align}
    \overline{\gE}_{2,i}^{i,j}&\coloneqq \{\text{the agent traverses all children of $n_{h_m}$ for all $m\in[i-1]$ (if $i>1$),} \notag\\
    &\qquad\text{then visits $j$ children of $n_{h_i}$ and arrives at $n_{h_i}$ at step $h_i+2j$, $l=2$}\},\notag\\
    \forall i'<i:\quad \overline{\gE}_{2,i}^{i',j}&\coloneqq \{\text{the agent traverses all children of $n_{h_m}$ for all $m\in[i]\setminus\{i'\}$}, \notag\\
    &\qquad\text{visits $j$ children of $n_{h_{i'}}$, and arrives at $n_{h_i}$ at step $h_i+2k$, $l=2$}\},
\end{align}
and $\forall i,j\in[k]$:
\begin{align}\label{eq:overline_gE_2_i_times_j}
    \overline{\gE}_{2,i}^{\times,j}&\coloneqq \{\text{the agent traverses all children of $n_{h_m}$ for all $m\in[i-1]$ (if $i>1$),} \notag\\
    &\text{then visits $j$ children of $n_{h_i}$ and arrives at the $j$-th child of $n_{h_i}$ visited at step $h_i+2j-1$, $l=2$}\}.
\end{align}

For any sequence of random variables $\{x_h\}_{h\in[H]}$ and policy $\pi$, we have
\begin{align}\label{eq:EE_sum_x_h_l=2}
    \EE_{\pi}\left[\sum_{h=1}^H x_h\Big|l=2\right]=\EE_{\pi}\left[x_1|l=2\right]+\sum_{i=1}^k \PP^{\pi}\left(\gE_{2,i}\right)\EE_{\pi}\left[\sum_{h=h_i}^{h_{i+1}-1} x_h\Big|l=2,\gE_{2,i}\right].
\end{align}
And for any $\pi=\pi^{(t)}$ ($t\in [T_1,s-1]$), we have by our policy computation (c.f.~\eqref{eq:pi_h_up_Ax=0_c=0_phase2}, \eqref{eq:1-pi_h_up_c=x_phase2}, \eqref{eq:pi_h_up_1<=Ax<=k-1_phase2}, \eqref{eq:pi_h_up_Ax=k_phase2}, \eqref{eq:pi_h_down_x_phase2}) we have
\begin{align}\label{eq:PP_gE_2_i}
    \PP^{\pi}\left(\gE_{2,1}\right)&=1-\pi_1(\up)\overset{\eqref{eq:pi_h_up_Ax=0_c=0_phase2}}\geq 1-\epsilon^{1.05+O\left(1/\ln^2 k\right)},\notag\\
    \forall i\in\{2,\cdots,k\}:\quad \PP^{\pi}\left(\gE_{2,i}\right)
    &=(1+o(\epsilon^{0.5}))\frac{k-i+1}{k}\prod_{j=1}^{i-1}\pi_{2,j}(\up).
\end{align}

With the above preparations, we are ready to simplify the gradients in Lemma~\ref{lm:gradients}.
All expectations here are conditioned on tree depth $l=2$ and policy $\pi^{(t)}$ unless otherwise specified, i.e., $\EE[\cdot]=\EE_{\pi^{(t)}}[\cdot|l=2]$. We also use $\PP^{(t)}(\cdot)$ as a shorthand for $\PP^{\pi^{(t)}}(\cdot)$.

We first compute $\delta a_{p,3}^{(t)}$. From \eqref{eq:e_ab_0_lb_T_1}, \eqref{eq:ea_b_1_lb_T_1}, \eqref{eq:e_ab_2_lb_T_1} and \eqref{eq:delta_a_p_3_simplified_balanced} we know that 
\small
\begin{align}\label{eq:delta_a_p_3_simplified_balanced_intermediate}
    \delta a_{p,3}^{(t)}&=\frac{1+o(1/k^2)}{k}\EE\left[\sum_{h=1}^H\mathbbm{1}\{c_h=0,|\gA_{h,\times}|\geq 1\}\pi_h^{(t)}(\up)\left(Q_h^{(t)}(\up)-G_h^{(t)}\right)\right]\notag\\
    &\quad +(1+o(1/k^2))\EE\left[\sum_{h=1}^H\mathbbm{1}\{c_h=0,|\gA_{h,\times}|=0\}\pi_h^{(t)}(\up)\frac{N_h(j)}{e^{a_{b,0}-a_{b,2}}}\left(Q_h^{(t)}(\up)-G_h^{(t)}\right)\Big|j\in[k]\right]\notag\\
    &\quad +(1+o(1/k^2))\EE\left[\sum_{h=1}^H\mathbbm{1}\{c_h=\times\}\pi_h^{(t)}(\up)\left(1-\pi_h^{(t)}(\up)\right)\frac{N_h(j)}{e^{a_{b,0}-a_{b,2}}}Q_h^{(t)}(\up)\Big|j\in[k]\right].
\end{align}
\normalsize
Applying \eqref{eq:EE_sum_x_h_l=2} to the first term of \eqref{eq:delta_a_p_3_simplified_balanced_intermediate}, and using Lemma~\ref{lm:phase_1.2} and Induction Hypothesis III, 
 we have 
\begin{align}\label{eq:delta_a_p_3_simplified_balanced_intermediate_first_term}
    &\EE\left[\sum_{h=1}^H\mathbbm{1}\{c_h=0,|\gA_{h,\times}|\geq 1\}\pi_h^{(t)}(\up)\left(Q_h^{(t)}(\up)-G_h^{(t)}\right)\right]\notag\\
    &=(1+o(1/k^2))\sum_{i=1}^k \PP\left(\gE_{2,i}\right)\sum_{j=1}^{k}\frac{(k-i+1)k-j}{(k-i+1)k}\EE\left[\pi_{h_i+2j}^{(t)}(\up)\left(Q_{h_i+2j}^{(t)}(\up)-G_{h_i+2j}^{(t)}\right)\Big|\overline{\gE}_{2,i}^{i,j}\right].
\end{align}
By our policy computation (c.f.~\eqref{eq:pi_h_up_Ax=0_c=0_phase2}, \eqref{eq:1-pi_h_up_c=x_phase2}, \eqref{eq:pi_h_up_1<=Ax<=k-1_phase2}, \eqref{eq:pi_h_up_Ax=k_phase2}, \eqref{eq:pi_h_down_x_phase2}) and the definition of $\pi_{2,i}(\up)$ and $Q_{2,i}(\up)$ (c.f.~\eqref{eq:pi_2_i_up_and_Q_2_i_up}), we have for all $i\in[k]$ (note $\overline{\gE}_{2,i}=\overline{\gE}_{2,i}^{i,k}$):
\begin{align}\label{eq:pi_up*(Q_up-G_up)_k}
    \EE\left[\pi_{h_i+2k}^{(t)}(\up)\left(Q_{h_i+2k}^{(t)}(\up)-G_{h_i+2k}^{(t)}\right)\Big|\overline{\gE}_{2,i}\right]=
    \pi_{2,i}^{(t)}(\up)\left(1-\pi_{2,i}^{(t)}(\up)\right)Q_{2,i}^{(t)}(\up).
\end{align}
By our policy computation we can also obtain  (recall 
$\prod_{i=a}^b x_i=1$ if $a>b$):
\begin{align}\label{eq:Q_2_i_up}
   \forall i\in[k-1]:\quad Q_{2,i}^{(t)}(\up)=\frac{1+o(\epsilon^{0.5})}{k-i}\sum_{j=1}^{k-i}\left(\prod_{m=1}^{j-1}\pi_{2,i+m}^{(t)}(\up)\right),
\end{align}
for all $i\in[k],j\in\{0,1,\ldots,k-1\}$:
\begin{align}\label{eq:G_h_i+2j}
    \EE\left[G_{h_i+2j}^{(t)}\Big|\overline{\gE}_{2,i}^{i,j}\right]&=(1+o(\epsilon^{0.5}))
    \left(\sum_{m=1}^{k-j}\frac{1}{(k-i+1)k-j}
    +\frac{(k-i)k}{(k-i+1)k-j}\pi_{2,i}^{(t)}(\up)Q_{2,i}^{(t)}(\up)\right),
\end{align}
and
\begin{align}\label{eq:Q_h_k+2j_up=0_gE_2_1}
    \forall j\in[k-1]:\quad \EE\left[Q_{h_k+2j}^{(t)}(\up)\Big|\overline{\gE}_{2,k}^{k,j}\right]=0,
\end{align}
for all $i\in[k],j\in\{0,1,\ldots,k-1\}$:
\begin{align}\label{eq:Q_h_i+2j_up_overline_gE_2_i}
    \EE\left[Q_{h_i+2j}(\up)\Big|\overline{\gE}_{2,i}^{i,j}\right]&=
    \frac{(1+o(\epsilon^{0.5}))k}{(k-i+1)k-j}
    \sum_{j'=1}^{k-i}\left(\prod_{m=1}^{j'-1}\pi_{h_{i+m}+2k}^{(t)}(\up)|\overline{\gE}_{2,i+m}^{i,j}\right).
\end{align}
Note that 
\begin{align}\label{eq:N_h_i+2k_times_overline_gE_2_i}
    \forall i\in[k-1]:\quad N_{h_i+2k,\times}|\overline{\gE}_{2,i}=ik,\quad h_i+2k|\overline{\gE}_{2,i}=i(2k+2);
\end{align}
and for all $i'\in[k-1]$, $2\leq i\leq k-1$, $j\in\{0,1,\ldots,k-1\}$, $i'<i$:
\begin{align}\label{eq:N_h_i+2k_times_overline_gE_2_i_i_j}
    N_{h_{i}+2k,\times}|\overline{\gE}_{2,i}^{i',j}=(i-1)k+j,\quad h_i+2k|\overline{\gE}_{2,i}^{i',j}=i(2k+2)-2(k-j).
\end{align}
By \eqref{eq:N_h_i+2k_times_overline_gE_2_i}, \eqref{eq:N_h_i+2k_times_overline_gE_2_i_i_j} and \eqref{eq:xi_h_balanced} we have for all $i\in[k-1]$:
\begin{align}
    \xi_{h_i+2k}|\overline{\gE}_{2,i}=\frac{ke^{-2a_{c,0}}a_{q,\times}-(k+2)a_{q,0}}{ke^{-2a_{c,0}}+k+2},
\end{align}
and for all $2\leq i\leq k-1$, $i'<i$ and $j\in\{0,1,\ldots,k-1\}$:
\begin{align}\label{eq:xi_h_i+2k_overline_gE_2_i_i_j}
    \xi_{h_i+2k}|\overline{\gE}_{2,i}^{i',j}&=\frac{\left(k-\frac{k-j}{i}\right)e^{-2a_{c,0}}a_{q,\times}-\left(k+2-\frac{k-j}{i}\right)a_{q,0}}{\left(k-\frac{k-j}{i}\right)e^{-2a_{c,0}}+k+2-\frac{k-j}{i}}\notag\\
    &=\xi_{h_i+2k}|\overline{\gE}_{2,i}-\frac{2\frac{k-j}{i}e^{-2a_{c,0}}}{\left(ke^{-2a_{c,0}}+k+2\right)\left(\left(k-\frac{k-j}{i}\right)e^{-2a_{c,0}}+k+2-\frac{k-j}{i}\right)}\left(a_{q,0}+a_{q,\times}\right).
\end{align}
From \eqref{eq:N_h_i+2k_times_overline_gE_2_i}, \eqref{eq:tilde_phi_h_i_Ax=k}, \eqref{eq:pi_h_simplified_balanced_phase2} 
we know that 
\begin{align}\label{eq:pi_2_i_up_equal}
    \forall i\in[k-1]:\quad \pi_{2,i}^{(t)}(\up)=\left(1+O\left(\frac{1}{\left(\ln (1/\epsilon)\right)^{1+o(1)}}\right)\right)\overline{\pi}_{2}^{(t)}(\up),
\end{align}
where $\overline{\pi}_{2}^{(t)}(\up)$ is defined in \eqref{eq:overline_pi_2}.
Then from \eqref{eq:pi_2_i_up_equal}, \eqref{eq:Q_2_i_up} 
we deduce
\begin{align}\label{eq:Q_2_i_up_simplified_balanced}
    \forall i\in[k-1]:\quad Q_{2,i}^{(t)}(\up)=\frac{1+o(1/k)}{k-i}\frac{1-\left(\overline{\pi}_{2}^{(t)}(\up)\right)^{k-i}}{1-\overline{\pi}_{2}^{(t)}(\up)},
\end{align}
and from \eqref{eq:pi_2_i_up_equal}, \eqref{eq:PP_gE_2_i} we deduce
\begin{align}\label{eq:PP_gE_2_i_simplified_balanced}
    \forall i\in[k]:\quad \PP\left(\gE_{2,i}\right)=(1+o(1/k^2))\frac{k-i+1}{k}\left(\overline{\pi}_{2}^{(t)}(\up)\right)^{i-1}.
\end{align}
Further, by \eqref{eq:xi_h_i+2k_overline_gE_2_i_i_j}, \eqref{eq:tilde_phi_h_i_Ax=k}, \eqref{eq:pi_h_simplified_balanced_phase2}, Lemma~\ref{lm:phase_1.2} and Induction Hypothesis III we have for all $2\leq i\leq k-1$, $i'<i$ and $j\in\{0,1,\ldots,k-1\}$:
\begin{align}\label{eq:pi_h_i+2k_overline_gE_2_i_i_j<=pi_2_i}
    \pi_{h_i+2k}^{(t)}(\up)|\overline{\gE}_{2,i}^{i',j}<\overline{\pi}_{2}^{(t)}(\up),
\end{align}
and there exists a (large enough) constant $C>0$ such that if 
\begin{align}\label{eq:a_p_3_condition_1}
    -\left(1-\frac{1}{\left(\ln (1/\epsilon)\right)^{1-C/N}}\right)\frac{a_{p,3}^{(t)}}{k}+4.1\ln k\leq -\left(1+\frac{1}{k}\right)a_{q,0},
\end{align}
then for all $2\leq i\leq k-1$, $i'<i$ and $j\in\{0,1,\ldots,k-1\}$:
\begin{align}\label{eq:pi_h_i+2k_overline_gE_2_i_i_j=pi_2_i_if}
    \pi_{h_i+2k}^{(t)}(\up)|\overline{\gE}_{2,i}^{i',j}=(1+o(1/k^3))\pi_{2,i}^{(t)}(\up).
\end{align}
And from \eqref{eq:pi_h_i+2k_overline_gE_2_i_i_j<=pi_2_i}, \eqref{eq:pi_h_i+2k_overline_gE_2_i_i_j=pi_2_i_if}, \eqref{eq:Q_2_i_up} and \eqref{eq:Q_h_i+2j_up_overline_gE_2_i} we know that for all $i\in[k-1],j\in\{0,1,\ldots,k-1\}$:
\begin{align}\label{eq:Q_h_i+2j_up_overline_gE_2_i_i_j<=Q_2_i}
    \EE\left[Q_{h_i+2j}^{(t)}(\up)\Big|\overline{\gE}_{2,i}^{i,j}\right]<
    \frac{(k-i)k}{(k-i+1)k-j} 
    Q_{2,i}^{(t)}(\up),
\end{align}
and if \eqref{eq:a_p_3_condition_1} holds, then for all $i\in[k-1],\,j\in\{0,1,\ldots,k-1\}$: 
\begin{align}\label{eq:Q_h_i+2j_up_overline_gE_2_i_i_j=Q_2_i_if}
    \EE\left[Q_{h_i+2j}^{(t)}(\up)\Big|\overline{\gE}_{2,i}^{i,j}\right]= (1+o(1/k^2))\frac{(k-i)k}{(k-i+1)k-j}Q_{2,i}^{(t)}(\up).
\end{align}
Therefore, by \eqref{eq:Q_h_i+2j_up_overline_gE_2_i_i_j<=Q_2_i} and \eqref{eq:G_h_i+2j}, \eqref{eq:Q_h_k+2j_up=0_gE_2_1} we have for all $i\in[k],j\in\{0,1,\ldots,k-1\}$ (recall we define $Q_{2,k}^{(t)}(\up)=0$ in \eqref{eq:pi_2_i_up_and_Q_2_i_up}):
\begin{align}\label{eq:Q_h_i+2j_up-G_ub}
    &\EE\left[Q_{h_i+2j}^{(t)}(\up)-G_{h_i+2j}^{(t)}\Big|\overline{\gE}_{2,i}^{i,j}\right]\notag\\
    &\leq -(1+o(1/k))\frac{k-j}{(k-i+1)k-j}+(1+o(1/k))\frac{(k-i)k}{(k-i+1)k-j}\left(1-\pi_{2,i}^{(t)}(\up)\right)Q_{2,i}^{(t)}(\up)\notag\\
    &=-(1+o(1/k))\frac{k-j}{(k-i+1)k-j}+(1+o(1/k))\frac{k}{(k-i+1)k-j}\left(1-\left(\pi_{2,i}^{(t)}(\up)\right)^{k-i}\right),
\end{align}
where the last line follows from \eqref{eq:Q_2_i_up_simplified_balanced} and \eqref{eq:pi_2_i_up_equal}.
We also have for all $i\in[k],j\in\{0,1,\ldots,k-1\}$:
\begin{align}\label{eq:Q_h_i+2j_up-G_lb}
    \EE\left[Q_{h_i+2j}^{(t)}(\up)-G_{h_i+2j}^{(t)}\Big|\overline{\gE}_{2,i}^{i,j}\right]\geq -1,
\end{align}
and when \eqref{eq:a_p_3_condition_1} holds, by \eqref{eq:Q_h_i+2j_up_overline_gE_2_i_i_j=Q_2_i_if} and \eqref{eq:G_h_i+2j} we have for all $i\in[k-1],j\in\{0,1,\ldots,k-1\}$:
\begin{align}\label{eq:Q_h_i+2j_up-G_lb_if}
    \EE\left[Q_{h_i+2j}^{(t)}(\up)-G_{h_i+2j}^{(t)}\Big|\overline{\gE}_{2,i}^{i,j}\right]=-(1+o(1))\frac{k-j}{(k-i+1)k-j}.
\end{align}
Also note that if \eqref{eq:a_p_3_condition_1} holds, then
\begin{align}\label{eq:1-pi_2_i_up_if}
    \forall i\in[k-1]:\quad 1-\pi_{2,i}^{(t)}(\up)=o(1/k^3),\quad Q_{2,i}^{(t)}(\up)\overset{\eqref{eq:Q_2_i_up}}=1+o(1),
\end{align}
and if \eqref{eq:a_p_3_condition_1} does not hold, then there exists a constant $c>0$ such that
\begin{align}\label{eq:1-pi_2_i_up_if_not}
    \forall i\in[k-1]:\quad 1-\pi_{2,i}^{(t)}(\up)\geq \frac{1}{\exp\left(\frac{c}{\ln^2 k}a_{q,0}\right)}.
\end{align}
Therefore,
\begin{itemize}
    \item if \eqref{eq:a_p_3_condition_1} does not hold, then by \eqref{eq:Q_h_i+2j_up-G_ub}, \eqref{eq:Q_h_i+2j_up-G_lb}, \eqref{eq:1-pi_2_i_up_if_not}, \eqref{eq:pi_h_up_1<=Ax<=k-1_phase2}, \eqref{eq:pi_h_up_Ax=k_phase2} and \eqref{eq:pi_up*(Q_up-G_up)_k} we know that for all $i\in[k-1]$:
\begin{align}\label{eq:delta_a_p_3_main_term_if_not}
    &\sum_{j=1}^{k}\frac{(k-i+1)k-j}{(k-i+1)k}\EE\left[\pi_{h_i+2j}^{(t)}(\up)\left(Q_{h_i+2j}^{(t)}(\up)-G_{h_i+2j}^{(t)}\right)\Big|\overline{\gE}_{2,i}^{i,j}\right]\notag\\
    &=(1+o(1/k^2))\frac{k-i}{k-i+1}\pi_{2,i}^{(t)}(\up)\left(1-\pi_{2,i}^{(t)}(\up)\right)Q_{2,i}^{(t)}(\up);
\end{align}
\item if \eqref{eq:a_p_3_condition_1} holds, then by \eqref{eq:Q_h_i+2j_up-G_ub}, \eqref{eq:Q_h_i+2j_up-G_lb}, \eqref{eq:Q_h_i+2j_up-G_lb_if}, \eqref{eq:1-pi_2_i_up_if}, \eqref{eq:pi_h_up_1<=Ax<=k-1_phase2}, \eqref{eq:pi_h_up_Ax=k_phase2} and \eqref{eq:pi_up*(Q_up-G_up)_k} we know that for all $i\in[k-1]$:
\begin{align}\label{eq:delta_a_p_3_main_term_if}
    &\sum_{j=1}^{k}\frac{(k-i+1)k-j}{(k-i+1)k}\EE\left[\pi_{h_i+2j}^{(t)}(\up)\left(Q_{h_i+2j}^{(t)}(\up)-G_{h_i+2j}^{(t)}\right)\Big|\overline{\gE}_{2,i}^{i,j}\right]\notag\\
    &=(1+o(1))\frac{k-i}{k-i+1}\left(1-\pi_{2,i}^{(t)}(\up)\right)-\sum_{j=1}^{k-1}\frac{\exp\left(-\left(1+O\left(\frac{1}{\ln^2 k}\right)\right)a_{q,0}\right)}{(k-j)\exp\left(\frac{a_{p,1}^{(t)}-a_{p,3}^{(t)}}{k-1}\right)}\frac{k-j}{(k-i+1)k}\notag\\
    &=\frac{k-i}{k-i+1}\left((1+o(1))\pi_{2,i}^{(t)}(\up)-\frac{k-1}{k(k-i)\exp\left(\frac{a_{p,1}^{(t)}-a_{p,3}^{(t)}}{k-1}+\left(1+O\left(\frac{1}{\ln^2 k}\right)\right)a_{q,0}\right)}\right).
\end{align}
\end{itemize}
By Lemma~\ref{lm:phase_1.2} and Induction Hypothesis III, we could also uniformly write the two cases above as ($i\in[k-1]$):
\begin{align}
    &\sum_{j=1}^{k}\frac{(k-i+1)k-j}{(k-i+1)k}\EE\left[\pi_{h_i+2j}^{(t)}(\up)\left(Q_{h_i+2j}^{(t)}(\up)-G_{h_i+2j}^{(t)}\right)\Big|\overline{\gE}_{2,i}^{i,j}\right]\notag\\
    &=\frac{k-i}{k-i+1}\pi_{2,i}^{(t)}(\up)\left((1+o(1))\left(1-\pi_{2,i}^{(t)}(\up)\right)-\frac{k-1}{k(k-i)\exp\left(\frac{a_{p,1}^{(t)}-a_{p,3}^{(t)}}{k-1}+\left(1+O\left(\frac{1}{\ln^2 k}\right)\right)a_{q,0}\right)}\right)Q_{2,i}^{(t)}(\up).
\end{align}
When $i=k$, by our gradient computation, \eqref{eq:pi_2_i_up_and_Q_2_i_up}, \eqref{eq:Q_h_k+2j_up=0_gE_2_1}, \eqref{eq:G_h_i+2j} we have
\begin{align}\label{eq:delta_a_p_3_main_term_if_k}
    \sum_{j=1}^{k}\frac{k-j}{k}\EE\left[\pi_{h_k+2j}^{(t)}(\up)\left(Q_{h_k+2j}^{(t)}(\up)-G_{h_k+2j}\right)\Big|\overline{\gE}_{2,k}^{k,j}\right]
    &=-\sum_{j=1}^{k-1}\frac{\exp\left(-\left(1+O\left(\frac{1}{\ln^2 k}\right)\right)a_{q,0}\right)}{(k-j)\exp\left(\frac{a_{p,1}^{(t)}-a_{p,3}^{(t)}}{k-1}\right)}\frac{k-j}{k}\notag\\
    &=-\frac{k-1}{k\exp\left(\frac{a_{p,1}^{(t)}-a_{p,3}^{(t)}}{k-1}+\left(1+O\left(\frac{1}{\ln^2 k}\right)\right)a_{q,0}\right)},
\end{align}
and 
\begin{align}\label{eq:Q_2_i>=1/k}
    \forall i\in[k-1]:\quad Q_{2,i}^{(t)}(\up)\geq \frac{1+o(1)}{k}.
\end{align}
Therefore, by \eqref{eq:delta_a_p_3_simplified_balanced_intermediate_first_term},   \eqref{eq:delta_a_p_3_main_term_if}, \eqref{eq:delta_a_p_3_main_term_if_not}, \eqref{eq:pi_2_i_up_equal}, \eqref{eq:delta_a_p_3_main_term_if_k} and \eqref{eq:Q_2_i>=1/k} we can rewrite the first term in \eqref{eq:delta_a_p_3_simplified_balanced_intermediate} as follows:
\begin{align}\label{eq:delta_a_p_3_first_term}
    &\frac{1}{k}\EE\left[\sum_{h=1}^H\mathbbm{1}\{c_h=0,|\gA_{h,\times}|\geq 1\}\pi_h^{(t)}(\up)\left(Q_h^{(t)}(\up)-G_h^{(t)}\right)\right]\notag\\
    &=\frac{1}{k}\Bigg[\sum_{i=1}^{k-1}\PP\left(\gE_{2,i}\right)\frac{k-i}{k-i+1}\pi_{2,i}^{(t)}(\up)\Bigg((1+o(1))\left(1-\pi_{2,i}^{(t)}(\up)\right)\notag\\
    &-\frac{k-1}{k(k-i)\exp\left(\frac{a_{p,1}^{(t)}-a_{p,3}^{(t)}}{k-1}+\left(1+O\left(\frac{1}{\ln^2 k}\right)\right)a_{q,0}\right)}\Bigg)Q_{2,i}^{(t)}(\up)\Bigg]-\frac{(k-1)\PP\left(\gE_{2,k}\right)}{k^2\exp\left(\frac{a_{p,1}^{(t)}-a_{p,3}^{(t)}}{k-1}+\left(1+O\left(\frac{1}{\ln^2 k}\right)\right)a_{q,0}\right)}\notag\\
    &\overset{\eqref{eq:PP_gE_2_i_simplified_balanced}}=\frac{1}{k}\sum_{i=1}^{k-1}\frac{k-i}{k}\left(\overline{\pi}_{2}^{(t)}(\up)\right)^{i}\Bigg((1+o(1))\left(1-\overline{\pi}_{2}^{(t)}(\up)\right)\notag\\
    &\qquad\qquad-\frac{1}{(k-i)\exp\left(\frac{a_{p,1}^{(t)}-a_{p,3}^{(t)}}{k-1}+\left(1+O\left(\frac{1}{\ln^2 k}\right)\right)a_{q,0}\right)}\Bigg)Q_{2,i}^{(t)}(\up)
\end{align}
where the last line follows from absorbing the last term in the second line into the other terms. 

For the second term in \eqref{eq:delta_a_p_3_simplified_balanced_intermediate}, by symmetry of the balanced data distribution we have
\begin{align}\label{eq:delta_a_p_3_simplified_balanced_intermediate_second_term}
    &\EE\left[\sum_{h=1}^H\mathbbm{1}\{c_h=0,|\gA_{h,\times}|=0\}\pi_h^{(t)}(\up)\frac{N_h(j)}{e^{a_{b,0}-a_{b,2}}}\left(Q_h^{(t)}(\up)-G_h^{(t)}\right)\Big|j\in[k]\right]\notag\\
    &=\frac{1}{e^{a_{b,0}-a_{b,2}}}\sum_{i=1}^{k}\frac{(i-1)k+i}{k}\PP\left(\gE_{2,i}\right)\left(\pi_{h_i}^{(t)}(\up)|\gE_{2,i}\right)\EE\left[Q_{h_i}^{(t)}(\up)-G_{h_i}^{(t)}\Big|\gE_{2,i}\right]\notag\\
    &\overset{\eqref{eq:PP_gE_2_i_simplified_balanced}}=\frac{1}{e^{a_{b,0}-a_{b,2}}}\sum_{i=1}^{k}\frac{(i-1)k+i}{k}\frac{k-i+1}{k}\left(\overline{\pi}_{2}^{(t)}(\up)\right)^{i-1}\left(\pi_{h_i}^{(t)}(\up)|\gE_{2,i}\right)\EE\left[Q_{h_i}^{(t)}(\up)-G_{h_i}^{(t)}\Big|\gE_{2,i}\right].
\end{align}
Similar as how we compute \eqref{eq:pi_h_up_Ax=0_c=0_phase2}, we can compute that
\begin{align}\label{eq:pi_h_1_up_gE_2_1_phase2}
    \pi_{h_1}^{(t)}(\up)|\gE_{2,1}=\frac{\exp\left(-\left(1+\frac{1}{k}\right)a_{q,0}+\frac{a_{p,0}^{(t)}}{k}+O\left(\frac{k}{(\ln(1/\epsilon))^{1+O(1/N)}}\right)a_{p,1}^{(t)}\right)}{k+\exp\left(-\left(1+\frac{1}{k}\right)a_{q,0}+\frac{a_{p,0}^{(t)}}{k}+O\left(\frac{k}{(\ln(1/\epsilon))^{1+O(1/N)}}\right)a_{p,1}^{(t)}\right)},
\end{align}
By \eqref{eq:overline_pi_2}, \eqref{eq:e2a_c_0_balanced_phase_1.2} in Lemma~\ref{lm:phase_1.2} and Induction Hypothesis III we have
\begin{align}\label{eq:overline_pi_2_ub}
    \overline{\pi}_{2}^{(t)}(\up)\geq \frac{\exp\left(-\left(1+\frac{1}{k}\right)a_{q,0}+\frac{a_{p,3}^{(t)}}{k}+c_1\frac{a_{q,\times}}{\ln^2 k}\right)}{k+\exp\left(-\left(1+\frac{1}{k}\right)a_{q,0}+\frac{a_{p,3}^{(t)}}{k}+c_1\frac{a_{q,\times}}{\ln^2 k}\right)}
\end{align}
for some constant $c_1>0$.
 Note that by \eqref{eq:delta_a_p_3_bound_phase1.2} and \eqref{eq:a_q_x_q_0_p_2_asymp_phase_1.2} we have
\begin{align}\label{eq:a_p_3_T_1_bound}
    |a_{p,3}^{(T_1)}|\lesssim \frac{a_{q,\times}}{k}.
\end{align}
Thus by \eqref{eq:pi_h_1_up_gE_2_1_phase2}, \eqref{eq:overline_pi_2_ub}, \eqref{eq:a_p_3_T_1_bound} and Induction Hypothesis III we have
\begin{align}\label{eq:overline_pi_2>>pi_h_1_up}
    \overline{\pi}_{2}^{(t)}(\up)\geq \exp\left(c_1'\frac{a_{q,\times}}{\ln^2 k}\right)\left(\pi_{h_1}^{(t)}(\up)|\gE_{2,1}\right)
\end{align}
for some constant $c_1'>0$. Therefore, from \eqref{eq:overline_pi_2>>pi_h_1_up}, \eqref{eq:pi_h_up_Ax=0_c=0_phase2}, \eqref{eq:delta_a_p_3_simplified_balanced_intermediate_second_term} and Lemma~\ref{lm:phase_1.2} we know that 
\begin{align}\label{eq:delta_a_p_3_simplified_balanced_second_term_bound}
    &\left|\EE\left[\sum_{h=1}^H\mathbbm{1}\{c_h=0,|\gA_{h,\times}|=0\}\pi_h^{(t)}(\up)\frac{N_h(j)}{e^{a_{b,0}-a_{b,2}}}\left(Q_h^{(t)}(\up)-G_h^{(t)}\right)\Big|j\in[k]\right]\right|\notag\\
    &=o\left(\overline{\pi}_{2}^{(t)}(\up)\left(1-\overline{\pi}_{2}^{(t)}(\up)\right)Q_{2,1}^{(t)}(\up)\right).
\end{align}
Using \eqref{eq:a_q_x>a_p_2_bound_T_15}, \eqref{eq:a_p2>a_q_0_bound_T_15} in Lemma~\ref{lm:phase_1.2}, and by a similar argument as how we obtain \eqref{eq:overline_pi_2>>pi_h_1_up}, we can compute that
\begin{align}\label{eq:overline_pi_2>>pi_x_up}
    \overline{\pi}_{2}^{(t)}(\up)\geq \exp\left(c_2\frac{a_{q,\times}}{\ln^2 k}\right)\left(\left(1-\pi_{h_i+2j-1}^{(t)}(\up)\right)|\overline{\gE}_{2,i}^{\times,j}\right)
\end{align}
for some constant $c_2>0$. And similar as for \eqref{eq:delta_a_p_3_simplified_balanced_second_term_bound}, by \eqref{eq:overline_pi_2>>pi_x_up}, \eqref{eq:1-pi_h_up_c=x_phase2} we can bound the last term in \eqref{eq:delta_a_p_3_simplified_balanced_intermediate} as follows:
\begin{align}\label{eq:delta_a_p_3_simplified_balanced_last_term_bound}
    &\left|\EE\left[\sum_{h=1}^H\mathbbm{1}\{c_h=\times\}\pi_h^{(t)}(\up)\left(1-\pi_h^{(t)}(\up)\right)\frac{N_h(j)}{e^{a_{b,0}-a_{b,2}}}Q_h^{(t)}(\up)\Big|j\in[k]\right]\right|\notag\\
    &=o\left(\overline{\pi}_{2}^{(t)}(\up)\left(1-\overline{\pi}_{2}^{(t)}(\up)\right)Q_{2,1}^{(t)}(\up)\right).
\end{align}
Plugging \eqref{eq:delta_a_p_3_first_term}, \eqref{eq:delta_a_p_3_simplified_balanced_second_term_bound}, 
and \eqref{eq:delta_a_p_3_simplified_balanced_last_term_bound} into \eqref{eq:delta_a_p_3_simplified_balanced_intermediate}, we obtain
\begin{align}\label{eq:delta_a_p_3_final_phase2}
    \delta a_{p,3}^{(t)}&=\frac{1}{k}\sum_{i=1}^{k-1}\frac{k-i}{k}\left(\overline{\pi}_{2}^{(t)}(\up)\right)^{i}\Bigg((1+o(1))\left(1-\overline{\pi}_{2}^{(t)}(\up)\right)\notag\\
    &\qquad\qquad-\frac{1}{(k-i)\exp\left(\frac{a_{p,1}^{(t)}-a_{p,3}^{(t)}}{k-1}+\left(1+O\left(\frac{1}{\ln^2 k}\right)\right)a_{q,0}\right)}\Bigg)Q_{2,i}^{(t)}(\up).
\end{align}
This gives \eqref{eq:delta_a_p_3_simplified_balanced_phase2}.

Now we bound $\delta a_{p,1}^{(t)}$.
By \eqref{eq:delta_a_p_1_simplified_balanced} and Lemma~\ref{lm:phase_1.2} we have 
\small
\begin{align}\label{eq:delta_a_p_1_simplified_balanced_intermediate}
    \delta a_{p,1}^{(t)}&=(1+o(1/k^2))\EE\left[\sum_{h=1}^H\mathbbm{1}\{c_h=0,|\gA_{h,\times}|\geq 1,j\in\gA_{h,\times}\}\frac{1}{|\gA_{h,\times}|}\pi_h^{(t)}(j)G_h^{(t)}\right]\notag\\
    &\quad +\left(1+o(1/k^2)\right)\EE\left[\sum_{h=1}^H\mathbbm{1}\{c_h=\times\}\frac{N_h(j)}{e^{a_{b,0}-a_{b,2}}}\pi_h^{(t)}(j)\pi_h^{(t)}(\up)Q_h^{(t)}(\up)\Big|j\in[k]\right]\notag\\
    &\quad-\left(1+o(1/k^2)\right)\EE\left[\sum_{h=1}^H\mathbbm{1}\{c_h=0,|\gA_{h,\times}|=0\}\frac{N_h(j)}{e^{a_{b,0}-a_{b,2}}}\pi_h^{(t)}(j)\left(Q_h^{(t)}(j)-G_h^{(t)}\right)\Big|j\in[k]\right]\notag\\
    &\quad -\left(1+o(1/k^2)\right)\EE\left[\sum_{h=1}^H\mathbbm{1}\{c_h=0,|\gA_{h,\times}|\geq 1,j\in[k]\setminus \gA_{h,\times}\}\frac{N_h(j)}{|\gA_{h,\times}|e^{a_{b,1}-a_{b,2}}}\pi_h^{(t)}(j)\left(Q_h^{(t)}(j)-G_h^{(t)}\right)\right].
\end{align}
\normalsize
Similar as in \eqref{eq:delta_a_p_3_simplified_balanced_intermediate_first_term}, for the first term we have
\begin{align}\label{eq:EE_sum_x_h_l=2_first_term}
    &\EE\left[\sum_{h=1}^H\mathbbm{1}\{c_h=0,|\gA_{h,\times}|\geq 1,j\in\gA_{h,\times}\}\frac{1}{|\gA_{h,\times}|}\pi_h^{(t)}(j)G_h^{(t)}\right]\notag\\
    &=(1+o(1/k^2))\sum_{i=1}^k\PP\left(\gE_{2,i}\right)\sum_{j=1}^{k}\frac{(k-i+1)k-j}{(k-i+1)k}\cdot\frac{1}{j}\EE\left[\mathbbm{1}\{m\in\gA_{h_i+2j,\times}\}\pi_{h_i+2j}^{(t)}(m)G_{h_i+2j}^{(t)}\Big|\overline{\gE}_{2,i}^{i,j}\right]\notag\\
    &=\frac{1+o(1/k^2)}{k}\sum_{i=1}^k\PP\left(\gE_{2,i}\right)\sum_{j=1}^{k}\frac{(k-i+1)k-j}{(k-i+1)k}\EE\left[\pi_{h_i+2j}^{(t)}(m)G_{h_i+2j}^{(t)}\Big|\overline{\gE}_{2,i}^{i,j},m\in\gA_{h_i+2j,\times}\right].
\end{align}
By \eqref{eq:pi_h_down_x_phase2}, \eqref{eq:G_h_i+2j} we have for all $i\in[k]$ (recall $\epsilon$ is small enough, $Q_{2,k}^{(t)}(\up)=0$):
\begin{align}\label{eq:EE_sum_x_h_l=2_first_term_<=k-1}
    &\sum_{j=1}^{k-1}\frac{(k-i+1)k-j}{(k-i+1)k}\EE\left[\pi_{h_i+2j}^{(t)}(m)G_{h_i+2j}^{(t)}\Big|\overline{\gE}_{2,i}^{i,j},m\in\gA_{h_i+2j,\times}\right]\notag\\
    &=(1+o(1))\frac{(k-i+1)k-k+1}{(k-i+1)k}\EE\left[\pi_{h_i+2(k-1)}^{(t)}(m)G_{h_i+2(k-1)}^{(t)}\Big|\overline{\gE}_{2,i}^{i,k-1},m\in\gA_{h_i+2(k-1),\times}\right]\notag\\
    &=\exp\left(-\left(1+O\left(\frac{1}{\left(\ln (1/\epsilon)\right)^{1+O(1/N)}}\right)\right)\frac{a_{p,1}^{(t)}+a_{p,2}^{(t)}}{k-1}\right)\left(\frac{1}{(k-i+1)k}+\frac{k-i}{k-i+1}\pi_{2,i}^{(t)}(\up)Q_{2,i}^{(t)}(\up)\right).
\end{align}
And by \eqref{eq:pi_h_j_in_Ax=k_phase2} we have for all $i\in[k]$:
\begin{align}\label{eq:EE_sum_x_h_l=2_first_term_k}
    \EE\left[\pi_{h_i+2k}^{(t)}(m)G_{h_i+2k}^{(t)}\Big|\overline{\gE}_{2,i},m\in\gA_{h_i+2k,\times}\right]&=(1+o(1/k^2))\frac{\pi_{2,i}^{(t)}(\up)\left(1-\pi_{2,i}^{(t)}(\up)\right)}{k}Q_{2,i}^{(t)}(\up).
\end{align}
Therefore, by 
 \eqref{eq:EE_sum_x_h_l=2_first_term}, \eqref{eq:EE_sum_x_h_l=2_first_term_<=k-1}, \eqref{eq:EE_sum_x_h_l=2_first_term_k}, \eqref{eq:pi_2_i_up_equal}, we can rewrite the first term in \eqref{eq:delta_a_p_1_simplified_balanced_intermediate} as follows:
\begin{align}
    &\EE\left[\sum_{h=1}^H\mathbbm{1}\{c_h=0,|\gA_{h,\times}|\geq 1,j\in\gA_{h,\times}\}\frac{1}{|\gA_{h,\times}|}\pi_h^{(t)}(j)G_h^{(t)}\right]\notag\\
    &=\frac{1+o(1)}{k}\sum_{i=1}^{k}\PP\left(\gE_{2,i}\right)\Bigg[\exp\left(-\left(1+O\left(\frac{1}{\left(\ln (1/\epsilon)\right)^{1+O(1/N)}}\right)\right)\frac{a_{p,1}^{(t)}+a_{p,2}^{(t)}}{k-1}\right)\notag\\
    &\qquad\qquad\cdot\left(\frac{1}{(k-i+1)k}+\frac{k-i}{k-i+1}\pi_{2,i}^{(t)}(\up)Q_{2,i}^{(t)}(\up)\right)+\frac{k-i}{k-i+1}\frac{\pi_{2,i}^{(t)}(\up)\left(1-\pi_{2,i}^{(t)}(\up)\right)}{k}Q_{2,i}^{(t)}(\up)\Bigg]\notag\\
    &\overset{\eqref{eq:PP_gE_2_i_simplified_balanced}}=\frac{1+o(1)}{k^2}\sum_{i=1}^{k}\left(\overline\pi_{2}^{(t)}(\up)\right)^{i-1}\Bigg[\exp\left(-\left(1+O\left(\frac{1}{\left(\ln (1/\epsilon)\right)^{1+O(1/N)}}\right)\right)\frac{a_{p,1}^{(t)}+a_{p,2}^{(t)}}{k-1}\right)\notag\\
    &\qquad\qquad\cdot\left(\frac{1}{k}+(k-i)\overline\pi_{2}^{(t)}(\up)Q_{2,i}^{(t)}(\up)\right)+\frac{k-i}{k}\overline\pi_{2}^{(t)}(\up)\left(1-\overline\pi_{2}^{(t)}(\up)\right)Q_{2,i}^{(t)}(\up)\Bigg].
\end{align}
Similar as how we bound the last two terms in \eqref{eq:delta_a_p_3_simplified_balanced_intermediate}, By Induction Hypothesis III and Lemma~\ref{lm:phase_1.2} we can compute that 
\begin{align}
    &\frac{1}{k^2}\sum_{i=1}^{k}\left(\overline\pi_{2}^{(t)}(\up)\right)^{i-1}\exp\left(-\left(1+O\left(\frac{1}{\left(\ln (1/\epsilon)\right)^{1+O(1/N)}}\right)\right)\frac{a_{p,1}^{(t)}+a_{p,2}^{(t)}}{k-1}\right)\left(\frac{1}{k}+(k-i)\overline\pi_{2}^{(t)}(\up)Q_{2,i}^{(t)}(\up)\right)\notag\\
    &=o\Bigg(\frac{1}{k^2}\sum_{i=1}^{k-1}\frac{k-i}{k}\left(\overline\pi_{2}^{(t)}(\up)\right)^{i}\left(1-\overline\pi_{2}^{(t)}(\up)\right)Q_{2,i}^{(t)}(\up)\Bigg),
\end{align}
and the last three terms in \eqref{eq:delta_a_p_1_simplified_balanced_intermediate} can also be bounded by
\begin{align}
    &o\Bigg(\frac{1}{k^2}\sum_{i=1}^{k-1}\frac{k-i}{k}\left(\overline\pi_{2}^{(t)}(\up)\right)^{i}\left(1-\overline\pi_{2}^{(t)}(\up)\right)Q_{2,i}^{(t)}(\up)\Bigg).
\end{align}
By the above two relations we have 
\begin{align}\label{eq:delta_a_p_1_simplified_balanced_final}
    \delta a_{p,1}^{(t)}&=\frac{1+o(1)}{k^2}\sum_{i=1}^{k-1}\frac{k-i}{k}\left(\overline\pi_{2}^{(t)}(\up)\right)^{i}\left(1-\overline\pi_{2}^{(t)}(\up)\right)Q_{2,i}^{(t)}(\up).
\end{align}
This gives \eqref{eq:delta_a_p_1_simplified_balanced_phase2}.

Now we compute $\delta a_{p,0}^{(t)}$. From \eqref{eq:delta_a_p_0_simplified_balanced} we know that 
\begin{align}\label{eq:delta_a_p_0_simplified_balanced_intermediate}
    \delta a_{p,0}^{(t)}&=(1+o(1/k^2))\EE\left[\sum_{h=1}^H\mathbbm{1}\{c_h=0,|\gA_{h,\times}|=0\}\pi_h^{(t)}(\up)\left(Q_h^{(t)}(\up)-G_h^{(t)}\right)\right]\notag\\
    &\quad +(1+o(1/k^2))\EE\left[\sum_{h=1}^H\mathbbm{1}\{c_h=\times\}\pi_h^{(t)}(\up)\left(1-\pi_h^{(t)}(\up)\right)Q_h^{(t)}(\up)\right]\notag\\
    &\quad +(1+o(1/k^2))\EE\left[\sum_{h=1}^H\mathbbm{1}\{c_h=0,|\gA_{h,\times}|\geq 1\}\frac{1}{|\gA_{h,\times}|e^{a_{b,1}-a_{b,0}}}\pi_h^{(t)}(\up)\left(Q_h^{(t)}(\up)-G_h^{(t)}\right)\right].
\end{align}
Applying \eqref{eq:EE_sum_x_h_l=2} to the first term of \eqref{eq:delta_a_p_0_simplified_balanced_intermediate}, we have 
\begin{align}\label{eq:delta_a_p_0_simplified_balanced_first_term}
    &\EE\left[\sum_{h=1}^H\mathbbm{1}\{c_h=0,|\gA_{h,\times}|=0\}\pi_h^{(t)}(\up)\left(Q_h^{(t)}(\up)-G_h^{(t)}\right)\right]\notag\\
    &=-\pi_{1}^{(t)}(\up)G_1^{(t)}+\sum_{i=1}^{k}\PP\left(\gE_{2,i}\right)\left(\pi_{h_i}^{(t)}(\up)|\gE_{2,i}\right)\EE\left[Q_{h_i}^{(t)}(\up)-G_{h_i}^{(t)}\Big|\gE_{2,i}\right]\notag\\
    &\overset{\eqref{eq:PP_gE_2_i_simplified_balanced}}=-\pi_{1}^{(t)}(\up)G_1^{(t)}+(1+o(1))\sum_{i=1}^{k}\frac{k-i+1}{k}\left(\overline\pi_{2}^{(t)}(\up)\right)^{i-1}\left(\pi_{h_i}^{(t)}(\up)|\gE_{2,i}\right)\EE\left[Q_{h_i}^{(t)}(\up)-G_{h_i}^{(t)}\Big|\gE_{2,i}\right].
\end{align}
Same as how we compute \eqref{eq:overline_pi_2>>pi_h_1_up}, we can compute that 
\begin{align}\label{eq:overline_pi_2_up>>pi_1}
    \overline\pi_{2}^{(t)}(\up)\geq \exp\left(c_3\frac{a_{q,\times}}{\ln^2 k}\right)\pi_1^{(t)}(\up)
\end{align}
for some constant $c_3>0$.
Therefore, similar as how we bound the first term of $\delta a_{p,3}^{(t)}$ in \eqref{eq:delta_a_p_3_simplified_balanced_second_term_bound}, by \eqref{eq:overline_pi_2_up>>pi_1}, \eqref{eq:pi_h_up_Ax=0_c=0_phase2}, Induction Hypothesis III and Lemma~\ref{lm:phase_1.2} we can bound the first term in \eqref{eq:delta_a_p_0_simplified_balanced_intermediate} as
\begin{align}\label{eq:delta_a_p_0_simplified_balanced_first_term_bound}
    &\left|\EE\left[\sum_{h=1}^H\mathbbm{1}\{c_h=0,|\gA_{h,\times}|=0\}\pi_h^{(t)}(\up)\left(Q_h^{(t)}(\up)-G_h^{(t)}\right)\right]\right|\notag\\
    &\leq\exp\left(-c_4\frac{\ln (1/\epsilon)}{\ln^2k}\right)\overline\pi_{2}^{(t)}(\up)\left(1-\overline\pi_{2}^{(t)}(\up)\right)Q_{2,1}^{(t)}(\up)
\end{align}
for some constant $c_4>0$.

Applying \eqref{eq:EE_sum_x_h_l=2} to the second term of \eqref{eq:delta_a_p_0_simplified_balanced_intermediate}, and using Lemma~\ref{lm:phase_1.2}, Induction Hypothesis III, we have (recall we define $\overline{\gE}_{2,i}^{\times,j}$ in \eqref{eq:overline_gE_2_i_times_j})
\begin{align}\label{eq:delta_a_p_0_simplified_balanced_second_term}
    &\EE\left[\sum_{h=1}^H\mathbbm{1}\{c_h=\times\}\pi_h^{(t)}(\up)\left(1-\pi_h^{(t)}(\up)\right)Q_h^{(t)}(\up)\right]\notag\\
    &=(1+o(1))\sum_{i=1}^{k}\PP\left(\gE_{2,i}\right)\sum_{j=1}^k\frac{(k-i+1)k-j}{(k-i+1)k}\EE\left[\pi_{h_i+2j-1}^{(t)}(\up)\left(1-\pi_{h_i+2j-1}^{(t)}(\up)\right)Q_{h_i+2j-1}^{(t)}(\up)\Big|\overline{\gE}_{2,i}^{\times,j}\right]\notag\\
    &=(1+o(1))\sum_{i=1}^{k}\PP\left(\gE_{2,i}\right)\sum_{j=1}^k\frac{(k-i+1)k-j}{(k-i+1)k}\left(1-\pi_{h_i+2j-1}^{(t)}(\up)|\overline{\gE}_{2,i}^{\times,j}\right)\EE\left[G_{h_i+2j}^{(t)}\Big|\overline{\gE}_{2,i}^{i,j}\right]\notag\\
    &=(1+o(1))\sum_{i=1}^{k}\left(\overline\pi_{2}^{(t)}(\up)\right)^{i-1}\sum_{j=1}^k\left(1-\pi_{h_i+2j-1}^{(t)}(\up)|\overline{\gE}_{2,i}^{\times,j}\right)\left(\frac{k-j}{k^2}+\frac{k-i}{k}\overline\pi_{2}^{(t)}(\up)Q_{2,i}^{(t)}(\up)\right),
\end{align}
where the last line follows from \eqref{eq:G_h_i+2j} and \eqref{eq:PP_gE_2_i_simplified_balanced}. Then similar as how we compute the last term of $\delta a_{p,3}^{(t)}$ in \eqref{eq:delta_a_p_3_simplified_balanced_last_term_bound}, by \eqref{eq:overline_pi_2>>pi_x_up} and \eqref{eq:1-pi_h_up_c=x_phase2} we can obtain the following bound: 
\begin{align}\label{eq:delta_a_p_0_simplified_balanced_second_term_bound}
    \left|\EE\left[\sum_{h=1}^H\mathbbm{1}\{c_h=\times\}\pi_h^{(t)}(\up)\left(1-\pi_h^{(t)}(\up)\right)Q_h^{(t)}(\up)\right]\right|\leq \exp\left(-c_5\frac{\ln (1/\epsilon)}{\ln^2k}\right)\overline\pi_{2}^{(t)}(\up)\left(1-\overline\pi_{2}^{(t)}(\up)\right)Q_{2,1}^{(t)}(\up)
\end{align}
for some constant $c_5>0$.

And similar as how we compute the first term of $\delta a_{p,3}^{(t)}$ in \eqref{eq:delta_a_p_3_first_term}, we can rewrite the last term in \eqref{eq:delta_a_p_0_simplified_balanced_intermediate} as
\begin{align}\label{eq:delta_a_p_0_simplified_balanced_third_term}
    &\EE\left[\sum_{h=1}^H\mathbbm{1}\{c_h=0,|\gA_{h,\times}|\geq 1\}\frac{1}{|\gA_{h,\times}|e^{a_{b,1}-a_{b,0}}}\pi_h^{(t)}(\up)\left(Q_h^{(t)}(\up)-G_h^{(t)}\right)\right]\notag\\
    &=\frac{1}{ke^{a_{b,1}-a_{b,0}}}\sum_{i=1}^{k-1}\frac{k-i}{k}\left(\overline{\pi}_{2}^{(t)}(\up)\right)^{i}\Bigg((1+o(1))\left(1-\overline{\pi}_{2}^{(t)}(\up)\right)\notag\\
    &\qquad\qquad-\frac{\ln k}{(k-i)\exp\left(\frac{a_{p,1}^{(t)}-a_{p,3}^{(t)}}{k-1}+\left(1+O\left(\frac{1}{\ln^2 k}\right)\right)a_{q,0}\right)}\Bigg)Q_{2,i}^{(t)}(\up).
\end{align}
Combining \eqref{eq:delta_a_p_0_simplified_balanced_intermediate} with \eqref{eq:delta_a_p_0_simplified_balanced_first_term_bound}, \eqref{eq:delta_a_p_0_simplified_balanced_second_term_bound} and \eqref{eq:delta_a_p_0_simplified_balanced_third_term}, and by \eqref{eq:ea_b_1_lb_T_1} we have (recall we assume $\ln (1/\epsilon)\geq e^k$, and thus $k=\ln (1/\epsilon)^{o(1)}$)
\begin{align}\label{eq:|delta_a_p_0|_phase_2_final_bound}
    |\delta a_{p,0}^{(t)}|&\lesssim \frac{1}{\left(\ln (1/\epsilon)\right)^{1+o(1)}}\cdot \frac{1}{k^2}\sum_{i=1}^{k}\frac{k-i}{k}\left(\overline\pi_{2}^{(t)}(\up)\right)^{i}\left(1-\overline\pi_{2}^{(t)}(\up)\right)Q_{2,i}^{(t)}(\up)\notag\\
    &\quad +\frac{1}{\left(\ln (1/\epsilon)\right)^{1+o(1)}}\cdot \frac{1}{k^3}\sum_{i=1}^{k-1}\frac{k-i}{k}\left(\overline{\pi}_{2}^{(t)}(\up)\right)^{i}\frac{1}{(k-i)\exp\left(\frac{a_{p,1}^{(t)}-a_{p,3}^{(t)}}{k-1}+\left(1+O\left(\frac{1}{\ln^2 k}\right)\right)a_{q,0}\right)}Q_{2,i}^{(t)}(\up)\notag\\
    &\lesssim \frac{1}{\left(\ln (1/\epsilon)\right)^{1+o(1)}}\left(\delta a_{p,1}^{(t)}-\frac{1}{k^2}\delta a_{p,3}^{(t)}\right).
\end{align}
This gives \eqref{eq:delta_a_p_0_simplified_balanced_phase2}.

Finally, we compute $\delta a_{p,\up}^{(t)}$. From \eqref{eq:delta_a_p_up_simplified_balanced} we know that 
\begin{align}\label{eq:delta_a_p_up_simplified_balanced_intermediate}
    \delta a_{p,\up}^{(t)}&=-(1+o(1/k^2))\EE\left[\sum_{h=1}^H\mathbbm{1}\{c_h=0,|\gA_{h,\times}|=0\}\frac{N_h(\up)}{e^{a_{b,0}-a_{b,2}}}\pi_h^{(t)}(\up)\left(Q_h^{(t)}(\up)-G_h^{(t)}\right)\right]\notag\\
    &\quad -(1+o(1/k^2))\EE\left[\sum_{h=1}^H\mathbbm{1}\{c_h=\times\}\frac{N_h(\up)}{e^{a_{b,0}-a_{b,2}}}\pi_h^{(t)}(\up)\left(1-\pi_h^{(t)}(\up)\right)Q_h^{(t)}(\up)\right]\notag\\
    &\quad -(1+o(1/k^2))\EE\left[\sum_{h=1}^H\mathbbm{1}\{c_h=0,|\gA_{h,\times}|\geq 1\}\frac{N_h(\up)}{|\gA_{h,\times}|e^{a_{b,1}-a_{b,2}}}\pi_h^{(t)}(\up)\left(Q_h^{(t)}(\up)-G_h^{(t)}\right)\right].
\end{align}
Notice that for all $i,j\in[k]$,
\begin{align}
   N_{h_i}(\up)|\gE_{2,i}=(i-1)(k+1),\,\, N_{h_i+2j-1}(\up)|\gE_{2,i}^{\times,j}=(i-1)(k+1)+j-1,\notag\\ N_{h_i+2j}(\up)|\gE_{2,i}^{i,j}=(i-1)(k+1)+j.
\end{align}
Thus same as how we rewrite the first term of \eqref{eq:delta_a_p_0_simplified_balanced_intermediate} in \eqref{eq:delta_a_p_0_simplified_balanced_first_term}, we can rewrite the first term in \eqref{eq:delta_a_p_up_simplified_balanced_intermediate} as
\begin{align}
    &\EE\left[\sum_{h=1}^H\mathbbm{1}\{c_h=0,|\gA_{h,\times}|=0\}\frac{N_h(\up)}{e^{a_{b,0}-a_{b,2}}}\pi_h^{(t)}(\up)\left(Q_h^{(t)}(\up)-G_h^{(t)}\right)\right]\notag\\
    &=\frac{1}{e^{a_{b,0}-a_{b,2}}}\sum_{i=1}^{k}(i-1)(k+1)\PP\left(\gE_{2,i}\right)\left(\pi_{h_i}^{(t)}(\up)|\gE_{2,i}\right)\EE\left[Q_{h_i}^{(t)}(\up)-G_{h_i}^{(t)}\Big|\gE_{2,i}\right]\notag\\
    &\overset{\eqref{eq:PP_gE_2_i_simplified_balanced}}=\frac{(1+o(1))(k+1)}{e^{a_{b,0}-a_{b,2}}}\sum_{i=1}^{k}(i-1)\frac{k-i+1}{k}\left(\overline\pi_{2}^{(t)}(\up)\right)^{i-1}\left(\pi_{h_i}^{(t)}(\up)|\gE_{2,i}\right)\EE\left[Q_{h_i}^{(t)}(\up)-G_{h_i}^{(t)}\Big|\gE_{2,i}\right].
\end{align}
And same as in \eqref{eq:delta_a_p_0_simplified_balanced_first_term_bound}, from the above expression we can compute that 
\begin{align}\label{eq:delta_a_p_up_simplified_balanced_first_term_bound}
    &\left|\EE\left[\sum_{h=1}^H\mathbbm{1}\{c_h=0,|\gA_{h,\times}|=0\}\frac{N_h(\up)}{e^{a_{b,0}-a_{b,2}}}\pi_h^{(t)}(\up)\left(Q_h^{(t)}(\up)-G_h^{(t)}\right)\right]\right|\notag\\
    &\leq\exp\left(-c_6\frac{\ln (1/\epsilon)}{\ln^2k}\right)\overline\pi_{2}^{(t)}(\up)\left(1-\overline\pi_{2}^{(t)}(\up)\right)Q_{2,1}^{(t)}(\up)
\end{align}
for some constant $c_6>0$.

Same as how we compute the second term of \eqref{eq:delta_a_p_0_simplified_balanced_intermediate} in \eqref{eq:delta_a_p_0_simplified_balanced_second_term}, we can rewrite the second term in \eqref{eq:delta_a_p_up_simplified_balanced_intermediate} as
\begin{align}
    &\EE\left[\sum_{h=1}^H\mathbbm{1}\{c_h=\times\}\frac{N_h(\up)}{e^{a_{b,0}-a_{b,2}}}\pi_h^{(t)}(\up)\left(1-\pi_h^{(t)}(\up)\right)Q_h^{(t)}(\up)\right]\notag\\
    &=\frac{1+o(1)}{e^{a_{b,0}-a_{b,2}}}\sum_{i=1}^{k}\left(\overline\pi_{2}^{(t)}(\up)\right)^{i-1}\sum_{j=1}^k\left((i-1)(k+1)+j-1\right)\left(1-\pi_{h_i+2j-1}^{(t)}(\up)|\overline{\gE}_{2,i}^{\times,j}\right)\notag\\
    &\hspace{7cm}\cdot\left(\frac{k-j}{k^2}+\frac{k-i}{k}\overline\pi_{2}^{(t)}(\up)Q_{2,i}^{(t)}(\up)\right).
\end{align}
And same as in \eqref{eq:delta_a_p_0_simplified_balanced_second_term_bound}, we can bound the above term as 
\begin{align}\label{eq:delta_a_p_up_simplified_balanced_second_term_bound}
    &\left|\EE\left[\sum_{h=1}^H\mathbbm{1}\{c_h=\times\}\frac{N_h(\up)}{e^{a_{b,0}-a_{b,2}}}\pi_h^{(t)}(\up)\left(1-\pi_h^{(t)}(\up)\right)Q_h^{(t)}(\up)\right]\right|\notag\\
    &\leq \exp\left(-c_7\frac{\ln (1/\epsilon)}{\ln^2k}\right)\overline\pi_{2}^{(t)}(\up)\left(1-\overline\pi_{2}^{(t)}(\up)\right)Q_{2,1}^{(t)}(\up)
\end{align}
for some constant $c_7>0$.

Same as how we rewrite the last term of \eqref{eq:delta_a_p_0_simplified_balanced_intermediate} in \eqref{eq:delta_a_p_0_simplified_balanced_third_term}, we can rewrite the last term in \eqref{eq:delta_a_p_up_simplified_balanced_intermediate} as
\begin{align}\label{eq:delta_a_p_up_simplified_balanced_third_term}
    &\EE\left[\sum_{h=1}^H\mathbbm{1}\{c_h=0,|\gA_{h,\times}|\geq 1\}\frac{N_h(\up)}{|\gA_{h,\times}|e^{a_{b,1}-a_{b,2}}}\pi_h^{(t)}(\up)\left(Q_h^{(t)}(\up)-G_h^{(t)}\right)\right]\notag\\
    &=\frac{1}{ke^{a_{b,1}-a_{b,2}}}\sum_{i=1}^{k-1}\frac{k-i}{k}\left(\overline\pi_{2}^{(t)}(\up)\right)^{i}\Bigg((1+o(1))(i(k+1)-1)\left(1-\overline\pi_{2}^{(t)}(\up)\right)\notag\\
    &\qquad\qquad-\frac{(i-1)(k+1)\ln k+k-1}{(k-i)\exp\left(\frac{a_{p,1}^{(t)}-a_{p,3}^{(t)}}{k-1}+\left(1+O\left(\frac{1}{\ln^2 k}\right)\right)a_{q,0}\right)}\Bigg)Q_{2,i}^{(t)}(\up).
\end{align}
Then similar as in \eqref{eq:|delta_a_p_0|_phase_2_final_bound}, combining \eqref{eq:delta_a_p_up_simplified_balanced_first_term_bound}, \eqref{eq:delta_a_p_up_simplified_balanced_second_term_bound}, \eqref{eq:delta_a_p_up_simplified_balanced_third_term} with \eqref{eq:delta_a_p_up_simplified_balanced_intermediate}, and using \eqref{eq:ea_b_1_lb_T_1}, \eqref{eq:e_ab_2_lb_T_1} we can bound $\delta a_{p,\up}^{(t)}$ as
\begin{align}\label{eq:|delta_a_p_up|_phase_2_final_bound}
    |\delta a_{p,\up}^{(t)}|&\lesssim \frac{1}{\left(\ln (1/\epsilon)\right)^{3+o(1)}}\cdot \frac{1}{k^2}\sum_{i=1}^{k}\frac{k-i}{k}\left(\overline\pi_{2}^{(t)}(\up)\right)^{i}\left(1-\overline\pi_{2}^{(t)}(\up)\right)Q_{2,i}^{(t)}(\up)\notag\\
    &+\frac{1}{\left(\ln (1/\epsilon)\right)^{3+o(1)}}\cdot \frac{1}{k^3}\sum_{i=1}^{k-1}\frac{k-i}{k}\left(\overline{\pi}_{2}^{(t)}(\up)\right)^{i}\frac{1}{(k-i)\exp\left(\frac{a_{p,1}^{(t)}-a_{p,3}^{(t)}}{k-1}+\left(1+O\left(\frac{1}{\ln^2 k}\right)\right)a_{q,0}\right)}Q_{2,i}^{(t)}(\up)\notag\\
    &\lesssim \frac{1}{\left(\ln (1/\epsilon)\right)^{3+o(1)}}\left(\delta a_{p,1}^{(t)}-\frac{1}{k^2}\delta a_{p,3}^{(t)}\right).
\end{align}
This gives \eqref{eq:delta_a_p_up_simplified_balanced_phase2}.



\section{Proof of Proposition~\ref{prop:optimal_policy_balanced} and Proposition~\ref{prop:optimal_policy_imbalanced}}
\label{sec:proof_optimal_policy} 

\subsection{Proof of Proposition~\ref{prop:optimal_policy_balanced}}\label{sec:proof_optimal_policy_balanced}

For a fixed tree $\gT$ and policy $\pi\in\Pi_\text{success}$, let $H^\pi(\gT)$ denote the expected number of steps to hit the goal when the goal is generated by the balanced latent process $p_1^a=\cdots=p_k^a=\frac{1}{k}$. Define
\begin{align}\label{eq:H_pi_l}
    \overline{H}^\pi(l)\coloneqq \EE_{\gT\sim p_{\text{env,tree}}^l}\left[H^\pi(\gT)\right].
\end{align}
We will show that $\overline{H}^\pi(l)$ is the same for all $\pi\in\Pi_\text{success}$ and $l\in[L]$.

Let $\gQ_{d,r}$ denote the collection of all posterior laws of the current unresolved subtree that can arise as follows: first sample an original tree from
$p_{\textnormal{env,tree}}^l$, then condition on a reachable state and on the event that the goal lies in the current unresolved subtree rooted at the node the agent is at, with remaining depth at most $d$, and with exactly $r$ unresolved children at the current node.
For $q\in\gQ_{d,r}$
and a continuation policy $\pi\in\Pi_{\textnormal{success}}$, let $H_q^\pi$ denote the expected remaining number of steps to hit the goal. 

We claim that for every $d\geq 1$, every $r\in[k]$, and every $q\in\gQ_{d,r}$, the quantity $H_q^\pi$
is independent of $\pi$ for all $\pi\in\Pi_{\textnormal{success}}$. We prove this by lexicographical induction on $(d,r)$: we regulate $(d',r')<(d,r)$ if either $d'<d$ or $d'=d$ and $r'<r$.

\paragraph{Base case: $d=1$.}
If $q\in\gQ_{1,r}$, then the agent is at a node whose children are all leaves.
Let $\gU_q\subseteq [k]$ be the set of currently unresolved children and $r=|\gU_q|$. Since the latent goal process is balanced and the
prior over child positions is permutation-invariant, conditional on the current state, the correct child is uniformly distributed
over $\gU_q$. And any policy in $\Pi_{\textnormal{success}}$ must explore the children in $\gU_q$ in some order before
terminating. Therefore, the expected remaining number of steps is
\[
H_q^\pi=\frac1r\sum_{i=1}^r \bigl(2(i-1)+1\bigr)=r,
\]
which is independent of $\pi\in\Pi_{\textnormal{success}}$.

\paragraph{Induction step.} Fix $d\geq 2$, $r\in[k]$, and assume the claim holds for every $(d',r')<(d,r)$. We show it holds for $(d,r)$.
Fix any $q\in \gQ_{d,r}$, and let $\gU_q\subseteq [k]$ be the set of currently unresolved children and $r=|\gU_q|$. If $r=1$, there is only one unresolved child, so every policy in $\Pi_{\textnormal{success}}$ must choose it next. If this child is a leaf, then $H_q^\pi=1$. Otherwise, after moving into that child, we obtain a new posterior $q'\in\gQ_{d-1,k}$. Hence 
$$H_q^\pi=1+H_{q'}^\pi,$$
and by induction hypothesis, the right-hand side is independent of $\pi\in\Pi_{\textnormal{success}}$. In the following we assume $2\leq r\leq k$. 

For $\gT\sim q$, for each $j\in\gU_q$, write $\gT_{j}$ for the child subtree rooted at the $j$-th child of the current node, and let $h(\gT_{j})$ be the number of steps needed to fully explore $\gT_{j}$ and return to its root. If child $j$ is chosen next and is incorrect (the child subtree doesn't contain the goal), let $q_j^\times$ denote the posterior after fully exploring $\gT_{j}$ and returning to the current node, and ruling out child $j$. If child $j$ is chosen next and is correct and nonterminal, let $q_j^\checkmark$ denote the posterior after moving down into that child.
For any $\pi\in\Pi_{\textnormal{success}}$, since $\pi$ must explore all children in $\gU_q$ before hitting the goal, we can let $a^\pi(q)\in \gU_q$ denote the first child that $\pi$ chooses from the current state (without loss of generality, we assume $\pi$ is deterministic; if $\pi$ is randomized, it can be handled by conditioning on the realized internal randomness and average afterward).  There are two cases: (i) with probability $1/r$, the chosen child is the correct one; (ii) with probability $1-1/r$, it's not. 

If the chosen child is correct, then after paying \(1\) step to move down into that child, there are two subcases. If \(\gT_{a^\pi(q)}\) is a leaf, then the goal is reached immediately and no continuation cost is incurred. If \(\gT_{a^\pi(q)}\) is not a leaf, then we obtain a posterior $q^\checkmark_{a^\pi(q)} \in \gQ_{d-1,k}$.
By induction hypothesis on depth, the continuation value
$H_{q^\checkmark_{a^\pi(q)}} \coloneqq H_{q^\checkmark_{a^\pi(q)}}^\pi$
is independent of \(\pi \in \Pi_{\textnormal{success}}\).
If the chosen child is incorrect, then we pay \(2 + h(\gT_{a^\pi(q)})\) steps to fully explore that child and return, after which we obtain a posterior
$q^\times_{a^\pi(q)} \in \gQ_{d,r-1}$. By induction hypothesis, the continuation value
$H_{q^\times_{a^\pi(q)}} \coloneqq H_{q^\times_{a^\pi(q)}}^\pi$
is also independent of $\pi \in \Pi_{\textnormal{success}}$. Therefore, the expected remaining number of steps is
\begin{align}
\label{eq:H_q_pi}
H_q^\pi
&=
\EE_{\gT\sim q}\Biggl[
\frac{1}{r}\,\mathbf{1}\!\left\{\gT_{a^\pi(q)} \text{ is a leaf}\right\}
+
\frac{1}{r}\,\mathbf{1}\!\left\{\gT_{a^\pi(q)} \text{ is not a leaf}\right\}
\left(1 + H_{q^\checkmark_{a^\pi(q)}}\right)
\notag\\
&\hspace{4em}
+
\frac{r-1}{r}\left(2 + h(\gT_{a^\pi(q)}) + H_{q^\times_{a^\pi(q)}}\right)
\Biggr].
\end{align}
Since by construction, every $q\in\gQ_{d,r}$ is invariant under permutations of the children within $\gU_q$, the random vector family $(\gT_j,q_j^\checkmark,q_j^\times)_{j\in\gU_q}$ is exchangeable under $q$, i.e., for any permutation $\sigma$ of $\gU_q$, 
\begin{align}
    (T_j,q_j^\checkmark,q_j^\times)_{j\in\gU_q} \overset{d}{=} (\gT_{\sigma(j)},q_{\sigma(j)}^\checkmark,q_{\sigma(j)}^\times)_{j\in\gU_q}.
\end{align}
Therefore the expectation on the right-hand side of \eqref{eq:H_q_pi} is the same no matter which child in $\gU_q$ is chosen first. As a result, $H_q^\pi$ is independent of $\pi\in\Pi_{\textnormal{success}}$. This proves the induction step.

Finally, for each $l \in [L]$, let $q_l$ be the posterior law at the root induced by
$p_{\textnormal{env,tree}}^l$. Then
\begin{align}
    \overline{H}^{\pi}(l)=H_{q_l}^{\pi}
\end{align}
is independent of $\pi \in \Pi_{\textnormal{success}}$. 
And by our definition of the environment distribution $\gP$ in \eqref{eq:env_distribution}, we have
\begin{align}
    \overline{H}^{\pi}(\gP)
    = \sum_{l=1}^L \omega_l \, \overline{H}^{\pi}(l).
\end{align}
Because each $\overline{H}^{\pi}(l)$ is independent of $\pi \in \Pi_{\textnormal{success}}$, the weighted sum
$\overline{H}^{\pi}(\gP)$ is also independent of $\pi \in \Pi_{\textnormal{success}}$. This proves the proposition.

\subsection{Proof of Proposition~\ref{prop:optimal_policy_imbalanced}}\label{sec:proof_optimal_policy_imbalanced}

\paragraph{Proof of part 1: $\pi^\star\in\Pi_\text{success}$ has the minimum expected trajectory length.} 
First of all, by definition, it's obvious that 
$\pi^\star\in\Pi_\text{success}$. For any fixed tree $\gT$ and policy $\pi\in\Pi_\text{success}$, we let $H^\pi(\gT)$ denote the expected number of steps to hit the goal when the goal is generated by the imbalanced goal distribution \eqref{eq:imbalanced_goal_distribution}. Same as \eqref{eq:H_pi_l}, for the imbalanced case we also define 
\begin{align*}
    \overline{H}^\pi(l)\coloneqq \EE_{\gT\sim p_{\text{env,tree}}^l}\left[H^\pi(\gT)\right].
\end{align*}
Then by the definition of the environment distribution $\gP$ in \eqref{eq:env_distribution}, we have
\begin{align}\label{eq:H_pi_gP}
    H^\pi(\gP)=\sum_{l=1}^L \omega_l \overline{H}^\pi(l).
\end{align}
Therefore, to show part 1, it's enough to show that, for every $l\in[L]$, $\pi^\star$ uniquely minimizes $\overline{H}^\pi(l)$ among all policies in $\Pi_\text{success}$. 

Fix $l\in[L]$. For $d\geq 1$ and nonempty $\gU\subseteq[k]$, let $\gQ_{d,\gU}$ be the collection of all posterior laws $q$ of the current unresolved subtree that can arise as follows:
\begin{enumerate}
    \item sample the original tree from $p_{\text{env,tree}}^l$;
    \item sample the goal from the goal process $p^a$;
    \item condition on a reachable state $s$ and on the event that the goal lies in the unresolved subtree rooted at the node the agent is currently at;
    \item require that the remaining depth of that unresolved subtree is at most $d$, and that the currently unresolved child set is $\gU$.
\end{enumerate}
For $q\in\gQ_{d,\gU}$ and $\pi\in\Pi_\text{success}$, we let $H^\pi_q$ denote the expected number of steps to hit the goal.

We claim that for every $d\geq 1$, every nonempty $\gU\subseteq[k]$, and every $q\in\gQ_{d,\gU}$, the unique minimizer of $H^\pi_q$ over $\pi\in\Pi_\text{success}$ is the ranked DFS policy: at the current node, explore the unresolved children in decreasing order of $p^a_j$ for $j\in\gU$; when the current node has no unresolved child left, takes $\up$; after moving to a child, apply the same rule recursively at the new node.

Let $r\coloneqq |\gU|$.
We prove the above claim by lexicographic induction on the tuple $(d,r)$: we regulate $(d',r')<(d,r)$ if either $d'<d$ or $d'=d$ and $r'<r$.

Fix any $q\in\gQ_{d,\gU}$, and let 
\begin{align}\label{eq:S_gU}
    S_\gU\coloneqq \sum_{j\in\gU} p^a_j.
\end{align}
For each $j\in\gU$, let $G_j$ be the event that the goal lies in the $j$-th child subtree of the current node. We'll use the following two facts:
\begin{itemize}
    \item Fact 1: posterior probability of the correct child 
    \begin{align*}
        \PP_q(G_j) = \frac{p^a_j}{S_\gU},\,\,\forall j\in\gU.
    \end{align*}
    This is guaranteed by our goal generation process.
    \item Fact 2: Under $q$, the unresolved child subtrees $(T_j)_{j\in\gU}$ are exchangeable in shape across $j\in\gU$. This is because the prior $p^l_{\text{env,tree}}$ is uniform over all full $k$-ary trees of depth $l$.
\end{itemize}

\paragraph{Base case: $d=1$.} Let $q\in\gQ_{1,\gU}$ and write $r\coloneqq |\gU|$. Then all unresolved children of the current node are leaves. Any policy $\pi\in\Pi_\text{success}$ must traverse all $r$ child subtrees in some order before goal is found (note that taking $\up$ now permanently exits the current unresolved subtree; under the conditioning defining $q$, this yields failure with probability 1; taking $\up$ at any node when it still has unexplored children leads to a positive failure probability). Thus any $\pi\in\Pi_\text{success}$ is determined by an order
$$\sigma=(\sigma(1),\cdots,\sigma(r)),$$
of the elements of $\gU$, where $\sigma(i)\in\gU$ is the child inspected in position $i$. Note it suffices to consider deterministic orders of $\gU$: a randomized policy $\pi\in\Pi_\text{success}$ induces a probability distribution $\mu$ over deterministic orders $\sigma$ of $\gU$. If $H_q(\sigma)$ denotes the expected remaining path length under order $\sigma$ and optimal continuation thereafter, then  
\begin{align*}
    H_q(\sigma) = \sum_{\sigma} \mu(\sigma) H_q(\sigma).
\end{align*}
Hence $H_q^\pi\geq \min_{\sigma} H_q(\sigma)$. Moreover, if the minimizing order is unique, then any optimal policy must put probability one on that order.
If the correct child is inspected in position $i$, then the number of steps until success is exactly $2(i-1)+1$. Therefore, the expected number of steps until success is
\begin{align*}
    H_q^\pi = \sum_{i=1}^r \PP_q(G_{\sigma(i)}) \left(2(i-1)+1\right)=\frac{1}{S_\gU}\sum_{i=1}^r p^a_{\sigma(i)} \left(2(i-1)+1\right),
\end{align*}
where the second equality follows from Fact 1. By rearrangement inequality and our assumption that $p_1>p_2>\cdots>p_k$ we know that the unique optimal order is 
$$\sigma(1), \cdots, \sigma(r) \quad \text{given by decreasing order of $p^a_j$ on $\gU$}.$$
This proves the claim for all $q\in\gQ_{1,\gU}$.

\paragraph{Induction step.} Fix $d\geq 2$, nonempty $\gU\subseteq[k]$, and assume the claim holds for every $(d',r')<(d,r)$. We prove it for $q\in\gQ_{d,\gU}$. 

We still let $r\coloneqq |\gU|$. If $r=1$, i.e. $\gU=\{j\}$ for some $j\in[k]$, then there is only one unresolved child. Any $\pi\in\Pi_\text{success}$ must choose $\down_j$ next. If that child is a leaf, success occurs in one step; if not, after moving down we obtain a posterior $q'\in\gQ_{d-1,[k]}$, and by induction hypothesis, the unique optimal continuation from there is the ranked DFS policy. So the claim holds for $r=1$. Below we consider $2\leq r\leq k$.

Note that under $q$, any $\pi\in\Pi_\text{success}$ must, at the current node, simply choose an order in which to inspect the children in $\gU$. So let 
$$\sigma=(\sigma(1),\cdots,\sigma(r))$$
be any deterministic order of $\gU$. We define two constants:
\begin{itemize}
    \item Let $A_q$ be the expected remaining steps conditioned on the child currently chosen is the correct child and, after moving into it, one continues optimally from there.
    \item Let $B_q$ be the expected steps one fully explores that child subtree and returns to the current node, conditioned on the child currently chosen is incorrect
\end{itemize}
By Fact 2, both $A_q$ and $B_q$ are independent of the identity of the chosen child $j\in\gU$. Let $I\in\{1,\cdots,r\}$ by the random rank of the correct child in the order $\sigma$. By Fact 1, we have 
\begin{align*}
    \PP(I=i)=\PP_q(G_{\sigma(i)})=\frac{p^a_{\sigma(i)}}{S_\gU},\,\,\forall i\in[r].
\end{align*}
Conditional on $I=i$, the first $i-1$ children are wrong, each costing $B_q$, and the $i$-th child is correct, costing $A_q$. Hence 
\begin{align*}
    H_q(\sigma) = \sum_{i=1}^r \frac{p^a_{\sigma(i)}}{S_\gU} \left((i-1)B_q + A_q\right).
\end{align*}
Since $A_q$ and $B_q/S_\gU$ do not depend on $\sigma$, $B_q>0$, and $p^a_1>p^a_2>\cdots>p^a_k$, we know that $H_q(\sigma)$ is minimized only if $\sigma(1), \cdots, \sigma(r)$ is the decreasing order of $p^a_j$ on $\gU$. So the unique optimal first-choice order at the current node is: among unresolved children, choose the one with largest $p^a_j$ first, then the next largest, etc.

It remains to justify the recursive part. (i) If the chosen child $j$ is correct and non-leaf, then after moving down we obtain a posterior in $\gQ_{d-1,[k]}$, and by induction hypothesis, the unique optimal continuation from there is again the ranked DFS policy. (ii) If the chosen child $j$ is incorrect, then after fully exploring it and returning, the posterior belongs to $\gQ_{d,\gU\setminus\{j\}}$, and by induction hypothesis on $(d,r-1)$, the unique optimal continuation is also the ranked DFS policy on the remaining unresolved children. This completes the induction.

Finally, we apply the claim at the root: for each fixed $l\in[L]$, let $q_l$ denote the initial posterior law at the root $p_{\text{env,tree}}^l$. we have 
\begin{align*}
    \overline{H}^\pi(l)=H_{q_l}^\pi.
\end{align*}
Since $q_l\in\gQ_{l,[k]}$, the claim implies that $\pi^\star$ uniquely minimizes $\overline{H}^\pi(l)$ over $\Pi_\text{success}$. By \eqref{eq:H_pi_gP} and our assumption that $\omega_l>0$ for all $l\in[L]$ we know that $\pi^\star$ uniquely minimizes $H^\pi(\gP)$ over $\Pi_\text{success}$.

\paragraph{Proof of part 2: $\pi^\star=\argmax_{\pi}J_\gamma^{\pi}(\gP)$ for all $\gamma\in\left(1-\Theta\left(\left(p^a_k\right)^{L}\right),1\right)$.} We show this by first showing for any $\pi\in\Pi_\text{success}\setminus\{\pi^\star\}$, $J_\gamma^{\pi}(\gP)<J_\gamma^{\pi^\star}(\gP)$, and then showing $\argmax_{\pi}J_\gamma^{\pi}(\gP)\subseteq\Pi_\text{success}$.

\paragraph{Step 1: show for any $\pi\in\Pi_\text{success}\setminus\{\pi^\star\}$, $J_\gamma^{\pi}(\gP)<J_\gamma^{\pi^\star}(\gP)$ for all $\gamma\in(0,1)$.} 
Fix $\gamma\in(0,1)$. We use the same posterior family $\gQ_{d,\gU}$ as in part 1, write $r\coloneqq |\gU|$, and consider any deterministic order $\sigma=(\sigma(1),\cdots,\sigma(r))$ in which a successful policy $\pi\in\Pi_\text{success}$ inspects the children in $\gU$ before eventually going up. Let $W_q(i)$ be the discounted return obtained if the correct child is inspected in position $i$, and after entering that child one continues recursively in the optimal way. By Fact 2 in part 1, $W_q(i)$ depends only on the rank $i$, not on the label $\sigma(i)$. Moreover, since delaying the correct child by one extra wrong child inserts at least two extra steps (down and back up), and $\gamma\in(0,1)$, we have
\begin{align}\label{eq:W_q_imbalanced}
    W_q(1)>W_q(2)>\cdots>W_q(r)>0.
\end{align}
Hence the discounted value of order $\sigma$ is 
\begin{align*}
    V_q(\sigma) = \sum_{i=1}^r \frac{p^a_{\sigma(i)}}{S_\gU} W_q(i).
\end{align*}
By \eqref{eq:W_q_imbalanced}, \eqref{eq:imbalanced_goal_distribution} and rearrangement inequality we know that $V_q(\sigma)$ is uniquely maximized by $\sigma(1), \cdots, \sigma(r)$ given by decreasing order of $p^a_j$ on $\gU$. Then similar as the proof of part 1, applying this recursively after entering the correct child proves, by lexicographic induction on ($d$, $|\gU|$), that
\begin{align*}
    \pi^\star =\arg\max_{\pi\in\Pi_\text{success}} J_\gamma^{\pi}(\gP),\,\,\forall \gamma\in(0,1).
\end{align*}

\paragraph{Step 2: show $\argmax_{\pi}J_\gamma^{\pi}(\gP)\subseteq\Pi_\text{success}$ for all $\gamma\in\left(1-\Theta\left(\left(p^a_k\right)^{L}\right),1\right)$.} We show this by fixing any policy $\pi\notin\Pi_\text{success}$ and showing $\pi\notin\argmax_{\pi}J_\gamma^{\pi}(\gP)$. Since $\pi\notin\Pi_\text{success}$, it either (i) takes an illegal action at some (reachable, non-terminal) state at some node with positive probability, when there's still untried actions at that node, or (ii) act legally when there's still legal actions to take (with probability 1), but has positive probability to not explore all children of a node before going up at some state. 
    
    We let $J_\gamma^\star(s)$ and $Q_\gamma^\star(s,a)$ denote the optimal value and Q-function with discount factor $\gamma$ at state $s$ and state-action pair $(s,a)$ respectively. We let $J^\star_\gamma(\gP)$ denote the optimal expected return with discount factor $\gamma$ under the training distribution $\gP$.
    For case (i), consider any reachable non-terminal state $s$ such that $\pi$ assigns positive probability $p>0$ to illegal actions at $s$. Let $h(s)$ be the timestep of $s$, and let $d_\gP^\pi(s)$ be the probability that $s$ is reached under $(\gP,\pi)$. Define $\widetilde{\pi}$ to agree with $\pi$ before reaching $s$, and at $s$ choose an action $a^\star\in\argmax_{a\in\gA}Q_\gamma^\star(s,a)$, following an optimal continuation thereafter. Then, conditional on reaching $s$, policy $\pi$ loses at least $pJ_\gamma^\star(s)$ in continuation value compared to $\widetilde{\pi}$. Therefore, 
    \begin{align*}
        J_\gamma^{\pi}(\gP) \leq  J_\gamma^{\widetilde{\pi}}(\gP) - d_\gP^\pi(s)\gamma^{h(s)-1}pJ_\gamma^\star(s)<J_\gamma^{\widetilde{\pi}}(\gP)\leq J_\gamma^\star(\gP).
    \end{align*}
Therefore, any policy that has positive probability to take illegal actions at a reachable, non-terminal state cannot be optimal.
    
    Thus it remains to consider case (ii), i.e., the policies that act legally everywhere but are not successful. 
    Without loss of generality, we assume at some reachable, non-terminal state $s$ where the agent is at node $u$ at level $h\in[l-1]$, $\pi$, for the first time, takes action $\up$ with probability 1 before exhausting all currently unresolved children at the current node $u$, Let $r$ be the number of still-untried children at $u$, and let $d$ be the remaining depth below $u$.

    Construct a comparison policy $\pi'$ that is identical to $\pi$ except at state $s$: instead of going up immediately, first explores the $r$ remaining children under $u$ in decreasing $p^a_j$-order, using $\pi^\star$ recursively inside each such child subtree, and only after that returns to $u$ and resumes $\pi$.
    We let $\gE_s$ denote the event that the goal is in the untried children under $u$, and let $p_s\coloneqq\PP(\gE_s|s)$ denote the probability that $\gE_s$ happens conditioned on the current state $s$. Let $R_\gamma(s)$ denote the expected return of $\pi'$ conditioned on $\gE_s$. Let $R'_\gamma(s)$ denote the expected return of $\pi$ conditioned on $\gE_s^c$. Let $n_s$ be the number of steps taken to finish exploring the remaining $r$ children under $u$ and return to $u$ under event $\gE_s$. Then the expected return of $\pi,\pi'$ at $s$, denoted by $G_{\gamma}^{\pi}(s)$ and $G_{\gamma}^{\pi'}(s)$ respectively, are
    \begin{align}\label{eq:G_imbalanced}
        G_{\gamma}^{\pi'}(s) &= p_s R_{\gamma}(s) + (1-p_s)\gamma^{n_s}R'_{\gamma}(s),\quad G_{\gamma}^{\pi}(s) = (1-p_s)R'_{\gamma}(s).
    \end{align}
We'll show that when $\gamma\in\Big(1-\Theta\left(\left(p^a_k\right)^{L}\right),1\Big)$, $G_{\gamma,l}^{\pi'}(s)>G_{\gamma,l}^{\pi}(s)$ for all $l\in\{2,\cdots,L\}$. 
To do so, below we separately bounding $p_s$, $n_s$, $R_\gamma(s)$, $R'_\gamma(s)$. 
We let $\gU=\{n_1,\cdots,n_r\}\subset[k]$ denote the set of untried children of $u$ with child indices $n_1<n_2<\cdots<n_r$, and write
\begin{align}\label{eq:P_r}
    P_r(\gU)\coloneqq\sum_{n_i\in \gU} p_{n_i}.
\end{align}

\begin{itemize}
\item \textbf{Bound $p_s$.} 
We define
\begin{align*}
    q_u\coloneqq \PP\left[\text{the goal is in the subtree rooted at } u | s\right].
\end{align*}
Then we have
\begin{align}\label{eq:q_u_imbalanced}
    q_u \geq \left(p^a_k\right)^{L-d}.
\end{align}
Recall we assume it's the first time for $\pi$ to take action $\up$ before exploring all remaining children under $u$. \eqref{eq:P_r} and \eqref{eq:q_u_imbalanced} allow us to bound $p_s$ as
\begin{align}\label{eq:p_s_imbalanced}
    p_s \geq \frac{q_uP_r(\gU)}{1-q_u+q_uP_r(\gU)}.
\end{align}
\item \textbf{Bound $n_s$.} 
Let $h_d$ be the (deterministic) number of steps needed to fully explore a perfect 
$k$-ary tree of depth $d$
and return to the root (DFS tour length). A child subtree is explored by: go $\down_i$ (1 step) for all $i\in[k]$, explore that subtree and return ($h_l-1$ steps), then go $\up$ (1 step), hence
\begin{align}\label{eq:h_d}
  h_0=0,\quad h_d=k(2+h_{d-1})=2k\frac{k^d-1}{k-1}.
\end{align}
Each wrong child subtree under $u$ has remaining depth at most $d-1$. The largest possible cost to fully explore one such child subtree and return to $u$ is attained by a perfect depth-($d-1$) subtree, and equals  $2+h_{d-1}$. Hence
\begin{align}\label{eq:n_s_imbalanced}
    n_s\leq r(2+h_{d-1}).
\end{align}
\item \textbf{Bound $R_\gamma(s)$.} 
Let $\tau^\star_d$ be the random hitting time of the ranked DFS policy $\pi^\star$ on a perfect $k$-ary subtree of depth $d$ conditioned on the event that the goal is in that subtree, with the clock started at the subtree root. It is easy to show by induction on $d$ that among all full $k$-ary subtrees of remaining depth at most $d$, the perfect depth-$d$ subtree gives the smallest discounted continuation value under $\pi^\star$: let $\tau^\star(\gT)$ denote the conditional hitting time on such a subtree $\gT$, then $\EE[\gamma^{\tau^\star(\gT)}] \geq \EE[\gamma^{\tau^\star_d}]$.
Then conditioned on $\gE_s$, recall we let $\pi'$ act the same as $\pi^\star$ in every untried subtree under $u$, we have
\begin{align}\label{eq:R_s_imbalanced}
    R_\gamma(s) \geq \frac{1}{P_r(\gU)}\sum_{i=1}^r p_{n_i}^a\EE\left[\gamma^{(2+h_{d-1})(i-1)+\tau_{d-1}^\star}\right]\geq \frac{p^a_k}{P_r(\gU)}\frac{1-\gamma^{(2+h_{d-1})r}}{1-\gamma^{2+h_{d-1}}}\EE\left[\gamma^{\tau_{d-1}^\star}\right].
\end{align}
\item \textbf{Bound $R'_\gamma(s)$.} Conditioned on $\gE_s^c$, if $\pi$ goes up immediately from $u$, then any eventual success can occur only after at least two more steps: one step up to the parent, and at least one step down into some other child subtree, because the goal is always a leaf. Therefore, 
\begin{align}\label{eq:R'_s_imbalanced}
    R'_\gamma(s) \leq \gamma^2.
\end{align}
\end{itemize}

\paragraph{Combine the bounds.} From \eqref{eq:n_s_imbalanced}, \eqref{eq:R_s_imbalanced}, \eqref{eq:R'_s_imbalanced} we deduce 
\begin{align*}
    G_{\gamma}^{\pi'}(s) - G_{\gamma}^{\pi}(s) &\overset{\eqref{eq:G_imbalanced}}= p_s R_{\gamma}(s) - (1-p_s)\left(1-\gamma^{n_s}\right)R'_{\gamma}(s) \notag\\
    &\geq p_s \frac{p^a_k}{P_r(\gU)}\frac{1-\gamma^{(2+h_{d-1})r}}{1-\gamma^{2+h_{d-1}}}\EE\left[\gamma^{\tau_{d-1}^\star}\right] -(1-p_s)\left(1-\gamma^{(2+h_{d-1})r}\right)\gamma^2.
\end{align*}
A sufficient condition for $G_{\gamma}^{\pi'}(s)>G_{\gamma}^{\pi}(s)$ is
\begin{align*}
    \frac{p_s}{1-p_s} > \frac{P_r(\gU)}{p^a_k}\frac{(1-\gamma^{(2+h_{d-1})})\gamma^2}{\EE[\gamma^{\tau_{d-1}^\star}]}.
\end{align*}
From \eqref{eq:p_s_imbalanced} we know that 
\begin{align*}
    \frac{p_s}{1-p_s} \geq \frac{q_uP_r(\gU)}{1-q_u}\overset{\eqref{eq:q_u_imbalanced}}\geq \frac{\left(p^a_k\right)^{L-d}P_r(\gU)}{1-\left(p^a_k\right)^{L-d}}.
\end{align*}
Combining the above two relations, we know that it's enough to require 
\begin{align}\label{eq:sufficient_condition_imbalanced}
    \frac{\left(p^a_k\right)^{L-d}}{1-\left(p^a_k\right)^{L-d}}>\frac{1}{p^a_k}\frac{(1-\gamma^{(2+h_{d-1})})\gamma^2}{\EE[\gamma^{\tau_{d-1}^\star}]}.
\end{align}
Now we bound $\EE[\gamma^{\tau_{d-1}^\star}]$.
Define $T_d^\star\coloneqq \EE[\tau_{d}^\star]$. Then we have 
\begin{align*}
    T_0^\star=0,\quad T^\star_d=T^\star_{d-1}+1+\sum_{j=1}^kp_j(j-1)(2+h_{d-1}),\,\,\forall d\geq 1.
\end{align*}
This further yields 
\begin{align}\label{eq:T_d_star}
    T_d^\star=d+\sum_{j=1}^k p_j(j-1)\sum_{i=0}^{d-1}(2+h_i)\overset{\eqref{eq:h_d}}=d+\frac{2\sum_{j=1}^k p_j(j-1)}{k-1}\left(\frac{k(k^d-1)}{k-1}-d\right).
\end{align}
By Bernoulli's inequality, we have
\begin{align} \label{eq:T_d_star_bound}
    \gamma^h\geq 1-h(1-\gamma),\,\,\forall h\geq 0, \quad
\Longrightarrow \quad 
    \EE[\gamma^{\tau_{d-1}^\star}] \geq 1-T_{d-1}^\star(1-\gamma),
\end{align}
and 
\begin{align}\label{eq:a_gamma_bound}
    (1-\gamma^{2+h_{d-1}})\gamma^2\leq 1-\gamma^{2+h_{d-1}}\leq (2+h_{d-1})(1-\gamma).
\end{align}
Substituting \eqref{eq:T_d_star_bound} and \eqref{eq:a_gamma_bound} into \eqref{eq:sufficient_condition_imbalanced}, a sufficient condition for $G_{\gamma}^{\pi'}(s)>G_{\gamma}^{\pi}(s)$ is
\begin{align*}
    \left(p^a_k\right)^{L-d+1}(1-T^\star_{d-1}(1-\gamma))> \left(1-\left(p^a_k\right)^{L-d}\right)(2+h_{d-1})(1-\gamma),
\end{align*}
which is equivalent to
\begin{align*}
    \gamma>1-\underbrace{\frac{\left(p^a_k\right)^{L-d+1}}{\left(p^a_k\right)^{L-d+1}T^\star_{d-1}+\left(1-\left(p^a_k\right)^{L-d}\right)(2+h_{d-1})}}_{\coloneqq\delta_d},\,\,\forall d\in[L].
\end{align*}
Define 
\begin{align}\label{eq:delta}
    \delta\coloneqq \min_{d\in[L]} \delta_d.
\end{align}
Then a sufficient condition for $G_{\gamma}^{\pi'}(s)>G_{\gamma}^{\pi}(s)$ is
\begin{align}\label{eq:sufficient_condition_imbalanced_final}
    \gamma>1-\delta.
\end{align}
Below we show 
\begin{align}\label{eq:delta_bound}
    \delta=\Theta\left(\left(p^a_k\right)^{L}\right).
\end{align}
When $d=1$, since $h_0=T_0^\star=0$, we have
\begin{align}\label{eq:delta_1}
    \delta_1=\frac{\left(p^a_k\right)^{L}}{2\left(1-\left(p^a_k\right)^{L-1}\right)}=\Theta\left(\left(p^a_k\right)^{L}\right).
\end{align}
For $d\geq 2$, by \eqref{eq:h_d} and \eqref{eq:T_d_star} we know that 
\begin{align*}
    h_{d-1}=\Theta\left(k^{d-1}\right),\,\,\,T_{d-1}^\star=O\left(k^{d-1}\right),
\end{align*}
so
\begin{align}\label{eq:delta_d}
    \delta_d=\Theta\left(\frac{\left(p^a_k\right)^{L-d+1}}{k^{d-1}}\right),\,\,\forall 2\leq d\leq L.
\end{align}
From \eqref{eq:delta}, \eqref{eq:delta_1}, \eqref{eq:delta_d} and the fact that $p_k<1/k$ we know that \eqref{eq:delta_bound} holds.

\section{Proof of Construction}\label{sec:proof_transformer_construction}
\paragraph{Notation.} We'll reuse some notations defined at the beginning of Appendix~\ref{sec:proof_convergence_balanced}, which we copy below for reader's convenience. We define
$\pi_h\coloneqq \pi_\theta(E^{(h)}),\,\forall h\in[H].$ and $N_h$ as the number of leaf nodes explored up to step $h$:
\begin{align}\label{eq:N_h_times_construction}
    N_{h,\times}\coloneqq \sum_{i=1}^h \mathbbm{1}\{c_i=\times\}.
\end{align}
At any step $h$, the agent is at node $n_h$, and we let
\begin{align}\label{eq:gA_h_times_construction}
\gA_{h,\times}\coloneqq \begin{cases}
\{i\in[k]|i \text{ is taken at } n_h\}, \quad \text{if } c_h=0,\\
\emptyset,\quad \text{if } c_h=\times,\\
\end{cases}
\end{align}
denote the set of downward actions that are taken at node $n_h$. Define 
\begin{align}\label{eq:N_h_i_construction}
    \forall i\in[k+1]:\quad N_h(i)\coloneqq \sum_{j=1}^{h-1} \mathbbm{1}\{a_j=i\}
\end{align}
as the number of times action $i$ is taken up to step $h$. Let $N_h(\up)\coloneqq N_h(k+1)$.

We'll make use of the following lemma from \citet{yang2025multi} to show the existence of our matrix constructions.
\begin{lm}[Restatement of Lemma 1 in \citet{yang2025multi}]\label{lm:matrix}
    Suppose matrices $X\in\R^{m_1\times n_1}, Y\in\R^{m_2\times n_2}$ have full column rank. Then for any matrix $A\in\R^{n_1\times n_2}$, there exists a matrix $M\in\R^{m_1\times m_2}$ such that 
    $$ X^\top M Y = A.$$ 
\end{lm}



\subsection{Proof of Theorem~\ref{thm:transformer_construction}}\label{sec:proof_construction_balanced}
First of all, from Lemma~\ref{lm:matrix}, the definition of $A_B, A_C, A_P, A_Q$ in \eqref{eq:induced_matrices_main} and the full column rank assumption on $U,V,Z$, we know that our construction \eqref{eq:A_B_C_P_Q} exists. We use $O(f(k))$ to denote a quantity that is bounded above by a constant multiple of $f(k)$ for any $k\geq 2$ (we don't require $k$ to be large).

Fix any full $k$-ary tree $\gT$ of depth $l\in[L]$ and any goal leaf $g(\gT)$. Consider any legal trajectory generated by policy $\pi$ on $\gT$. 
From the proof of Lemma~\ref{lm:gradients} we know that the policy expressions in Lemma~\ref{lm:gradients} still hold here with $a_{b,2}=a_{p,\up}=a_{p,0}=a_{p,2}=0$, i.e., we have 
\begin{align}\label{eq:pi_h_construction}
    \pi_h(\up)&=\frac{1}{Z_h}\exp\left(
        \left(1+\frac{1}{k}\right)\xi_h\right),\notag\\
    \forall i\in[k]:\quad \pi_h(i)&=\frac{1}{Z_h}\exp\left(\frac{-\mathbbm{1}\{i\in\gA_{h,\times}\}a_{p,1}(e^{a_{b,1}}-1)-N_h(i)a_{p,1}}{e^{a_{b,0}}+|\gA_{h,\times}|e^{a_{b,1}}+h-1-|\gA_{h,\times}|}\right),
\end{align}
where
\begin{align}\label{eq:xi_h_construction}
    \xi_h&=\begin{cases}
        \frac{N_{h,\times}e^{-a_{c,0}}a_{q,\times}-(h-N_{h,\times})e^{a_{c,0}}a_{q,0}}{N_{h,\times}e^{-a_{c,0}}+(h-N_{h,\times})e^{a_{c,0}}}, & \text{if } c_h=0,\\
        \frac{N_{h,\times}e^{a_{c,1}}a_{q,\times}-(h-N_{h,\times})e^{-a_{c,1}}a_{q,0}}{N_{h,\times}e^{a_{c,1}}+(h-N_{h,\times})e^{-a_{c,1}}}, & \text{if } c_h=\times,
    \end{cases}
\end{align}
and $Z_h$ is the normalizing factor:
\begin{align}
    Z_h&\coloneqq \exp\left(\left(1+\frac{1}{k}\right)\xi_h\right)+\sum_{i=1}^{k} \exp\left(\frac{\varphi_h(i)}{e^{a_{b,0}}+|\gA_{h,\times}|e^{a_{b,1}}+h-1-|\gA_{h,\times}|}\right).
\end{align}

Note that the trajectory length $H$ is upper bounded by $2N$, and $N_{h,\times}\leq \floor{\frac{h}{2}}$, as under our setting, $c_1=0$, and two wrong-leaf observations cannot occur in consecutive timesteps.
Therefore, by our choice of $a_{q,0},a_{q,\times},a_{c,0}$ in \eqref{eq:construction_balanced} and \eqref{eq:xi_h_construction}, we have 
\begin{align}\label{eq:xi_h_construction_simplified}
    \xi_h&=\begin{cases}
        -\left(1+O\left(\frac{1}{\ln^2 k}\right)\right)a_{q,0}, & \text{if } c_h=0,\\
        \left(1+O\left(\frac{1}{\ln^2 k}\right)\right)a_{q,\times}, & \text{if } c_h=\times.
    \end{cases}
\end{align}

For each step $h\in[H]$, if the trajectory has not terminated yet, only four cases would occur: (i) $c_h=0,|\gA_{h,\times}|=0$: the agent enters a non-leaf node for the first time; (ii) $c_h=\times,|\gA_{h,\times}|=0$: the agent is at a non-goal leaf; (iii) $c_h=0,|\gA_{h,\times}|\in[k-1]$: the agent is at a non-leaf node and has already explored part of its children; (iv) $c_h=0,|\gA_{h,\times}|=k$: the agent is at a non-leaf node and has explored all $k$ children.

For case (i), from \eqref{eq:pi_h_construction}, \eqref{eq:xi_h_construction_simplified} we know that when $c_h=0,|\gA_{h,\times}|=0$,
\begin{align}\label{eq:pi_h_up_(i)}
    \pi_h(\up)&=\frac{\exp\left(-\left(1+O\left(\frac{1}{\ln^2 k}\right)\right)a_{q,0}\right)}{\exp\left(-\left(1+O\left(\frac{1}{\ln^2 k}\right)\right)a_{q,0}\right)+\sum_{i=1}^{k}\exp\left(\frac{-N_h(i)a_{p,1}}{e^{a_{b,0}}+h-1}\right)}\notag\\
    &\leq \frac{\exp\left(-\left(1+O\left(\frac{1}{\ln^2 k}\right)\right)a_{q,0}\right)}{\exp\left(-\left(1+O\left(\frac{1}{\ln^2 k}\right)\right)a_{q,0}\right)+\exp\left(\frac{-\min_{i\in[k]}\{N_h(i)\} a_{p,1}}{e^{a_{b,0}}+h-1}\right)}.
\end{align}
Note that
\begin{align}
    \min_{i\in[k]}\{N_h(i)\}\leq\frac{h-1-N_h(\up)}{k}.
\end{align}
Thus by our choice of $a_{p,1},a_{q,0},a_{b,0}$ in \eqref{eq:construction_balanced} and \eqref{eq:pi_h_up_(i)}, we have
\begin{align}
    \pi_h(\up)&\leq\exp\left(-\left(1+O\left(\frac{1}{\ln k}\right)\right)a_{q,0}\right)\leq \frac{\epsilon}{2N}
\end{align}
when constant $C$ is large enough in \eqref{eq:construction_balanced}. This suggests under our construction, when the agents enter a new non-leaf node, it will explore its children instead of going up with probability at least $1-\frac{\epsilon}{2N}$.

Similarly, for case (ii), all actions other than $\up$ are illegal, and from \eqref{eq:pi_h_construction}, \eqref{eq:xi_h_construction_simplified} and our choice of $a_{q,\times},a_{p,1}$ in \eqref{eq:construction_balanced} we know that when $c_h=\times$,
\begin{align}
    1-\pi_h(\up)&=\frac{\sum_{i=1}^{k}\exp\left(\frac{-N_h(i)a_{p,1}}{e^{a_{b,0}}+h-1}\right)}{\exp\left(\left(1+O\left(\frac{1}{\ln^2 k}\right)\right)a_{q,\times}\right)+\sum_{i=1}^{k}\exp\left(\frac{-N_h(i)a_{p,1}}{e^{a_{b,0}}+h-1}\right)}\leq \frac{\epsilon}{2N}
\end{align}
when constant $C$ is large enough in \eqref{eq:construction_balanced}. This suggests the agent will go up at any non-goal leaf with probability at least $1-\frac{\epsilon}{2N}$.

For case (iii), $|\gA_{h,\times}|\in[k-1]$, from \eqref{eq:pi_h_construction} and our choice of $a_{p,1},a_{b,0},a_{b,1}$ in \eqref{eq:construction_balanced} we know that when $|\gA_{h,\times}|\in[k-1]$,
\begin{align}\label{eq:pi_h_i_construction_case_iii}
    \forall i\in\gA_{h,\times}:\quad \pi_h(i)&=\frac{1}{Z_h}\exp\left(\frac{-a_{p,1}\left(e^{a_{b,1}}-1+N_h(i)\right)}{e^{a_{b,0}}+|\gA_{h,\times}|e^{a_{b,1}}+h-1-|\gA_{h,\times}|}\right)\notag\\
    &\leq \frac{\exp\left(\frac{-a_{p,1}\left(e^{a_{b,1}}-1+N_h(i)\right)}{e^{a_{b,0}}+|\gA_{h,\times}|e^{a_{b,1}}+h-1-|\gA_{h,\times}|}\right)}{\sum_{j\in\gA_{h,\times}}\exp\left(\frac{-a_{p,1}N_h(j)}{e^{a_{b,0}}+|\gA_{h,\times}|e^{a_{b,1}}+h-1-|\gA_{h,\times}|}\right)}\notag\\
    &\leq \exp\left(-\frac{1}{|\gA_{h,\times}|}\left(1+O\left(\frac{1}{\ln k}\right)\right)a_{p,1}\right)\notag\\
    &\leq \exp\left(-\frac{1}{k-1}\left(1+O\left(\frac{1}{\ln k}\right)\right)a_{p,1}\right)\leq \frac{\epsilon}{2kN}
\end{align}
when constant $C$ is large enough in \eqref{eq:construction_balanced}. Analogously, we have
\begin{align}\label{eq:pi_h_up_construction_case_iii}
    \pi_h(\up)&\leq \frac{\exp\left(-\left(1+O\left(\frac{1}{\ln^2 k}\right)\right)a_{q,0}\right)}{\sum_{j\in\gA_{h,\times}}\exp\left(\frac{-a_{p,1}N_h(j)}{e^{a_{b,0}}+|\gA_{h,\times}|e^{a_{b,1}}+h-1-|\gA_{h,\times}|}\right)}\notag\\
    &=\exp\left(-\left(1+O\left(\frac{1}{\ln^2 k}\right)\right)a_{q,0}\right)\leq \frac{\epsilon}{2kN},
\end{align}
where the second line uses our choice of $a_{q,0},a_{p,1},a_{b,0},a_{b,1}$ in \eqref{eq:construction_balanced} as well as the fact that 
\begin{align}\label{eq:N_h_j_bound}
    N_h(j)\leq \frac{N}{k},\,\,\forall j\in[k].
\end{align}
\eqref{eq:pi_h_i_construction_case_iii} and \eqref{eq:pi_h_up_construction_case_iii} together indicate in case (iii), the agent will choose to visit the unexplored children with probability at least $1-\frac{\epsilon}{2N}$.

For case (iv), from \eqref{eq:pi_h_construction} and our choice of $a_{p,1},a_{b,0},a_{b,1}$ in \eqref{eq:construction_balanced} we deduce when $|\gA_{h,\times}|=k$,
\begin{align}
    1-\pi_h(\up)&=\frac{\sum_{i=1}^{k}\exp\left(\frac{-a_{p,1}\left(e^{a_{b,1}}-1+N_h(i)\right)}{e^{a_{b,0}}+ke^{a_{b,1}}+h-1-k}\right)}{\exp\left(-\left(1+O\left(\frac{1}{\ln^2 k}\right)\right)a_{q,0}\right)+\sum_{i=1}^{k}\exp\left(\frac{-a_{p,1}\left(e^{a_{b,1}}-1+N_h(i)\right)}{e^{a_{b,0}}+ke^{a_{b,1}}+h-1-k}\right)}\notag\\
    &\leq \frac{k\exp\left(-\left(1+O\left(\frac{1}{\ln k}\right)\right)\frac{a_{p,1}}{k}\right)}{\exp\left(-\left(1+O\left(\frac{1}{\ln^2 k}\right)\right)a_{q,0}\right)}\leq \frac{\epsilon}{2N}
\end{align}
when constant $C$ is large enough in \eqref{eq:construction_balanced}. This suggests the agent will go up at any non-leaf node after fully exploring its children with probability at least $1-\frac{\epsilon}{2N}$.

Note that the trajectory length can be bounded by $H=2N$. From the above discussion on the four cases, we know that given any $\Phi=(\gT,g(\gT))$ where $\gT$ is a full $k$-ary tree with depth $l\in[L]$, the agent will perform DFS on $\gT$ with probability at least $1-\epsilon$ until the goal is found. This completes the proof of Theorem~\ref{thm:transformer_construction}.

\subsection{Proof of Theorem~\ref{thm:transformer_construction_imbalanced}}\label{sec:proof_construction_imbalanced}
First, from Lemma~\ref{lm:matrix}, the definition of $A_B, A_C, A_P, A_Q$ in \eqref{eq:induced_matrices_main} and the full column rank assumption on $U,V,Z$, we know that our construction \eqref{eq:A_B_C_P_Q_imbalanced} exists.

Fix any full $k$-ary tree $\gT$ of depth $l\in[L]$ and any goal leaf $g(\gT)$. Consider any legal trajectory generated by policy $\pi$ on $\gT$.  To better align with the node/action/label numbering, we let $A_P,A_Q$'s column indices be 0-indexed, and $A_P,A_Q$'s row indices be 1-indexed, and we let $\times$ and $\checkmark$ denote index 1 and 2 respectively, e.g., $A_Q(:,0)$ and $A_Q(:,\times)$ denote the first and second columns of $A_Q$ respectively. We let $\up$ denote the action index $k+1$.

By \eqref{eq:pi_theta_simplified}, \eqref{eq:induced_matrices_main} and \eqref{eq:A_B_C_P_Q_imbalanced} we have 
\begin{align}\label{eq:pi_h_simplified_imbalanced}
    \pi_h&=\softmax\left(\mu_h+\nu_h\right),
\end{align}
where 
\begin{align}
    \mu_h&\coloneqq \frac{A_P(:,0)e^{a_{b,0}}+\sum_{j\in\gA_{h,\times}}A_P(:,j)e^{a_{b,1}}+\sum_{i=1}^k\left(N_h(i)-\mathbbm{1}\{i\in\gA_{h,\times}\}\right)A_P(:,i)}{e^{a_{b,0}}+|\gA_{h,\times}|e^{a_{b,1}}+h-1-|\gA_{h,\times}|},
    \end{align}
and
\begin{align}
    \nu_h&\coloneqq
    \begin{cases}
        \frac{N_{h,\times}e^{-a_{c,0}}A_Q(:,\times)+(h-N_{h,\times})e^{a_{c,0}}A_Q(:,0)}{N_{h,\times}e^{-a_{c,0}}+(h-N_{h,\times})e^{a_{c,0}}}, & \text{if } c_h=0\\
        \frac{N_{h,\times}e^{a_{c,1}}A_Q(:,\times)+(h-N_{h,\times})e^{-a_{c,1}}A_Q(:,0)}{N_{h,\times}e^{a_{c,1}}+(h-N_{h,\times})e^{-a_{c,1}}}, & \text{if } c_h=\times
    \end{cases}.
\end{align}
Combining \eqref{eq:constraints_imbalanced} with the above two expressions, we can further simplify $\mu_h$ and $\nu_h$ as
\begin{align}\label{eq:mu_h_simplified_imbalanced}
    \forall i\in[k]:\quad \mu_h(i)&=\frac{-\mathbbm{1}\{i\in\gA_{h,\times}\}a_{p,1}(e^{a_{b,1}}-1)-N_h(i)a_{p,1}}{e^{a_{b,0}}+|\gA_{h,\times}|e^{a_{b,1}}+h-1-|\gA_{h,\times}|},\quad \mu_h(\up)=0,
\end{align}
and
\begin{align}\label{eq:nu_h_simplified_imbalanced} 
    \forall i\in[k]:\quad \nu_h(i)&=\begin{cases}
        \left(1+O\left(\frac{1}{\ln^2 k}\right)\right)\lambda_i, & \text{if } c_h=0 \\
        O\left(\frac{1}{k\ln^2 k}\right)\lambda_i, & \text{if } c_h=\times
    \end{cases},\notag\\
    \nu_h(\up)&=\begin{cases}
        -\left(1+O\left(\frac{1}{\ln^2 k}\right)\right)a_{q,0}, & \text{if } c_h=0 \\
        \left(1+O\left(\frac{1}{k\ln^2 k}\right)\right)a_{q,\times}, & \text{if } c_h=\times
    \end{cases}.
\end{align}

Same as in the proof of Theorem~\ref{thm:transformer_construction} in Appendix~\ref{sec:proof_construction_balanced}, for each step $h\in[H]$, if the trajectory has not terminated yet, only four cases would occur: (i) $c_h=0,|\gA_{h,\times}|=0$: the agent enters a non-leaf node for the first time; (ii) $c_h=\times,|\gA_{h,\times}|=0$: the agent is at a non-goal leaf; (iii) $c_h=0,|\gA_{h,\times}|\in[k-1]$: the agent is at a non-leaf node and has already explored part of its children; (iv) $c_h=0,|\gA_{h,\times}|=k$: the agent is at a non-leaf node and has explored all $k$ children.

For case (i), from \eqref{eq:pi_h_simplified_imbalanced}, \eqref{eq:mu_h_simplified_imbalanced}, \eqref{eq:nu_h_simplified_imbalanced}, our choice of $a_{p,1},a_{b,0},a_{q,0},\lambda_i$ in \eqref{eq:constraints_imbalanced} and the fact that 
$$N_h(i)\leq N/k\,\,\forall i\in[k]$$ 
we know that when $c_h=0,|\gA_{h,\times}|=0$,
\begin{align}
    1-\pi_h(1)&\leq \frac{\exp\left(\mu_h(\up)+\nu_h(\up)\right)+\sum_{i=2}^k\exp\left(\mu_h(i)+\nu_h(i)\right)}{\exp\left(\mu_h(1)+\nu_h(1)\right)}\notag\\
    &\leq \frac{\exp\left(\exp\left(-\left(1+O\left(\frac{1}{\ln^2 k}\right)\right)a_{q,0}\right)+\sum_{i=2}^k\exp\left(\left(1+O\left(\frac{1}{\ln^2 k}\right)\right)\lambda_i\right)\right)}{\exp\left(\left(1+O\left(\frac{1}{\ln k}\right)\right)\lambda_1\right)}\leq \frac{\epsilon}{2N}
\end{align}
when constant $C$ is large enough in \eqref{eq:constraints_imbalanced}. This suggests when entering a non-leaf node for the first time, the agent will act the same as the optimal policy $\pi^\star$ and explore its first child with probability at least $1-\frac{\epsilon}{2N}$.

Similarly, For case (ii), from \eqref{eq:pi_h_simplified_imbalanced},\eqref{eq:mu_h_simplified_imbalanced} \eqref{eq:nu_h_simplified_imbalanced}, our choice of $a_{p,1},a_{b,0},\lambda_i,a_{q,\times}$ in \eqref{eq:constraints_imbalanced} we know that when $c_h=\times$,
\begin{align}
    1-\pi_h(\up)&\leq \frac{\sum_{i=1}^k\exp\left(\mu_h(i)+\nu_h(i)\right)}{\exp\left(\mu_h(\up)+\nu_h(\up)\right)} \leq \frac{k}{\exp\left(1+O\left(\frac{1}{\ln k}\right)a_{q,\times}\right)}\leq \frac{\epsilon}{2N}
\end{align}
given constant $C$ large enough in \eqref{eq:constraints_imbalanced}. This suggests when at a non-goal leaf, the agent will act the same as the optimal policy $\pi^\star$ and go up with probability at least $1-\frac{\epsilon}{2N}$.

For case (iii) where the agent is at a node whose $i\coloneqq |\gA_{h,\times}|\in[k-1]$ children have been explored, we let $m$ be the smallest index among legal action set $[k]\setminus\gA_{h,\times}$. Then from \eqref{eq:pi_h_simplified_imbalanced} we deduce
\begin{align}\label{eq:pi_h_m_simplified_imbalanced}
    &1-\pi_h(m)\notag\\
    &\leq \frac{\exp\left(\mu_h(\up)+\nu_h(\up)\right)+\sum_{j\in\gA_{h,\times}}\exp\left(\mu_h(j)+\nu_h(j)\right)+\sum_{j\in[k]\setminus\gA_{h,\times}\cup\{m\}}\exp\left(\mu_h(j)+\nu_h(j)\right)}{\exp\left(\mu_h(m)+\nu_h(m)\right)}.
\end{align}
From \eqref{eq:mu_h_simplified_imbalanced}, \eqref{eq:nu_h_simplified_imbalanced}, our choice of $a_{p,1},a_{b,0},a_{b,1},\lambda_i,a_{q,0}$ in \eqref{eq:constraints_imbalanced} we know that
\begin{align}\label{eq:mu_h_j_nu_h_j_simplified_imbalanced}
    \forall j\in\gA_{h,\times}:\quad \mu_h(j)+\nu_h(j)&=-\frac{1}{i}\left(1+O\left(\frac{1}{\ln k}\right)\right)a_{p,1}+\lambda_j,\notag\\
    \forall j\in[k]\setminus\gA_{h,\times}:\quad \mu_h(j)+\nu_h(j)&=O\left(\frac{1}{k\ln^2 k}\right)a_{p,1}+\lambda_j,\notag\\
    \mu_h(\up)+\nu_h(\up)&=-\left(1+O\left(\frac{1}{\ln^2 k}\right)\right)a_{q,0}.
\end{align}
Substituting \eqref{eq:mu_h_j_nu_h_j_simplified_imbalanced} into \eqref{eq:pi_h_m_simplified_imbalanced}, and by our choice of index $m$ and $a_{p,1},a_{q,0},\lambda_i$ in \eqref{eq:constraints_imbalanced} we get
\begin{align}
    1-\pi_h(m)&\leq \frac{\epsilon}{2N}
\end{align}
when constant $C$ is large enough in \eqref{eq:constraints_imbalanced}. This implies when at a node whose children are only partially visited, the agent will choose the child $m$ with the smallest index among the unexplored children (i.e., $m=\argmax_{j\in[k]\setminus\gA_{h,\times}}p^a_j$), same as what $\pi^\star$ does, with probability at least $1-\frac{\epsilon}{2N}$.

For case (iv) where the agent is at a node whose $k$ children all have been explored ($|\gA_{h,\times}|=k$), from \eqref{eq:mu_h_simplified_imbalanced}, \eqref{eq:nu_h_simplified_imbalanced}, \eqref{eq:constraints_imbalanced} we know that 
\begin{align}
    \forall i\in[k]:\quad \mu_h(i)+\nu_h(i)=-\frac{1}{k}\left(1+O\left(\frac{1}{\ln k}\right)\right)a_{p,1}+\lambda_i,\notag\\
    \mu_h(\up)+\nu_h(\up)=-\left(1+O\left(\frac{1}{\ln^2 k}\right)\right)a_{q,0}.
\end{align}
By the above relation, \eqref{eq:pi_h_simplified_imbalanced} and our choice of $a_{p,1},a_{q,0},\lambda_i$ in \eqref{eq:constraints_imbalanced} we have 
\begin{align}
    1-\pi_h(\up)&\leq \frac{\sum_{i=1}^k\exp\left(\mu_h(i)+\nu_h(i)\right)}{\exp\left(\mu_h(\up)+\nu_h(\up)\right)}\leq \frac{\epsilon}{2N}
\end{align}
when constant $C$ is large enough in \eqref{eq:constraints_imbalanced}. This implies when at a node whose children all have been explored, the agent will again act the same as $\pi^\star$ and go up with probability at least $1-\frac{\epsilon}{2N}$.

Putting the results of the above four cases together, and using the fact that the trajectory length is upper bounded by $H=2N$, we know that \eqref{eq:R_pi_Phi_imbalanced} in Theorem~\ref{thm:transformer_construction_imbalanced} holds under our construction.


\end{document}